\documentclass[lettersize,journal]{IEEEtran}

\usepackage{graphicx}
\usepackage{subfig}
\usepackage{xspace}
\usepackage{multirow}
\usepackage{booktabs}
\usepackage{enumitem}
\usepackage{amsmath}
\usepackage{microtype}
\usepackage{tikz}
\usepackage{amssymb}
\usepackage{url}
\usepackage{hyperref}
\usepackage{makecell}
\usepackage{dblfloatfix}
\usetikzlibrary{spy}

\AtBeginDocument{}

\newif\ifarxivversion
\arxivversiontrue

\title{Cinematic Compositing Using Character-Environment-Harmonized Video Generation Models}
\author{Tianyi Xiang, Mingming He, Li Ma, and Jing Liao}
\newcommand{\ZoomInBox}[5]{
    \begin{tikzpicture}[spy using outlines={#1}]
        
        \node[anchor=south west,inner sep=0] (image) at (0,0) {\includegraphics[width=#5]{#4}};

        \begin{scope}[x={(image.south east)}, y={(image.north west)}] 
            
            \spy on (#2) in node[fill=white] at (#3);
        \end{scope}
    \end{tikzpicture}
}

\begin{document}
\IEEEtitleabstractindextext{
\begin{abstract}
Cinematic compositing aims to integrate green-screen characters into novel environments while maintaining physical and photometric realism. Previous methods often fail to capture the complex bidirectional interactions between characters and their surroundings, which we characterize as Character-to-Environment (C2E) physical interaction and Environment-to-Character (E2C) lighting harmonization. To address this, we propose an end-to-end video diffusion framework that jointly models C2E and E2C interactions, specifically handling the challenges of interactive props. Our approach introduces a tri-mask-guided architecture with RGB-D joint denoising to ensure physically consistent interactions among the character, props, and environment. We further develop an efficient prior-driven data curation pipeline to construct high-quality relighting pairs without expensive rendering. Finally, a reference-conditioned mechanism enables controllable environment synthesis and precise prop replacement. Extensive experiments demonstrate that our framework significantly outperforms existing methods in cinematic-quality dynamic video compositing.
\ifarxivversion
Project page: \url{https://cehcomposition.github.io/demo/}
\fi
\end{abstract}
}
\maketitle
\IEEEdisplaynontitleabstractindextext
\newcommand{\revision}[1]{\textcolor{black}{#1}}
\begin{figure*}
  \includegraphics[width=\linewidth]{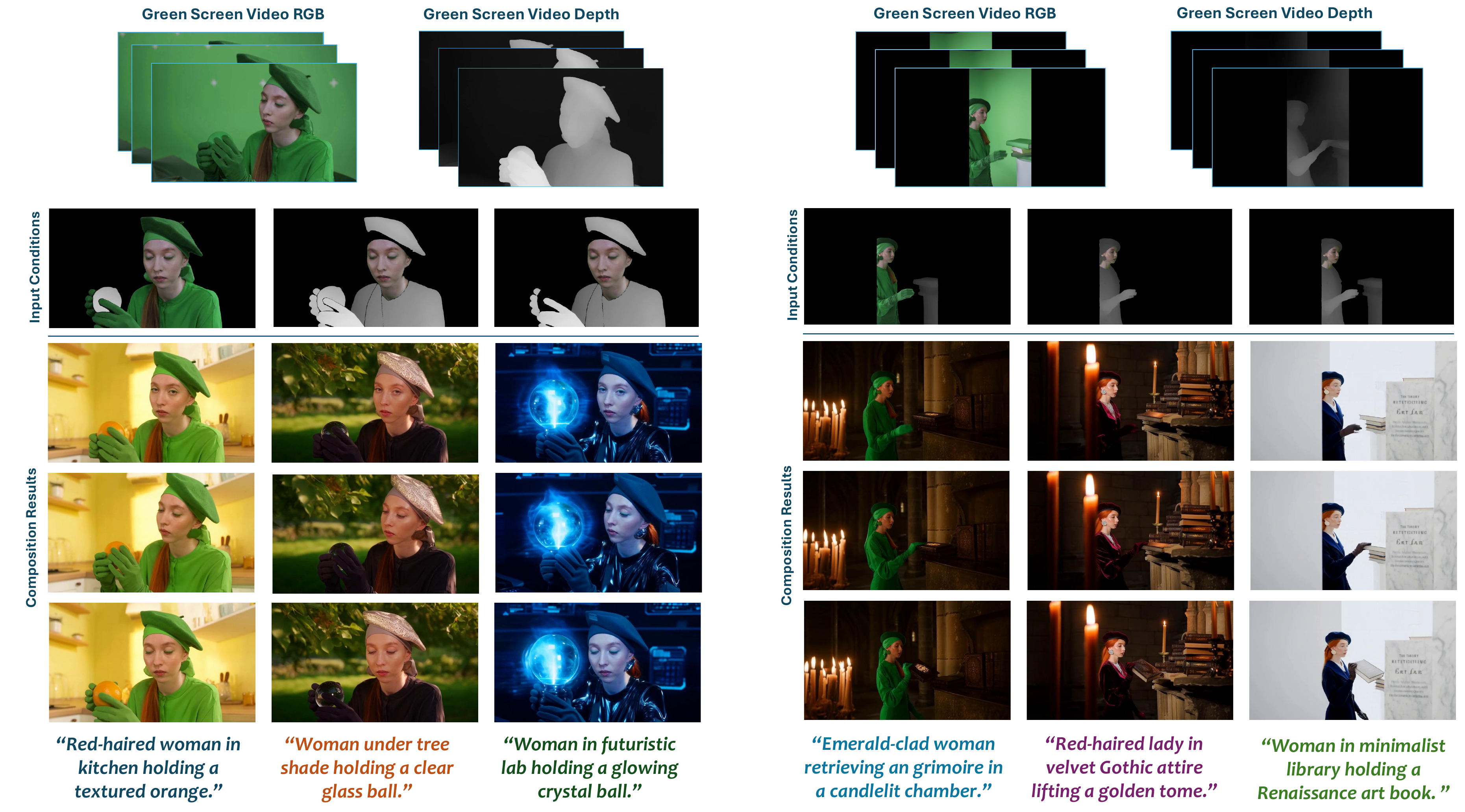}
  \caption{\textbf{Input and compositing results of our method.} We achieve cinematic-quality video compositing with bidirectional character-environment interaction. Given green-screen footage, predicted depth, and a text prompt, our method synthesizes realistic backgrounds while ensuring physical interaction and lighting harmonization. By adaptively combining conditions across different regions (RGB, depth, or unconditioned), our unified framework handles diverse production scenarios. These include performance and prop \textit{relighting} (e.g., adjusting light on faces or clothes), prop \textit{replacement} (e.g., the orange or glass ball), and prop/background \textit{generation} (e.g., the glowing crystal ball and background scene). Note that text prompts are simplified due to space constraints.}
  \label{fig:teaser}
\end{figure*}

\section{Introduction}
\label{sec:intro}

Cinematic compositing is the process of integrating dynamic foreground characters, typically captured against a green screen, into separately created background environments to produce visually coherent content. As a vital component of the VFX pipeline~\cite{hu2025animate}, it provides filmmakers with substantial creative control and enables the simulation of inaccessible environments. Traditionally, compositing relies on alpha-channel matting to perform pixel-wise blending. However, such approaches fail to model the complex interactions required between characters and their surroundings, often resulting in a lack of both physical and photometric realism.

Achieving film-quality compositing requires accurately modeling the bidirectional interaction between the character and the environment, which we define along two primary dimensions: \textit{Character-to-Environment} (C2E) and \textit{Environment-to-Character} (E2C). C2E describes physical interactions in which the character’s geometry and motion influence the surrounding environment, including occlusions, spatial contact, and environmental response. Conversely, E2C refers to lighting harmonization, describing how environmental illumination is realistically cast onto and reflected by the character.

Beyond direct interaction, props serve as a crucial mediator between the character and the environment, necessitating a specialized handling approach. Depending on the capture conditions and whether the character holds a physical object during capture, we account for three key scenarios: preserving and relighting real props, replacing green-screen proxies with synthesized objects, and generating entirely virtual props. Effectively managing these scenarios is essential for bridging the gap between isolated foreground captures and integrated cinematic narratives. \revision{However, existing methods typically employ a single binary mask to indicate whether pixels should be preserved or regenerated. This is insufficient for diverse prop-handling scenarios, since it cannot simultaneously preserve the RGB content of a real prop while allowing relighting, nor can it discard a proxy's appearance while retaining its captured geometry for interaction-aware replacement. Therefore, a more nuanced, production-oriented formulation is required to handle the distinct intents of preserving, replacing, or generating props.}

Given a green-screen foreground capture of a character and optional props, our goal is to synthesize a fully composited video that integrates a newly generated dynamic environment with a realistically relit character and any associated props. While recent advances in generative video models for inpainting and relighting show promise, they address these tasks in isolation. For example, video inpainting models~\cite{jiang2025vace,bian2025videopainter,zhang2024avid} can synthesize plausible backgrounds but lack E2C harmonization, either restricting background illumination changes or leaving the foreground lighting inconsistent with the new environment. Conversely, relighting models~\cite{zhang2025scaling,zhou2025light,fang2025relightvid} modify subject appearance but cannot generate dynamic background responses to character motion. Furthermore, existing methods lack a unified treatment of props, failing to adaptively relight, replace, or synthesize objects across diverse interaction scenarios. Consequently, neither single-task approaches nor naïvely cascaded pipelines can achieve the physically grounded bidirectional harmonization required for cinematic realism because background synthesis and foreground relighting are optimized independently.

Therefore, in this work, we propose an end-to-end generative framework that simultaneously addresses both C2E physical interaction and E2C lighting harmonization using a unified video diffusion model. To model C2E interactions, we design a tri-mask-guided generative architecture that adaptively handles diverse prop scenarios. \revision{Unlike conventional binary masks, the tri-mask explicitly maps the real production intents to modality-specific RGB and depth conditions.}
We also introduce an RGB-D joint denoising strategy to explicitly incorporate geometry information as a learning objective, allowing the model to better understand 3D spatial relations and human motion. To address E2C harmonization, we propose an efficient, prior-driven data curation pipeline. Specifically, we leverage video generation priors to synthesize multi-illumination scenes and subsequently filter them based on realism and aesthetic scoring, producing high-quality paired data for training video relighting and harmonization. Finally, we extend the conditioning mechanism for reference injection, enabling flexible control over target environments and props via reference examples.

Through extensive experiments, we demonstrate that our framework establishes a new state of the art for cinematic-quality dynamic video compositing. Our contributions are summarized as follows:

\begin{itemize}
\item \revision{We formulate green-screen compositing as bidirectional character--environment generation that jointly models C2E physical interaction and E2C lighting harmonization within a unified framework.}
\item \revision{We introduce a production-oriented tri-mask-guided architecture that converts distinct prop-handling intents into RGB and depth conditions, together with joint RGB-D denoising that improves the synthesis of physically consistent background interactions.}
\item We propose a prior-driven data construction pipeline that efficiently generates high-quality multi-illumination data without requiring expensive rendering pipelines.
\item We design a reference-conditioned mechanism that enables customizable environment synthesis and fine-grained control over interactive props.
\end{itemize}

\begin{figure*}[!t]
    \centering
    \includegraphics[width=0.99\linewidth]{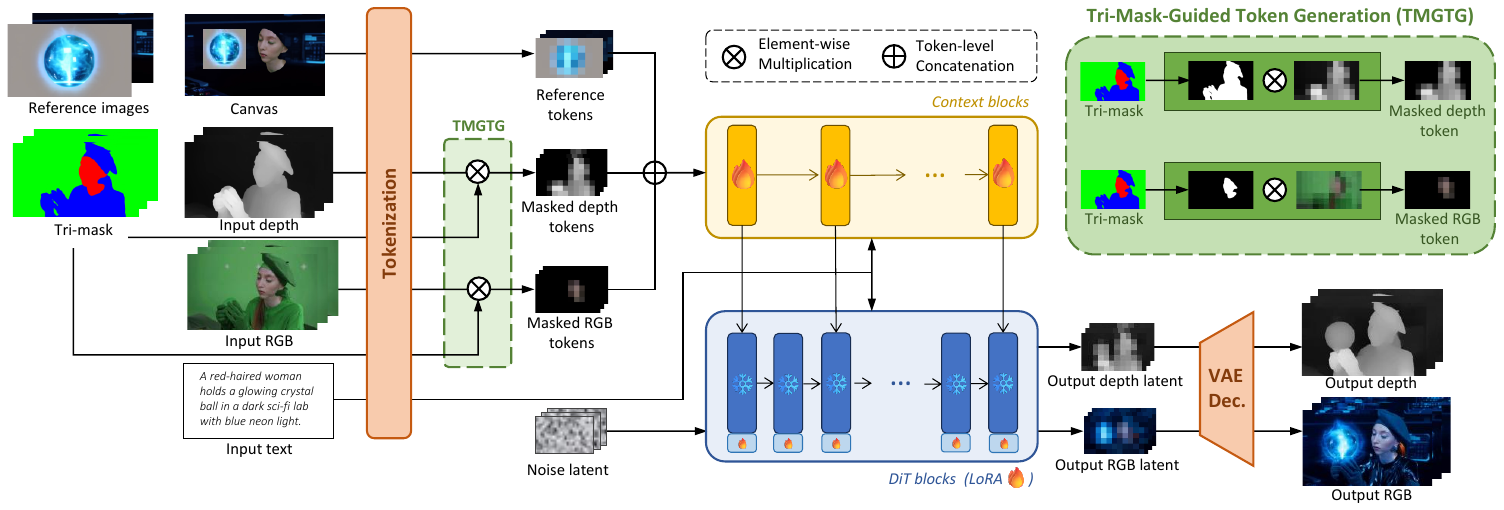}
    \vspace{-1em}
    \caption{\textbf{Overview of our framework.} We achieve bidirectional character–environment interaction by addressing physical interaction and lighting harmonization within a single diffusion-based framework. The architecture employs a Tri-Mask-Guided Token Generation (TMGTG) module, which uses a tri-mask to adaptively determine whether a region is conditioned by input RGB, depth, or remains unconditioned. These masked tokens, alongside reference and text conditions, are fed into context blocks to guide a joint RGB-D denoising process via a DiT-based backbone. By jointly denoising color and geometry in the latent space, the model synthesizes physically plausible backgrounds and interactive objects that are spatially and photometrically consistent with the actor.}
    \label{fig:overview}
\end{figure*}

\section{Related Work}
\label{sec:related}

\vspace{1mm}\noindent\textbf{Video matting and compositing.} High-fidelity foreground isolation from the captured environment is a prerequisite for cinematic compositing. Classical matting solutions such as BGMv2 \cite{lin2021real} and RVM \cite{lin2021robust} prioritize real-time efficiency and temporal stability, making them suitable for live applications. More recent advancements like MatAnyone \cite{yang2025matanyone} and MatAnyone 2 \cite{yang2026matanyone2} utilize memory propagation and learned quality evaluators to achieve high-quality extraction with fine-grained boundary details in complex scenes.

\vspace{1mm}\noindent\textbf{Video generation and background synthesis.} Beyond using alpha matting for direct foreground and background blending, background generation can be achieved using the recent generative video models which are designed for video inpainting and editing. Some unified video editing frameworks offer all-in-one solutions. For example, VACE \cite{jiang2025vace} integrates reference guidance, inpainting, and outpainting into a single multimodal interface. Given a foreground character sequence, it can inpaint a dynamic background for each frame. For long sequences, VideoPainter \cite{bian2025videopainter} and AVID \cite{zhang2024avid} employ temporal MultiDiffusion and context-aware encoders to prevent error accumulation. To address computational overhead, EditCtrl \cite{litman2026editctrl} introduces a local context module that restricts diffusion processing to masked tokens, allowing for efficient high-resolution editing.

Background synthesis has shifted toward subject-aware generation. ActAnywhere \cite{pan2024actanywhere} proposes environment generation conditioned on subject segmentation, while \cite{yao2025beyond} introduce explicit camera-pose control to ensure background motion aligns with complex foreground trajectories. Unlike general video inpainting, these models attempt to synthesize realistic environmental responses, including subtle interactions, shadows, and secondary lighting. However, the generated environmental illumination remains constrained by the initial foreground lighting. Consequently, these methods struggle to achieve realistic or desired environmental illumination if the target environment's radiance significantly deviates from the original foreground capture. Recent training-free methods, AnyPortal~\cite{gao2025anyportal} and FlowPortal~\cite{gao2026flowportal}, combine image and video diffusion priors for video background replacement and foreground relighting, using refinement projection and residual-corrected flow, respectively, to preserve foreground detail and temporal consistency. Despite their strong temporal consistency, these methods still perform composition by directly replacing the background and compositing a relit foreground, without explicitly modeling dynamic foreground--background interactions. As a result, when the target background contains objects that should interact with the actor, this decoupled process can introduce substantial artifacts around contact regions and foreground boundaries.

\vspace{1mm}\noindent\textbf{Character interactive motion.} Realistic interaction between characters and their surroundings is vital for visual coherence. Frameworks such as Animate Anyone 2 \cite{hu2025animate} and AnchorCrafter \cite{xu2026anchorcrafter} integrate environmental affordance and interactive object features to ensure physical plausibility. MimicMotion \cite{zhang2024mimicmotion} and ByteLOOM \cite{liu2025byteloom} further refine these interactions by introducing confidence-aware pose guidance and geometric priors to prevent artifacts such as hand-object interpenetration. However, these methods primarily focus on character or interactive object generation and struggle to synthesize dynamic backgrounds with physically consistent environmental responses to character motion.

\vspace{1mm}\noindent\textbf{Video relighting and harmonization.} To achieve lighting consistency between foreground characters and background environments, video relighting or harmonization methods have emerged as potential solutions. Earlier harmonization works adjust appearance via parametric shading maps \cite{wang2023semi}, scene depth~\cite{bao2022interactive}, diffusion priors \cite{ren2024relightful}, intrinsic properties~\cite{guo2026pnprorl}, or 3D Gaussian representation~\cite{chen2026relightable}. Recent video relighting generally follows two paradigms: training-free methods like LightCtrl \cite{peng2026lightctrl} and Light-A-Video \cite{zhou2025light} adapt image-based priors (e.g., IC-Light \cite{zhang2025scaling}) to video, while 4D-aware approaches such as \cite{hu2025ex4d,xiao2026relitlive,liu2026lightx} ensure consistency across viewpoints. HarmoVid~\cite{choi2026relightful} improves video portrait harmonization by training a video diffusion model on deflickered real and synthetic data and using asymmetric alpha-mask conditioning for temporally stable lighting and clean boundaries. Modern unified models (e.g., \cite{he2025unirelight,liu2025tclight,damo2025unilumos} further enhance physical plausibility by jointly modeling intrinsics and lighting.

However, acquiring high-quality real-world multi-illumination data for training data-driven methods remains challenging. Existing approaches often rely on game engines to construct paired datasets. For example, RelightVid~\cite{fang2025relightvid} and RelightMaster~\cite{bian2025relightmaster} utilize high-quality synthetic datasets rendered using game engines. Similarly, MoCha~\cite{xu2026mocha} also employs rendered paired videos to internalize complex light transport effects. In contrast, our method adopts a prior-driven data curation pipeline that eliminates the need for expensive rendering while preserving high visual fidelity.

\section{Method}
\label{sec:method}

\begin{figure}
    \centering
    \includegraphics[width=0.95\linewidth]{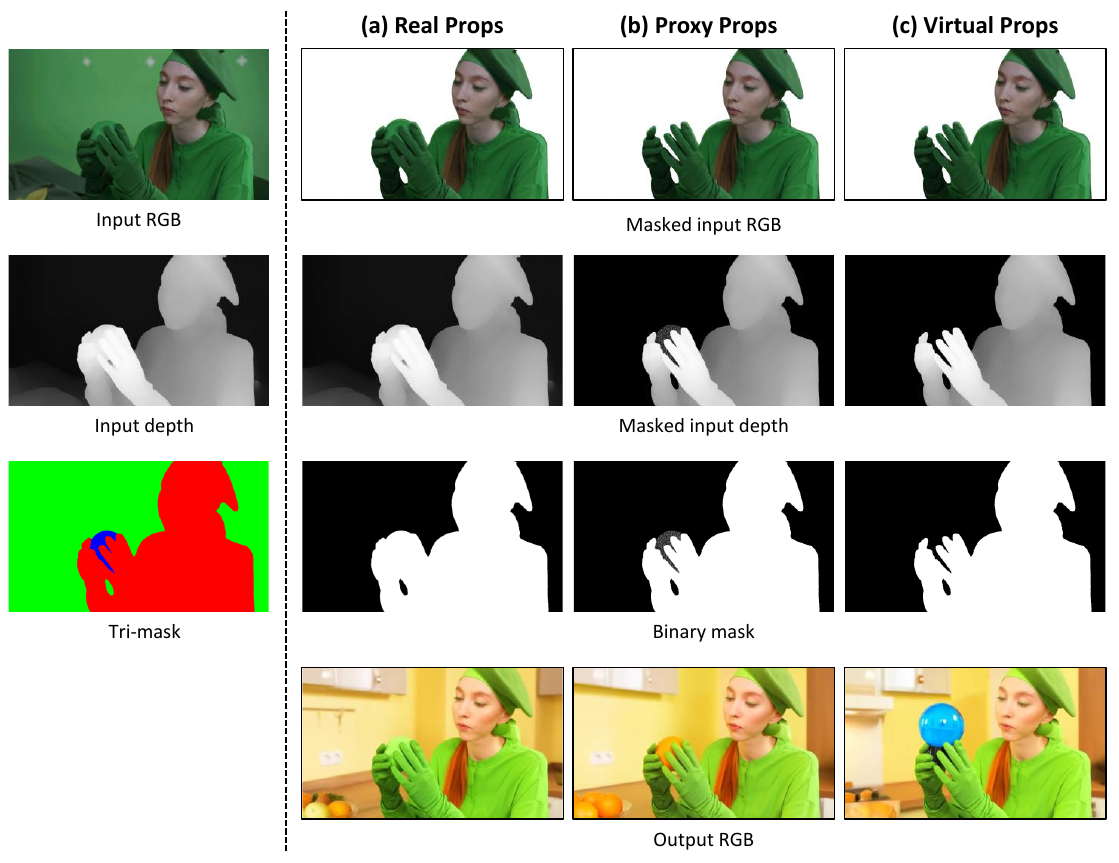}
    \vspace{-1em}
    \caption{\textbf{Multi-granularity foreground modeling.} We visualize the input, condition and output of three scenarios:  (a) Real Props (\textcolor{red}{Red} mask): Preserving and relighting existing physical objects.  (b) Proxy Props (\textcolor{blue}{Blue} mask): Replacing green-screen proxies with new objects; by utilizing sparse depth, the model ignores fine-grained material and texture details to adapt to different objects during replacement. (c) Virtual Props (\textcolor{green}{Green} mask): Synthesizing entirely new interactive elements.}
    \label{fig:trimask}
    \vspace{-10pt}
\end{figure}

In this section, we detail the formalization, data curation strategy, and architectural design of our framework. We first formulate green-screen compositing as a tri-mask-guided generative task, focusing on bidirectional character-environment interactions. Next, we describe our unified five-tuple training data structure and creation, which enables the model to simultaneously learn physical interactions and lighting harmonization. Finally, we present our tri-mask-guided RGB-D diffusion architecture, along with a reference injection mechanism designed for the precise and controllable generation of background environments and interactive objects.

\subsection{Problem Formulation}
\label{sec:problem}
Given a green-screen video $\mathbf{V}_{gs}$ and a mask $\mathbf{M}_t$, our goal is to synthesize a \textbf{composited} video $\mathbf{V}_{comp}$ that integrates the character into a newly generated environment while simultaneously enabling character-to-environment (C2E) physical interaction and environment-to-character (E2C) lighting harmonization.

Standard binary masks used in inpainting enforce an all-or-nothing preserve-or-generate decision, which fails to capture the full complexity of real-world green-screen captures. For example, in practical shooting pipelines, props typically fall into three distinct categories: real props to be preserved, green proxy props to be replaced, and virtual props to be generated. Standard binary masks cannot properly distinguish between these three cases, each requiring a fundamentally different processing strategy. To address this limitation, we introduce a tri-mask formulation as shown in Figure~\ref{fig:trimask}:
\begin{itemize}
    \item $\mathbf{M}_t=1$: \textbf{Preserve-and-relight regions} (e.g. character face, real props) where RGB content is retained but photometrically harmonized to match the target environment
    \item $\mathbf{M}_t=0$: \textbf{Geometry-preserving regions} (e.g. green-screen proxy props) where depth anchors are kept while RGB appearance and material properties are fully regenerated
    \item $\mathbf{M}_t=-1$: \textbf{Full-generation regions} (e.g. background, virtual interactive props) where content is created entirely from scratch without physical capture counterpart
\end{itemize}
This formulation supports three spatially adaptive generation modes, unifying all practical green-screen compositing cases in a single interface and offering flexibility for diverse shooting setups.

The generation of the composited video $\mathbf{V}_{comp}$ is guided by the text prompt $\mathcal{T}$, as well as optional reference images to specify desired environment or props. In our diffusion-based video generation framework, we introduce joint depth and RGB denoising to enhance the model's spatial understanding and improve the synthesis of physically consistent interactions. Next, we introduce our training data curation pipeline, followed by tri-mask-guided RGB-D diffusion, and reference injection methods.

\subsection{Training Data Curation}
\label{sec:data}
The tri-mask formulation naturally defines each training sample as a five-tuple
\begin{equation}
\mathcal{X} = \left(\mathbf{V}_{gt}, \mathbf{V}_{aug}, \mathbf{D}, \mathbf{M}_t, \mathcal{T}\right),
\end{equation}
where $\mathbf{V}_{gt}$ is the ground-truth video, $\mathbf{V}_{aug}$ is its lighting-augmented counterpart, $\mathbf{D}$ is the video depth, $\mathbf{M}_t$ is the tri-mask, and $\mathcal{T}$ is the text prompt. These five elements correspond directly to the three region types in Sec.~\ref{sec:problem}: regions with $\mathbf{M}_t=1$ use $\mathbf{V}_{aug}$ to learn relighting, regions with $\mathbf{M}_t=0$ use $\mathbf{D}$ to learn physically plausible interaction while regenerating RGB appearance, and regions with $\mathbf{M}_t=-1$ are mainly guided by $\mathcal{T}$ and the surrounding context for open-ended generation.
During training, $\mathbf{V}_{aug}$ and $\mathbf{V}_{gt}$ respectively serve as the input and supervision counterparts of the green-screen input $\mathbf{V}_{gs}$ and the composited output $\mathbf{V}_{comp}$ defined in Sec.~\ref{sec:problem}.

\subsubsection{Overall Design}
We construct the five-tuple from two complementary sources. A \emph{base subset} built from HOIGen1M~\cite{liu2025hoigen} provides realistic human motion and contact cues for C2E interaction modeling, while an additional \emph{multi-illumination subset} enriches the supervision for E2C harmonization. The former emphasizes realistic interaction structure, whereas the latter expands the illumination distribution with more cinematic color cast, contrast, dynamics, and lighting direction. Despite their different roles, both subsets are converted into the same representation so that the model always receives a unified training interface.

Under this design, the five elements of $\mathcal{X}$ are aligned with the three tri-mask states. $\mathbf{V}_{gt}$ serves as direct supervision for regions that should be preserved but relit from $\mathbf{V}_{aug}$, $\mathbf{D}$ provides geometric guidance for regions whose depth remains trustworthy while RGB must be regenerated, and $\mathcal{T}$ mainly drives the open-ended generation regions. This organization allows the model to jointly learn physically grounded composition and foreground lighting harmonization within a single data formulation.

\subsubsection{Implementation Details}
The base subset is curated from the large-scale dataset HOIGen1M with strict identity-preserving filters. Since our task requires the character to remain temporally and semantically consistent, we retain only clips with a single stable foreground character, verified by face detection confidence, face pose, person detection consistency, and spatial agreement between face and person localization. After filtering, we obtain approximately $2\times 10^5$ valid clips, each trimmed to 81 frames to match the training setup of our backbone model.

For each retained clip, we construct the conditions required by the tri-mask formulation. To generate $\mathbf{V}_{aug}$, we use IC-Light~\cite{zhang2025scaling} conditioned on diverse background images from~\cite{quattoni2009recognizing,sun2021deep,xiang2025let,li2022bridging} to perturb the original lighting of $\mathbf{V}_{gt}$.
We then parse the HOIGen1M captions with Qwen3-32B~\cite{yang2025qwen3} to identify interacted objects and segment them with Grounded SAM 2~\cite{ravi2025sam,liu2024grounding}.

Monocular video depth is estimated by Video Depth Anything~\cite{chen2025video}. For each detected interaction object, we randomly assign one of the three tri-mask states, allowing each detected object to be treated during training as a relit real prop, a proxy prop with reliable depth but regenerated RGB, or a fully generated object composited together with the background, with equal probability.

To enrich illumination diversity beyond HOIGen1M, we construct a multi-illumination subset from the high-quality and diverse background scene images in BG-20K~\cite{li2022bridging}. We first use BLIP-2~\cite{li2023blip} to extract scene keywords from each background image. Conditioned on these keywords, a large language model generates a text prompt describing a person performing a simple action in an environment consistent with the background scene, while specifying diverse and cinematic illumination. We feed these prompts to Wan 2.2 T2V A14B~\cite{wan2025wan} to synthesize videos with varied lighting conditions while retaining the intended scene semantics. The generated clips are filtered using the same identity-preserving criteria as the base subset, together with an additional aesthetic-quality filter~\cite{schuhmann2022improved}.
The retained clips are treated as $\mathbf{V}_{gt}$ in the five-tuple format, and then are processed by the same IC-Light-based augmentation and annotation pipeline to obtain lighting-augmented variants, person and object masks, and video depth.
Thus, the two subsets are aligned in the five-tuple representation and can be merged together during training.

\subsection{Tri-Mask-Guided RGB-D Diffusion}
Recent video inpainting models such as VACE~\cite{jiang2025vace} have demonstrated strong performance on mask-guided video generation, making them a natural starting point for our task. However, they are not directly suitable for green-screen compositing for two reasons. First, their conditioning interface assumes a binary mask, which cannot express the three distinct foreground handling modes required in our setting. Second, they operate with RGB denoising only, and therefore cannot explicitly exploit depth cues that are crucial for modelling interaction geometry.

To address the first limitation, we replace binary-mask conditioning with tri-mask-guided conditioning. As shown in Figure~\ref{fig:trimask}, the tri-mask is converted into modality-specific masks for RGB and depth. Let $\mathbf{M}_{rgb}$ and $\mathbf{M}_{d}$ denote the RGB-preservation mask and the depth-preservation mask, respectively. We define them from the tri-mask $\mathbf{M}_t \in \{-1,0,1\}^{T \times H \times W}$ as
\begin{equation}
\mathbf{M}_{rgb} = \mathbb{I}[\mathbf{M}_t = 1],
\qquad
\mathbf{M}_{d} = \mathbb{I}[\mathbf{M}_t \geq 0],
\end{equation}
where $\mathbb{I}[\cdot]$ is the indicator function. Therefore, regions with $\mathbf{M}_t=1$ preserve RGB and depth as relit foreground structure, regions with $\mathbf{M}_t=0$ regenerate RGB while retaining depth as a geometric anchor, and regions with $\mathbf{M}_t=-1$ regenerate both RGB and depth.

The resulting masked inputs are written as
\begin{equation}
\widetilde{\mathbf{V}}_{rgb} = \mathbf{M}_{rgb} \odot \mathbf{V}_{gs},
\qquad
\widetilde{\mathbf{D}} = \mathbf{M}_{d} \odot \mathbf{D},
\end{equation}
where $\odot$ denotes element-wise multiplication. In the proxy-prop case, the captured stand-in may have a material appearance that is inconsistent with the desired generated object. To reduce this material-text discrepancy, we randomly drop a subset of valid depth entries in geometry-preserving regions and obtain a new depth mask $\mathbf{M}^{s}_{d}$ which is partially sparse. This process can be formulated as:
\begin{equation}
\mathbf{M}^{s}_{d} = \mathbf{M}_{d} \odot (\mathbf{1} - \mathbf{R} \odot \mathbb{I}[\mathbf{M}_t = 0]),
\end{equation}
where $\mathbf{R}$ is a random binary mask whose entries independently equal $1$ with probability $p_R$. The actual depth condition then becomes $\widetilde{\mathbf{D}} = \mathbf{M}^{s}_{d} \odot \mathbf{D}$ in those regions. During training, we either use $\mathbf{M}_{d}$ or $\mathbf{M}^{s}_{d}$ with equal probability to produce $\widetilde{\mathbf{D}}$, which preserves coarse geometry without over-constraining the final appearance.

To address the second limitation, we perform RGB-D joint denoising rather than RGB-only denoising. Our network is built on a pre-trained text-to-video Diffusion Transformer with a parallel context branch inspired by VACE~\cite{jiang2025vace}. The context branch encodes the tri-mask-conditioned RGB and depth inputs, while the base branch processes the noisy latent trajectory. Let $\mathbf{z}^{rgb}_t \in \mathbb{R}^{L \times C}$ and $\mathbf{z}^{d}_t \in \mathbb{R}^{L \times C}$ denote the noisy RGB and depth latent sequences at diffusion step $t$, where $L$ is the spatiotemporal token length of a single modality. We concatenate them along the sequence dimension to form the joint trajectory:
\begin{equation}
\mathbf{x}_t = \operatorname{Concat}_{seq}\!\left(\mathbf{z}^{rgb}_t, \mathbf{z}^{d}_t\right) \in \mathbb{R}^{2L \times C}.
\end{equation}
Similarly, after tri-mask decoupling, we obtain RGB and depth context sequences $\mathbf{C}_{rgb}$ and $\mathbf{C}_{d}$, and build the joint condition as
\begin{equation}
\mathbf{C} = \operatorname{Concat}_{seq}\!\left(\mathbf{C}_{rgb}, \mathbf{C}_{d}\right) \in \mathbb{R}^{2L \times C_c}.
\end{equation}
We do not concatenate the two modalities along channels, because doing so would alter the input dimensionality expected by the pre-trained backbone.

Before patch embedding, we split the joint sequence back into two aligned halves,
\begin{equation}
\mathbf{x}^{rgb}_t = \mathbf{x}_t[1{:}L], \qquad \mathbf{x}^{d}_t = \mathbf{x}_t[L{+}1{:}2L],
\end{equation}
and feed them through the same patch embedder separately:
\begin{equation}
\mathbf{h}^{rgb}_0 = \mathcal{E}(\mathbf{x}^{rgb}_t) + \mathbf{e}_{rgb}, \qquad
\mathbf{h}^{d}_0 = \mathcal{E}(\mathbf{x}^{d}_t) + \mathbf{e}_{d},
\end{equation}
where $\mathcal{E}(\cdot)$ denotes the patch embedding layer, and $\mathbf{e}_{rgb}$ and $\mathbf{e}_{d}$ are learnable modality embeddings. The RGB embedding is initialized to zero to preserve the original RGB prior, while the depth embedding is randomly initialized so the model can explicitly distinguish geometric tokens from visual tokens. We further modify 3D RoPE so that $\mathbf{h}^{rgb}_0$ and $\mathbf{h}^{d}_0$ share identical temporal positional indices. After embedding, the two streams are concatenated back along the sequence dimension,
\begin{equation}
\mathbf{h}_0 = \operatorname{Concat}_{seq}\!\left(\mathbf{h}^{rgb}_0, \mathbf{h}^{d}_0\right),
\end{equation}
and jointly denoised by the transformer blocks under condition $\mathbf{C}$. This design preserves the pre-trained DiT interface while enabling unified reasoning over appearance and geometry in a single latent process.

During training, we adopt a mask-guided RGB-D velocity loss that emphasizes foreground and interaction-critical regions without destabilizing the pre-trained optimization scale. Let $\hat{\mathbf{v}} = \mathbf{v}_\theta(\mathbf{x}_t, \mathbf{C}, \mathcal{T})$ denote the predicted velocity for the concatenated RGB-D latent trajectory. We compute the squared velocity error and normalize it separately inside and outside the object-aware foreground mask, yielding
\begin{equation}
\mathcal{L}_{\mathrm{in}} = \frac{\sum \ell(\hat{\mathbf{v}}, \mathbf{v}) \odot \mathbf{M}_f}{\sum \mathbf{M}_f + \delta}, \qquad
\mathcal{L}_{\mathrm{out}} = \frac{\sum \ell(\hat{\mathbf{v}}, \mathbf{v}) \odot (1-\mathbf{M}_f)}{\sum (1-\mathbf{M}_f) + \delta},
\end{equation}
and combine them as
\begin{equation}
\mathcal{L}_{\mathrm{mask}} = \alpha \, \mathcal{L}_{\mathrm{in}} + (1-\alpha) \, \mathcal{L}_{\mathrm{out}}, \qquad
\alpha = \frac{s}{1+s},
\end{equation}
where $\mathbf{M}_f := \mathbb{I}[\mathbf{M}_t \geq  0]$ is the foreground mask, $s$ is the foreground emphasis scale, $\ell$ is the squared error function, and $\delta = 10^{-8}$. This formulation improves learning on character and interaction regions while keeping the overall loss magnitude well behaved. For efficient adaptation, we fully train the inserted context blocks and fine-tune the base DiT with LoRA modules~\cite{hu2022lora}.

\subsection{Reference Injection (Optional)}
Our framework supports both reference-free and reference-guided generation for the background and the interactive props.
Unlike general reference-based generation, our setting often provides not only the appearance of an object reference image but also its approximate spatial location in the first frame. We therefore inject references with an explicit spatial prior instead of using appearance tokens alone.

Each reference image is first encoded by the VAE encoder and then concatenated to the conditional token sequence. In addition, we introduce an RGB canvas token to explicitly represent the intended first-frame layout. The canvas is constructed by compositing elements from back to front according to depth order: we first place the background, then the foreground character, and finally each reference object at its first-frame location. We use the depth map of the first frame to guide the compositing process. The composed canvas is then encoded by the same VAE encoder and concatenated after the individual reference tokens.

This design provides the network with both appearance cues and an explicit spatial anchor. As a result, the model can better disambiguate where each reference object should appear and how it should interact with the character and environment, leading to more precise reference controllability in green-screen compositing.

\section{Experiments}
\label{sec:exp}

\newcommand{\qualicompwidth}{0.325\linewidth}
\newcommand{\qualrow}[3]{
\multicolumn{3}{l}{\scriptsize #1} \\
\includegraphics[width=\qualicompwidth]{figures/quali_comp/#2_frame026/#3} &
\includegraphics[width=\qualicompwidth]{figures/quali_comp/#2_frame056/#3} &
\includegraphics[width=\qualicompwidth]{figures/quali_comp/#2_frame071/#3} \\}
\begin{figure*}[t]
\centering
\begin{minipage}[t]{0.49\textwidth}\centering
\setlength{\tabcolsep}{0.01em}\renewcommand{\arraystretch}{0.7}
\begin{tabular}{ccc}
\qualrow{Green Screen}{pexel_9488551_p0}{01_input}
\qualrow{Tri-mask}{pexel_9488551_p0}{11_trimask}
\qualrow{FLUX 2 + WanAnimate}{pexel_9488551_p0}{04_animate}
\qualrow{FLUX 2 + MimicMotion}{pexel_9488551_p0}{05_mimicmotion}
\qualrow{VACE}{pexel_9488551_p0}{07_vace}
\qualrow{VACE + ICLight}{pexel_9488551_p0}{08_vace_relighting}
\qualrow{FlowPortal}{pexel_9488551_p0}{09_flowportal}
\qualrow{AnyPortal}{pexel_9488551_p0}{10_anyportal}
\qualrow{Ours}{pexel_9488551_p0}{06_ours}
\multicolumn{3}{p{0.98\linewidth}}{\scriptsize \textit{A stylish performer adjusts his bright bucket hat on a concert stage, while dramatic magenta and golden spotlights sweep across the scene.}}
\end{tabular}
\end{minipage}\hfill
\begin{minipage}[t]{0.49\textwidth}\centering
\setlength{\tabcolsep}{0.01em}\renewcommand{\arraystretch}{0.7}
\begin{tabular}{ccc}
\qualrow{Green Screen}{pexel_6134541_new}{01_input}
\qualrow{Tri-mask}{pexel_6134541_new}{11_trimask}
\qualrow{FLUX 2 + WanAnimate}{pexel_6134541_new}{04_animate}
\qualrow{FLUX 2 + MimicMotion}{pexel_6134541_new}{05_mimicmotion}
\qualrow{VACE}{pexel_6134541_new}{07_vace}
\qualrow{VACE + ICLight}{pexel_6134541_new}{08_vace_relighting}
\qualrow{FlowPortal}{pexel_6134541_new}{09_flowportal}
\qualrow{AnyPortal}{pexel_6134541_new}{10_anyportal}
\qualrow{Ours}{pexel_6134541_new}{06_ours}
\multicolumn{3}{p{0.98\linewidth}}{\scriptsize \textit{Two young men play chess on a large grey sofa in a modern room, as warm golden-orange twilight streams through a window overlooking the city skyline.}}
\end{tabular}
\end{minipage}

\caption{\textbf{Qualitative comparison on representative green-screen examples.} Our method better preserves actor identity, interaction geometry, and lighting coherence than the baselines.}
\label{fig:qual_main}
\end{figure*}

\subsection{Implementation Details}
We build our model on top of Wan 2.1 VACE‑14B~\cite{wan2025wan,jiang2025vace}, which also serves as the pre‑training initialization. The learning rate is set to $2 \times 10^{-5}$ with a total batch size of $8$. We first train the model on 17-frame clips for 15K iterations, and after convergence, continue fine-tuning on 65-frame clips for another 2K iterations. The LoRA rank used in the DiT blocks is set to 32. During training, we sample the base dataset with probability 0.8 and the multi-illumination dataset with probability 0.2. $p_R$ is randomly sampled from a uniform distribution over $[0.85, 0.95]$. The mask-guided velocity loss uses $s=10$, and all training samples are resized to a spatial resolution of $480 \times 832$.

\subsection{Benchmarks and Baselines}
\noindent \textbf{Benchmarks.} We evaluate our method on two benchmarks. The first is a \emph{synthetic benchmark}. Following the data‑filtering criteria in Sec.~\ref{sec:data}, we sample 100 video clips from the base dataset and exclude them from the training set. These samples are constructed by extracting pseudo green‑screen foregrounds from real videos, such that their original videos serve as ground truth for objective evaluation. Note that input videos of this benchmark are augmented using the same strategy introduced in Sec.~\ref{sec:data} to test the model’s lighting‑harmonization capability. The second is a \emph{real-world benchmark} comprising 20 real-world foreground videos, including both footage captured in physical green-screen studios and foregrounds matted from videos filmed without a green screen. For each video, we use Gemini 3.1 Pro to generate a prompt describing a novel scene with distinct interactive objects and lighting conditions. This ground-truth-free benchmark comprehensively evaluates the practical utility of each method in real-world applications.

\noindent \textbf{Baselines.} Since existing video generation methods cannot directly fulfill the green-screen compositing task, we construct baselines by combining existing approaches. We compare our method against six baselines grouped into four categories: first‑frame editing (Flux 2 \cite{flux-2-2025}) followed by animation (WanAnimate \cite{cheng2025wan} or MimicMotion \cite{zhang2024mimicmotion}), video inpainting (VACE \cite{jiang2025vace}), cascaded video inpainting combined with relighting (VACE+ICLight), and native background-replacement methods (AnyPortal \cite{gao2025anyportal} and FlowPortal \cite{gao2026flowportal}). For the cascaded VACE+ICLight baseline, we first use VACE to generate a composite result, remove foregrounds with Attentive Eraser \cite{sun2025attentive} to obtain a clean background, and then apply IC‑Light\cite{zhang2025scaling} to relight the foreground to match the background. AnyPortal~\cite{gao2025anyportal} is a zero-shot framework that combines the temporal prior of video diffusion models with image-diffusion-based relighting for consistent video background replacement. FlowPortal~\cite{gao2026flowportal} further targets video relighting with background replacement using a training-free residual-corrected flow formulation, decoupled lighting conditions, and foreground-background masking.

\subsection{Quantitative Comparison}
\subsubsection{Metrics}
For the synthetic benchmark with ground-truth video available, we adopt three ground-truth-based evaluation metrics, including Identity Preservation, which measures the cosine similarity between the main face in the ground‑truth video and the detected faces in the generated video using InsightFace~\cite{insightface}; Foreground Similarity, which evaluates the structural similarity (SSIM) within the foreground mask to quantify the accuracy of character relighting; and Background Similarity, which removes the foreground region via gray filling and computes the semantic similarity between generated and ground‑truth backgrounds using a video encoder (i.e., VideoCLIP \cite{xu2021videoclip}), reflecting the alignment between the generated scene and the prompt.

For the real-world benchmark without ground-truth video, we adopt five ground-truth-free metrics, including Prompt-Video Consistency score, which measures how accurately the video matches the semantic meaning and intent (interactive motion, environment lighting, etc.) of its text prompt using ViCLIP \cite{wang2024internvid}, and four VBench metrics \cite{huang2024vbench}: Background Consistency, Subject Consistency, Temporal Flickering, and Imaging Quality. We also conduct a user study based on this benchmark to evaluate the comprehensive quality of generated videos.

\subsubsection{Quantitative Results on Synthetic Benchmark}
Table~\ref{tab:quant_main} reports the quantitative results on the synthetic benchmark. Our method achieves the best Identity Preservation (0.637), Foreground Similarity (0.642), and Background Similarity (0.704), demonstrating its ability to jointly preserve the character, harmonize foreground lighting, and synthesize the target environment. The first-frame editing baselines perform poorly on identity and foreground similarity because animation errors propagate the edited appearance across frames. VACE preserves identity but leaves foreground lighting unchanged, while VACE+ICLight improves Foreground Similarity from 0.539 to 0.629 at the cost of lower Identity Preservation due to its cascaded relighting process. Among the native background-replacement methods, FlowPortal consistently outperforms AnyPortal, yet both remain substantially behind our method, indicating that their zero-shot or training-free formulations struggle to preserve character details and model complex environment lighting. Overall, our unified modeling of bidirectional character--environment interaction provides the best performance across all three ground-truth-based metrics.

\subsubsection{Quantitative Results on Real-world Benchmark}
Table~\ref{tab:user_study} also reports the ground-truth-free results on the real-world benchmark. Our method achieves the best Prompt--Video Consistency (0.209), Subject Consistency (0.954), and Background Consistency (0.934), ties for the best Temporal Flickering score (0.958), and ranks second in Imaging Quality (0.679), demonstrating the best overall balance between prompt alignment, visual quality, and temporal coherence. AnyPortal attains the highest Imaging Quality but performs poorly in prompt alignment and temporal and background consistency, whereas FlowPortal achieves competitive prompt and subject consistency but the lowest Imaging Quality. Although VACE obtains strong consistency scores by largely preserving the input foreground, our method achieves better prompt alignment and spatial consistency while also harmonizing the character with the generated environment.
\begin{table}[t]
	\centering
	\small
	\setlength{\tabcolsep}{5pt}
	\caption{\textbf{Quantitative comparison on synthetic benchmark using ground-truth-based metrics.} \textbf{Bold} numbers denote the best results.}
	\scalebox{0.95}{\begin{tabular}{lccc}
		\hline
		Method & Identity $\uparrow$ & Foreground $\uparrow$ & Background $\uparrow$ \\
		\hline
		FLUX 2+WanAni. & 0.375 & 0.272 & 0.601 \\
		FLUX 2+MimicM. & 0.275 & 0.251 & 0.573 \\
		VACE & 0.606 & 0.539 & 0.688 \\
		VACE+ICLight & 0.551 & 0.629 & 0.688 \\
		AnyPortal & 0.358 & 0.274 & 0.577 \\
		FlowPortal & 0.391 & 0.295 & 0.616 \\ \hline
		Ours & \textbf{0.637} & \textbf{0.642} & \textbf{0.704} \\
		\hline
	\end{tabular}}
	\label{tab:quant_main}
\end{table}

\subsubsection{User Study on Real-world Benchmark}
To further evaluate our method on real-world green-screen footage, we conduct a user study with 62 participants who have prior knowledge of AIGC or computer graphics. For each example, participants are shown the input green-screen video, the target prompt, and anonymized outputs from all compared methods, whose display order is independently randomized. For each of five criteria---identity preservation, interaction realism, lighting realism, prompt alignment, and overall quality---participants select the best output among all methods, and we separately compute the selection rate of each method. As shown in Table~\ref{tab:user_study}, our method achieves the highest selection rate across all five criteria, with substantial margins over the baselines.

\begin{table*}[t]
	\centering
	\small
	\setlength{\tabcolsep}{2.5pt}
	\caption{\textbf{User study and ground-truth-free metric results on the real-world  benchmark.} Metrics marked with $^\dagger$ are user-study preference percentages (\%). The remaining columns report the ground-truth-free metrics. Top-2 results are highlighted in \textbf{bold} and \underline{underlined}.}
	\scalebox{0.85}{\begin{tabular}{lccccc|ccccc}
		\hline
		Method & \makecell{Identity$^\dagger$\\Preservation} & \makecell{Interaction$^\dagger$\\Realism} & \makecell{Lighting$^\dagger$\\Realism} & \makecell{Prompt$^\dagger$\\Alignment} & \makecell{Overall$^\dagger$\\Quality} & \makecell{Prompt--Video\\Consistency} $\uparrow$ & \makecell{Imaging\\Quality} $\uparrow$ & \makecell{Temporal\\Flickering} $\uparrow$ & \makecell{Subject\\Consistency} $\uparrow$ & \makecell{Background\\Consistency} $\uparrow$ \\
		\hline
		FLUX 2+WanAni.  & 3.23\% & 4.84\% & 4.84\% & 3.23\% & 3.23\% & 0.197 & 0.669 & 0.954 & 0.939 & 0.928 \\
		FLUX 2+MimicM. & 0.00\% & 1.61\% & 3.23\% & 0.00\% & 0.00\% & 0.186 & 0.677 & 0.952 & 0.940 & 0.927 \\
		VACE           & \underline{27.42\%} & \underline{24.19\%} & 12.90\% & 12.90\% & \underline{20.97\%} & \underline{0.204} & 0.673 & \textbf{0.958} & \underline{0.952} & \underline{0.932} \\
		VACE+ICLight   & 0.00\% & 4.84\% & 8.06\% & 1.61\% & 1.61\% & 0.201 & 0.664 & 0.955 & 0.941 & 0.927 \\
		AnyPortal      & 1.61\% & 3.23\% & 8.06\% & 6.45\% & 4.84\% & 0.195 & \textbf{0.682} & 0.945 & 0.948 & 0.923 \\
		FlowPortal     & 6.45\% & 4.84\% & \underline{14.52\%} & \underline{22.58\%} & 9.68\% & \underline{0.204} & 0.610 & 0.955 & \underline{0.952} & 0.926 \\ \hline
		Ours           & \textbf{61.29\%} & \textbf{56.45\%} & \textbf{48.39\%} & \textbf{53.23\%} & \textbf{59.68\%} & \textbf{0.209} & \underline{0.679} & \textbf{0.958} & \textbf{0.954} & \textbf{0.934} \\
		\hline
	\end{tabular}}
    
	\label{tab:user_study}
\end{table*}

\begin{figure}[!t]
    \centering
    \setlength{\tabcolsep}{0.3pt}
    \renewcommand{\arraystretch}{0.85}
    \scalebox{0.95}{
    \begin{tabular}{ccc}
        \multicolumn{3}{l}{\scriptsize Green Screen Video} \\
        \includegraphics[width=0.33\linewidth]{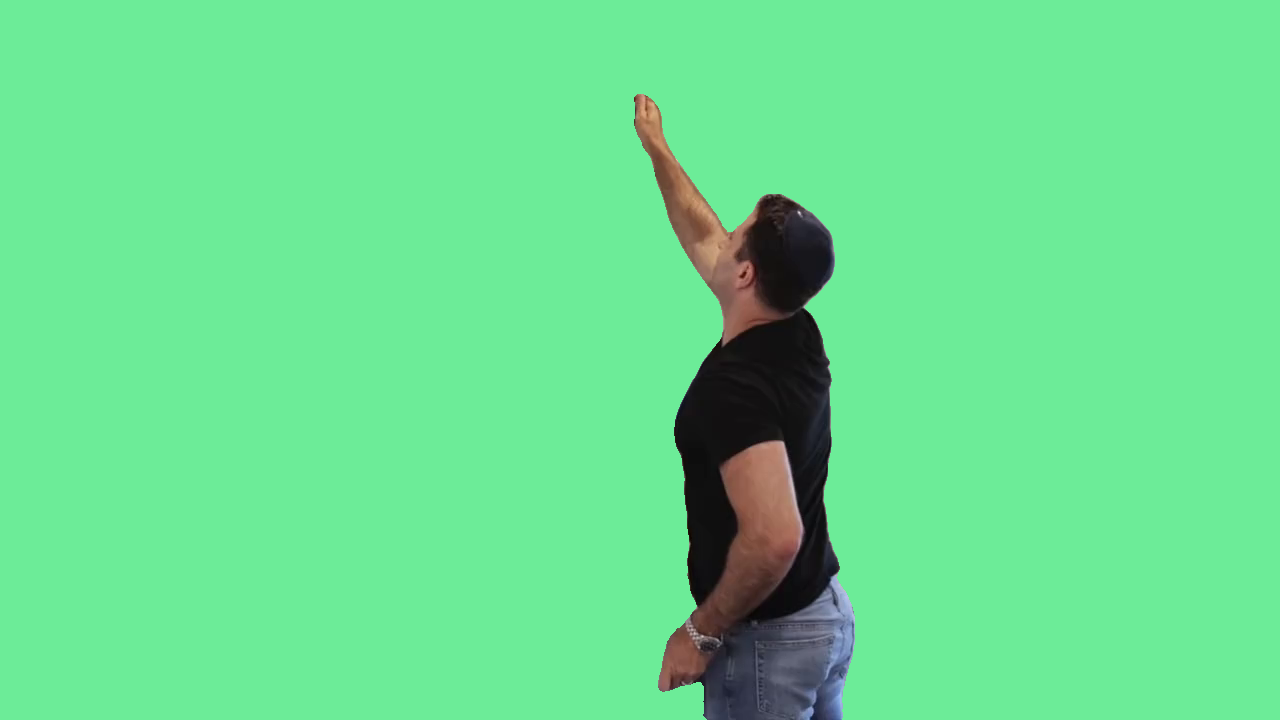} &
        \includegraphics[width=0.33\linewidth]{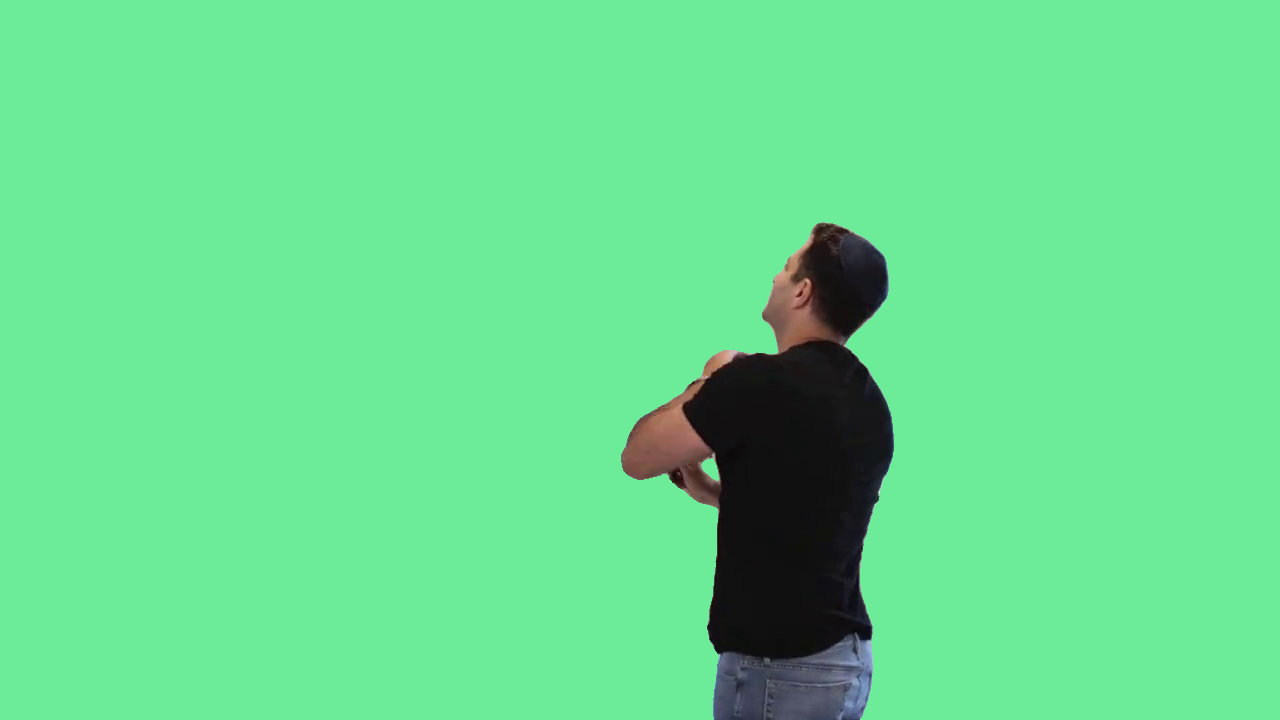} &
        \includegraphics[width=0.33\linewidth]{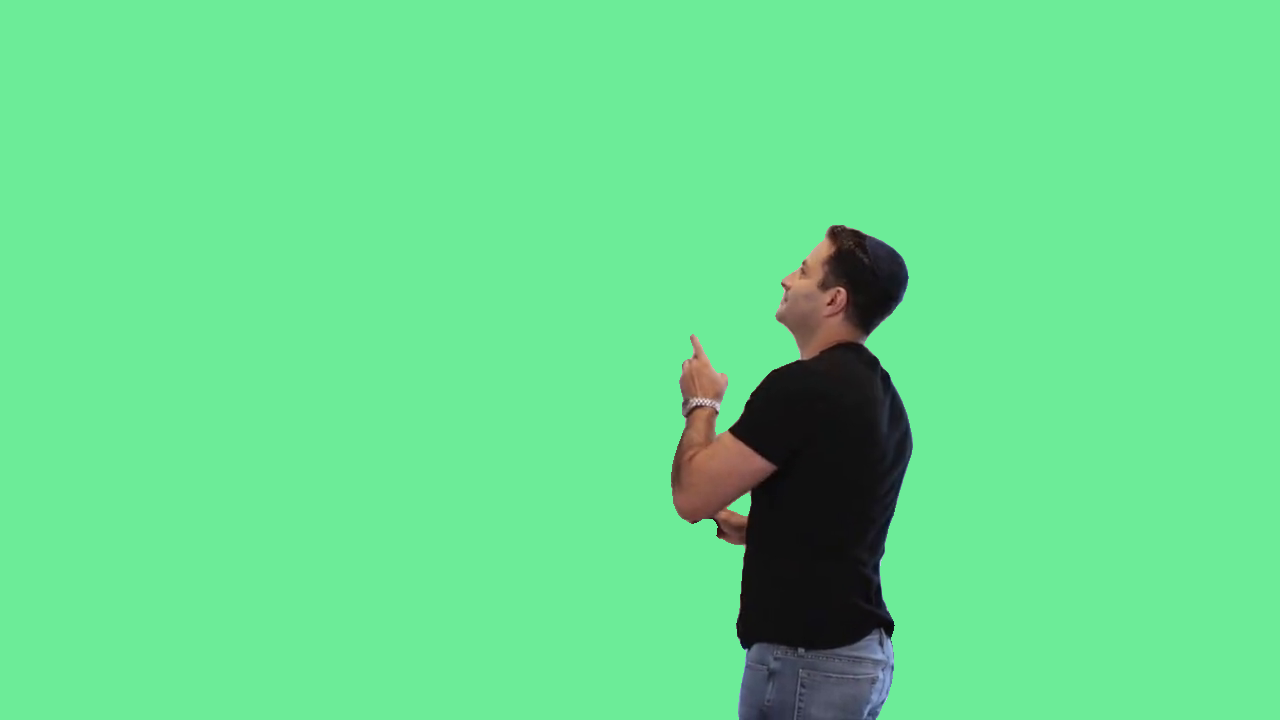} \\
        \multicolumn{3}{l}{\scriptsize Foreground Depth} \\
        \includegraphics[width=0.33\linewidth]{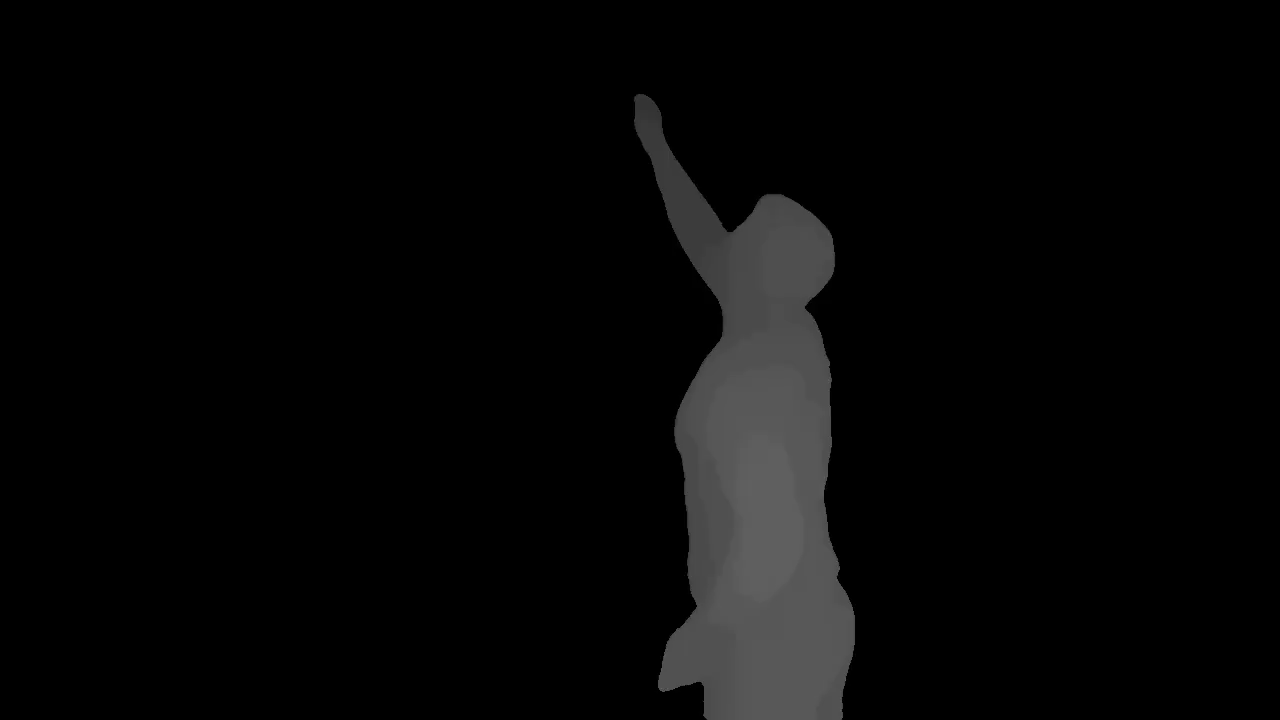} &
        \includegraphics[width=0.33\linewidth]{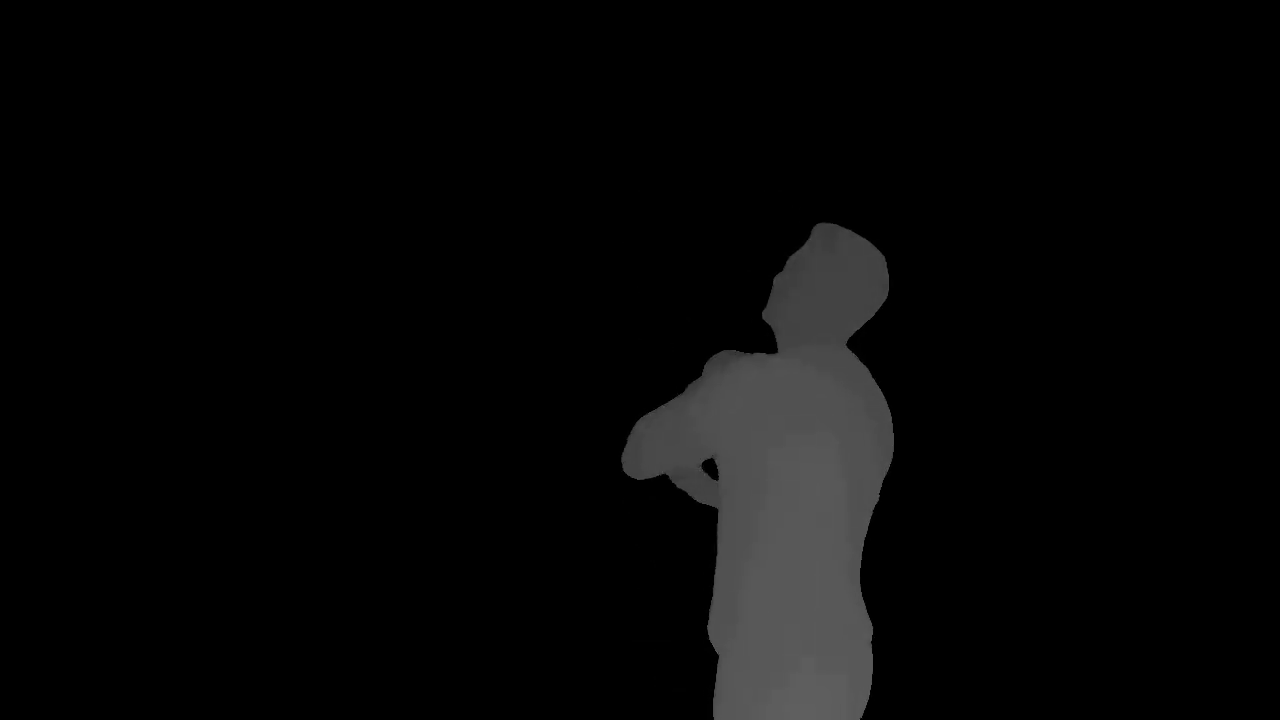} &
        \includegraphics[width=0.33\linewidth]{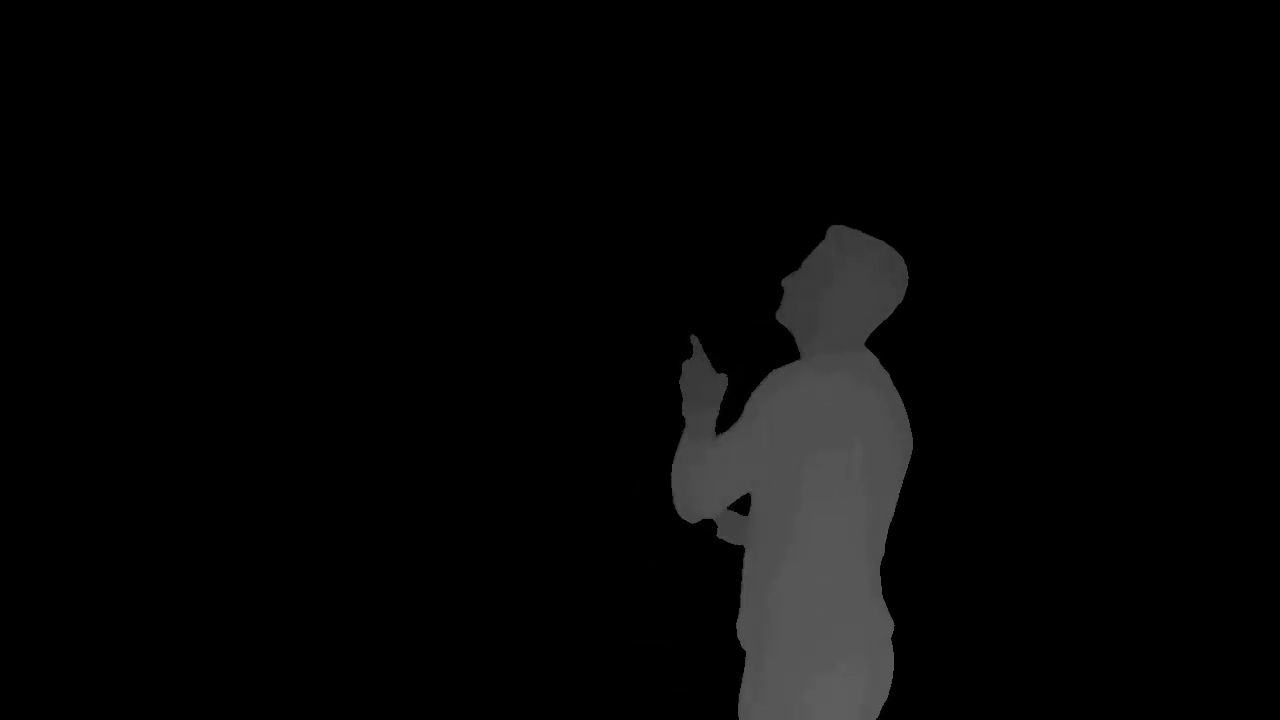} \\ \hline
        \multicolumn{3}{l}{\scriptsize Input: Green Screen; RGB-only Denoising} \\
        \includegraphics[width=0.33\linewidth]{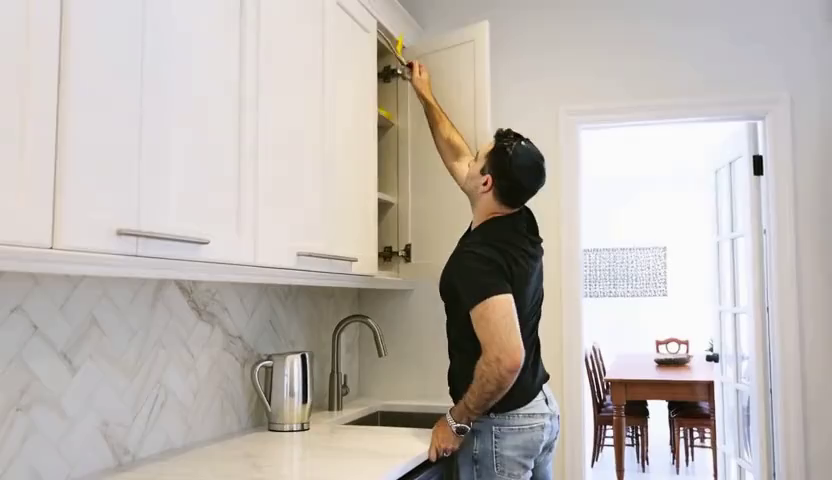} &
        \includegraphics[width=0.33\linewidth]{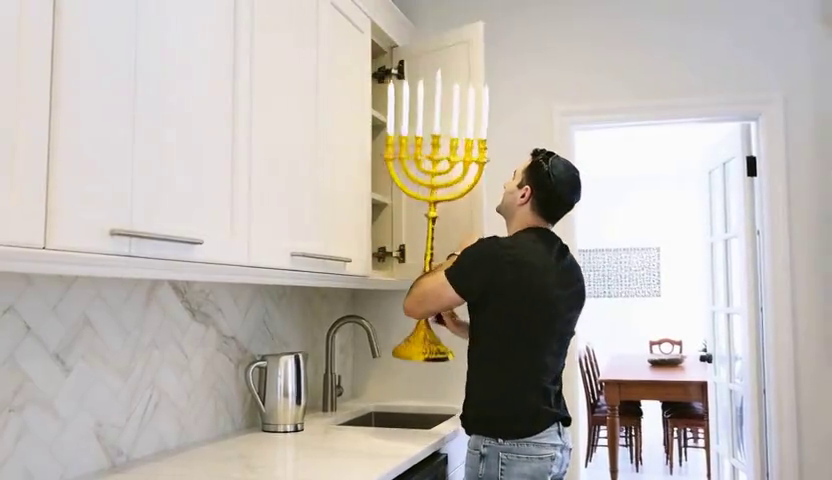} &
        \ZoomInBox{rectangle,red,magnification=2,size=0.83cm}{1.6,1.03}{0.55,1.05}{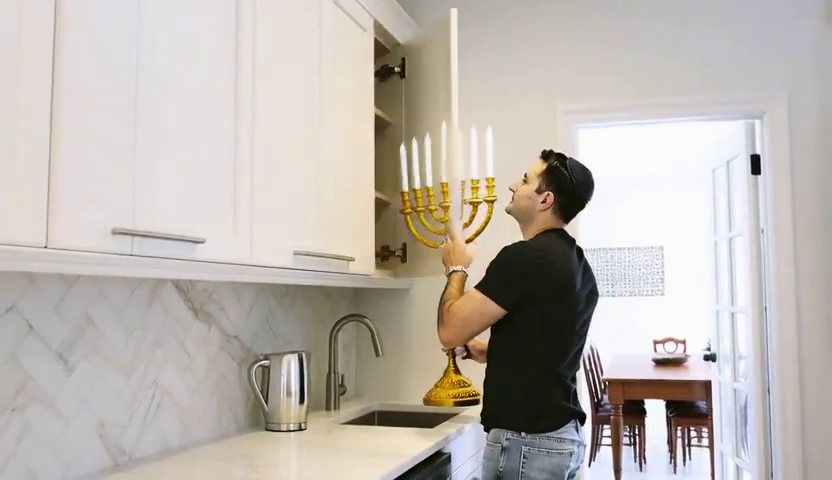}{0.33\linewidth} \\
        \multicolumn{3}{l}{\scriptsize Input: Green Screen + Foreground Depth; RGB-only Denoising} \\
        \ZoomInBox{rectangle,red,magnification=2,size=0.86cm}{1.25,1.3}{0.55,0.5}{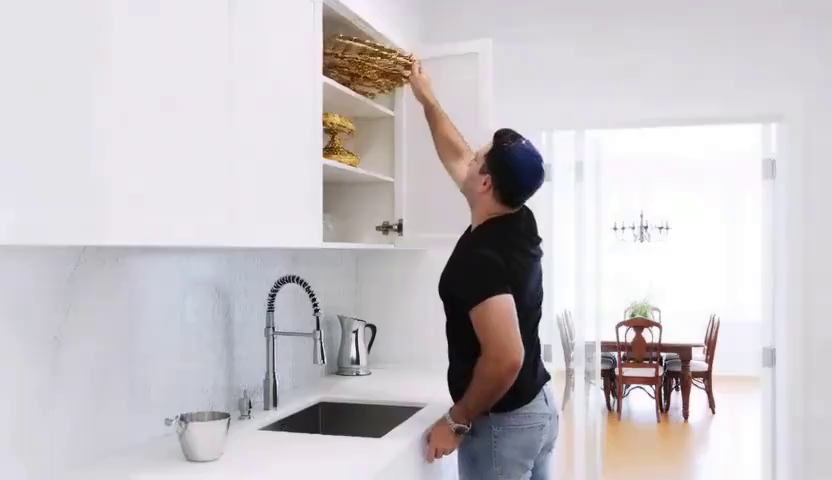}{0.33\linewidth} &
        \includegraphics[width=0.33\linewidth]{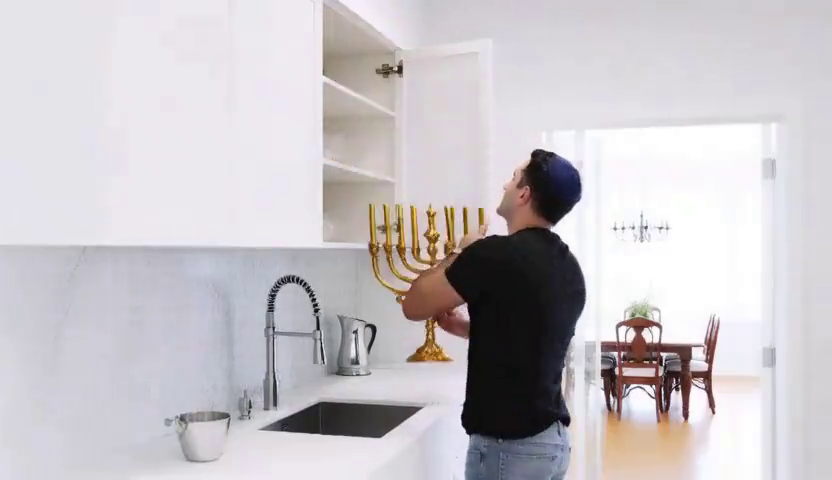} &
        \includegraphics[width=0.33\linewidth]{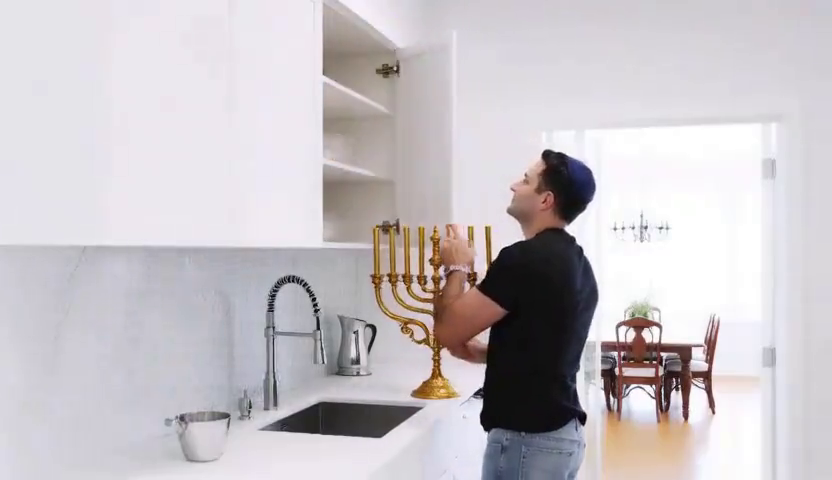} \\
        \multicolumn{3}{l}{\makecell[l]{\scriptsize Input: Green Screen + Foreground Depth; RGB-D Joint Denoising}} \\
        \includegraphics[width=0.33\linewidth]{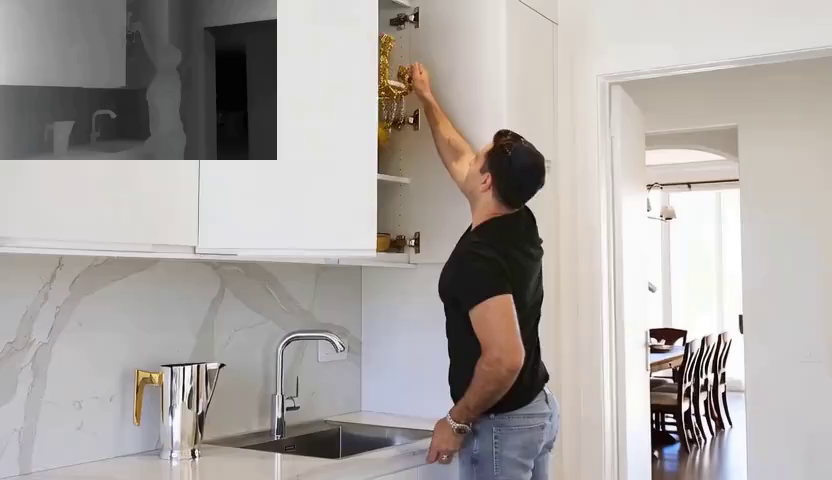} &
        \includegraphics[width=0.33\linewidth]{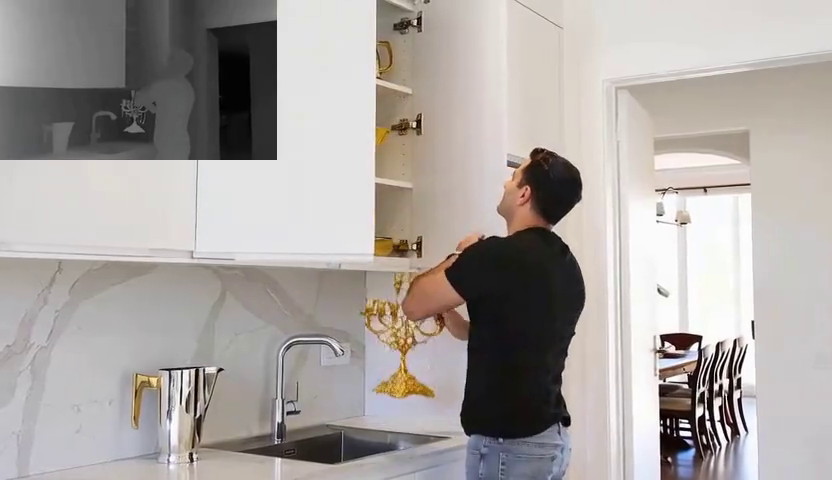} &
        \includegraphics[width=0.33\linewidth]{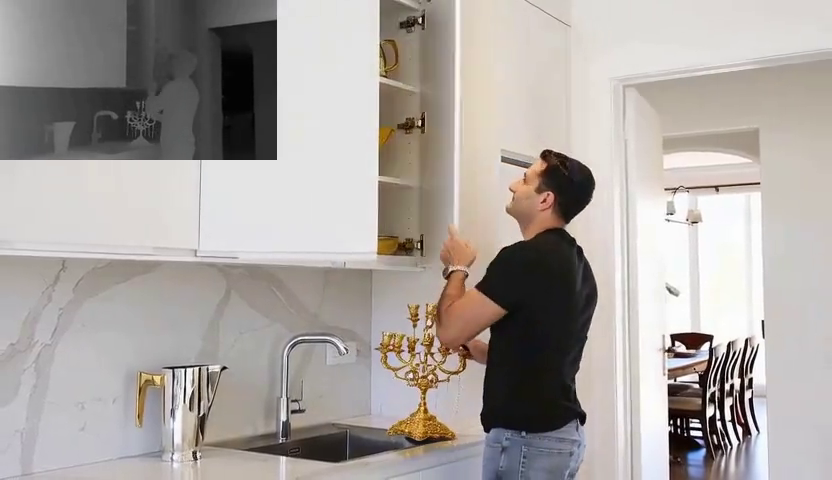} \\

        \multicolumn{3}{p{0.95\linewidth}}{
        \small \textit{A man in a modern white kitchen retrieves an ornate gold menorah from an upper cabinet, closes the door, and stands near a sink and marble-patterned backsplash.}} \\
    \end{tabular}
    }
    \caption{\textbf{Qualitative ablation on input conditioning and denoising formulation.} For the joint RGB-D denoising, we also visualize depth denoise result on the top left corner for each frames.}
    \label{fig:ablation_rgbd}
\end{figure}

\begin{figure}[h]
    \centering
    \setlength{\tabcolsep}{0.3pt}
    \renewcommand{\arraystretch}{0.85}
    \scalebox{1}{
    \begin{tabular}{ccc}
        \multicolumn{3}{l}{\scriptsize Green Screen Video} \\
        \includegraphics[width=0.315\linewidth]{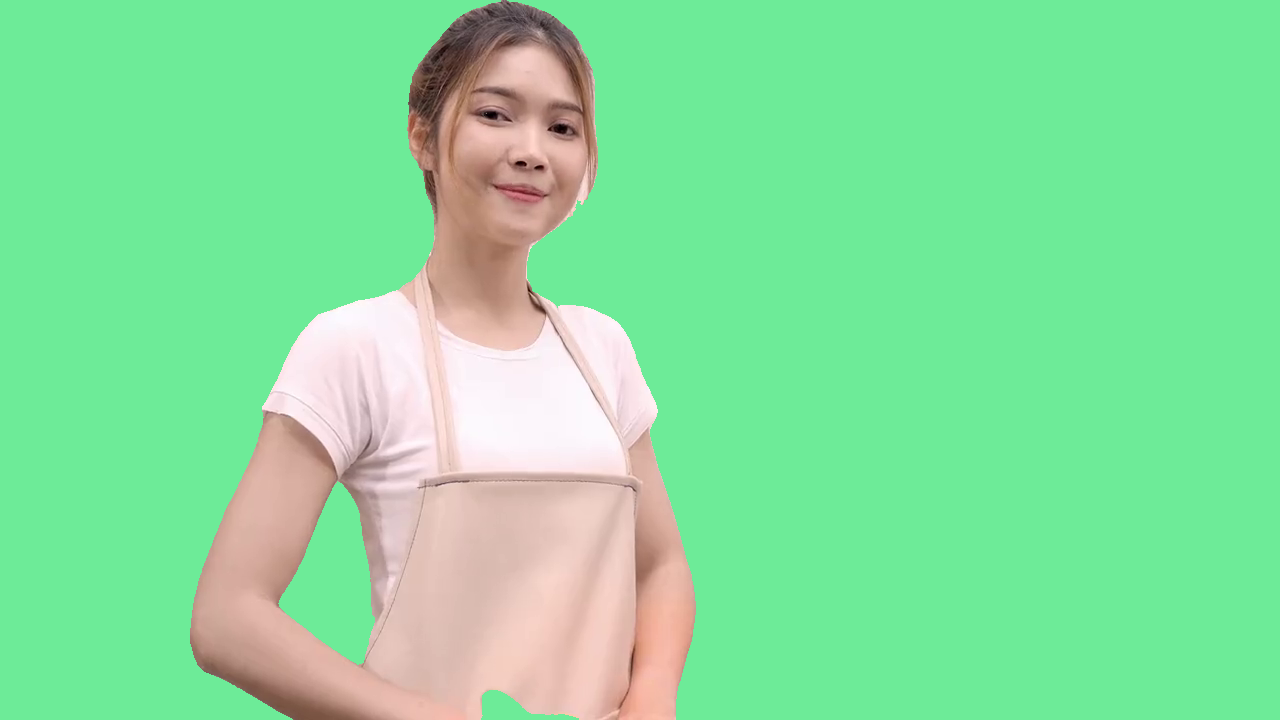} &
        \includegraphics[width=0.315\linewidth]{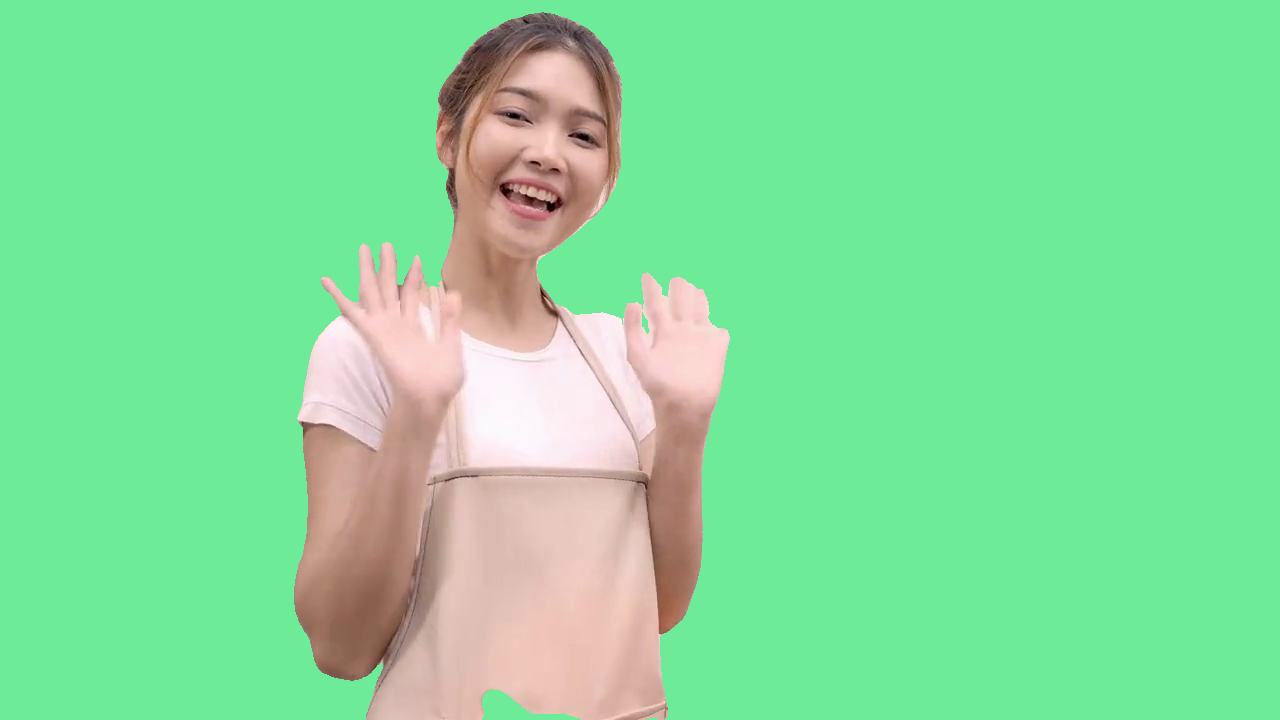} &
        \includegraphics[width=0.315\linewidth]{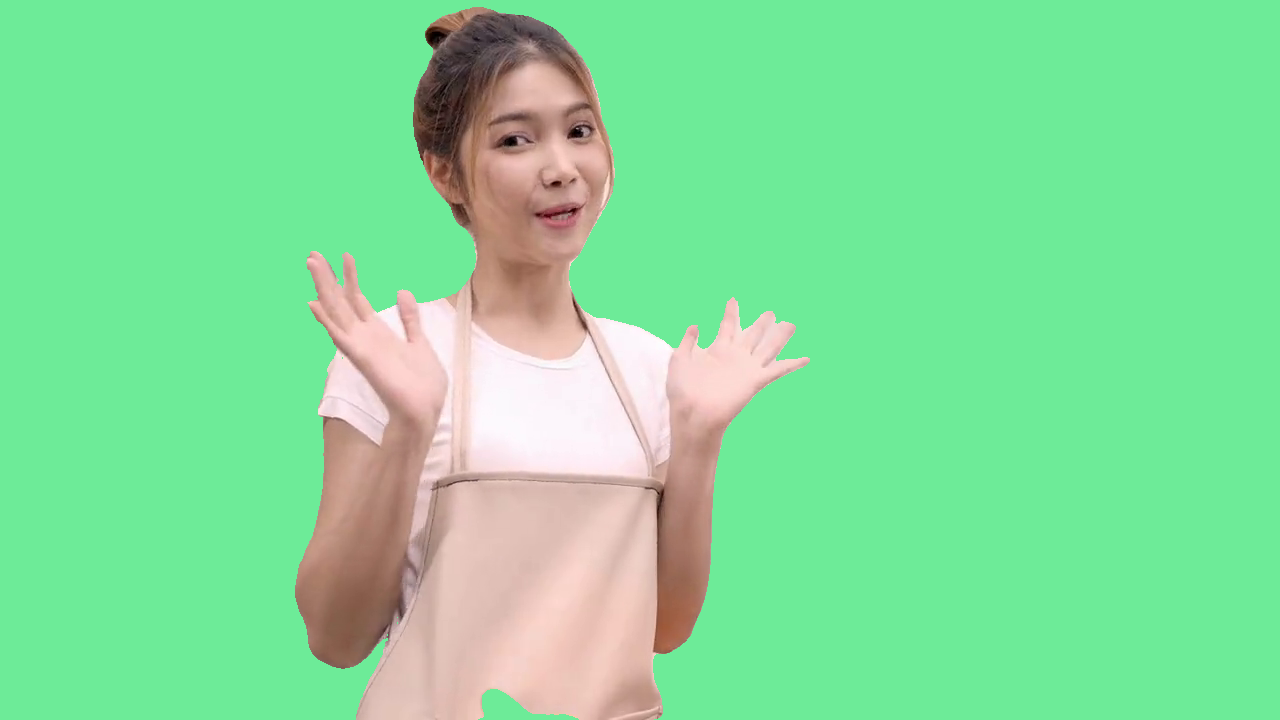} \\ \hline

        \multicolumn{3}{l}{\scriptsize w/o Multi-illumination Data} \\
        \includegraphics[width=0.315\linewidth]{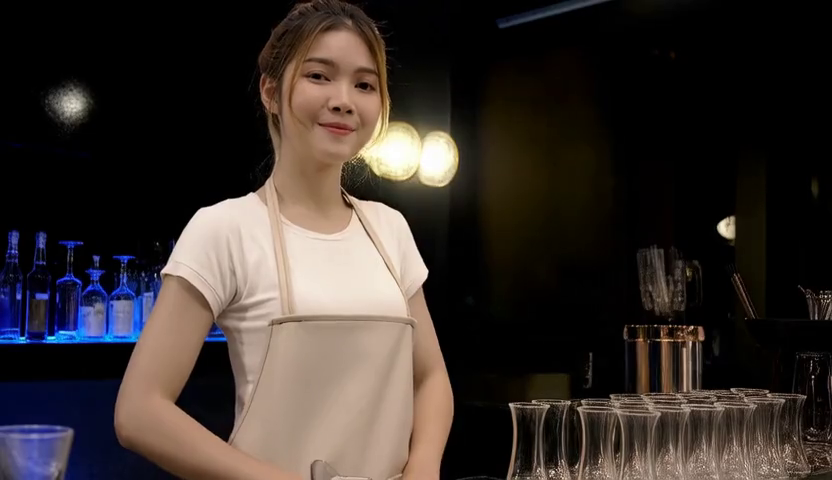} &
        \includegraphics[width=0.315\linewidth]{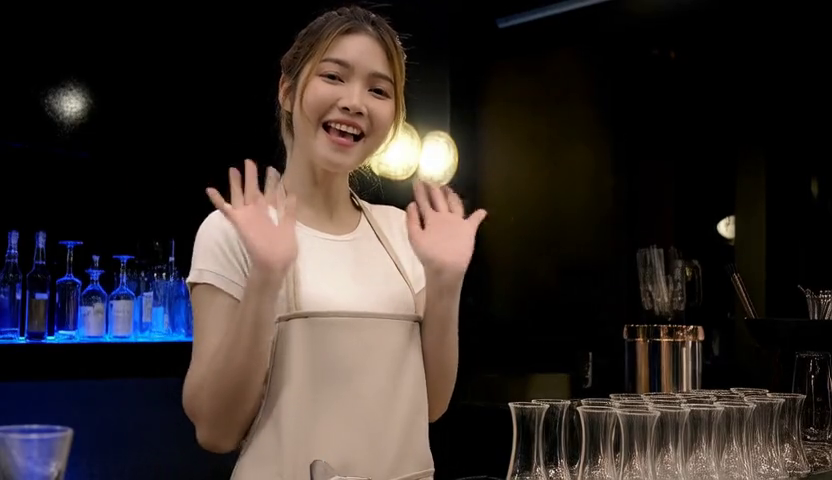} &
        \includegraphics[width=0.315\linewidth]{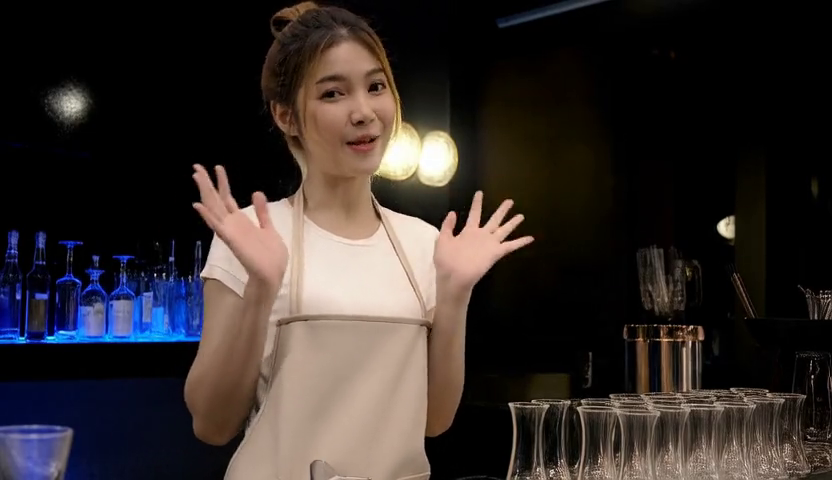} \\

        \multicolumn{3}{l}{\scriptsize w/ Multi-illumination Data} \\
        \includegraphics[width=0.315\linewidth]{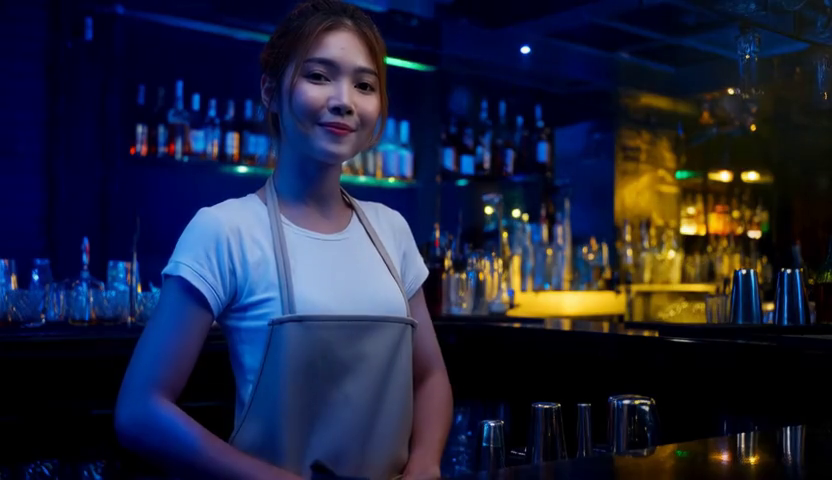} &
        \includegraphics[width=0.315\linewidth]{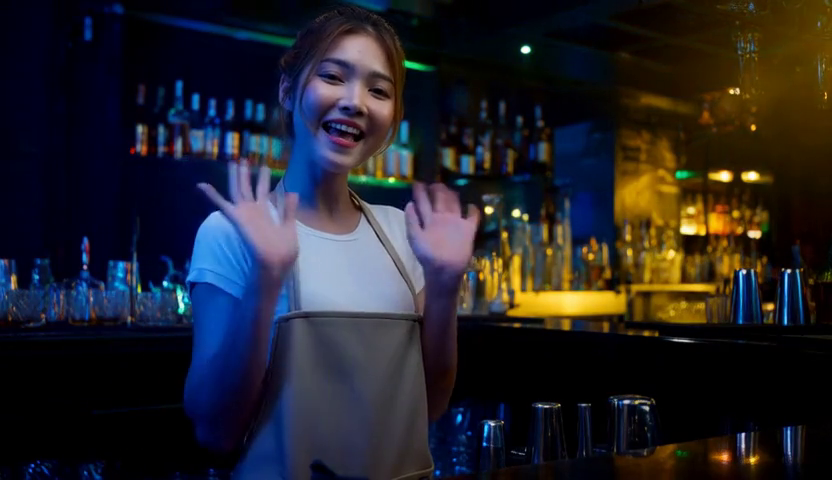} &
        \includegraphics[width=0.315\linewidth]{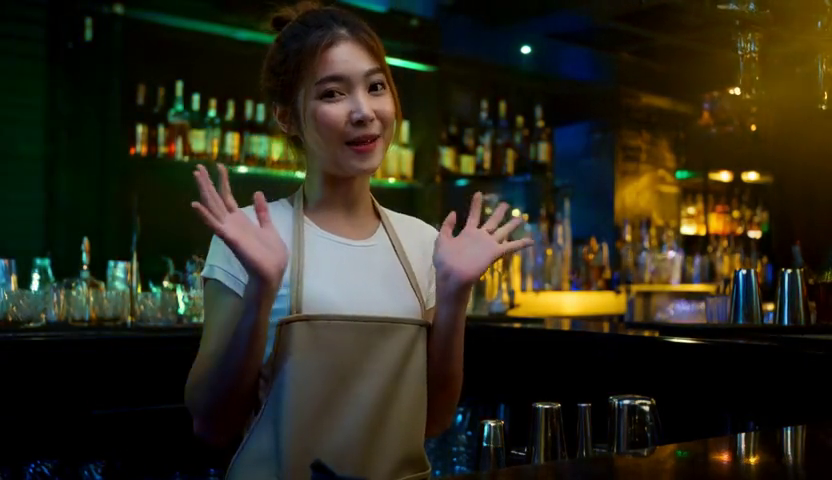} \\

        \multicolumn{3}{p{0.95\linewidth}}{
        \small \textit{A cheerful woman stands behind a bar counter and waves at the camera under dramatic blue and golden lighting.}} \\
    \end{tabular}
    }
    \caption{\textbf{Qualitative ablation on the training data composition.} Multi-illumination data improves both the background aesthetic quality and the harmony between foreground relighting and background appearance.}
    \label{fig:ablation_data}
\end{figure}

\begin{table*}[t]
	\centering
	\small
	
	\caption{\textbf{Ablation on RGBD conditioning and joint denoising.} All other training and inference settings are kept fixed. Our joint RGB-D denoising improves foreground and background quality while maintaining identity and prompt--video consistency.}
	\scalebox{0.85}{\begin{tabular}{lcccc}
		\hline
		\makecell[l]{Input \&  Denoising Objective} & Identity $\uparrow$ & Foreground $\uparrow$ & Background $\uparrow$ & \makecell{Prompt--Video Consistency $\uparrow$} \\
		\hline
		\makecell[l]{RGB \&  RGB-Only Denoising} & \textbf{0.638} & 0.636 & 0.694 & 0.196 \\
		\makecell[l]{RGB + Foreground Depth \&  RGB-Only Denoising} & 0.635 & 0.638  & 0.692  & 0.201 \\
		\makecell[l]{RGB + Foreground Depth \&  RGB-D Joint Denoising} & 0.637 & \textbf{0.642} & \textbf{0.704} & \textbf{0.209} \\
		\hline
	\end{tabular}}
    
	\label{tab:ablation_rgbd}
\end{table*}

\begin{table}[t]
	\centering
	\small
	\caption{\textbf{Ablation on the multi-illumination data.} Incorporating multi-illumination data yields consistent improvements across all evaluation metrics.}
	\setlength{\tabcolsep}{2pt}
	\scalebox{0.75}{\begin{tabular}{lcccc}
		\hline
		Training data & Identity $\uparrow$ & Foreground $\uparrow$ & Background $\uparrow$ & P--V Cons. $\uparrow$ \\
		\hline
		Base Data & 0.635 & 0.635 & 0.699 & 0.202 \\
		+ Multi-illumination Data & \textbf{0.637} & \textbf{0.642} & \textbf{0.704} & \textbf{0.209} \\
		\hline
	\end{tabular}}
    
	\label{tab:ablation_data}
\end{table}

\subsection{Qualitative Comparison}
Figure~\ref{fig:qual_main} presents qualitative comparisons on representative real‑world green‑screen examples. FLUX 2 + WanAnimate and FLUX 2 + MimicMotion drive first‑frame editing results with character motion, yet they struggle to preserve temporal consistency and are highly sensitive to motion estimation errors. This often leads to identity degradation, character motion distortion, or blurry backgrounds. VACE can inpaint the background conditioned on foreground information but suffers from two key limitations for the green‑screen compositing task. First, it leaves the foreground unchanged, resulting in inconsistent lighting between foreground and background. Second, VACE only denoises RGB inputs and lacks depth awareness, leading to unrealistic interactions between generated characters and environments. These issues make the character appear suspended rather than naturally integrated with the environment. Although the cascaded VACE+ICLight baseline partially mitigates illumination mismatch, its two‑stage pipeline fails to achieve bidirectional character‑environment harmonization and additionally introduces identity drift and compositing artifacts. \revision{The native background-replacement methods exhibit complementary limitations. AnyPortal produces substantial shifts in human identity, indicating insufficient preservation of foreground appearance. FlowPortal can synthesize visually appealing lighting effects, but its generated interaction motions remain unnatural. Moreover, neither method can generate dynamic illumination that evolves over time with the actor and the synthesized environment.} In contrast, our unified generative framework simultaneously realizes C2E physical interaction and E2C lighting harmonization, and leverages RGB‑D joint denoising to enhance the spatial and physical realism of interactions, thus producing the most natural compositing results.

\subsection{Ablation Study}
\label{sec:ablation}
We ablate two key designs in our framework: joint RGB-D denoising with foreground-depth conditioning and the multi‑illumination dataset used in addition to the base dataset. In both ablation studies, we keep all other settings unchanged and modify only the ablated setting. For each ablation, we report quantitative results on the synthetic benchmark using the GT-based metrics adopted in the main comparison and the prompt-video consistency score. We also provide qualitative comparison on representative example.

\subsubsection{RGBD Conditioning and Joint Denoising}

We investigate the respective roles of foreground-depth conditioning and the RGB-D joint denoising objective through three settings: (i) green-screen RGB input with RGB-only denoising, (ii) green-screen RGB plus foreground-depth input with RGB-only denoising, and (iii) green-screen RGB plus foreground-depth input with RGB-D joint denoising. \revision{As reported in Table~\ref{tab:ablation_rgbd}, merely adding foreground depth as an input condition yields only marginal changes. It slightly improves Foreground Similarity (0.636 to 0.638) and Prompt--Video Consistency (0.196 to 0.201). Thus, providing depth cues alone is insufficient for the model to effectively exploit interaction geometry.}

\revision{In contrast, introducing explicit depth supervision through joint RGB-D denoising achieves the best Foreground Similarity (0.642), Background Similarity (0.704), and Prompt--Video Consistency (0.209), while maintaining comparable Identity Preservation (0.637). This comparison shows that the improvements arise not simply from adding depth to the input, but from incorporating geometry into the generative objective.} Figure~\ref{fig:ablation_rgbd} further illustrates this effect. \revision{With RGB-only denoising, foreground depth provides spatial cues but does not directly constrain the generated geometry, leading to implausible foreground--environment interactions. Jointly supervising RGB and depth instead promotes coherent object contacts and background structures throughout denoising.}

\subsubsection{Base Data Only vs. Base + Multi-illumination Data}
We next investigate the contribution of the multi‑illumination dataset. Specifically, we compare a model trained only on the base dataset against our full training strategy that combines both datasets. As shown in Table~\ref{tab:ablation_data}, training with the multi-illumination dataset achieves better performance across all metrics. Qualitative examples in Figure~\ref{fig:ablation_data} further confirm this trend. Without multi‑illumination data, the model fails to generate visually rich illumination effects, whereas with this dataset, character lighting naturally varies dynamically with the surrounding environment.

\subsection{Reference-Guided Generation}
In Figure~\ref{fig:reference_result_new}, we show results with reference guidance, where user-provided reference images specify the target objects to be generated. For each example, we also show the pre-generation canvas, which visualizes both the provided reference images and the approximate object location in the first frame. These examples demonstrate that our model can incorporate visual references while preserving coherent object interactions and scene-level lighting.
\providecommand{\refresultwidth}{0.25\linewidth}
\begin{figure*}[t]
	\centering
	\begin{minipage}[t]{0.49\textwidth}
		\centering
		\setlength{\tabcolsep}{0.01em}
		\renewcommand{\arraystretch}{0.8}
		\begin{tabular}{c|ccc}
			\scriptsize Origin video frame & & \scriptsize Green Screen Video & \\
 			\includegraphics[width=\refresultwidth]{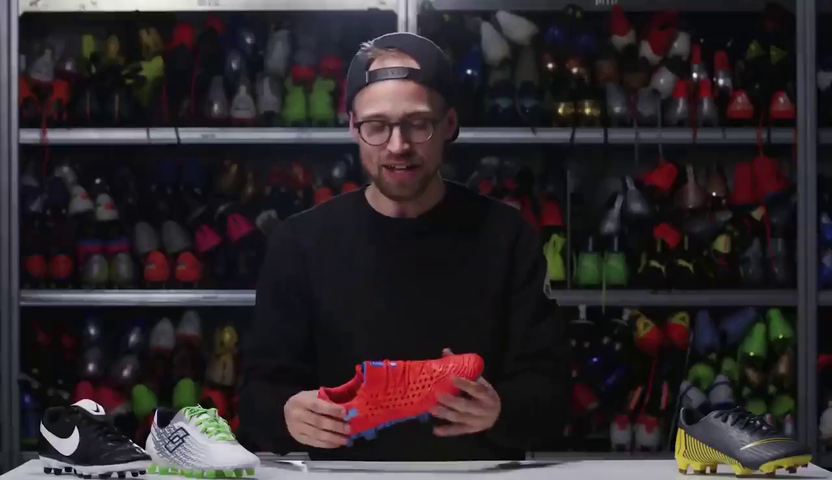} &
			\includegraphics[width=\refresultwidth]{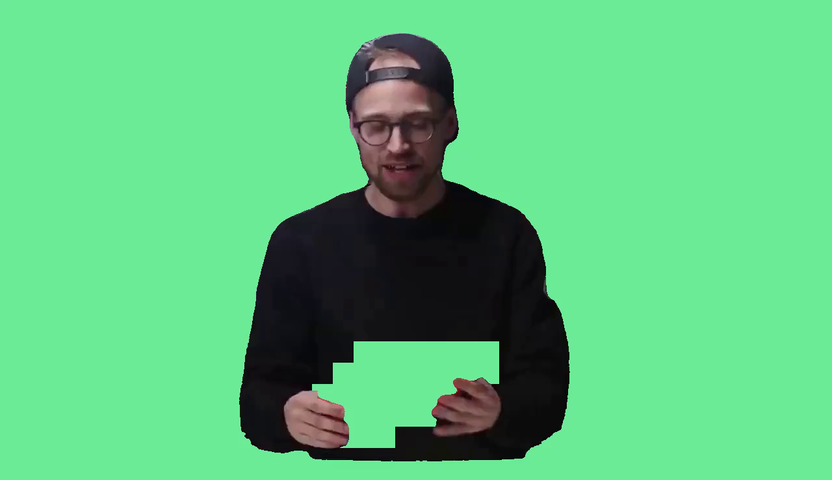} &
			\includegraphics[width=\refresultwidth]{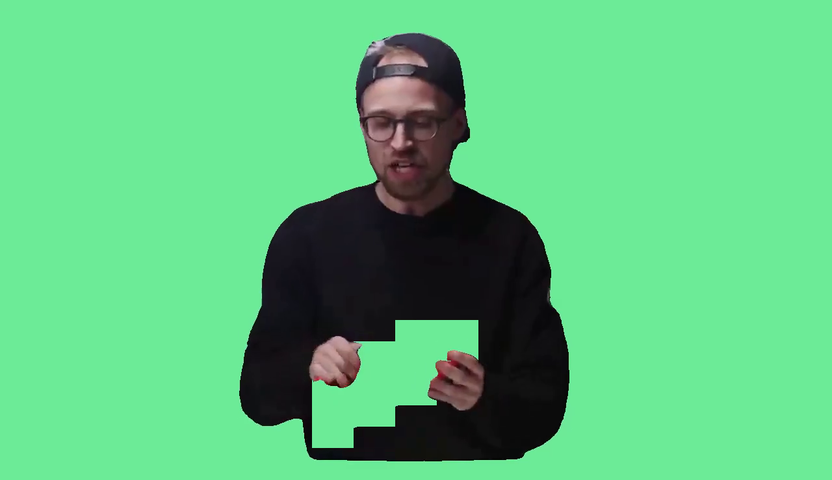} &
			\includegraphics[width=\refresultwidth]{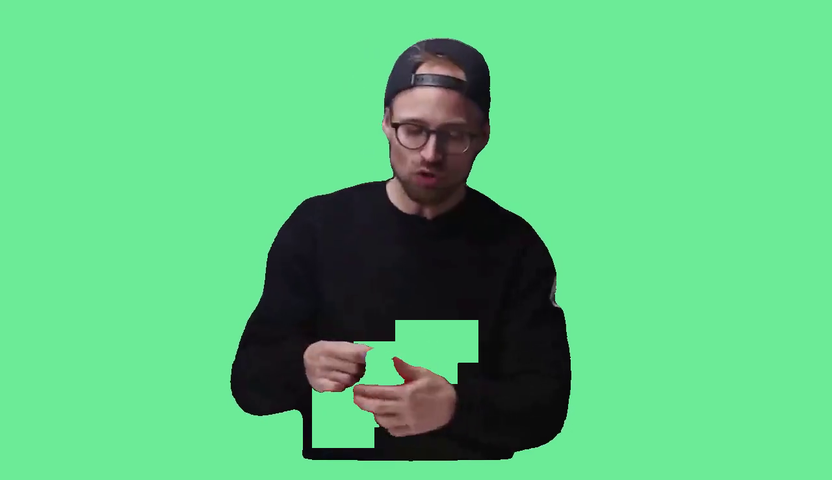} \\
			
			\scriptsize Given Canvas & & \scriptsize Result \\
			\includegraphics[width=\refresultwidth]{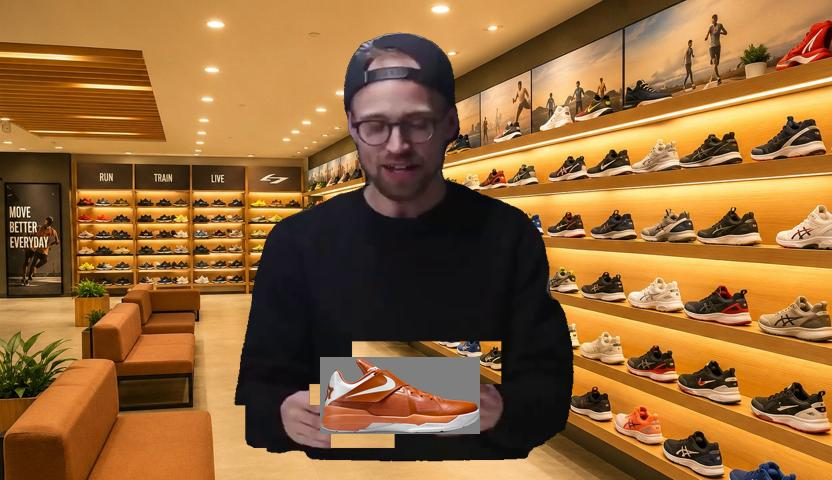} &
			\includegraphics[width=\refresultwidth]{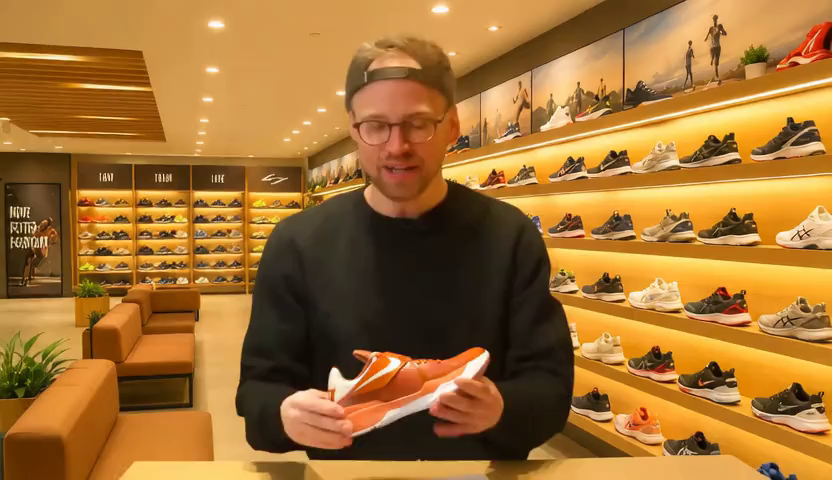} &
			\includegraphics[width=\refresultwidth]{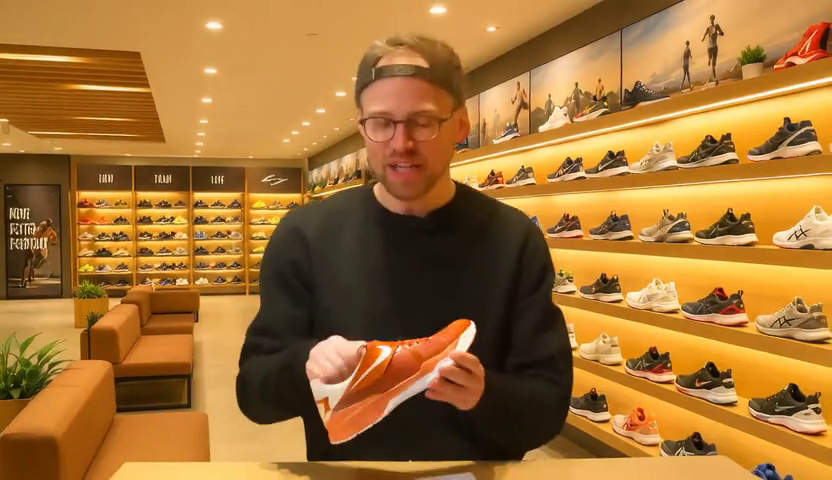} &
			\includegraphics[width=\refresultwidth]{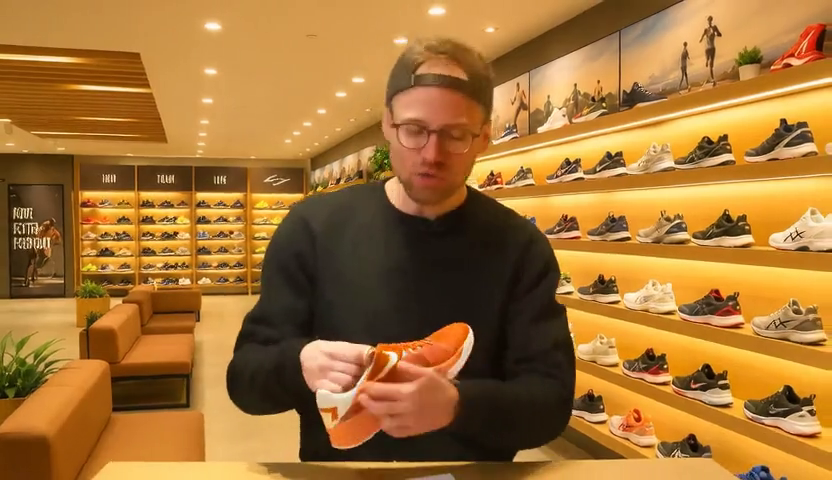} \\
			\multicolumn{4}{p{0.95\linewidth}}{\small Prompt: \textit{A man in glasses and a black shirt reviews orange shoes inside a warm modern shoe store. Wooden sneaker displays and golden retail lighting illuminate his face and the product.}} \\
		\end{tabular}
	\end{minipage}\hfill
	\begin{minipage}[t]{0.49\textwidth}
		\centering
		\setlength{\tabcolsep}{0.01em}
		\renewcommand{\arraystretch}{0.8}
		\begin{tabular}{c|ccc}
			\scriptsize Origin video frame & & \scriptsize Green Screen Video & \\
 			\includegraphics[width=\refresultwidth]{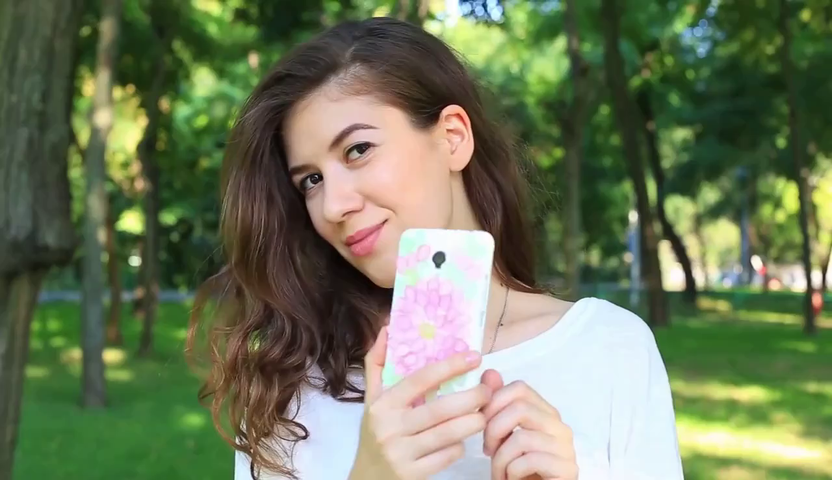} &
			\includegraphics[width=\refresultwidth]{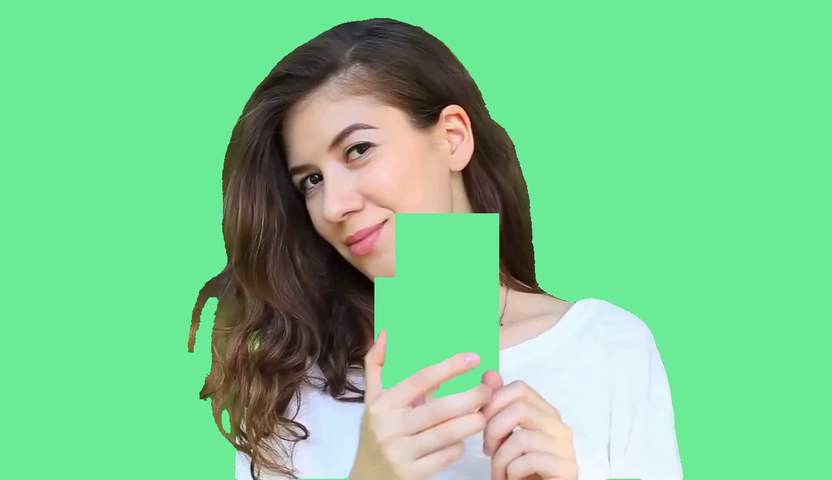} &
			\includegraphics[width=\refresultwidth]{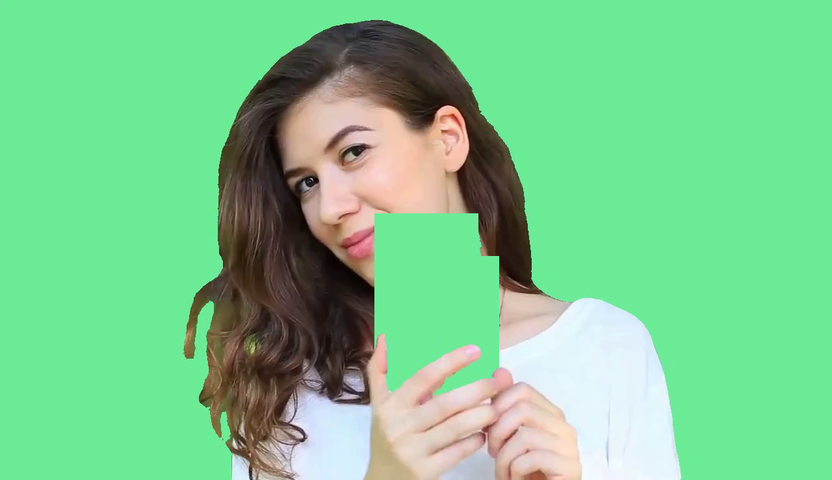} &
			\includegraphics[width=\refresultwidth]{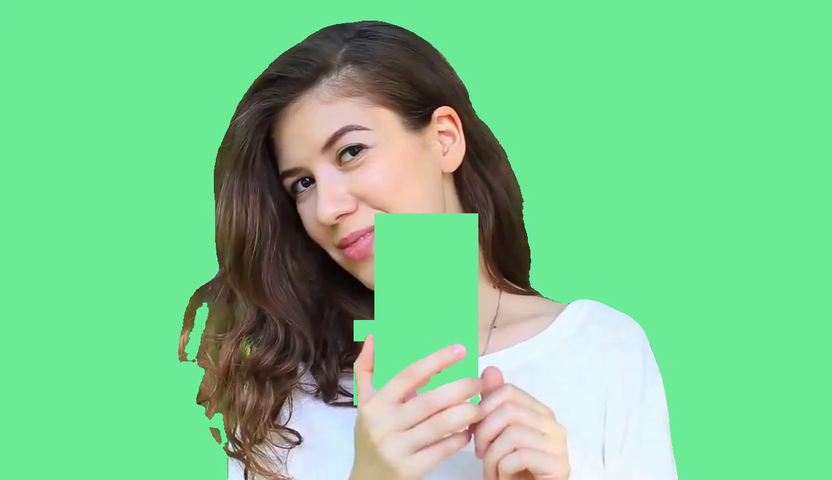} \\
			
			\scriptsize Given Canvas & & \scriptsize Result \\
			\includegraphics[width=\refresultwidth]{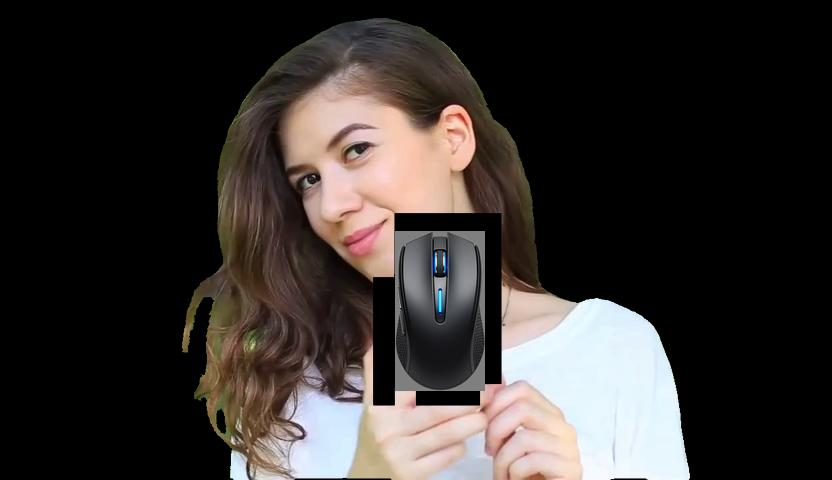} &
			\includegraphics[width=\refresultwidth]{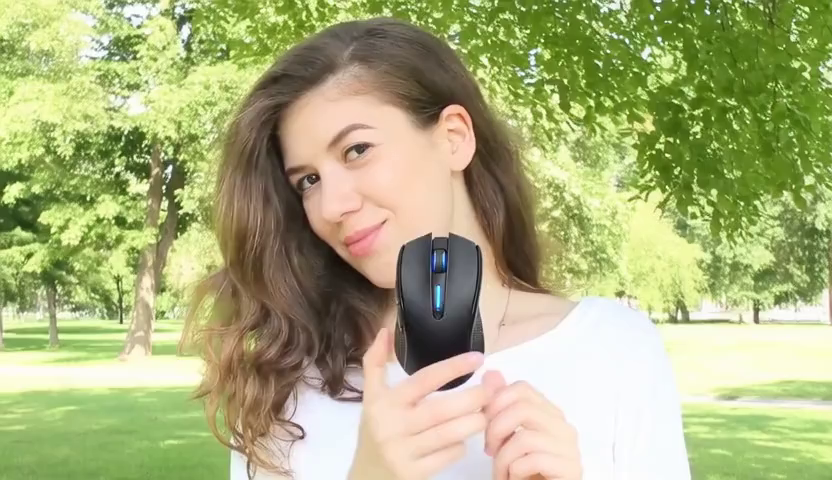} &
			\includegraphics[width=\refresultwidth]{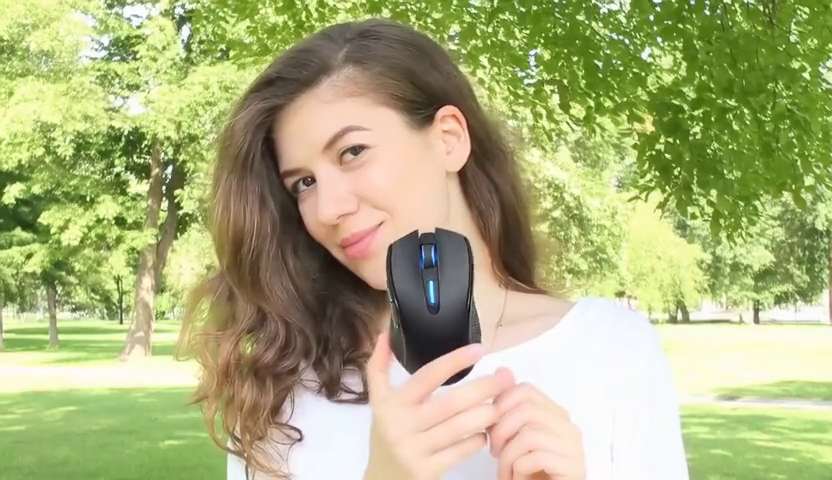} &
			\includegraphics[width=\refresultwidth]{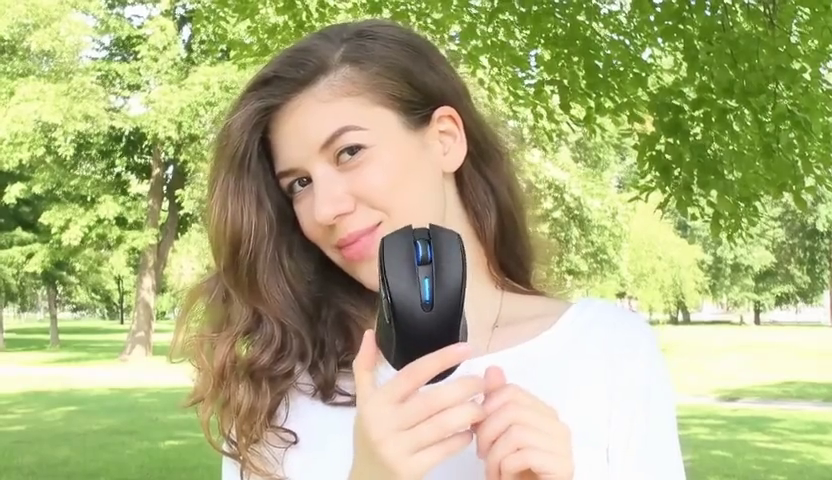} \\
			\multicolumn{4}{p{0.95\linewidth}}{\small Prompt: \textit{A young woman with long wavy hair stands in a sunny park, holding a black gaming mouse. She smiles toward the camera against a bright green outdoor background.}} \\
		\end{tabular}
	\end{minipage}

	\vspace{0.4em}

	\begin{minipage}[t]{0.49\textwidth}
		\centering
		\setlength{\tabcolsep}{0.01em}
		\renewcommand{\arraystretch}{0.8}
		\begin{tabular}{c|ccc}
			\scriptsize Origin video frame & & \scriptsize Green Screen Video & \\
 			\includegraphics[width=\refresultwidth]{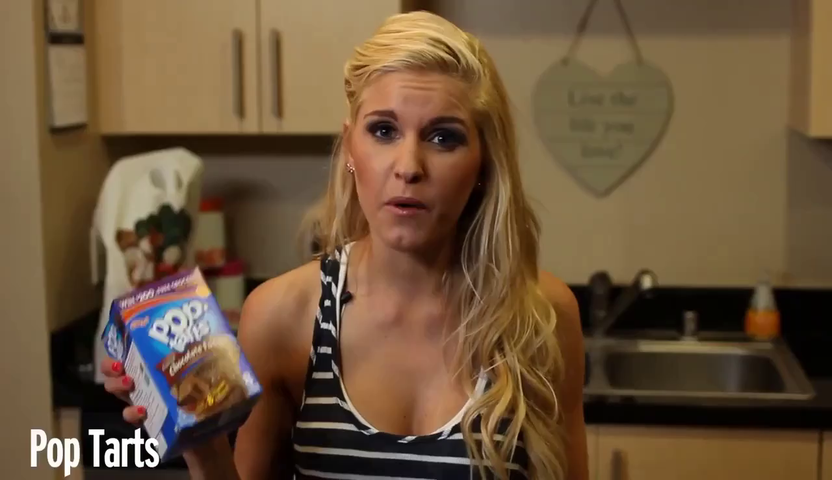} &
			\includegraphics[width=\refresultwidth]{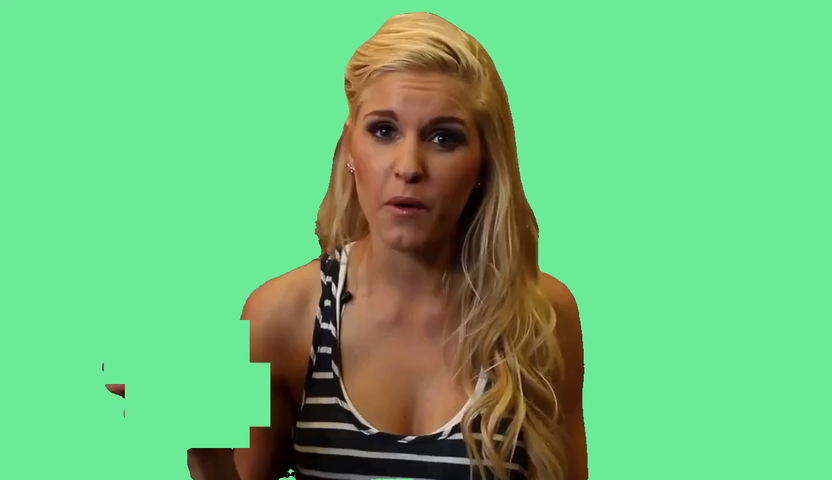} &
			\includegraphics[width=\refresultwidth]{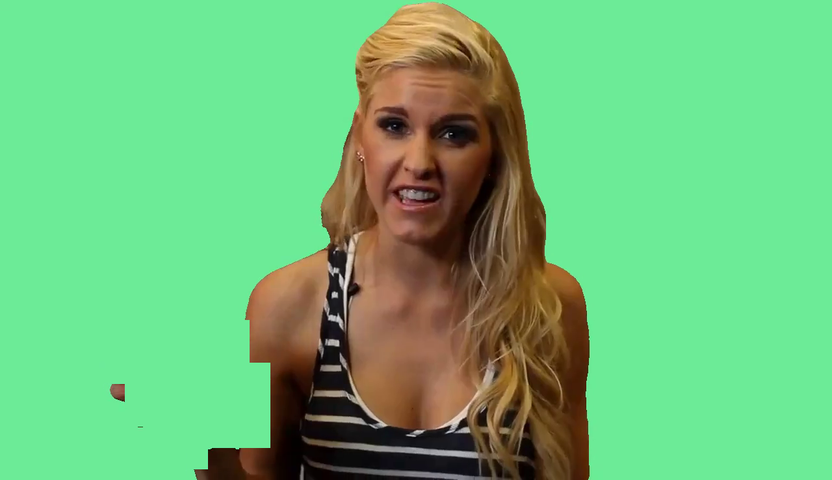} &
			\includegraphics[width=\refresultwidth]{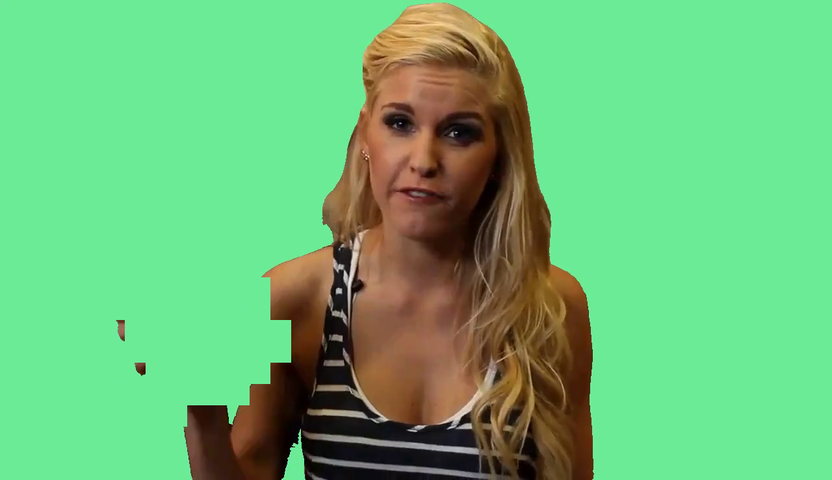} \\
			
			\scriptsize Given Canvas & & \scriptsize Result \\
			\includegraphics[width=\refresultwidth]{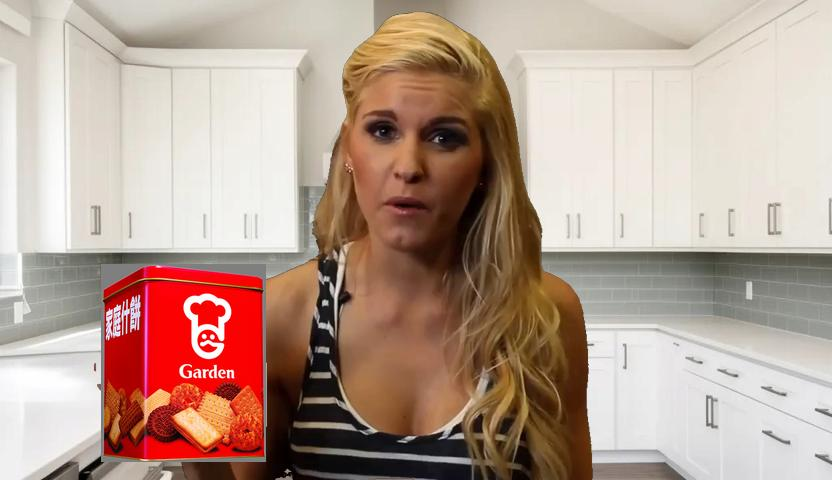} &
			\includegraphics[width=\refresultwidth]{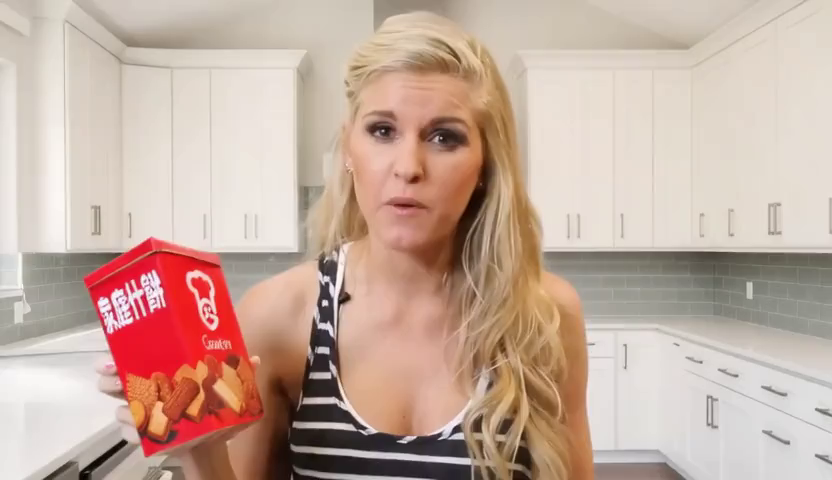} &
			\includegraphics[width=\refresultwidth]{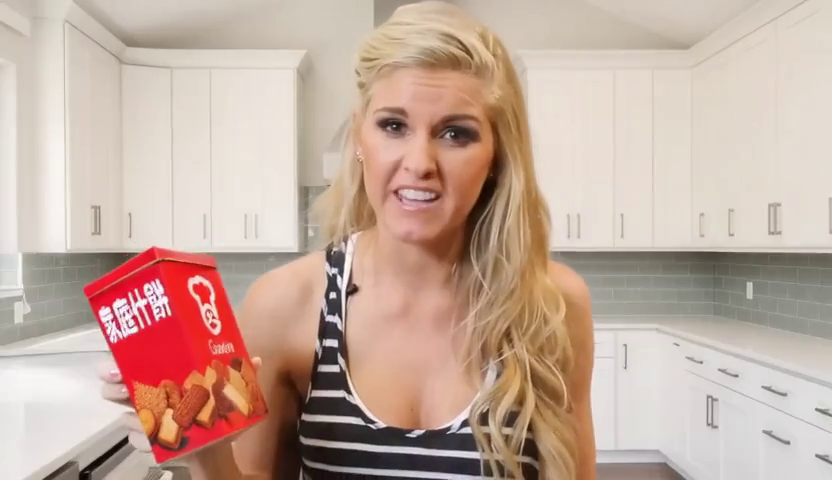} &
			\includegraphics[width=\refresultwidth]{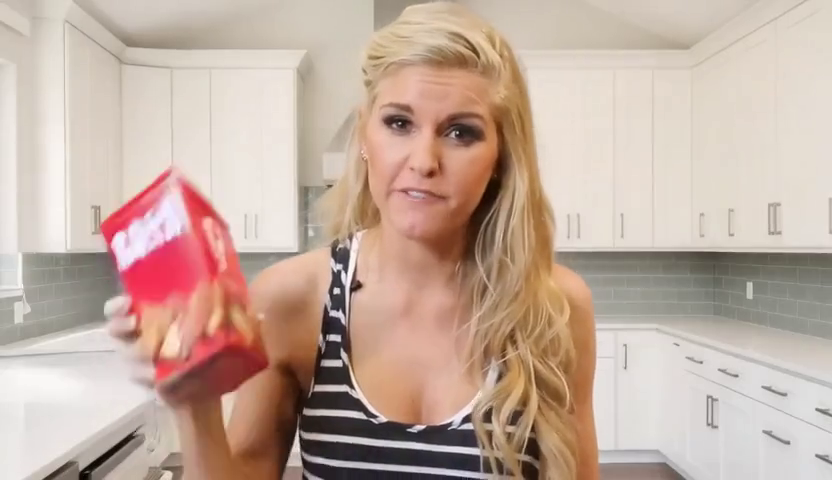} \\
			\multicolumn{4}{p{0.95\linewidth}}{\small Prompt: \textit{A blonde woman in a striped tank top stands in a bright kitchen, holding a red box of Garden biscuits. She presents the product to the camera with kitchen decorations behind her.}} \\
		\end{tabular}
	\end{minipage}\hfill
	\begin{minipage}[t]{0.49\textwidth}
		\centering
		\setlength{\tabcolsep}{0.01em}
		\renewcommand{\arraystretch}{0.8}
		\begin{tabular}{c|ccc}
			\scriptsize Origin video frame & & \scriptsize Green Screen Video & \\
 			\includegraphics[width=\refresultwidth]{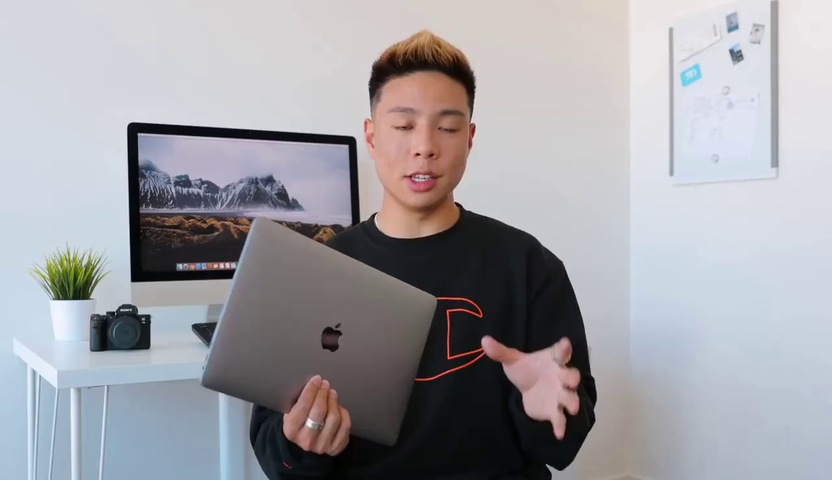} &
			\includegraphics[width=\refresultwidth]{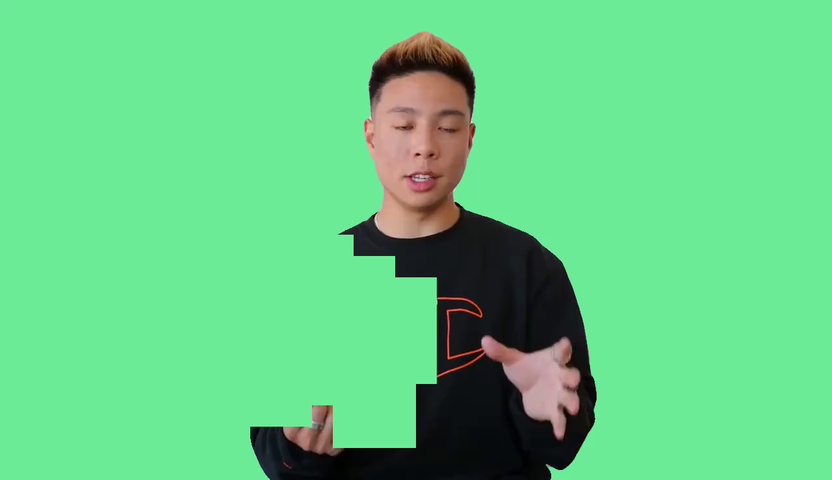} &
			\includegraphics[width=\refresultwidth]{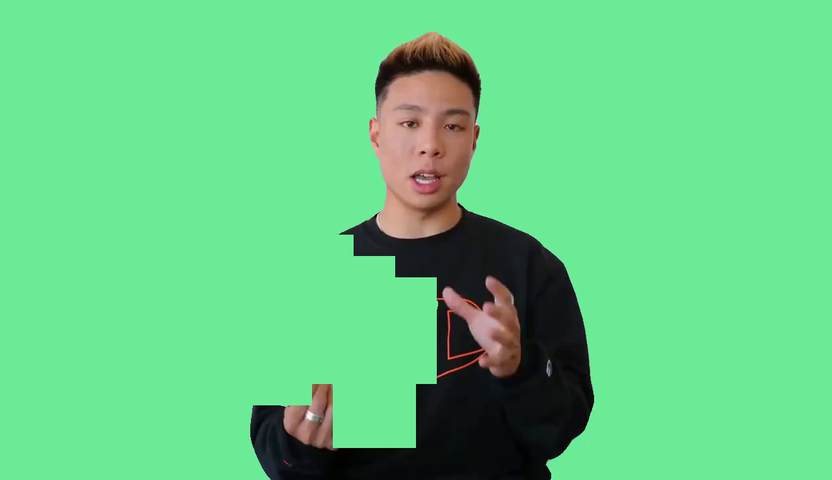} &
			\includegraphics[width=\refresultwidth]{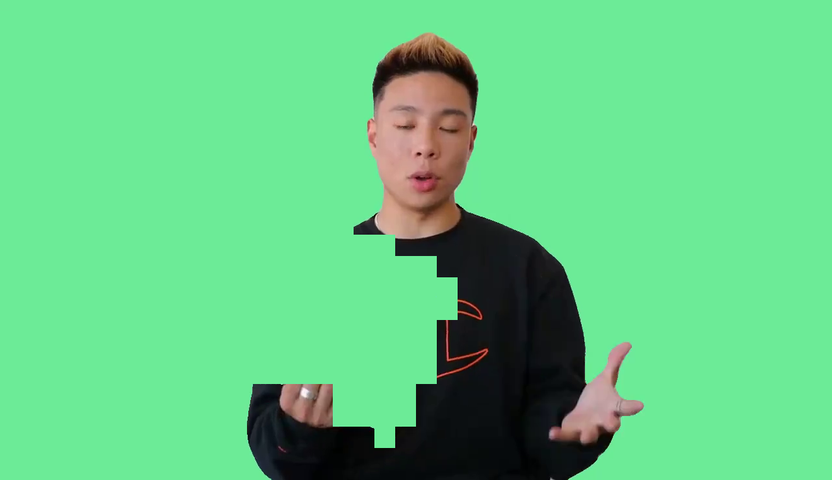} \\
			
			\scriptsize Given Canvas & & \scriptsize Result \\
			\includegraphics[width=\refresultwidth]{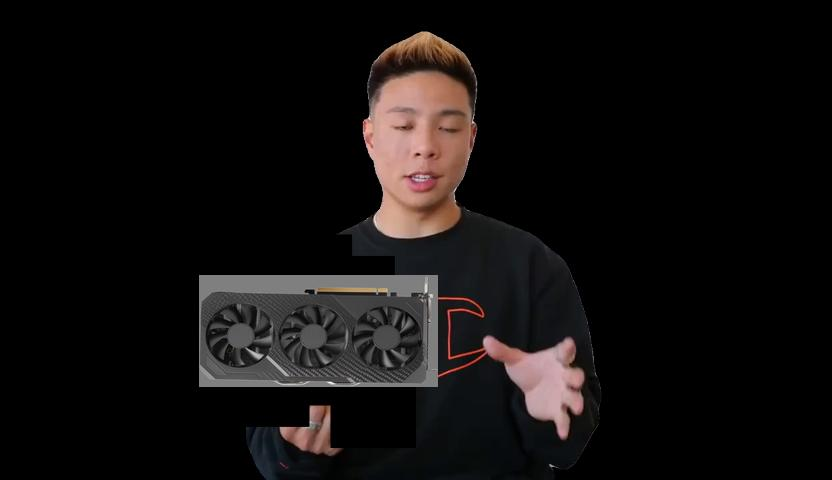} &
			\includegraphics[width=\refresultwidth]{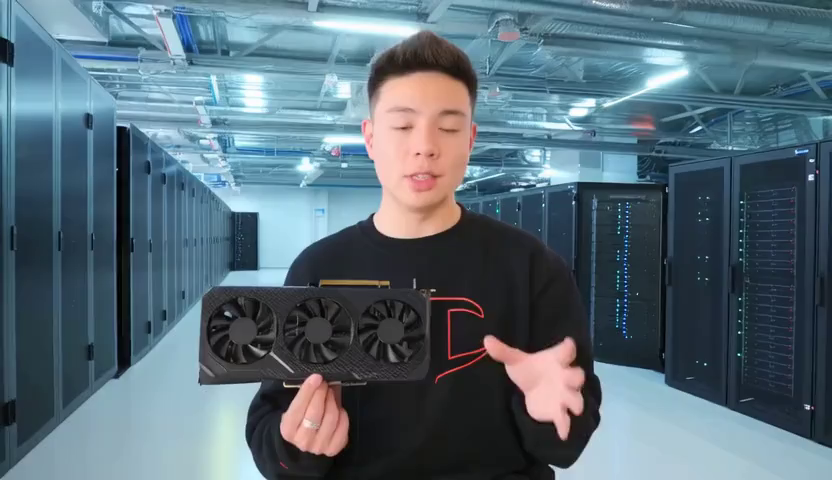} &
			\includegraphics[width=\refresultwidth]{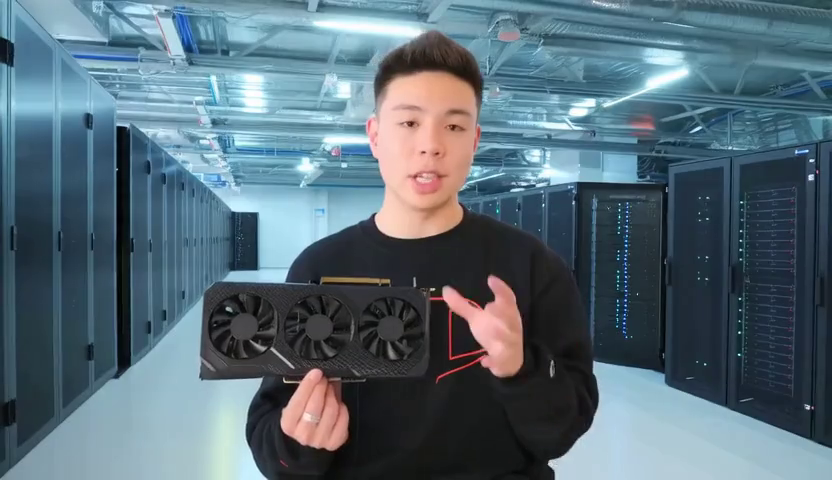} &
			\includegraphics[width=\refresultwidth]{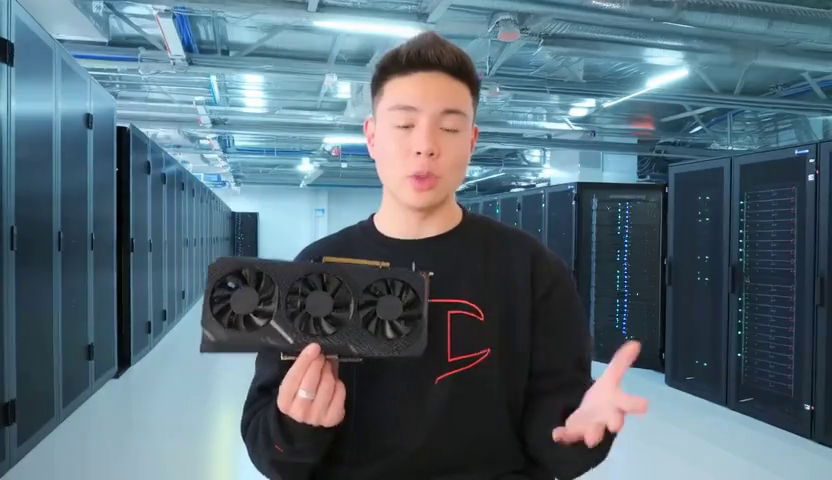} \\
			\multicolumn{4}{p{0.95\linewidth}}{\small Prompt: \textit{A young man in a black sweatshirt presents a black graphics card inside a bright server room. Rows of server racks, cables, and blinking lights form a clean high-tech background.}} \\
		\end{tabular}
	\end{minipage}
	
	\caption{\textbf{Results with reference guidance.} We present several examples where reference images guide the generation of specific objects. Note that prompts have been simplified for illustrative clarity.}
	\label{fig:reference_result_new}
\end{figure*}

\section{Conclusion}
\label{sec:conclusion}
In this work, we present an end-to-end generative framework for cinematic green-screen video compositing, which unifies character-to-environment physical interaction and environment-to-character lighting harmonization in a single video diffusion model. Our framework incorporates a tri-mask-guided architecture for adaptive prop handling, an RGB-D joint denoising strategy for improved 3D spatial understanding, a prior-driven data curation pipeline for efficient multi-illumination training data generation, and a reference-conditioned mechanism for customizable environment and prop control. Extensive experiments demonstrate that our method achieves state-of-the-art performance in cinematic-quality dynamic compositing.

\noindent \textbf{Limitations.} Limited by computational resources, our current implementation cannot support 4K ultra-high-resolution or long-form video synthesis. Developing efficient architectures and scalable training strategies to enable high-resolution long-video compositing will be our primary future direction.

\bibliographystyle{IEEEtran}
\bibliography{reference_abbreviated}

@inproceedings{yang2026matanyone2,
  author = {Yang, Peiqing and Zhou, Shangchen and Hao, Kai and Tao, Qingyi},
  booktitle = {CVPR},
  title = {{MatAnyone 2}: Scaling Video Matting via a Learned Quality Evaluator},
  year = {2026}
}

@inproceedings{yang2025matanyone,
  title={MatAnyone: Stable video matting with consistent memory propagation},
  author={Yang, Peiqing and Zhou, Shangchen and Zhao, Jixin and Tao, Qingyi and Loy, Chen Change},
  booktitle={CVPR},
  pages={7299--7308},
  year={2025}
}

@inproceedings{jiang2025vace,
  author = {Jiang, Zeyinzi and Han, Zhen and Mao, Chaojie and Zhang, Jingfeng and Pan, Yulin and Liu, Yu},
  booktitle = {ICCV},
  title = {{VACE}: All-in-One Video Creation and Editing},
  year = {2025}
}

@inproceedings{peng2026lightctrl,
  author = {Peng, Yizuo and Chen, Xuelin and Zhang, Kai and Cun, Xiaodong},
  booktitle = {ICLR},
  title = {{LightCtrl}: Training-free Controllable Video Relighting},
  year = {2026}
}

@inproceedings{xu2026mocha,
  author = {Xu, Zhengbo and Ma, Jie and Wang, Ziheng and Peng, Zhan and Liang, Jun and Li, Jing},
  booktitle = {CVPR},
  title = {{MoCha}: End-to-End Video Character Replacement without Structural Guidance},
  year = {2026}
}

@inproceedings{hu2025animate,
  author = {Hu, Li and Wang, Guangyuan and Shen, Zhen and Gao, Xin and Meng, Dechao and Zhuo, Lian and Zhang, Peng and Zhang, Bang and Bo, Liefeng},
  booktitle = {ICCV},
  title = {{Animate Anyone 2}: High-Fidelity Character Image Animation with Environment Affordance},
  year = {2025}
}

@inproceedings{bian2025videopainter,
  author = {Bian, Yuxuan and Zhang, Zhaoyang and Ju, Xuan and Mingdeng Cao and Liangbin Xie and Ying Shan and Qiang Xu},
  booktitle = {ACM SIGGRAPH},
  title = {{VideoPainter}: Any-length Video Inpainting and Editing with Plug-and-Play Context Control},
  year = {2025}
}

@article{yao2025beyond,
  author = {Yao, Mingshuai and others},
  journal = {arXiv:2504.02004},
  title = {Beyond Static Scenes: Camera-controllable Background Generation for Human Motion},
  year = {2025}
}

@InProceedings{zhang2024mimicmotion,
  title = 	 {{MimicMotion}: High-Quality Human Motion Video Generation with Confidence-aware Pose Guidance},
  author =       {Zhang, Yuang and Gu, Jiaxi and Wang, Li-Wen and Wang, Han and Cheng, Junqi and Zhu, Yuefeng and Zou, Fangyuan},
  booktitle = 	 {ICML},
  pages = 	 {74896--74910},
  year = 	 {2025},
  volume = 	 {267},
}

@inproceedings{lin2021robust,
  author = {Lin, Shanchuan and others},
  booktitle = {WACV},
  title = {Robust High-Resolution Video Matting with Temporal Guidance},
  year = {2022}
}

@inproceedings{lin2021real,
  author = {Lin, Shanchuan and Andrey Ryabtsev and Soumyadip Sengupta and Brian Curless and Steve Seitz and Ira Kemelmacher-Shlizerman},
  booktitle = {CVPR},
  title = {Real-Time High-Resolution Background Matting},
  year = {2021}
}

@article{hu2025ex4d,
  author = {Hu, Tao and Haoyang Peng and Xiao Liu and Yuewen Ma},
  journal = {arXiv:2506.05554},
  title = {{EX-4D}: EXtreme Viewpoint 4D Video Synthesis via Depth Watertight Mesh},
  year = {2025}
}

@article{fang2025relightvid,
  author = {Fang, Ye and Zeyi Sun and Shangzhan Zhang and Tong Wu and Yinghao Xu and Pan Zhang and Jiaqi Wang and Gordon Wetzstein and Dahua Lin},
  journal = {arXiv:2501.16330},
  title = {{RelightVid}: Temporal-Consistent Diffusion Model for Video Relighting},
  year = {2025}
}

@inproceedings{bian2025relightmaster,
  author = {Bian, W. and others},
  booktitle = {CVPR},
  title = {{RelightMaster}: Precise Video Relighting with Multi-plane Light Images},
  year = {2025}
}

@inproceedings{zhang2025scaling,
  author = {Lvmin Zhang and Anyi Rao and Maneesh Agrawala},
  booktitle = {ICLR},
  title = {Scaling In-the-Wild Training for Diffusion-based Illumination Harmonization and Editing by Imposing Consistent Light Transport},
  year = {2025}
}

@article{pan2024actanywhere,
  author = {Pan, Boxiao and Xu, Zhan and Huang, Chun-Hao P and Singh, Krishna K and Zhou, Yang and Guibas, Leonidas J and Yang, Jimei},
  journal = {NeurIPS},
  pages = {29754--29776},
  title = {Actanywhere: Subject-aware video background generation},
  volume = {37},
  year = {2024}
}

@inproceedings{zhang2024avid,
  author = {Zhang, Zhixing and Wu, Bichen and Wang, Xiaoyan and Luo, Yaqiao and Zhang, Luxin and Zhao, Yinan and Vajda, Peter and Metaxas, Dimitris and Yu, Licheng},
  booktitle = {CVPR},
  pages = {7162--7172},
  title = {Avid: Any-length video inpainting with diffusion model},
  year = {2024}
}

@inproceedings{litman2026editctrl,
  title={Editctrl: Disentangled local and global control for real-time generative video editing},
  author={Litman, Yehonathan and Liu, Shikun and Seyb, Dario and Milef, Nicholas and Zhou, Yang and Marshall, Carl and Tulsiani, Shubham and Leak, Caleb},
  booktitle={CVPR},
  pages={8965--8975},
  year={2026}
}

@article{xu2026anchorcrafter,
  author = {Xu, Ziyi and Huang, Ziyao and Cao, Juan and Zhang, Yong and Cun, Xiaodong and Shuai, Qing and Wang, Yuchen and Bao, Linchao and Tang, Fan},
  journal = {TVCG},
  title = {AnchorCrafter: Animate cyber-anchors selling your products via human-object interacting video generation},
  year = {2026}
}

@article{liu2025byteloom,
  author = {Liu, Bangya and Gong, Xinyu and Zhao, Zelin and Song, Ziyang and Lu, Yulei and Wu, Suhui and Zhang, Jun and Banerjee, Suman and Zhang, Hao},
  journal = {arXiv:2512.22854},
  title = {ByteLoom: Weaving Geometry-Consistent Human-Object Interactions through Progressive Curriculum Learning},
  year = {2025}
}

@article{cheng2025wan,
  author = {Cheng, Gang and Gao, Xin and Hu, Li and Hu, Siqi and Huang, Mingyang and Ji, Chaonan and Li, Ju and Meng, Dechao and Qi, Jinwei and Qiao, Penchong and others},
  journal = {arXiv:2509.14055},
  title = {Wan-animate: Unified character animation and replacement with holistic replication},
  year = {2025}
}

@inproceedings{zhou2025light,
  author = {Zhou, Yujie and Bu, Jiazi and Ling, Pengyang and Zhang, Pan and Wu, Tong and Huang, Qidong and Li, Jinsong and Dong, Xiaoyi and Zang, Yuhang and Cao, Yuhang and others},
  booktitle = {ICCV},
  pages = {13315--13325},
  title = {Light-a-video: Training-free video relighting via progressive light fusion},
  year = {2025}
}

@inproceedings{wang2023semi,
  author = {Wang, Ke and Gharbi, Micha{\"e}l and Zhang, He and Xia, Zhihao and Shechtman, Eli},
  booktitle = {CVPR},
  title = {Semi-supervised Parametric Real-world Image Harmonization},
  year = {2023}
}

@inproceedings{ren2024relightful,
  author = {Ren, Mengwei and Xiong, Wei and Yoon, Jae Shin and Shu, Zhixin and Zhang, Jianming and Jung, HyunJoon and Gerig, Guido and Zhang, He},
  booktitle = {CVPR},
  title = {{Relightful Harmonization}: Lighting-aware Portrait Background Replacement},
  year = {2024}
}

@inproceedings{he2025unirelight,
  author = {He, Kai and Liang, Ruofan and Munkberg, Jacob and Hasselgren, Jon and Vijaykumar, Nandita and Keller, Alexander and Fidler, Sanja and Gilitschenski, Igor and Gojcic, Zan and Wang, Zian},
  booktitle = {NeurIPS},
  title = {{UniRelight}: Learning Joint Decomposition and Synthesis for Video Relighting},
  year = {2025}
}

@inproceedings{liu2025tclight,
  author = {Liu, Yang and Luo, Chuanchen and Tang, Zimo and Li, Yingyan and Yang, Yuran and Ning, Yuanyong and Fan, Lue and Peng, Junran and Zhaoxiang Zhang},
  booktitle = {NeurIPS},
  title = {{TC-Light}: Temporally Coherent Generative Rendering for Realistic World Transfer},
  year = {2025}
}

@inproceedings{damo2025unilumos,
  author = {Liu, Pengwei and Yuan, Hangjie and Dong, Bo and Xing, Jiazheng and Wang, Jinwang and Zhao, Rui and Chen, Weihua and Wang, Fan},
  booktitle = {NeurIPS},
  title = {{UniLumos}: Fast and Unified Image and Video Relighting with Physics-Plausible Feedback},
  year = {2025}
}

@inproceedings{liu2026lightx,
  author = {Liu, Tianqi and Chen, Zhaoxi and Huang, Zihao and Xu, Shaocong and Zhang, Saining and Ye, Chongjie and Bohan Li and Zhiguo Cao and Wei Li and Hao Zhao and Ziwei Liu},
  booktitle = {ICLR},
  title = {{Light-X}: Generative 4D Video Rendering with Camera and Illumination Control},
  year = {2026}
}

@article{xiao2026relitlive,
  author = {Xiao, Weiqing and Li, Hong and Yang, Xiuyu and Chen, Houyuan and Wen, Yi and Liu, Tianqi and Xu, Shaocong and Ye, Chongjie and Zhao, Hao and Wang, Beibei},
  journal = {arXiv:2605.06658},
  title = {{Relit-LiVE}: Relight Video by Jointly Learning Environment Video},
  year = {2026}
}

@misc{flux-2-2025,
  author = {Black Forest Labs},
  title = {{FLUX.2: Frontier Visual Intelligence}},
  url = {https://bfl.ai/blog/flux-2},
  year = {2025}
}

@inproceedings{sun2025attentive,
  author = {Sun, Wenhao and Dong, Xue-Mei and Cui, Benlei and Tang, Jingqun},
  booktitle = {AAAI},
  number = {19},
  pages = {20734--20742},
  title = {Attentive Eraser: Unleashing Diffusion Model's Object Removal Potential via Self-Attention Redirection Guidance},
  volume = {39},
  year = {2025}
}

@inproceedings{xu2021videoclip,
  author = {Xu, Hu and Ghosh, Gargi and Huang, Po-Yao and Okhonko, Dmytro and Aghajanyan, Armen and Metze, Florian and Zettlemoyer, Luke and Feichtenhofer, Christoph},
  booktitle = {EMNLP},
  pages = {6787--6800},
  title = {VideoCLIP: Contrastive Pre-training for Zero-shot Video-Text Understanding},
  year = {2021}
}

@article{wan2025wan,
  author = {WanTeam and Wang, Ang and Ai, Baole and Wen, Bin and Mao, Chaojie and Xie, Chen-Wei and Chen, Di and Yu, Feiwu and Zhao, Haiming and Yang, Jianxiao and others},
  journal = {arXiv:2503.20314},
  title = {Wan: Open and Advanced Large-Scale Video Generative Models},
  year = {2025}
}

@inproceedings{liu2025hoigen,
  author = {Liu, Kun and Liu, Qi and Liu, Xinchen and Li, Jie and Zhang, Yongdong and Luo, Jiebo and He, Xiaodong and Liu, Wu},
  booktitle = {CVPR},
  pages = {24001--24010},
  title = {Hoigen-1m: A large-scale dataset for human-object interaction video generation},
  year = {2025}
}

@article{yang2025qwen3,
  author = {Yang, An and Li, Anfeng and Yang, Baosong and Zhang, Beichen and Hui, Binyuan and Zheng, Bo and Yu, Bowen and Gao, Chang and Huang, Chengen and Lv, Chenxu and others},
  journal = {arXiv:2505.09388},
  title = {Qwen3 technical report},
  year = {2025}
}

@inproceedings{ravi2025sam,
  title={Sam 2: Segment anything in images and videos},
  author={Ravi, Nikhila and Gabeur, Valentin and Hu, Yuan-Ting and Hu, Ronghang and Ryali, Chaitanya and Ma, Tengyu and Khedr, Haitham and R{\"a}dle, Roman and Rolland, Chloe and Gustafson, Laura and others},
  booktitle={ICLR},
  volume={2025},
  pages={28085--28128},
  year={2025}
}

@inproceedings{liu2024grounding,
  title={Grounding dino: Marrying dino with grounded pre-training for open-set object detection},
  author={Liu, Shilong and Zeng, Zhaoyang and Ren, Tianhe and Li, Feng and Zhang, Hao and Yang, Jie and Jiang, Qing and Li, Chunyuan and Yang, Jianwei and Su, Hang and others},
  booktitle={ECCV},
  pages={38--55},
  year={2024},
}

@article{li2022bridging,
  author = {Li, Jizhizi and Zhang, Jing and Maybank, Stephen J and Tao, Dacheng},
  journal = {IJCV},
  number = {2},
  pages = {246--266},
  title = {Bridging composite and real: towards end-to-end deep image matting},
  volume = {130},
  year = {2022}
}

@inproceedings{li2023blip,
  title={Blip-2: Bootstrapping language-image pre-training with frozen image encoders and large language models},
  author={Li, Junnan and Li, Dongxu and Savarese, Silvio and Hoi, Steven},
  booktitle={ICML},
  pages={19730--19742},
  year={2023},
}

@inproceedings{chen2025video,
  title={Video depth anything: Consistent depth estimation for super-long videos},
  author={Chen, Sili and Guo, Hengkai and Zhu, Shengnan and Zhang, Feihu and Huang, Zilong and Feng, Jiashi and Kang, Bingyi},
  booktitle={CVPR},
  pages={22831--22840},
  year={2025}
}

@misc{schuhmann2022improved,
  author = {Christoph Schuhmann},
  title = {Improved Aesthetic Prediction using CLIP Embeddings and Hypernetwork},
  url = {https://github.com/christophschuhmann/improved-aesthetic-predictor},
  year = {2022}
}

@misc{insightface,
  title = {InsightFace},
  url = {https://github.com/deepinsight/insightface},
}

@inproceedings{hu2022lora,
  title={Lo{RA}: Low-Rank Adaptation of Large Language Models},
  author = {Hu, Edward J. and Shen, Yelong and Wallis, Phillip and Allen-Zhu, Zeyuan and Li, Yuanzhi and Wang, Shean and Wang, Lu and Chen, Weizhu},
  booktitle={ICLR},
  year={2022},
}

@inproceedings{huang2024vbench,
  title={Vbench: Comprehensive benchmark suite for video generative models},
  author={Huang, Ziqi and He, Yinan and Yu, Jiashuo and Zhang, Fan and Si, Chenyang and Jiang, Yuming and Zhang, Yuanhan and Wu, Tianxing and Jin, Qingyang and Chanpaisit, Nattapol and others},
  booktitle={CVPR},
  pages={21807--21818},
  year={2024}
}

@inproceedings{gao2025anyportal,
  author = {Gao, Wenshuo and Lan, Xicheng and Yang, Shuai},
  booktitle = {ICCV},
  pages = {18990--18999},
  title = {ANYPORTAL: Zero-Shot Consistent Video Background Replacement},
  year = {2025}
}

@inproceedings{gao2026flowportal,
  author = {Gao, Wenshuo and Fan, Junyi and Zeng, Jiangyue and Yang, Shuai},
  booktitle = {CVPR},
  pages = {2025--2034},
  title = {FlowPortal: Residual-corrected flow for training-free video relighting and background replacement},
  year = {2026}
}

@article{bao2022interactive,
  title={Interactive lighting editing system for single indoor low-light scene images with corresponding depth maps},
  author={Bao, Zhongyun and Fu, Gang and Duan, Lian and Xiao, Chunxia},
  journal={Visual Informatics},
  volume={6},
  number={4},
  pages={90--99},
  year={2022},
}

@article{guo2026pnprorl,
  title={PNProRL: Self-Supervised Neural Relighting via Photometric Perception and Progressive Optimization},
  author={Guo, Chenhao and Yang, Zhulun and Ding, Xin and Yang, You and Liu, Qiong},
  journal={TVCG},
  year={2026},
}

@article{chen2026relightable,
  title={Relightable and Animatable Gaussian Head Avatar from Monocular Videos},
  author={Chen, Zhuo and Yan, Yichao and Gao, Jingnan and Su, Zhuo and Wen, Chao and Li, Zhaohu and Cheng, Yuhao and Lee, Xueying and Leng, Yutong and Zeng, Yikun and others},
  journal={TVCG},
  year={2026},
}

@inproceedings{choi2026relightful,
  title={Relightful Video Portrait Harmonization},
  author={Choi, Jun Myeong and Yoon, Jae Shin and Qi, Luchao and Sengupta, Roni and Lee, Joon-Young},
  booktitle={CVPR},
  pages={23356--23366},
  year={2026}
}

@inproceedings{wang2024internvid,
  title={Internvid: A large-scale video-text dataset for multimodal understanding and generation},
  author={Wang, Yi and He, Yinan and Li, Yizhuo and Li, Kunchang and Yu, Jiashuo and Ma, Xin and Li, Xinhao and Chen, Guo and Chen, Xinyuan and Wang, Yaohui and others},
  booktitle={ICLR},
  volume={2024},
  pages={42055--42079},
  year={2024}
}

@inproceedings{quattoni2009recognizing,
  title={Recognizing indoor scenes},
  author={Quattoni, Ariadna and Torralba, Antonio},
  booktitle={CVPR},
  pages={413--420},
  year={2009},
}

@inproceedings{sun2021deep,
  title={Deep video matting via spatio-temporal alignment and aggregation},
  author={Sun, Yanan and Wang, Guanzhi and Gu, Qiao and Tang, Chi-Keung and Tai, Yu-Wing},
  booktitle={CVPR},
  pages={6975--6984},
  year={2021}
}

@article{xiang2025let,
  title={Let Your Light Shine: Foreground Portrait Matting via Deep Flash Priors},
  author={Xiang, Tianyi and Xu, Yangyang and Hu, Qingxuan and Zi, Chenyi and Zhao, Nanxuan and Wang, Junle and He, Shengfeng},
  journal={TMLR},
  year={2025}
}

\appendix

\section{Additional Results and Applications}
Figure~\ref{fig:more_result} shows additional results across diverse characters, motions, scenes, and lighting conditions, demonstrating the generality of our model. Figures~\ref{fig:more_result_0610_a} and~\ref{fig:more_result_0610_b} further show real-world video applications where different tri-masks enable flexible region preservation and replacement while maintaining C2E physical interaction and E2C lighting harmonization.

\newcommand{\moreresultwidth}{0.235\linewidth}
\begin{figure*}[p]
	\centering
	\setlength{\tabcolsep}{0.5pt}
	\renewcommand{\arraystretch}{0.95}
	\scalebox{0.95}{\begin{tabular}{c@{\hspace{2pt}}cccc}
		\raisebox{0.05\height}{\rotatebox{90}{\scriptsize Green Screen}} &
		\includegraphics[width=\moreresultwidth]{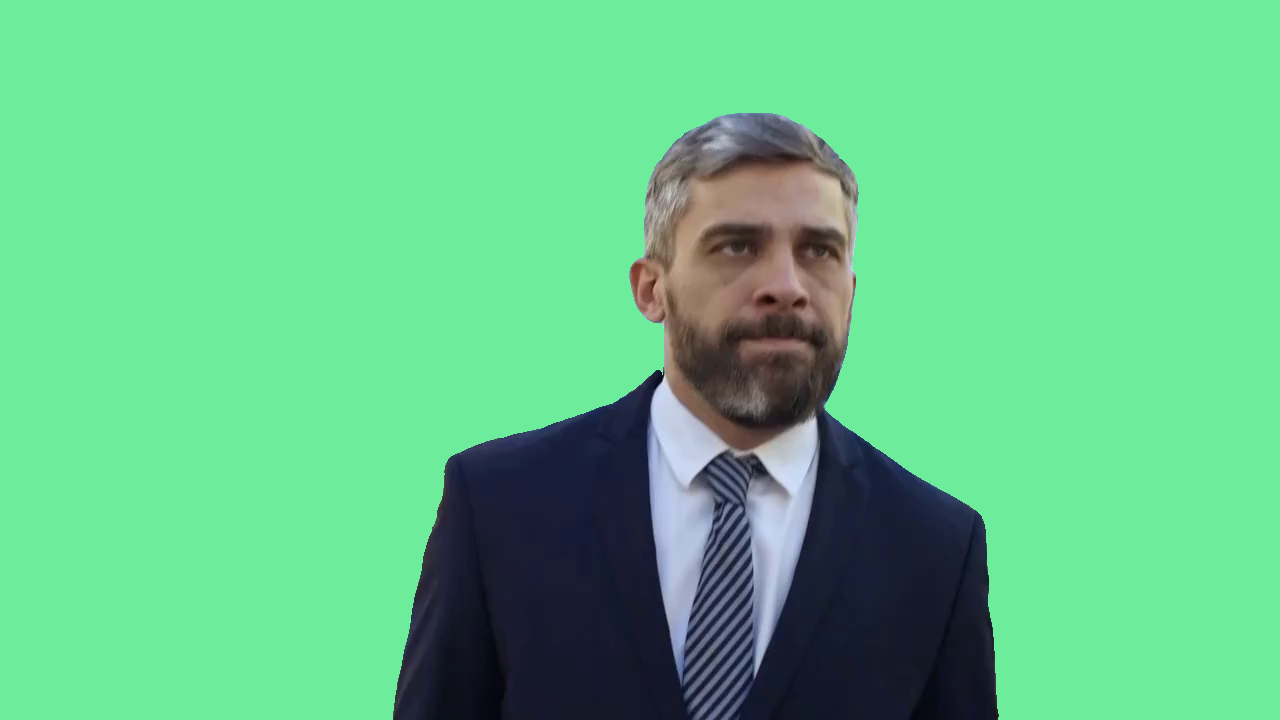} &
		\includegraphics[width=\moreresultwidth]{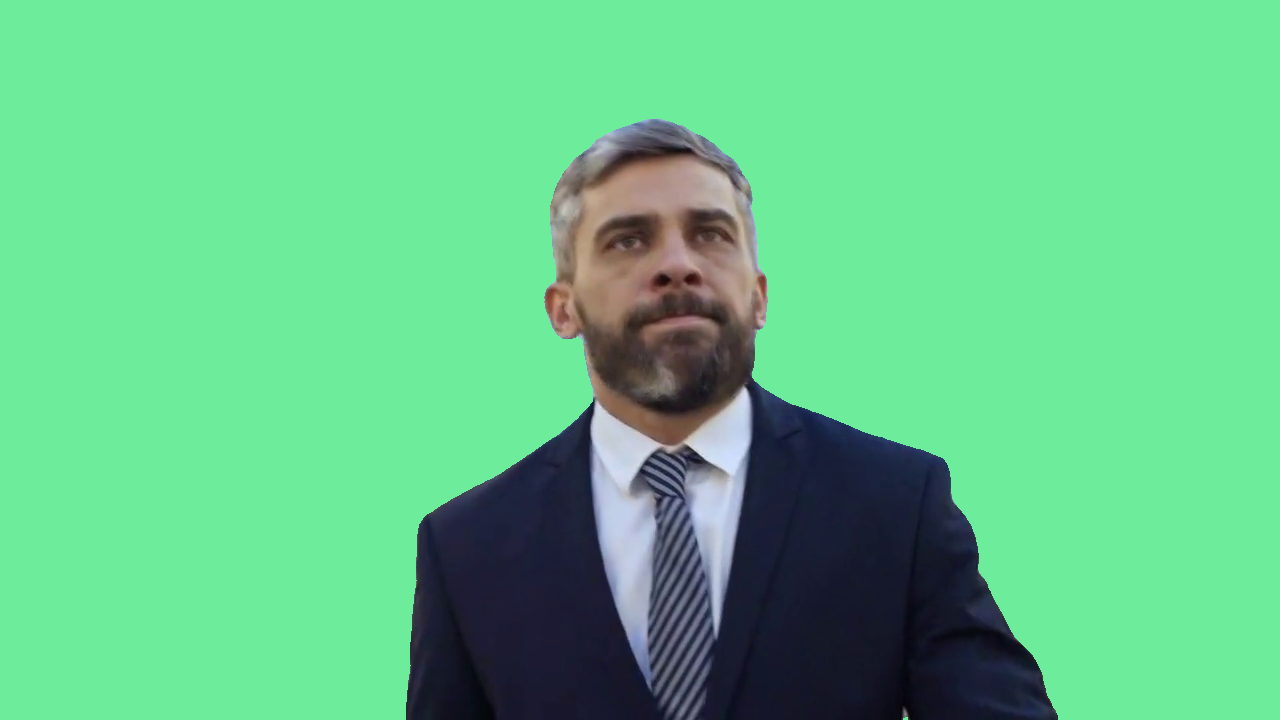} &
		\includegraphics[width=\moreresultwidth]{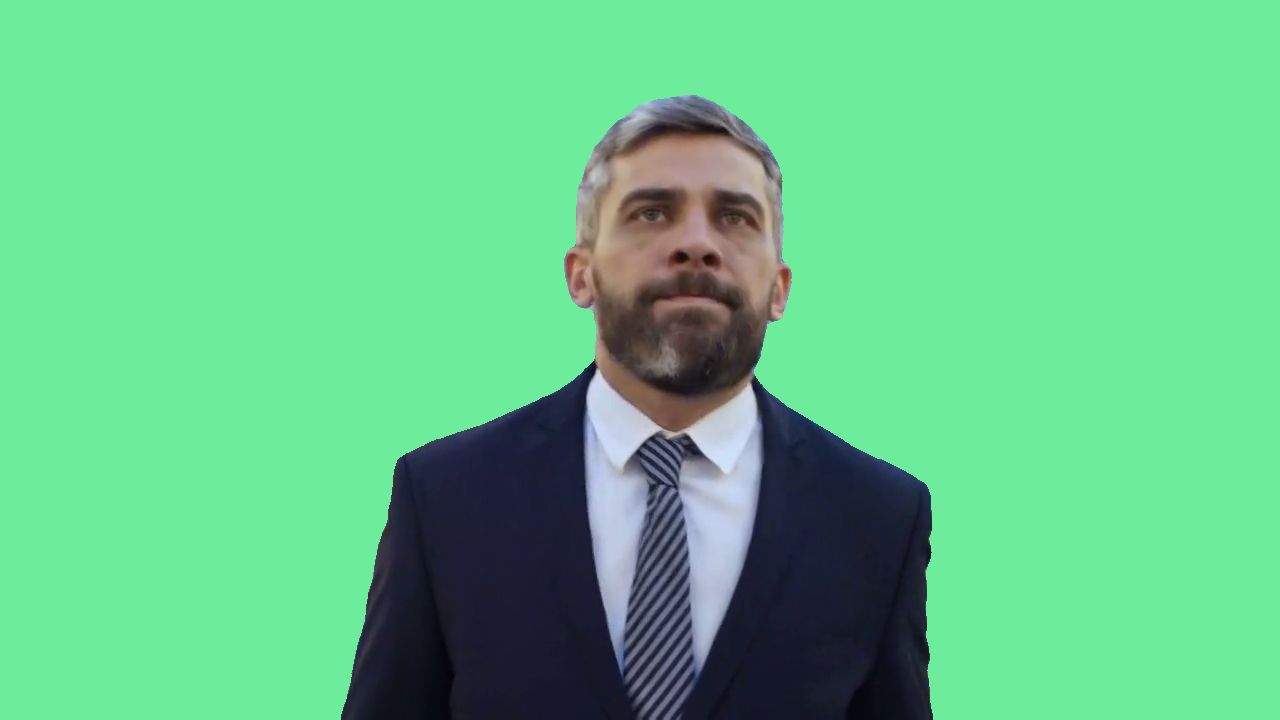} &
		\includegraphics[width=\moreresultwidth]{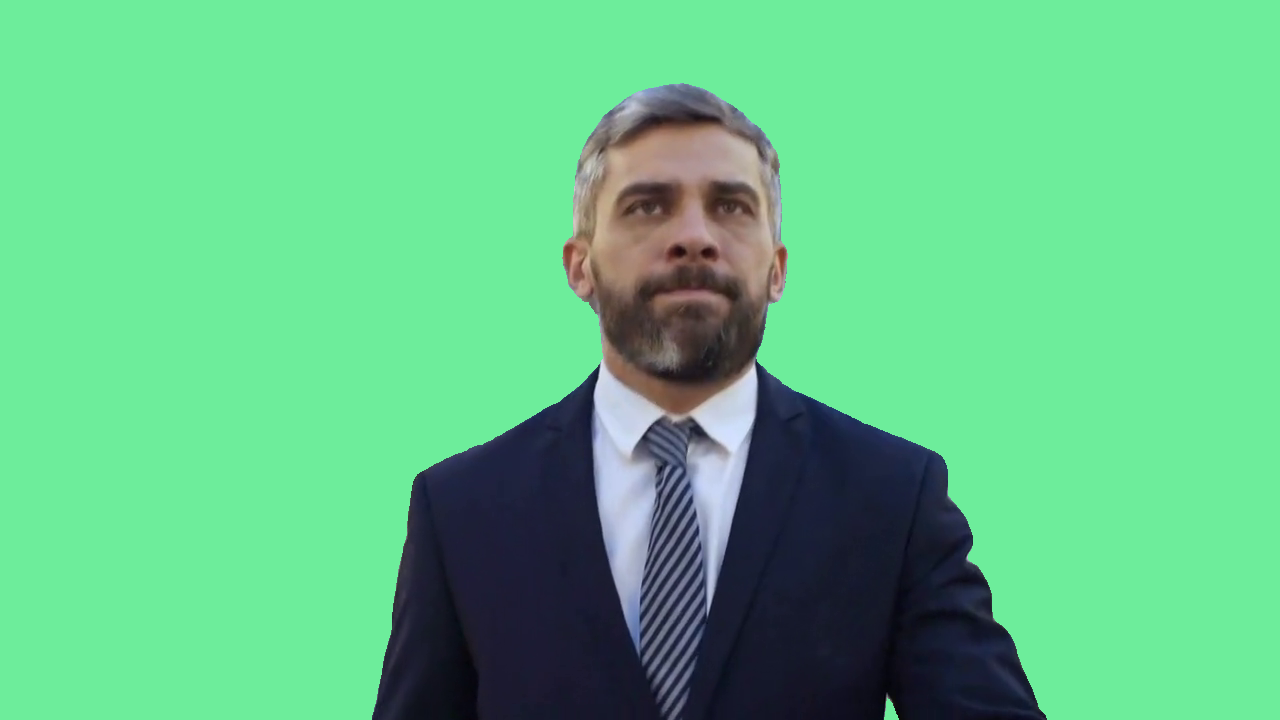} \\

		\raisebox{0.8\height}{\rotatebox{90}{\scriptsize Result}} &
		\includegraphics[width=\moreresultwidth]{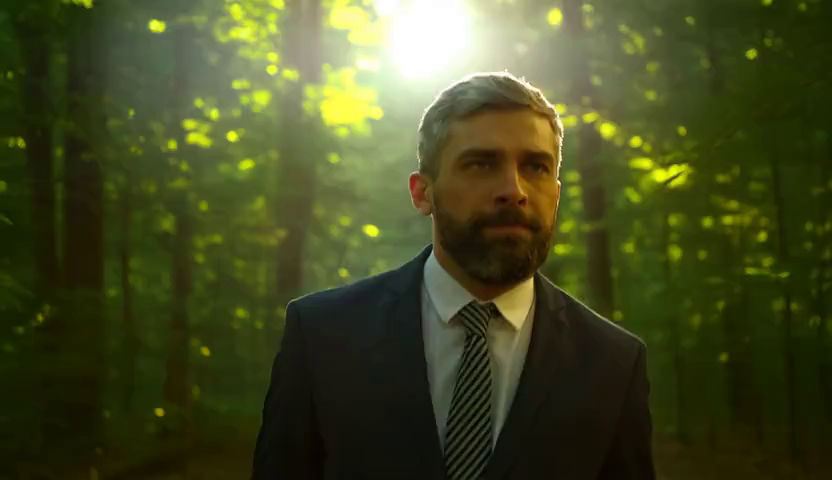} &
		\includegraphics[width=\moreresultwidth]{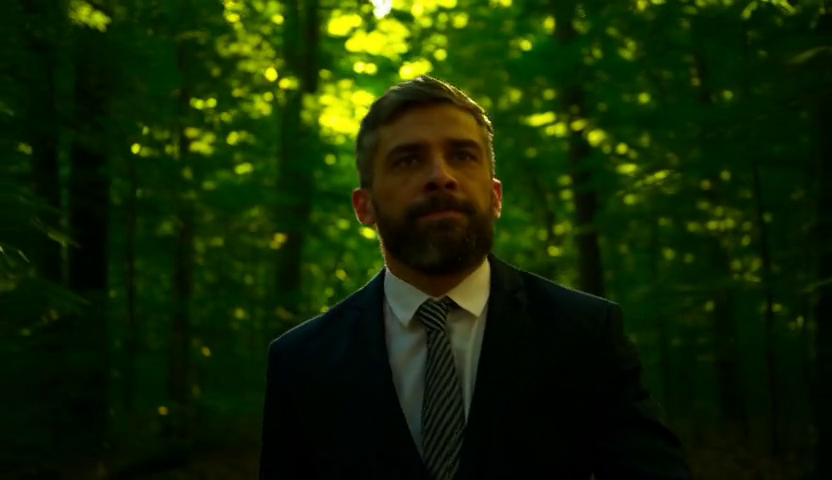} &
		\includegraphics[width=\moreresultwidth]{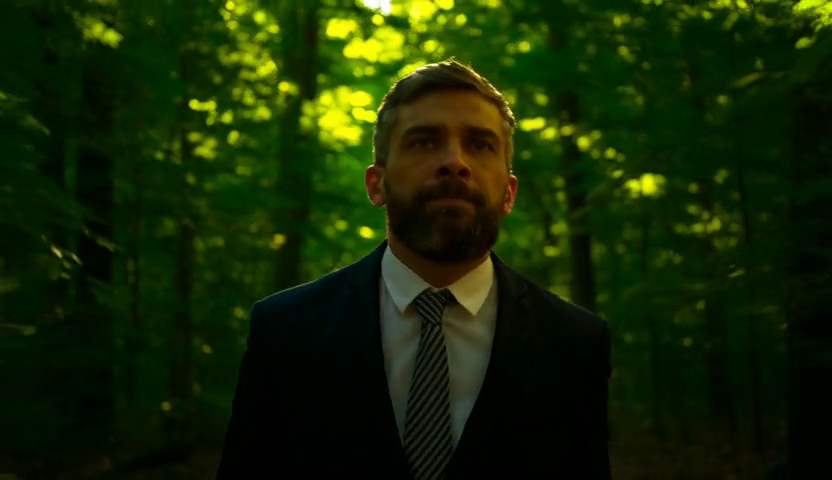} &
		\includegraphics[width=\moreresultwidth]{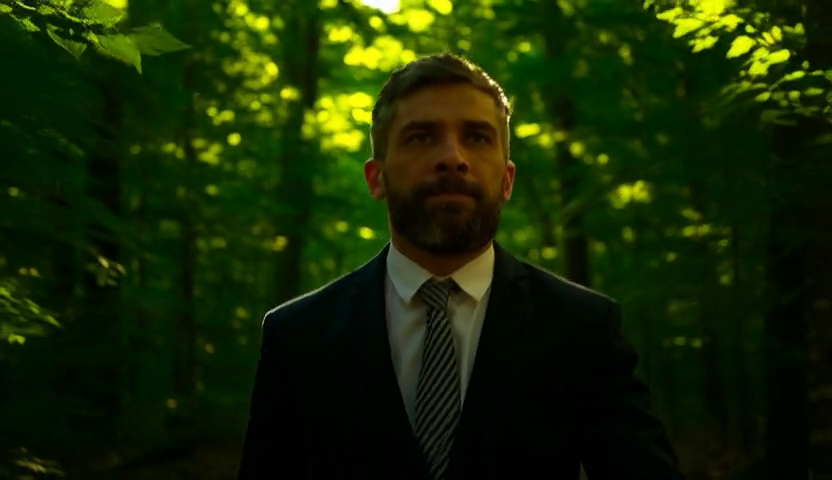} \\
		\multicolumn{5}{p{0.94\textwidth}}{\small Prompt: \textit{A bearded businessman walks toward the camera through a forest, while dappled sunlight through dense foliage casts shifting chartreuse highlights and deep leaf shadows across his face and suit.}} \\
		\raisebox{0.05\height}{\rotatebox{90}{\scriptsize Green Screen}} &
		\includegraphics[width=\moreresultwidth]{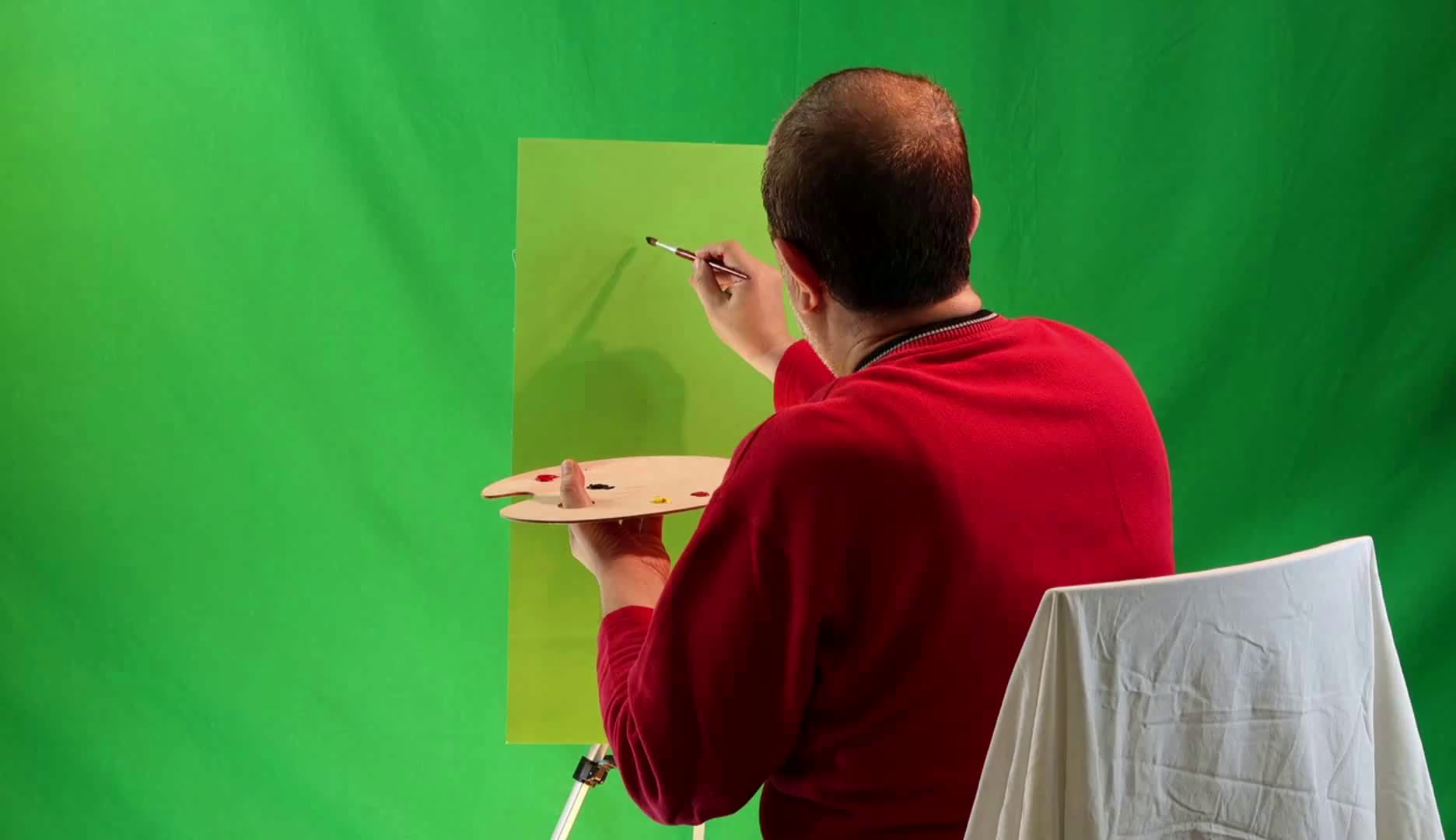} &
		\includegraphics[width=\moreresultwidth]{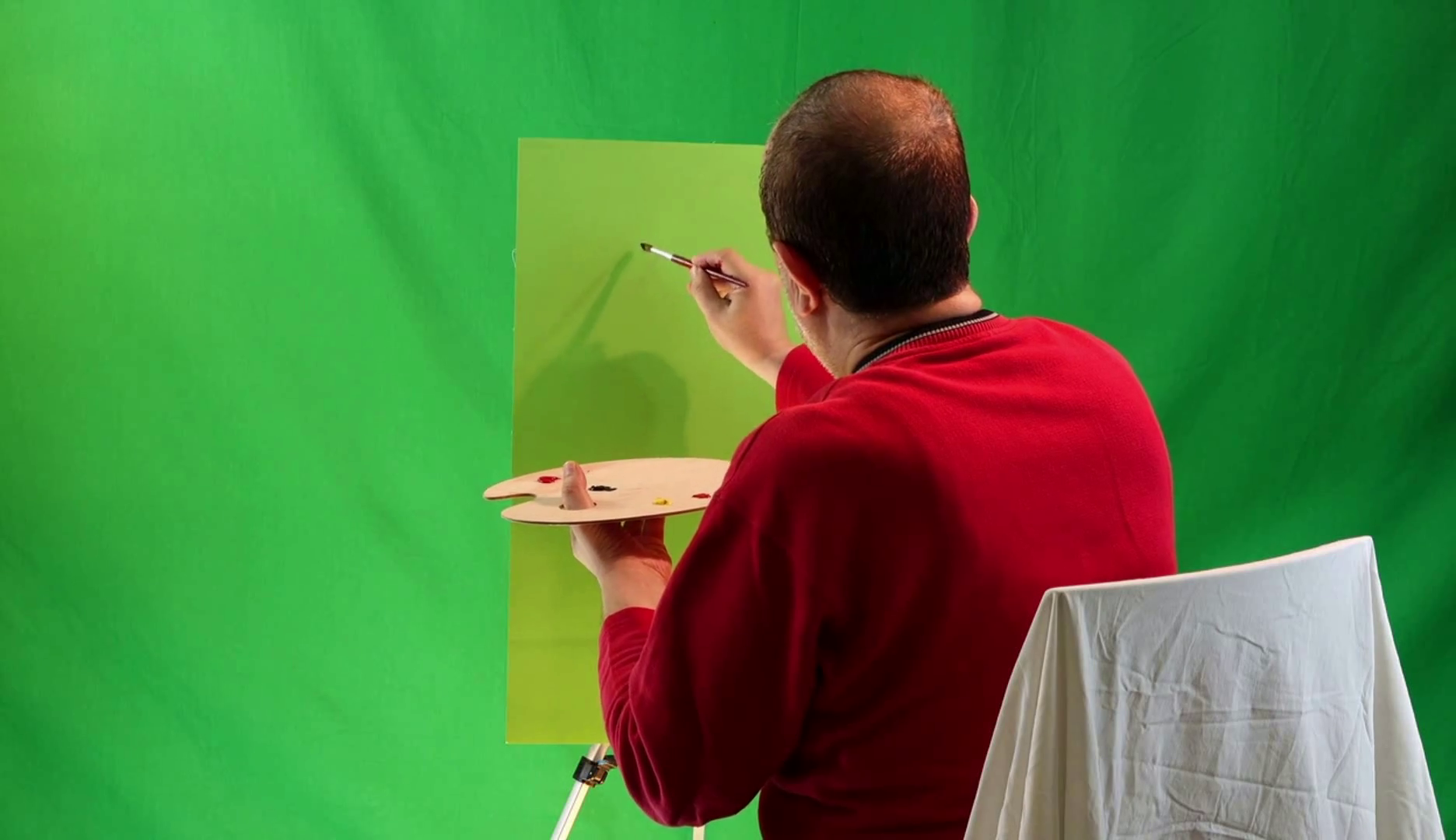} &
		\includegraphics[width=\moreresultwidth]{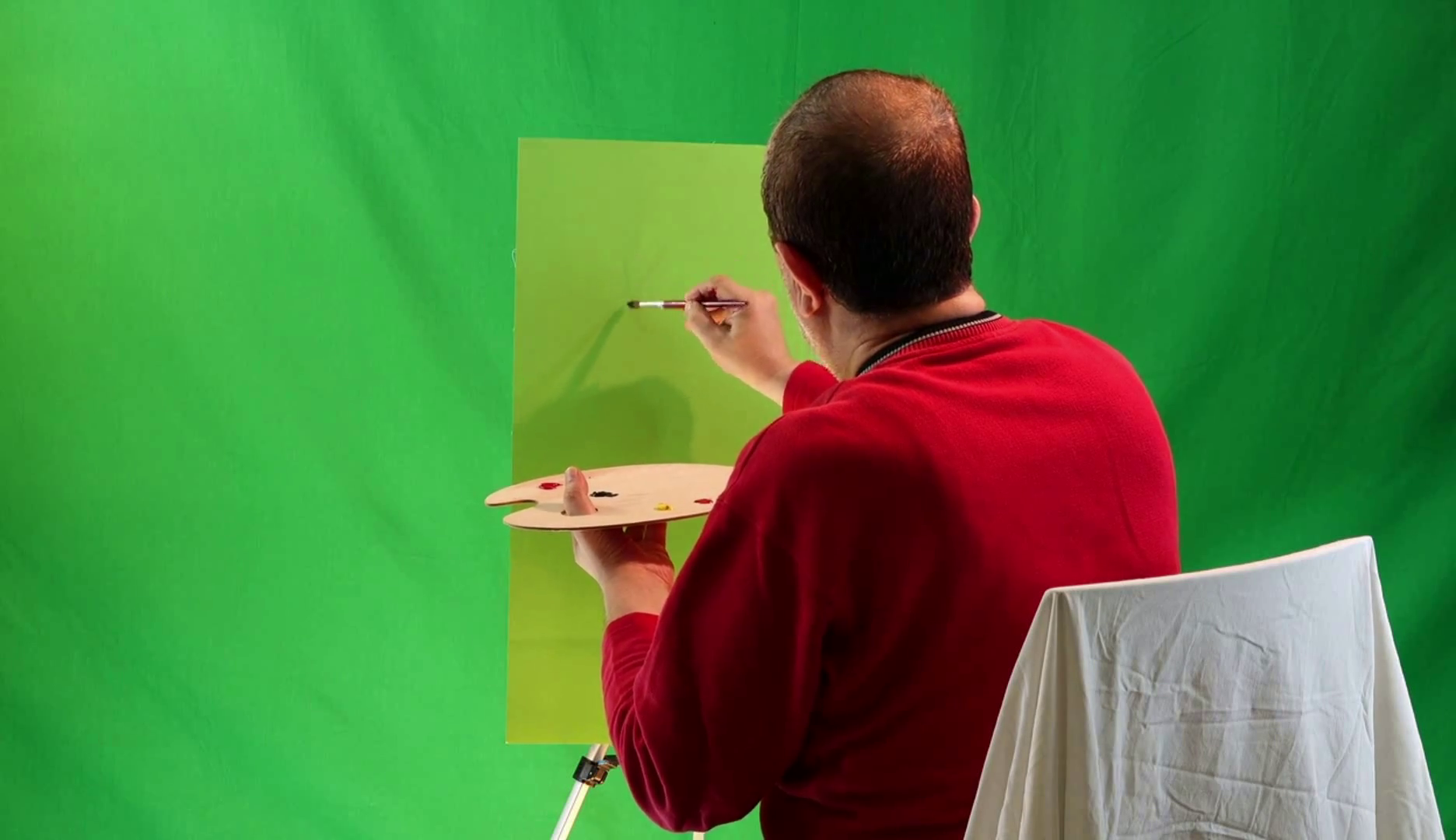} &
		\includegraphics[width=\moreresultwidth]{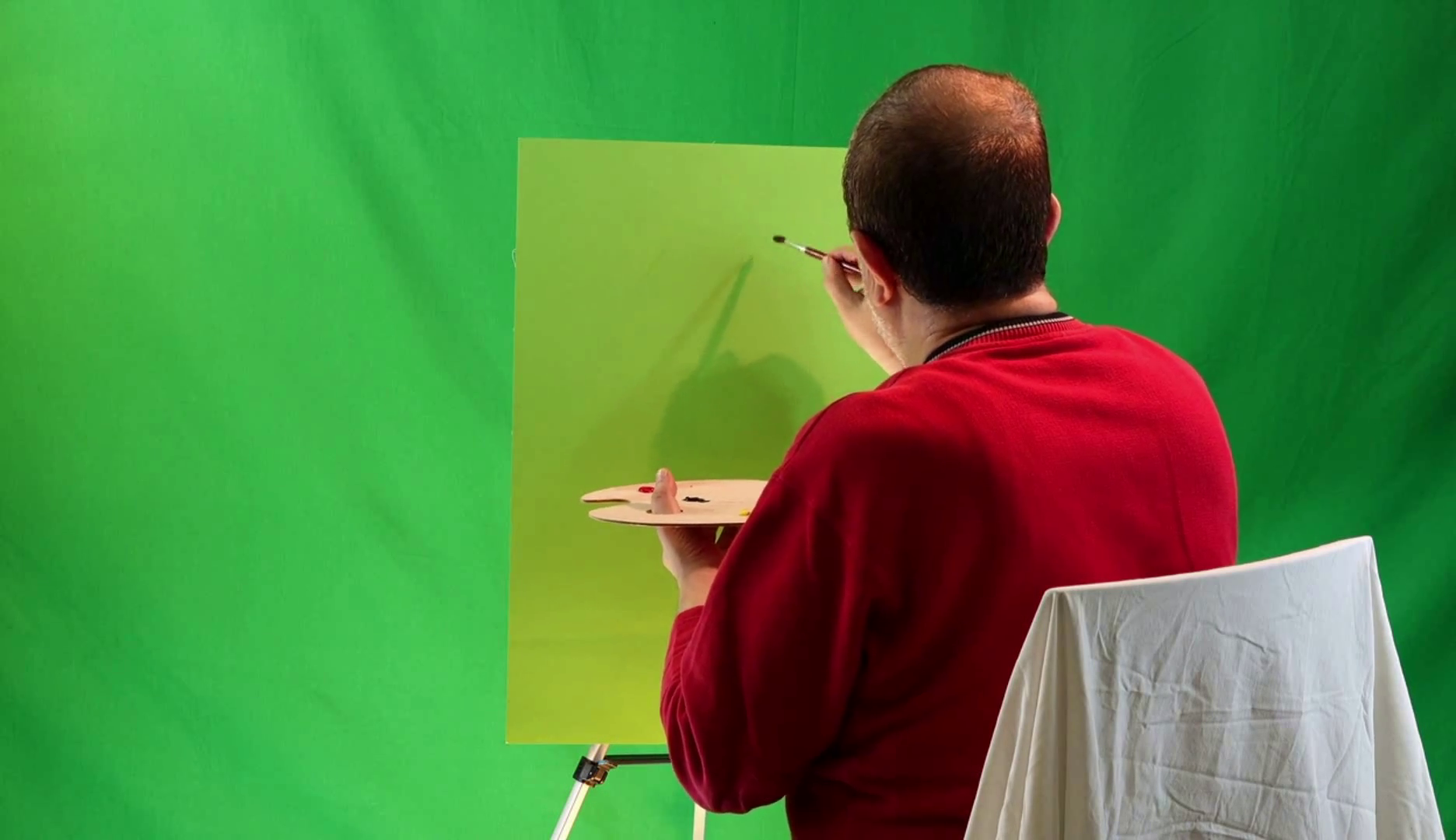} \\

		\raisebox{0.8\height}{\rotatebox{90}{\scriptsize Result}} &
		\includegraphics[width=\moreresultwidth]{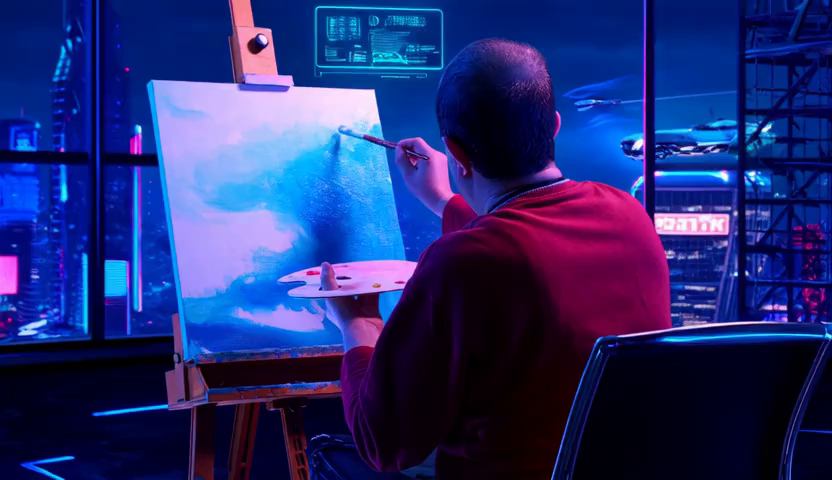} &
		\includegraphics[width=\moreresultwidth]{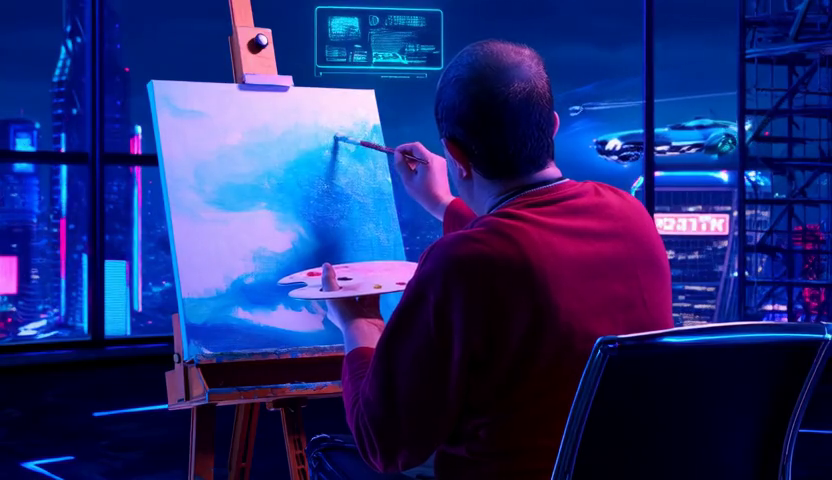} &
		\includegraphics[width=\moreresultwidth]{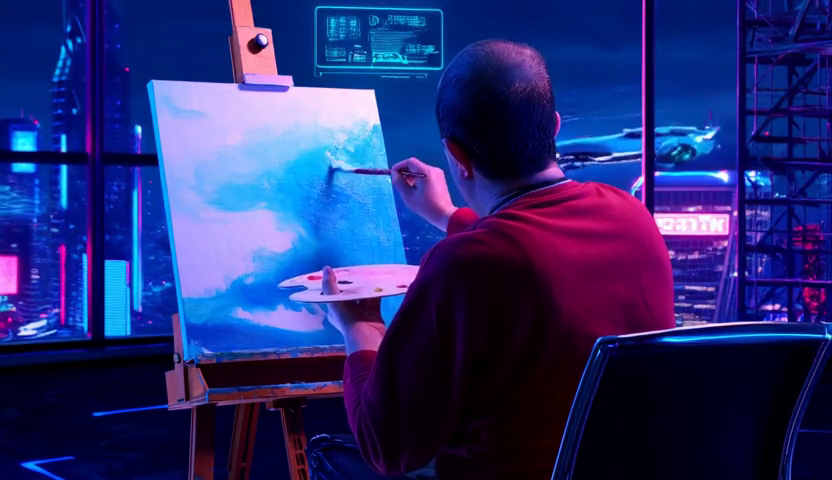} &
		\includegraphics[width=\moreresultwidth]{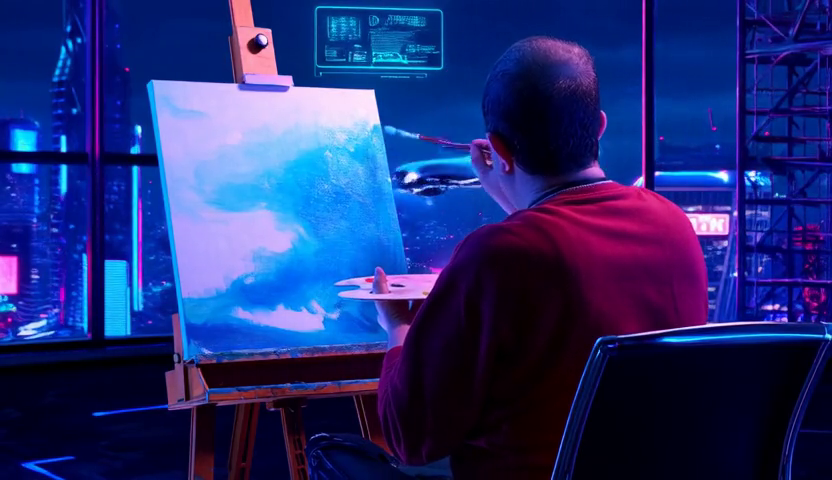} \\
		\multicolumn{5}{p{0.94\textwidth}}{\small Prompt: \textit{A man in a red long-sleeved outfit paints carefully at an easel inside a sleek penthouse studio, holding a palette while overlooking a neon-lit cyberpunk city at night. Electric blue and magenta lights reflect across chrome and glass surfaces as flying vehicles move between towering skyscrapers outside. Holographic displays flicker softly, blending traditional artistry with futuristic technology.}} \\
		\raisebox{0.05\height}{\rotatebox{90}{\scriptsize Green Screen}} &
		\includegraphics[width=\moreresultwidth]{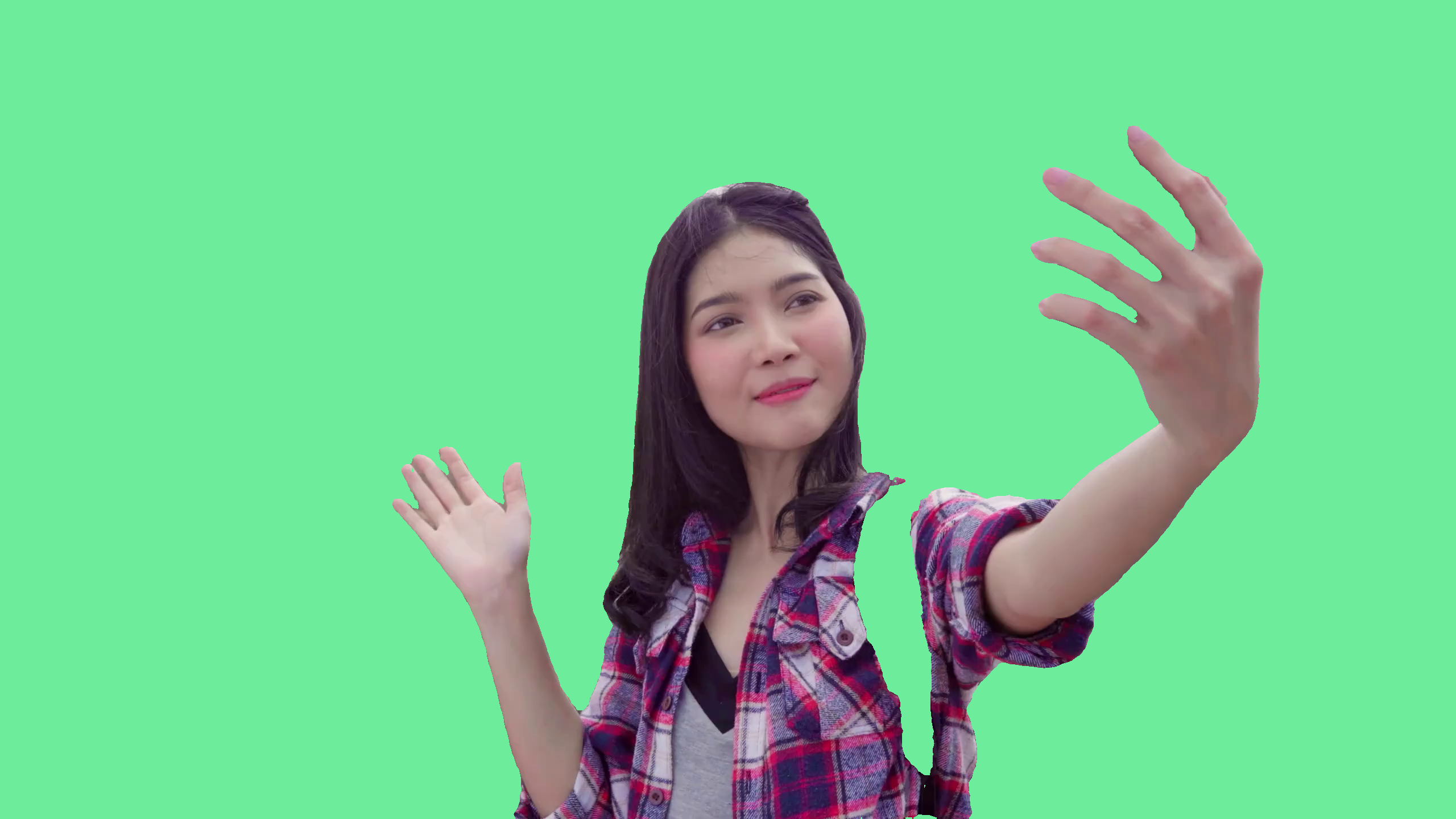} &
		\includegraphics[width=\moreresultwidth]{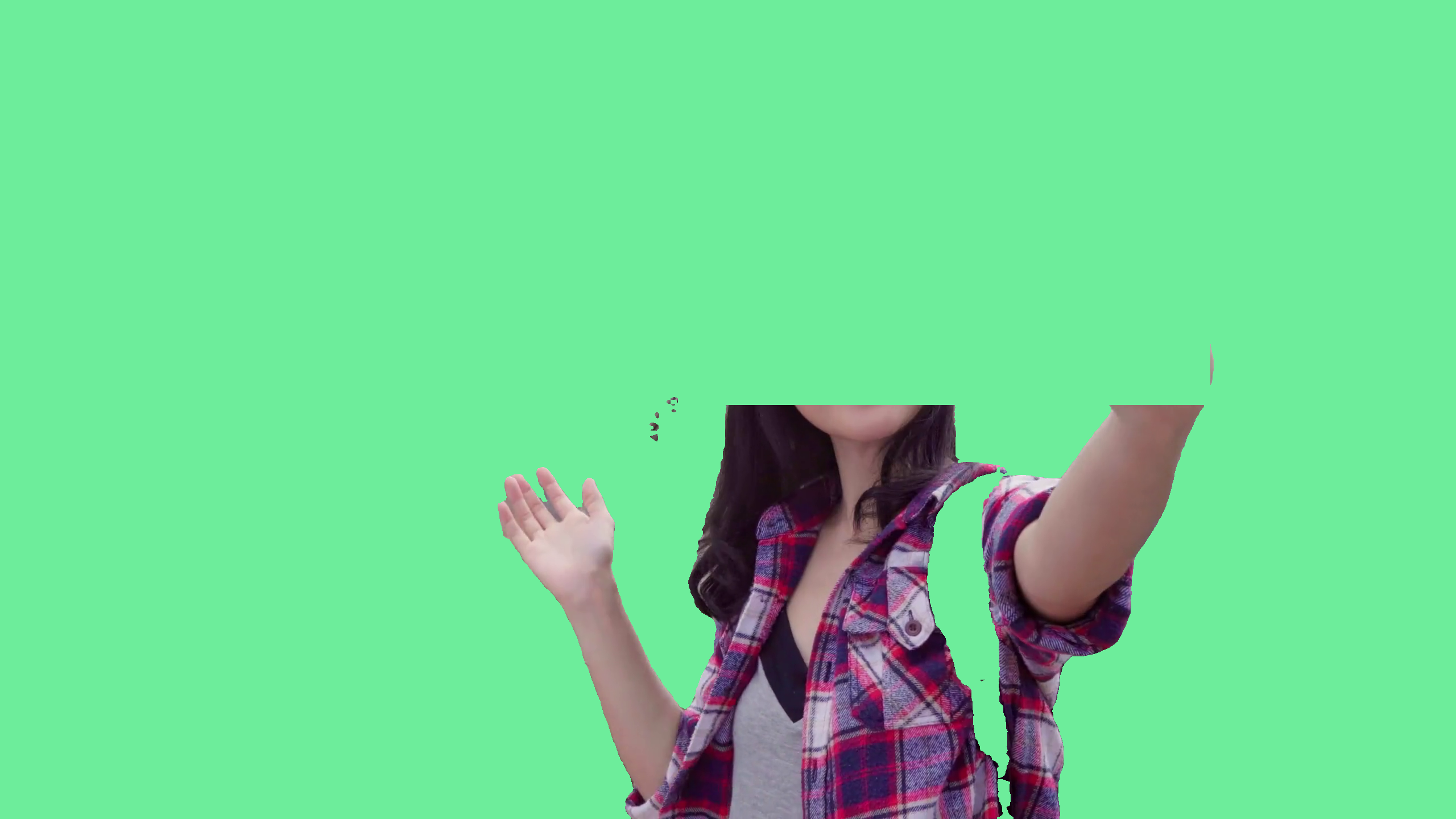} &
		\includegraphics[width=\moreresultwidth]{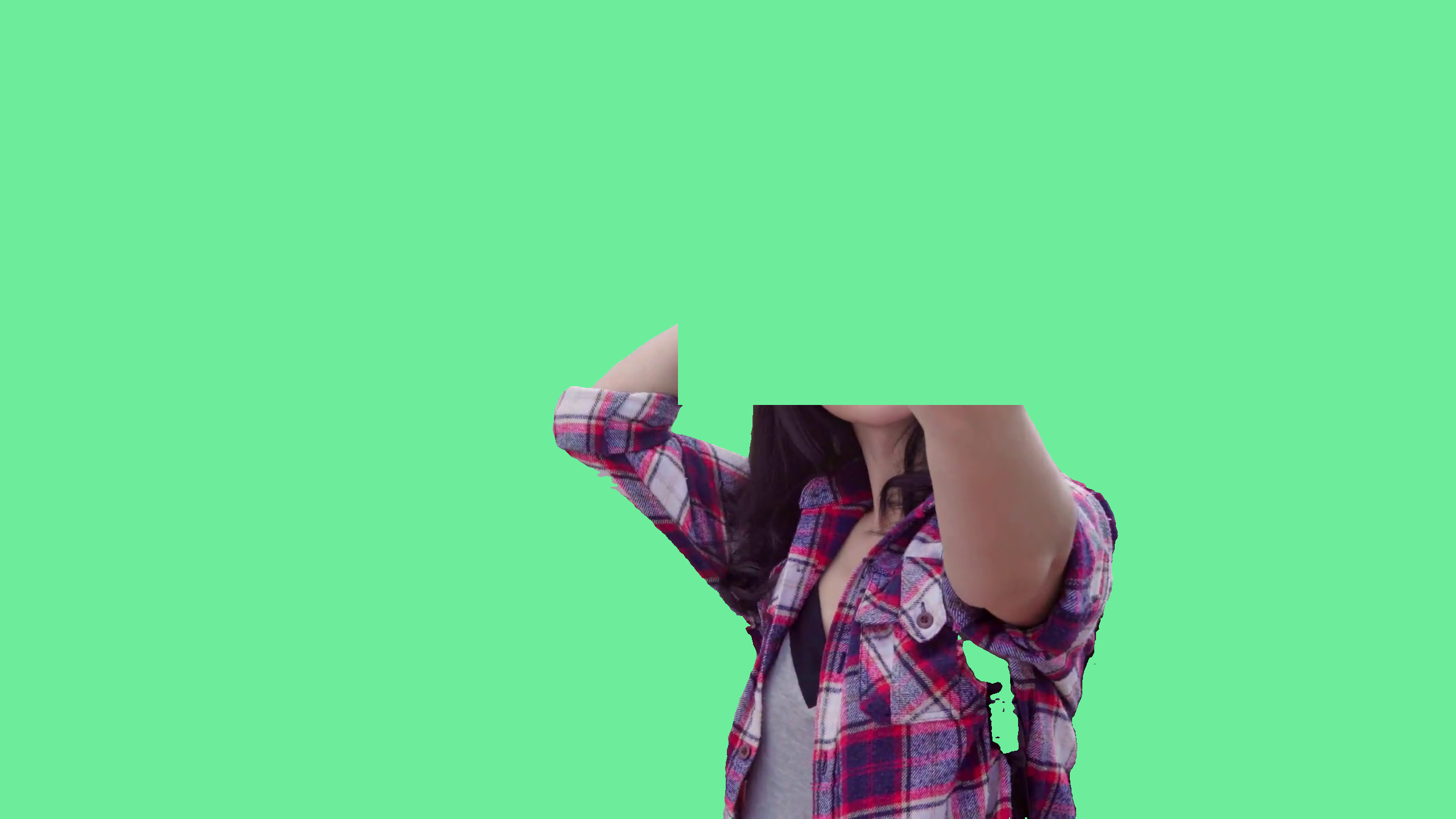} &
		\includegraphics[width=\moreresultwidth]{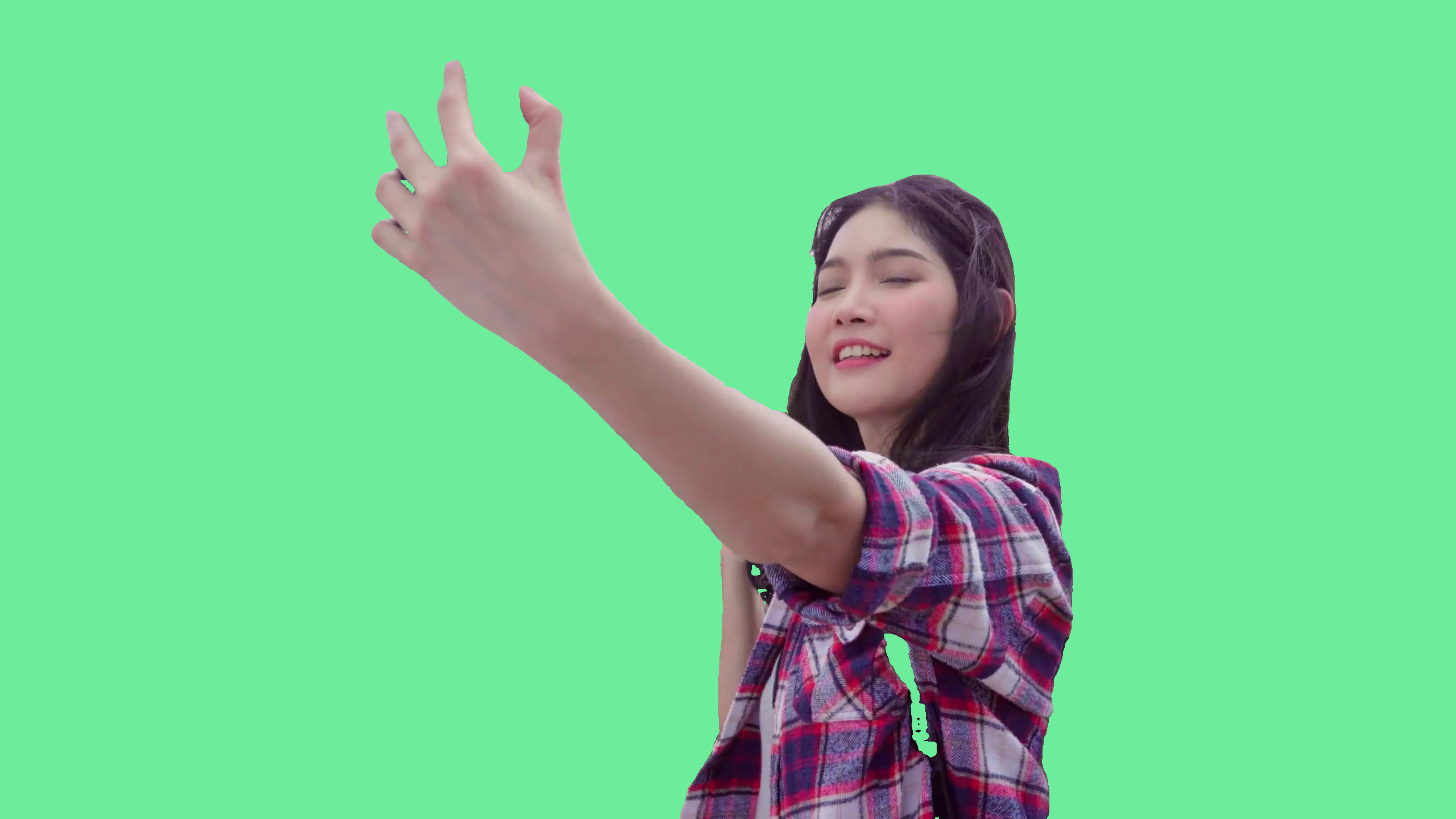} \\

		\raisebox{0.8\height}{\rotatebox{90}{\scriptsize Result}} &
		\includegraphics[width=\moreresultwidth]{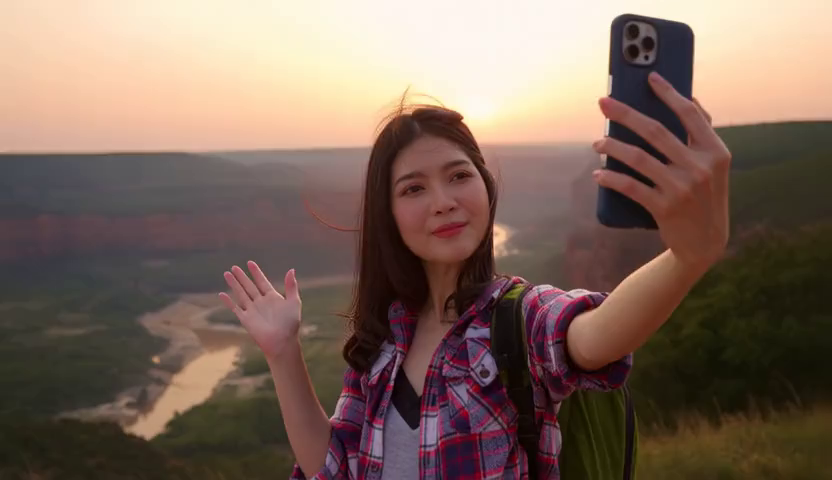} &
		\includegraphics[width=\moreresultwidth]{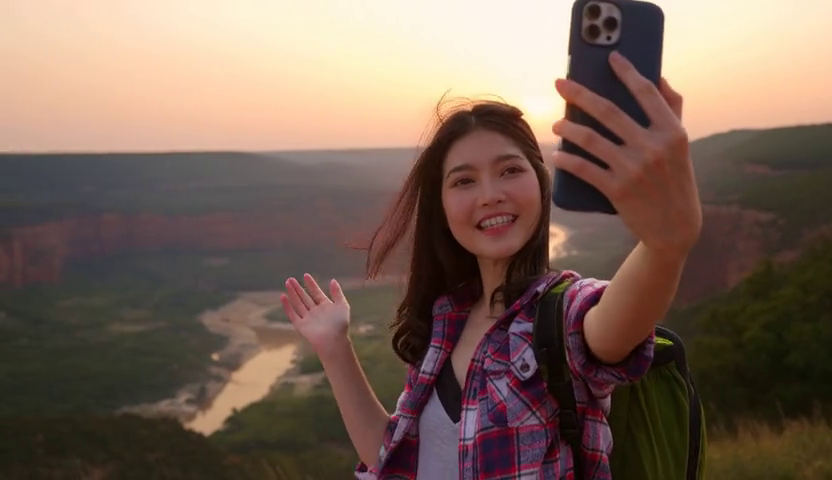} &
		\includegraphics[width=\moreresultwidth]{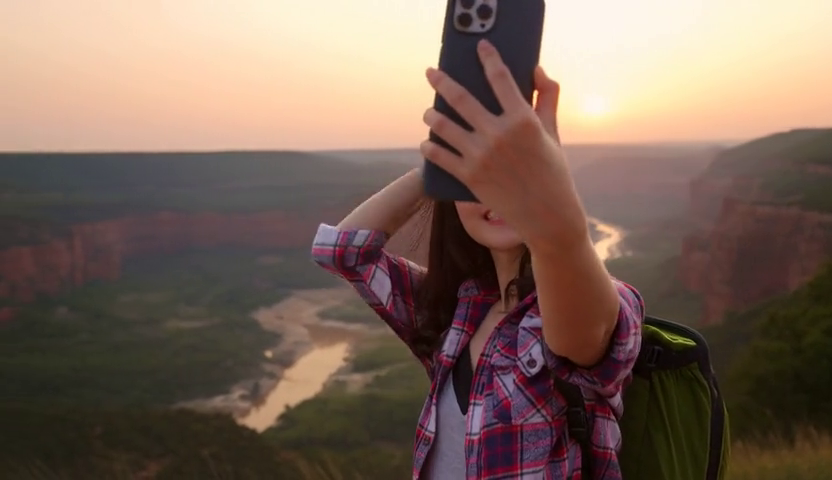} &
		\includegraphics[width=\moreresultwidth]{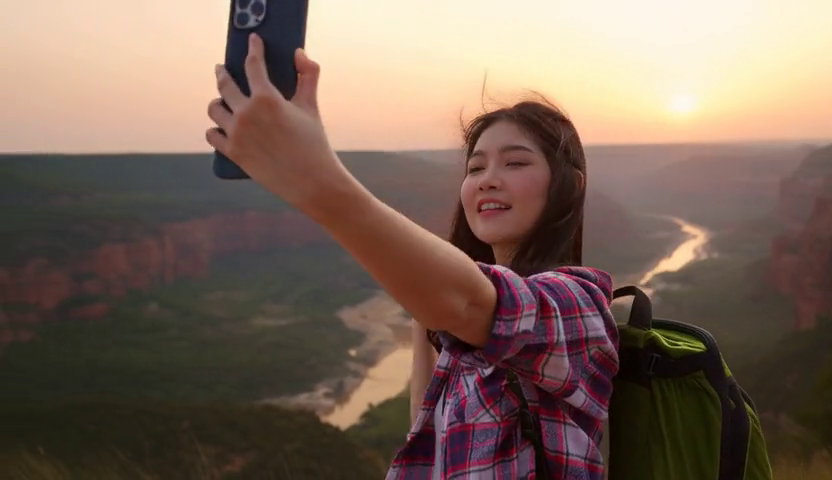} \\
		\multicolumn{5}{p{0.94\textwidth}}{\small Prompt: \textit{A cheerful young woman with long dark hair, wearing a plaid shirt and green backpack, records a selfie video with her smartphone in front of a majestic river valley at dusk. She waves, smiles, and adjusts her hair playfully while golden sunset light casts warm highlights and soft shadows across the scene. The atmosphere feels serene, cinematic, and natural.}} \\
		\raisebox{0.05\height}{\rotatebox{90}{\scriptsize Green Screen}} &
		\includegraphics[width=\moreresultwidth]{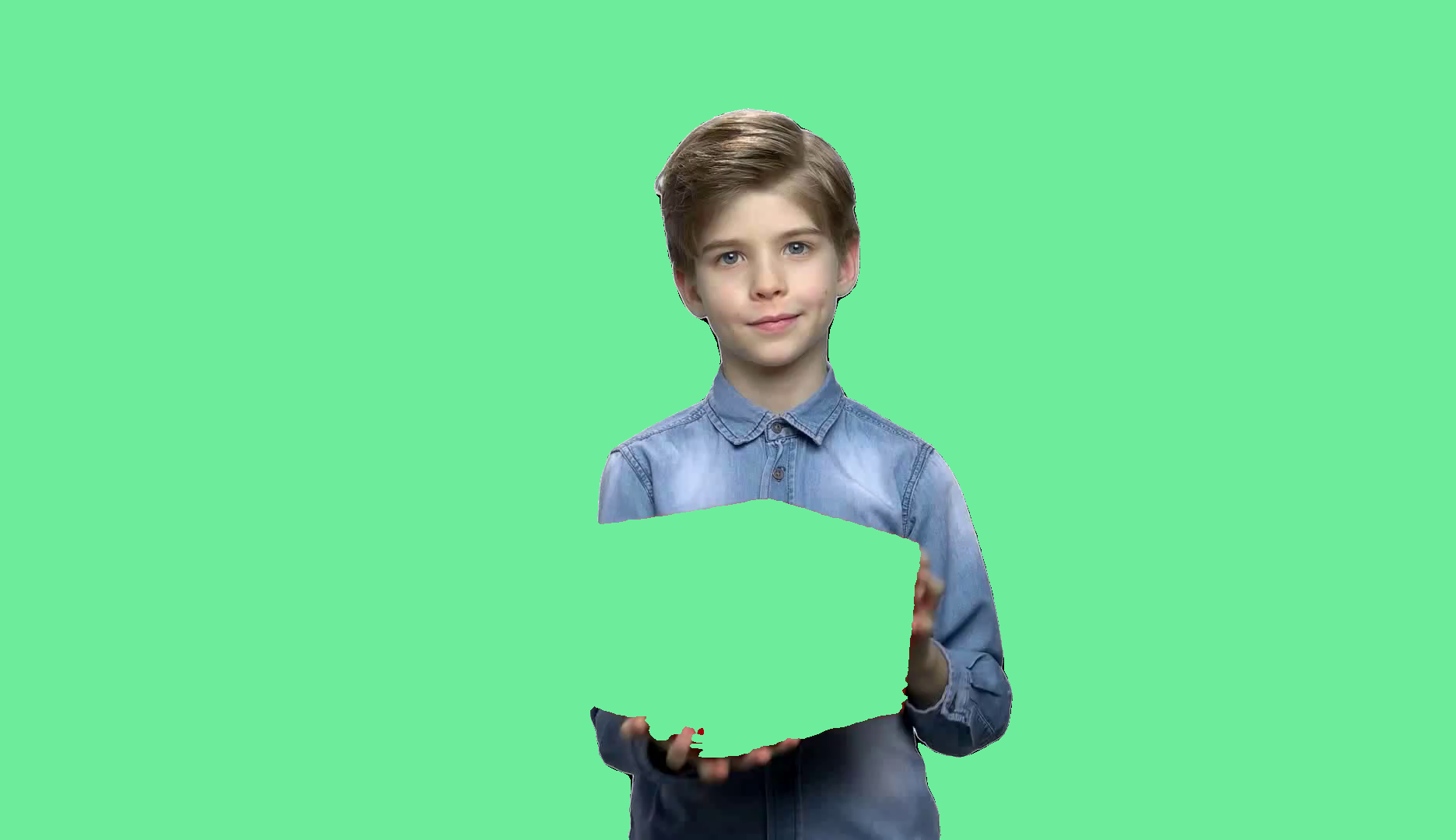} &
		\includegraphics[width=\moreresultwidth]{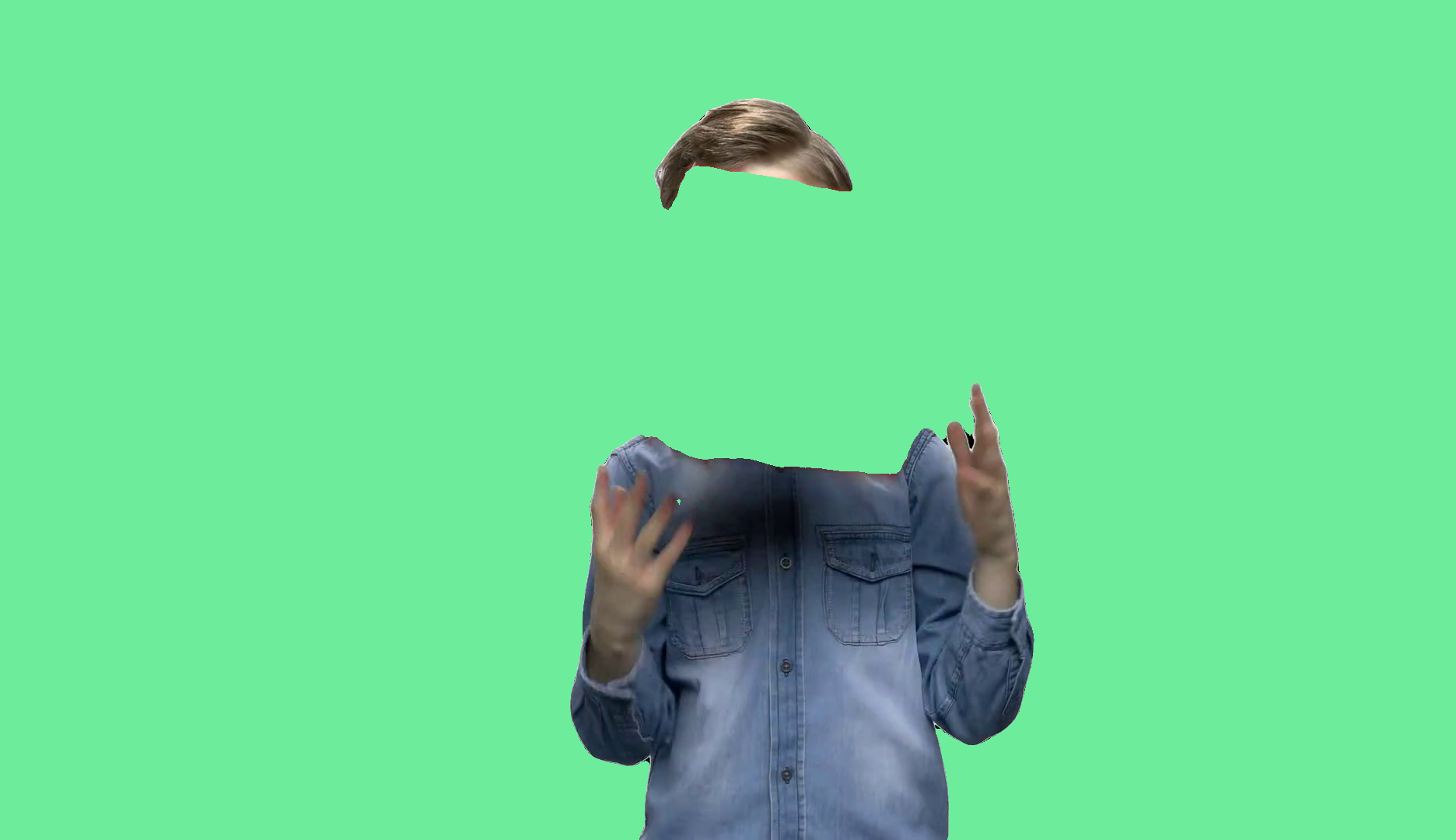} &
		\includegraphics[width=\moreresultwidth]{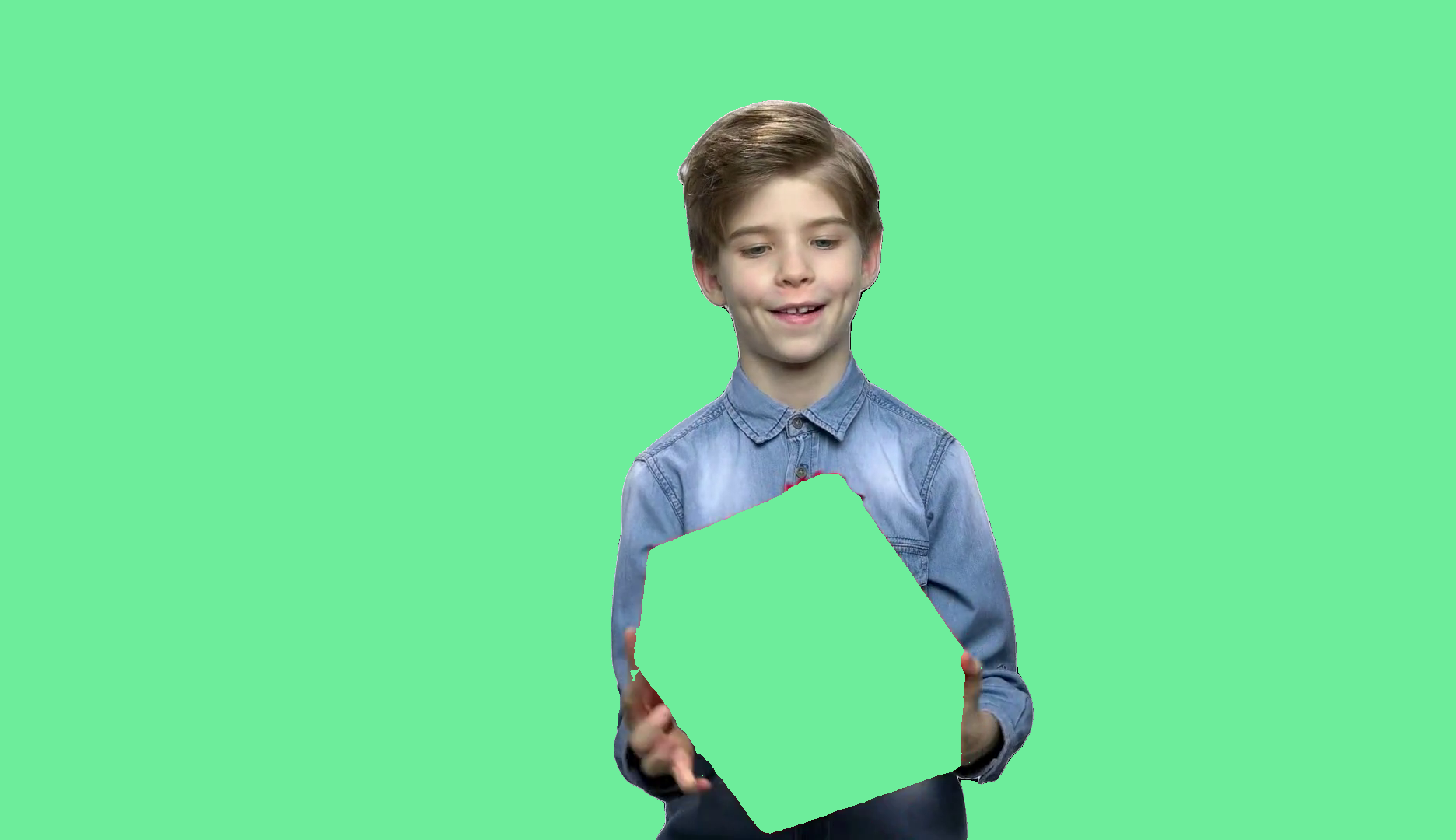} &
		\includegraphics[width=\moreresultwidth]{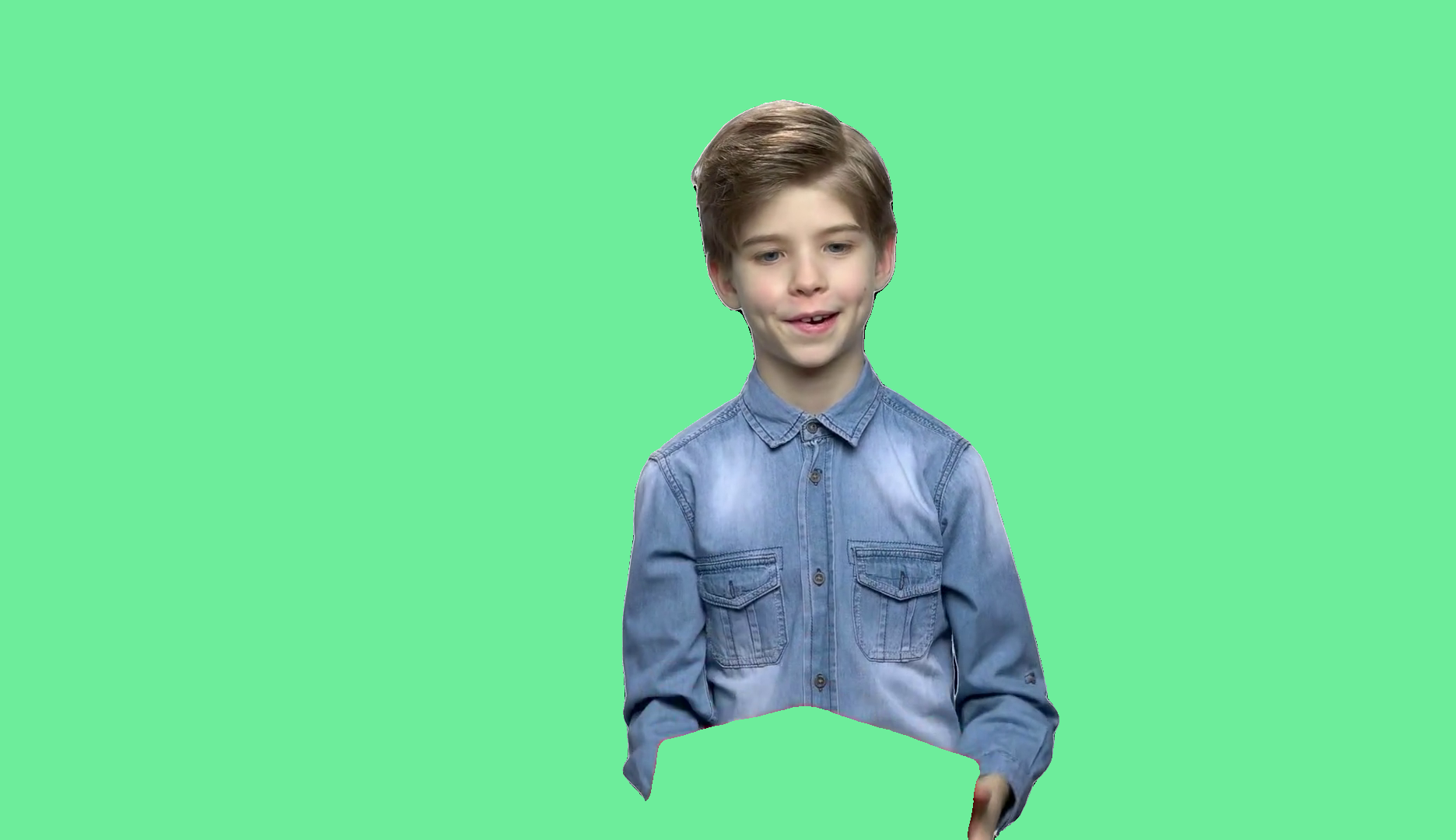} \\

		\raisebox{0.8\height}{\rotatebox{90}{\scriptsize Result}} &
		\includegraphics[width=\moreresultwidth]{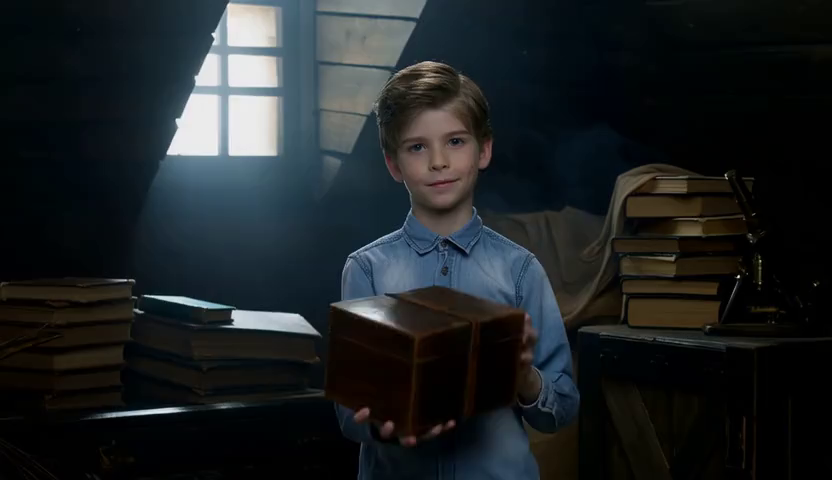} &
		\includegraphics[width=\moreresultwidth]{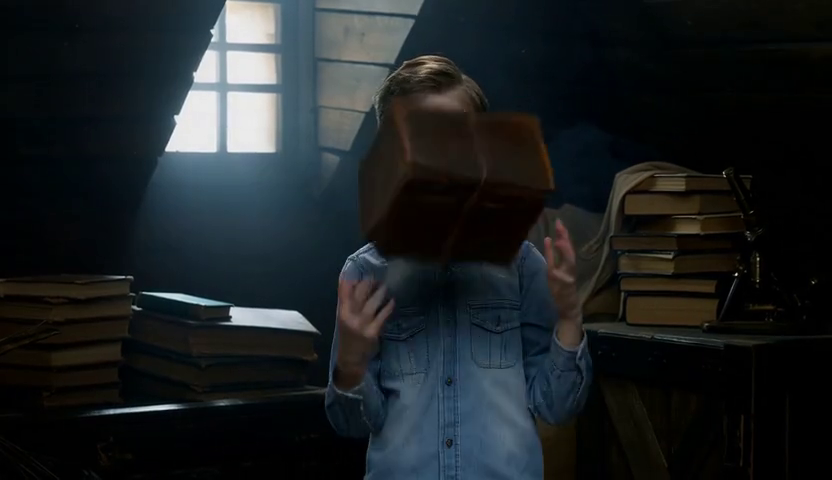} &
		\includegraphics[width=\moreresultwidth]{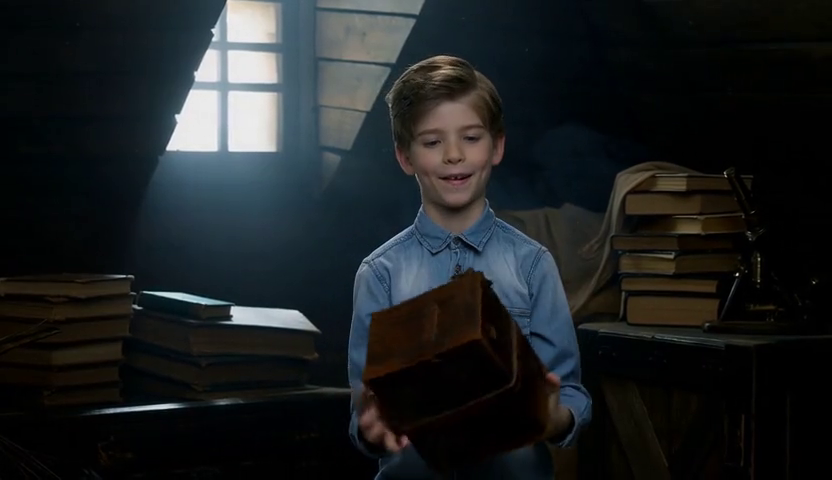} &
		\includegraphics[width=\moreresultwidth]{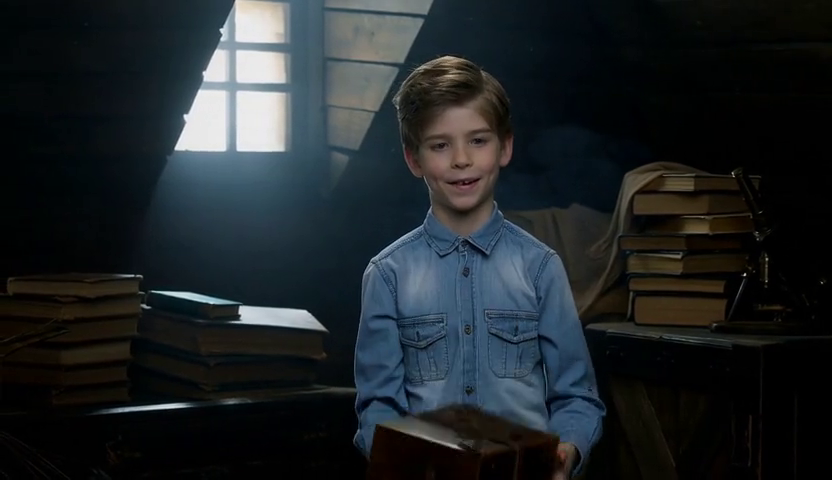} \\
		\multicolumn{5}{p{0.94\textwidth}}{\small Prompt: \textit{A curious young boy in a blue denim shirt tosses an antique wooden chest with rusted iron latches inside a dusty attic filled with old books and relics. Shafts of light stream through a cobwebbed window, illuminating floating dust and mysterious surroundings. Moody cinematic lighting and deep shadows create a sense of wonder and historical discovery.}} \\
	\end{tabular}}
	\caption{\textbf{Additional qualitative results.} We show more examples of our method. Our approach generates realistic backgrounds while maintaining physical interaction consistency and lighting harmonization.}
	
	\label{fig:more_result}
\end{figure*}

\newcommand{\moreresultwidthnew}{0.155\linewidth}

\newcommand{\moreresultmasklabel}{\scriptsize Tri-mask}
\begin{figure*}[p]
	\centering
	\setlength{\tabcolsep}{0.5pt}
	\renewcommand{\arraystretch}{0.95}
	\scalebox{1}{\begin{tabular}{c@{\hspace{2pt}}cccccc}
		\raisebox{0.05\height}{\rotatebox{90}{\scriptsize Input Video}} &
		\includegraphics[width=\moreresultwidthnew]{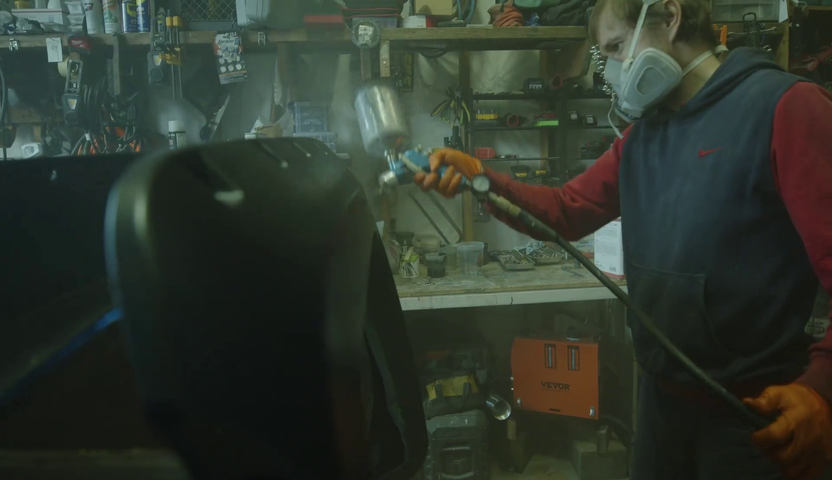} &
		\includegraphics[width=\moreresultwidthnew]{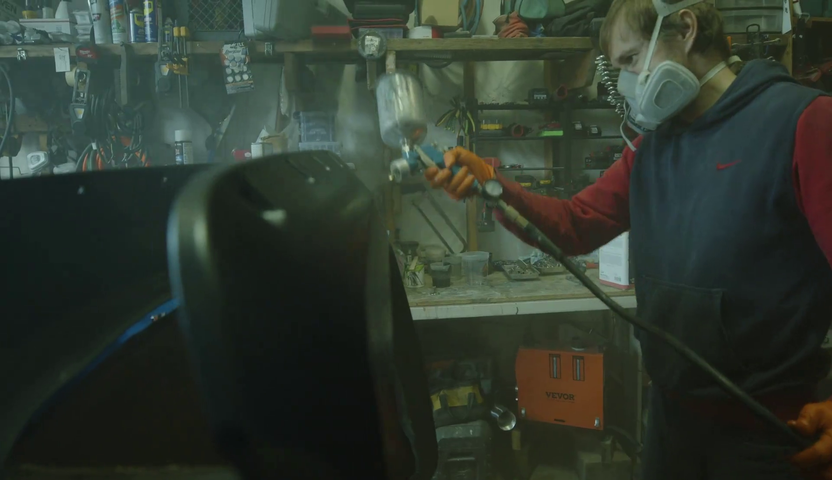} &
		\includegraphics[width=\moreresultwidthnew]{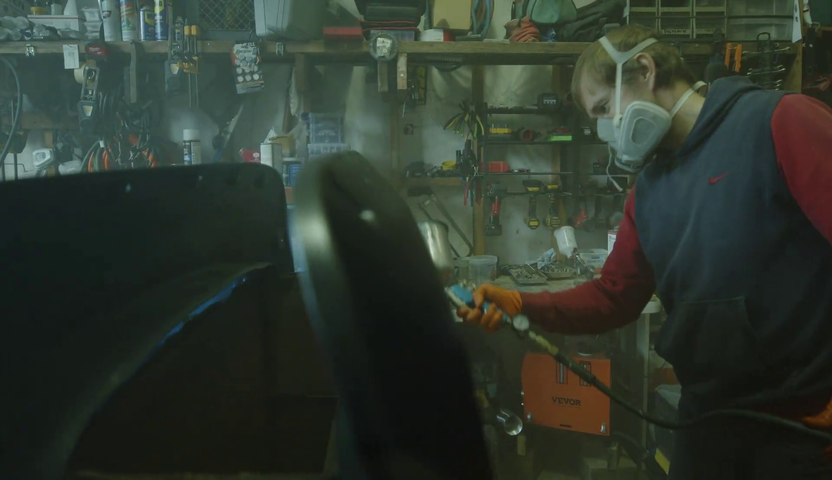} &
		\includegraphics[width=\moreresultwidthnew]{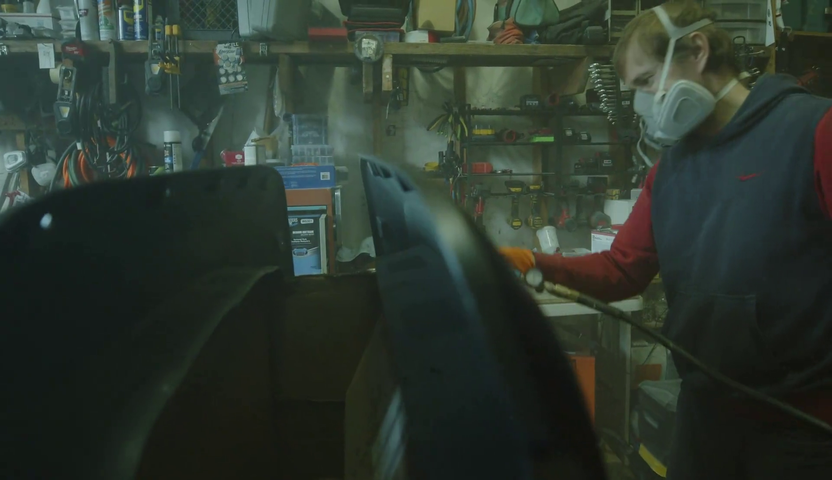} &
		\includegraphics[width=\moreresultwidthnew]{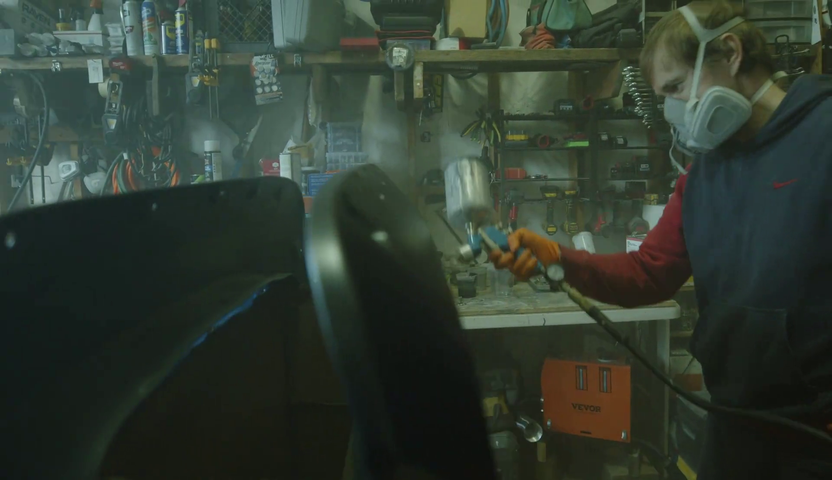} &
		\includegraphics[width=\moreresultwidthnew]{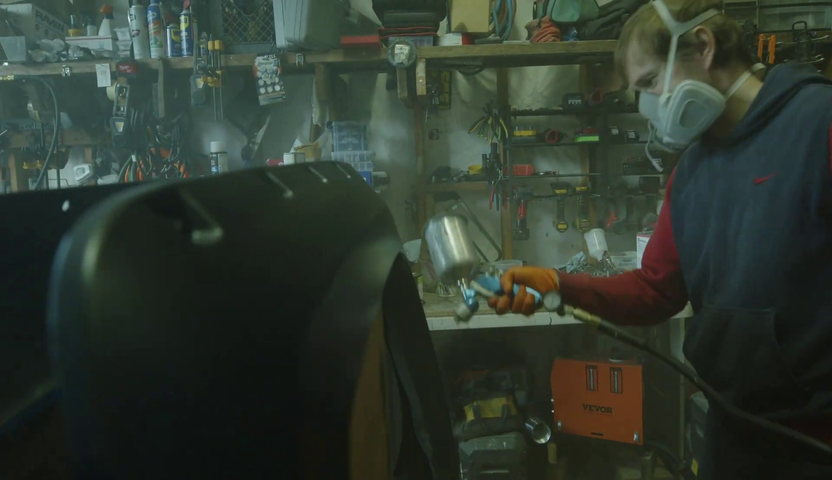} \\
		\raisebox{0.05\height}{\rotatebox{90}{\moreresultmasklabel}} &
		\includegraphics[width=\moreresultwidthnew]{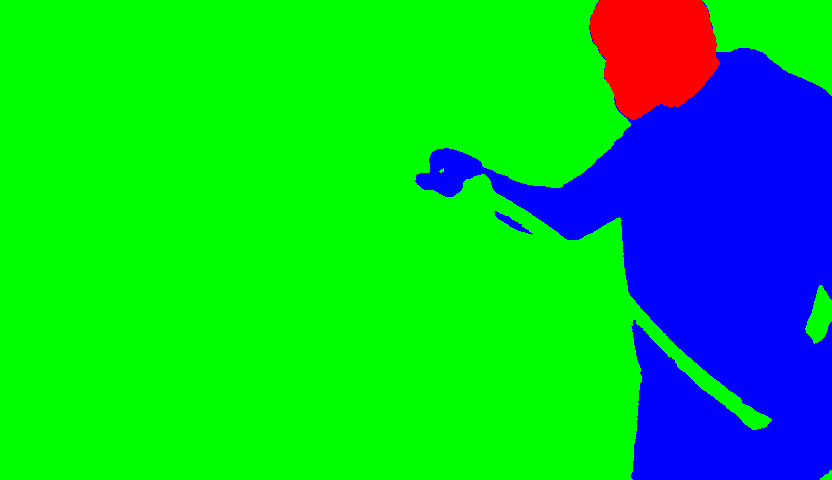} &
		\includegraphics[width=\moreresultwidthnew]{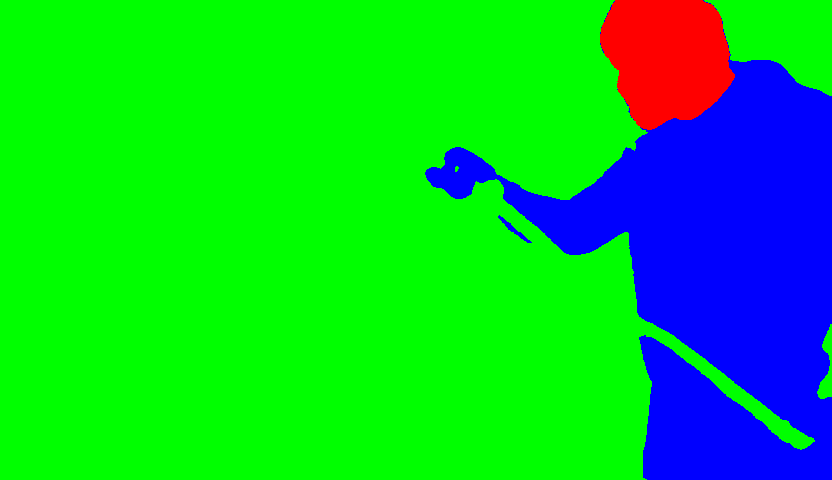} &
		\includegraphics[width=\moreresultwidthnew]{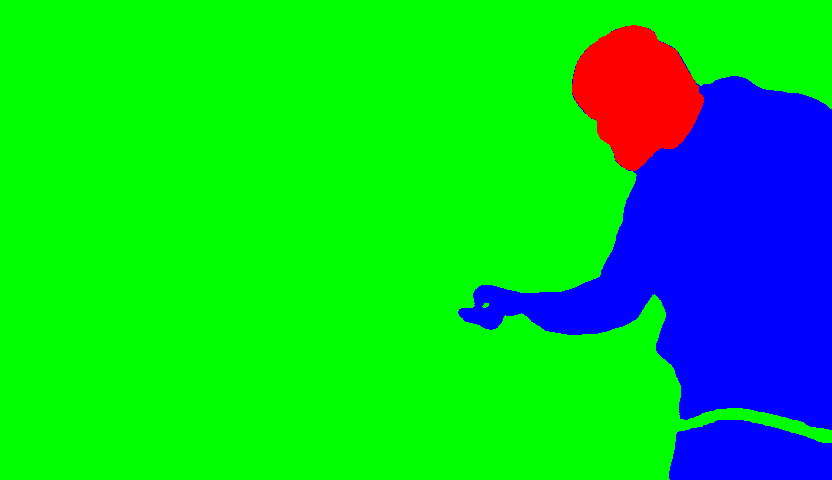} &
		\includegraphics[width=\moreresultwidthnew]{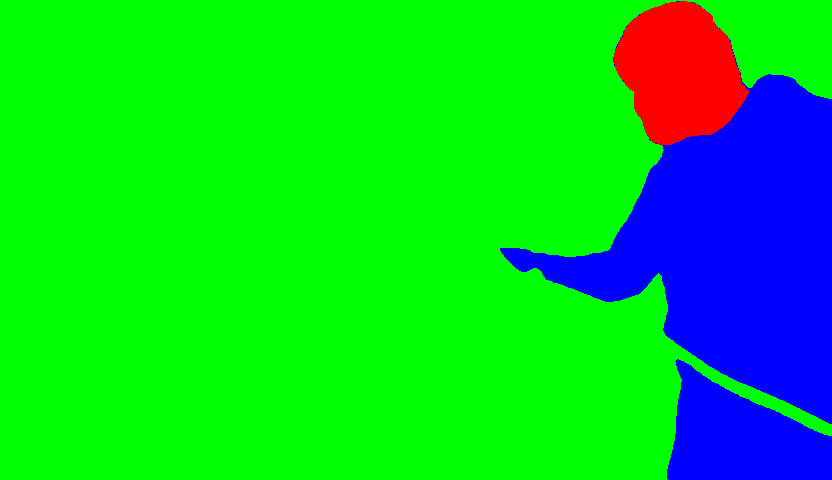} &
		\includegraphics[width=\moreresultwidthnew]{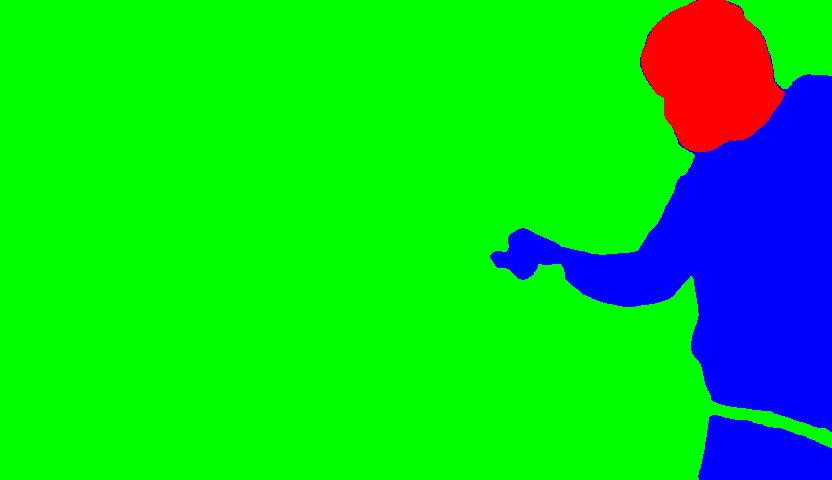} &
		\includegraphics[width=\moreresultwidthnew]{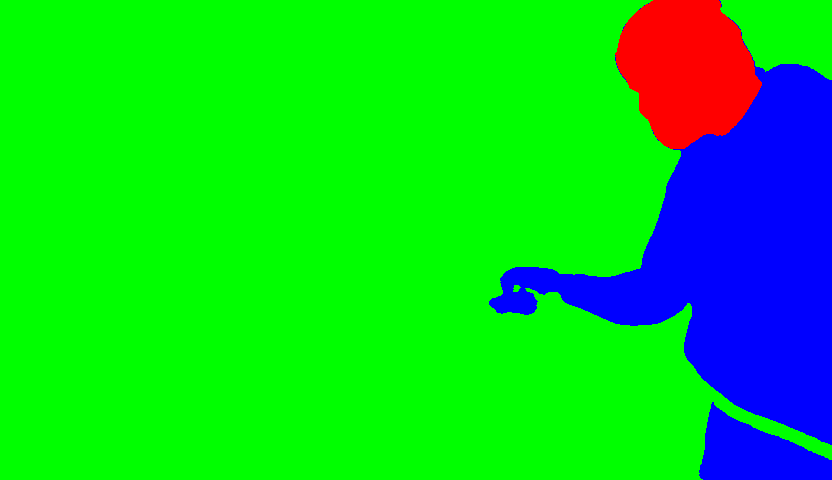} \\
		\raisebox{0.05\height}{\rotatebox{90}{\scriptsize Result}} &
		\includegraphics[width=\moreresultwidthnew]{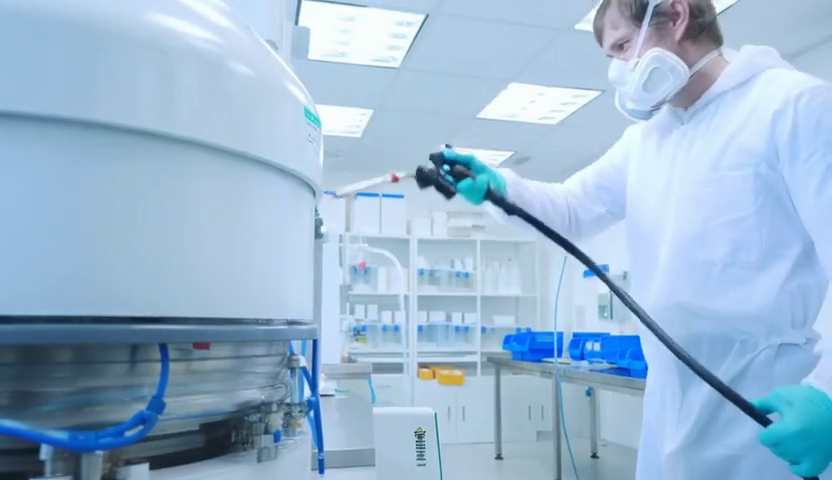} &
		\includegraphics[width=\moreresultwidthnew]{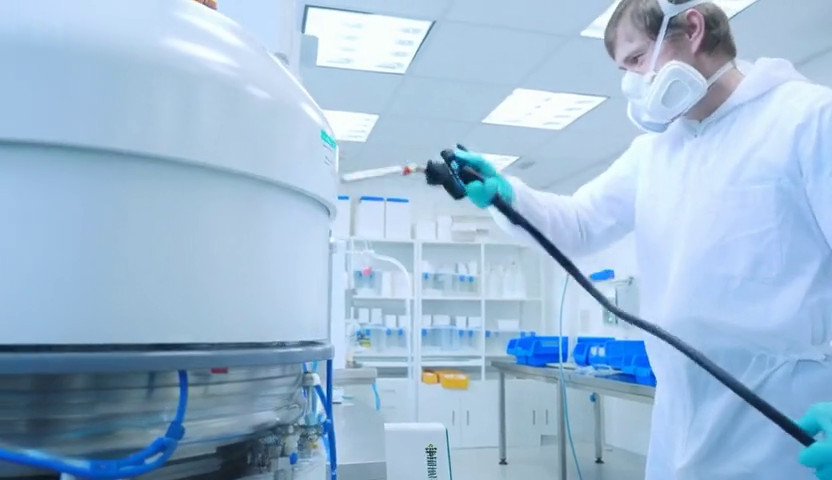} &
		\includegraphics[width=\moreresultwidthnew]{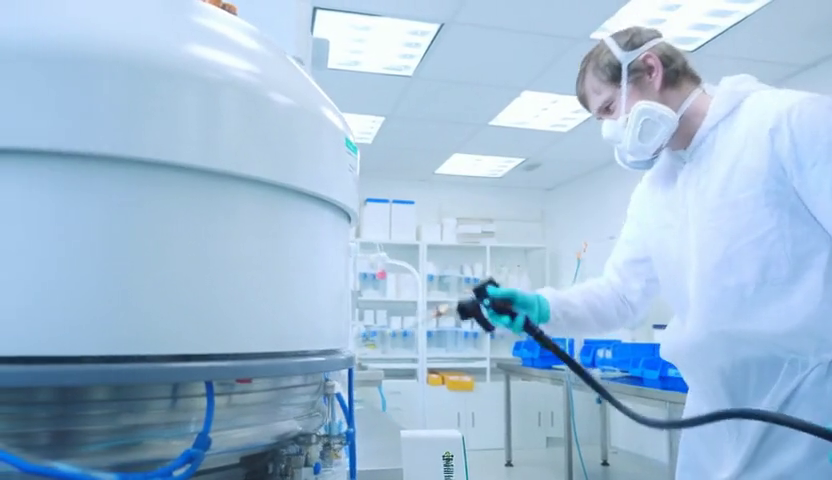} &
		\includegraphics[width=\moreresultwidthnew]{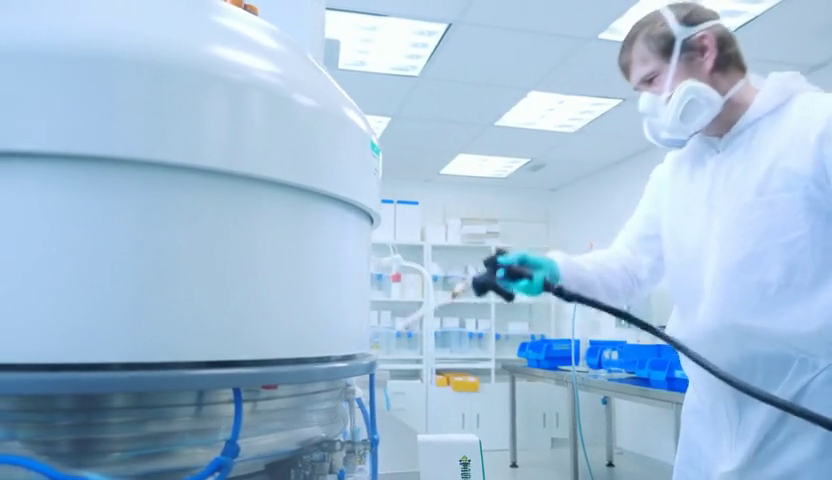} &
		\includegraphics[width=\moreresultwidthnew]{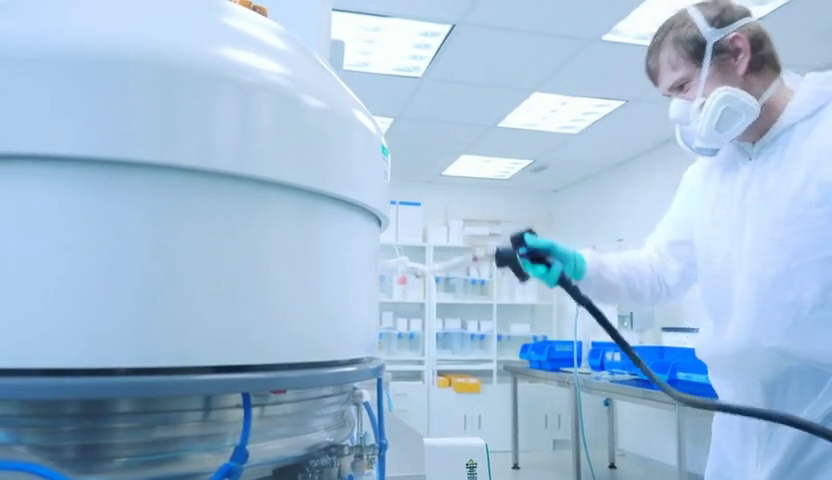} &
		\includegraphics[width=\moreresultwidthnew]{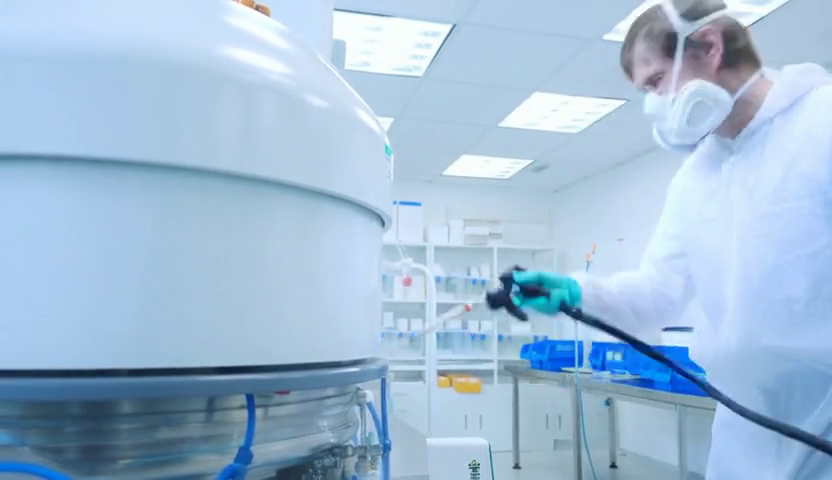} \\
		\multicolumn{7}{p{0.94\textwidth}}{\small Prompt: \textit{A male laboratory technician in protective gear sanitizes a high-tech metallic bioreactor capsule inside a brightly lit biological laboratory, with clean fluorescent lighting and misty disinfectant haze.}} \\
		\raisebox{0.05\height}{\rotatebox{90}{\scriptsize Input Video}} &
		\includegraphics[width=\moreresultwidthnew]{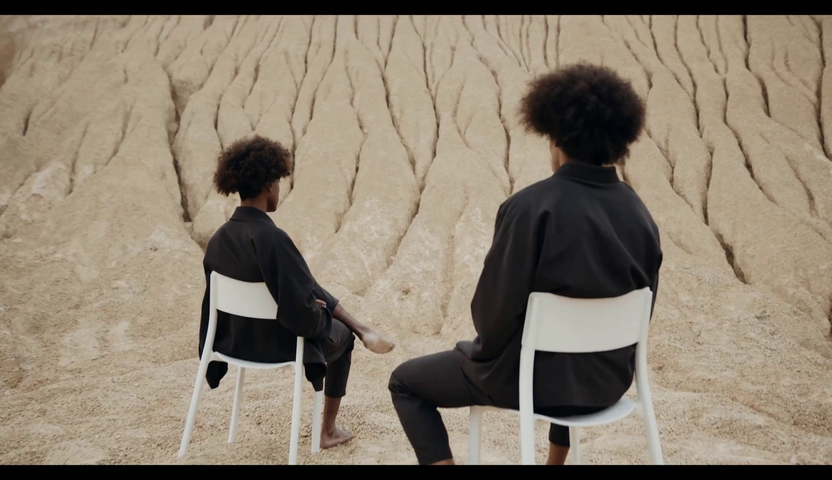} &
		\includegraphics[width=\moreresultwidthnew]{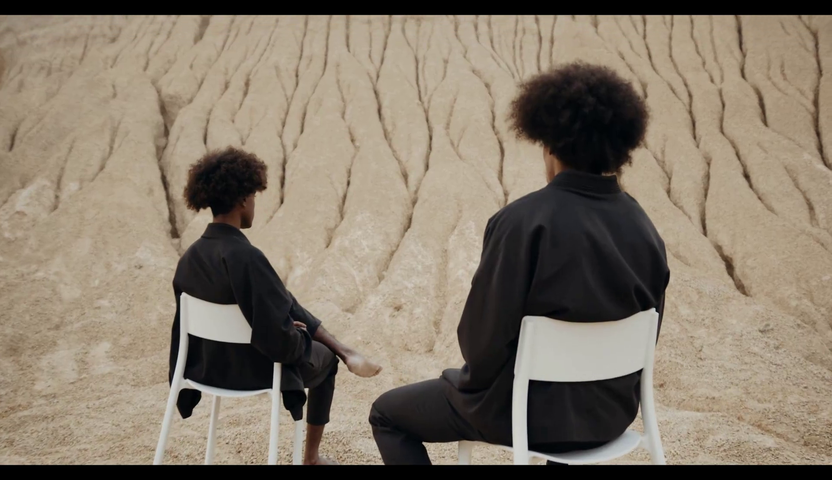} &
		\includegraphics[width=\moreresultwidthnew]{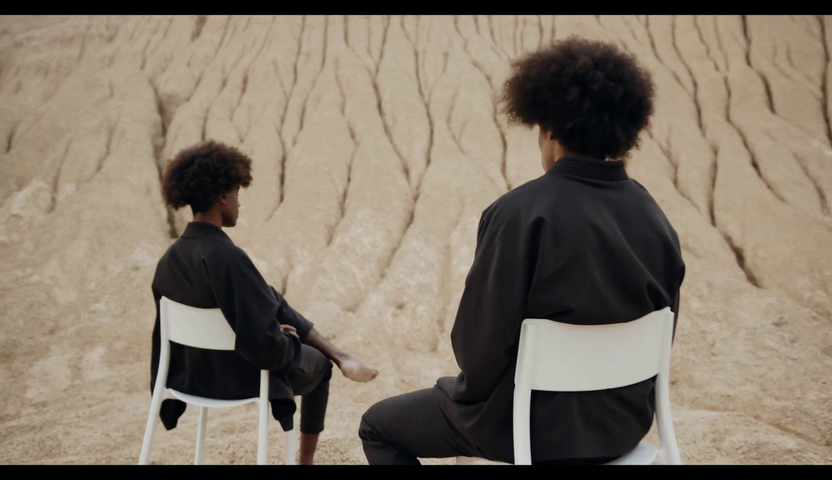} &
		\includegraphics[width=\moreresultwidthnew]{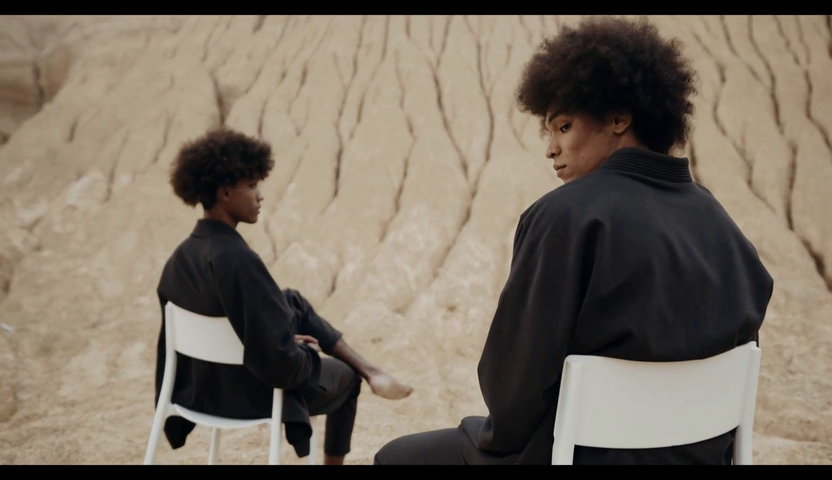} &
		\includegraphics[width=\moreresultwidthnew]{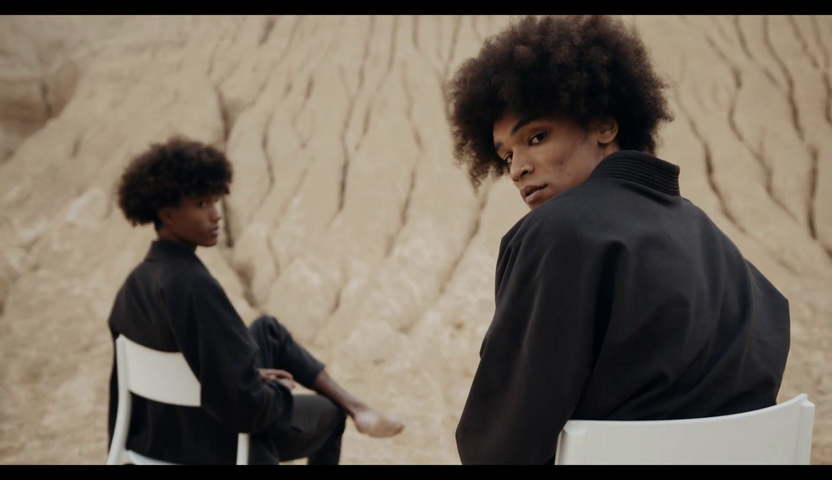} &
		\includegraphics[width=\moreresultwidthnew]{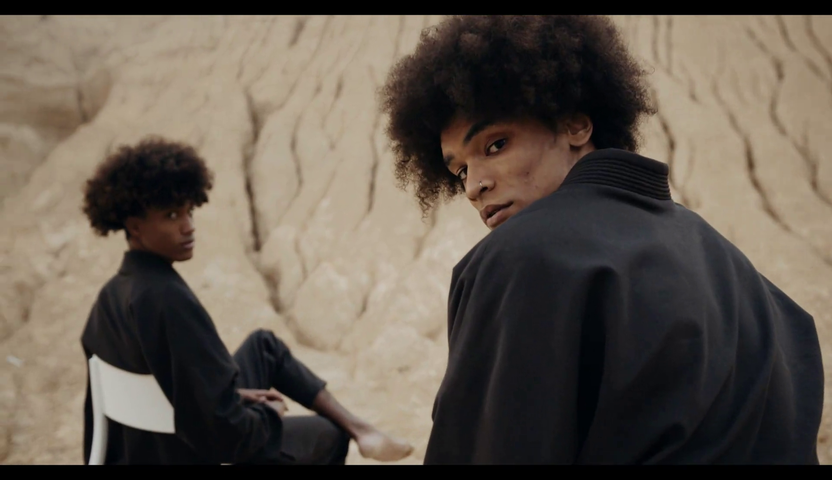} \\
		\raisebox{0.05\height}{\rotatebox{90}{\moreresultmasklabel}} &
		\includegraphics[width=\moreresultwidthnew]{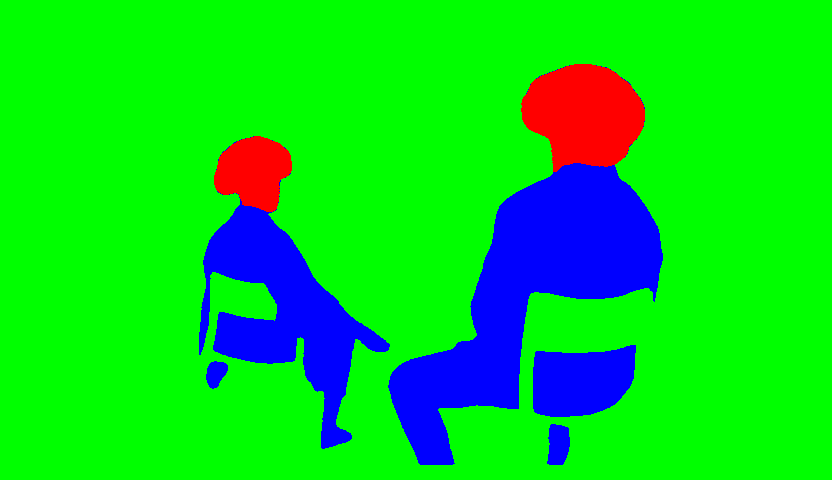} &
		\includegraphics[width=\moreresultwidthnew]{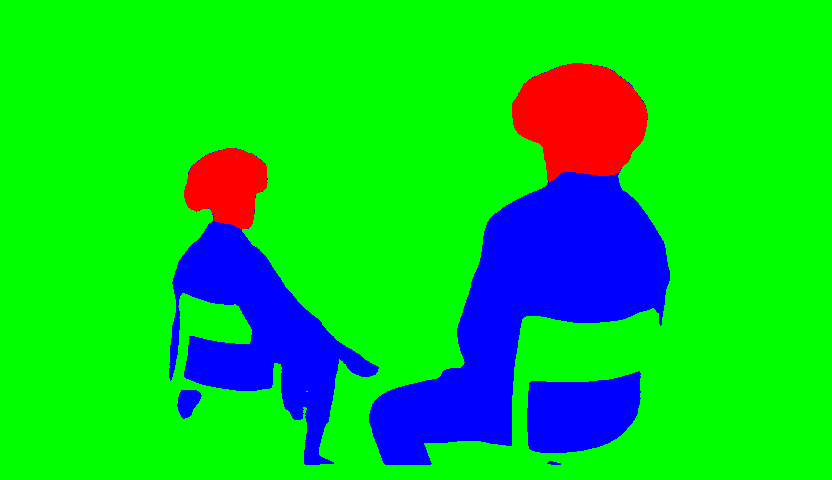} &
		\includegraphics[width=\moreresultwidthnew]{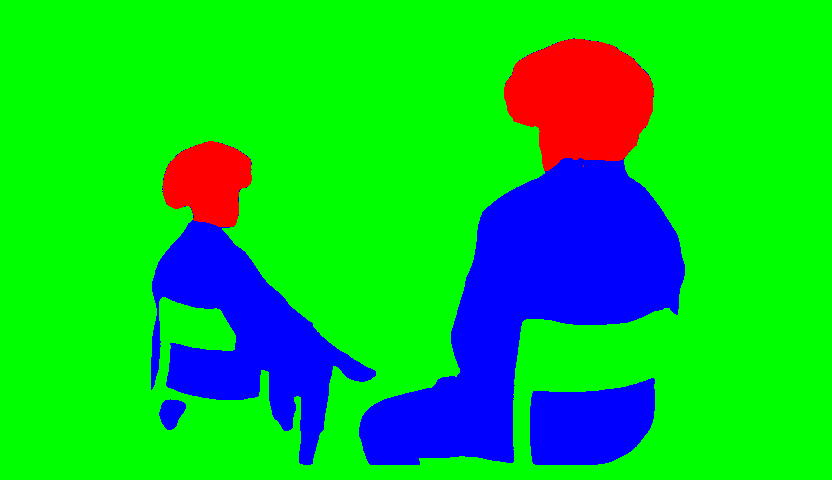} &
		\includegraphics[width=\moreresultwidthnew]{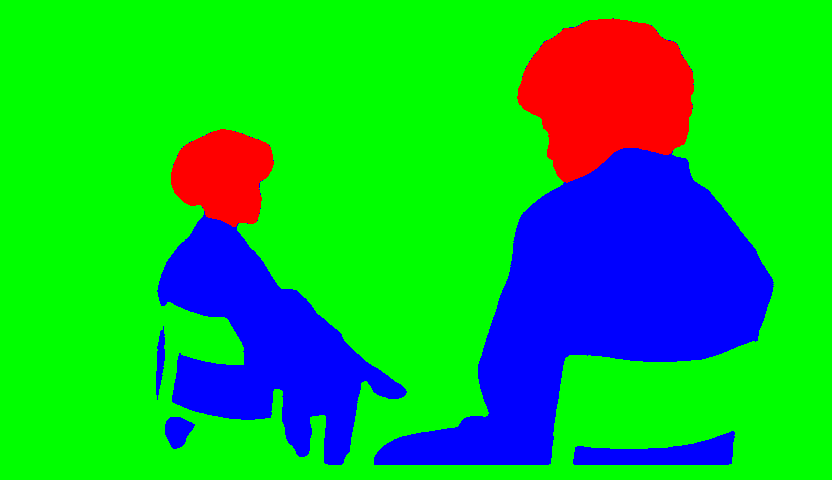} &
		\includegraphics[width=\moreresultwidthnew]{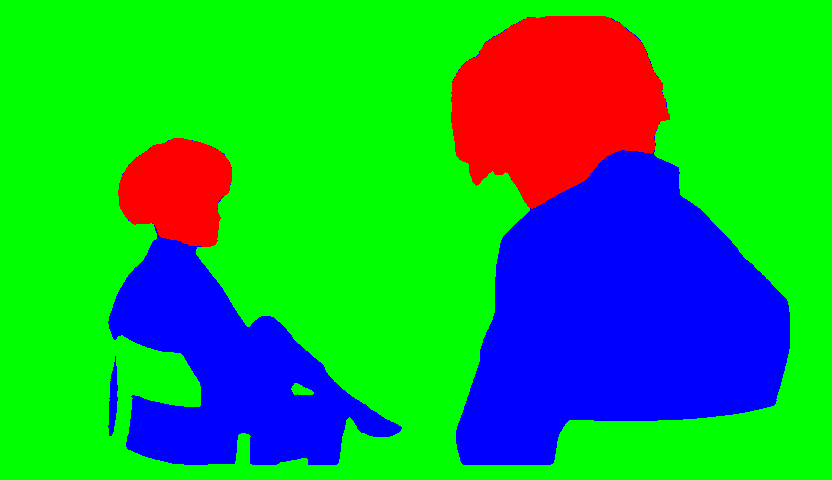} &
		\includegraphics[width=\moreresultwidthnew]{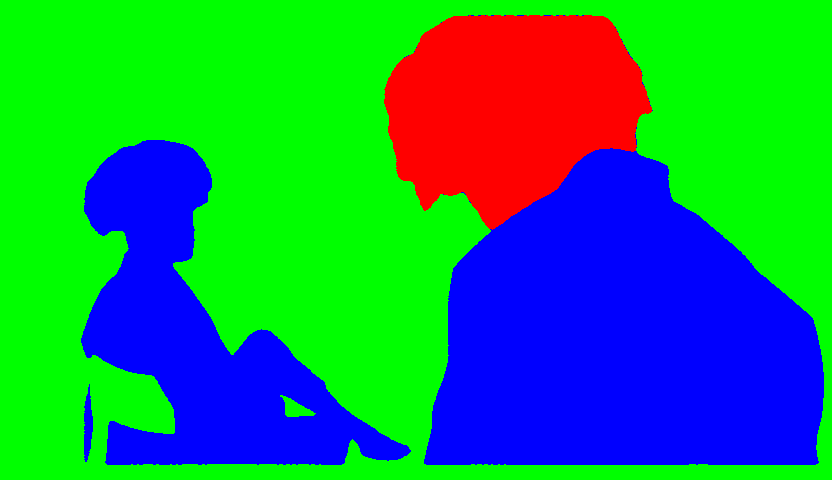} \\
		\raisebox{0.05\height}{\rotatebox{90}{\scriptsize Result}} &
		\includegraphics[width=\moreresultwidthnew]{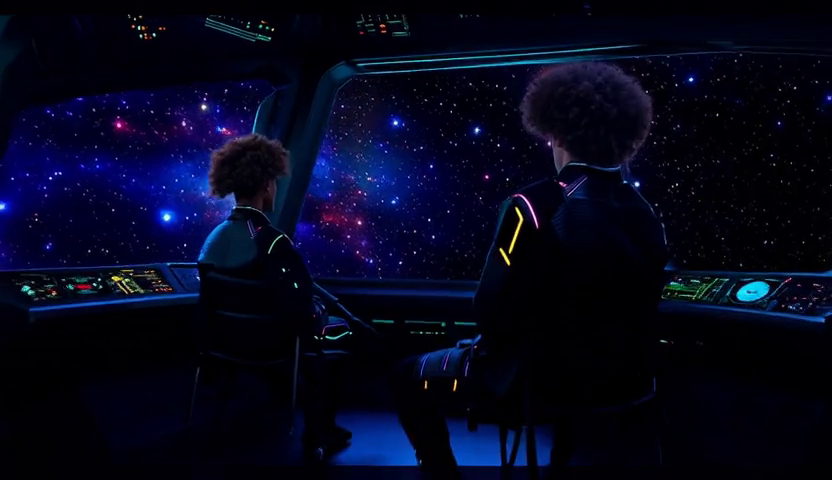} &
		\includegraphics[width=\moreresultwidthnew]{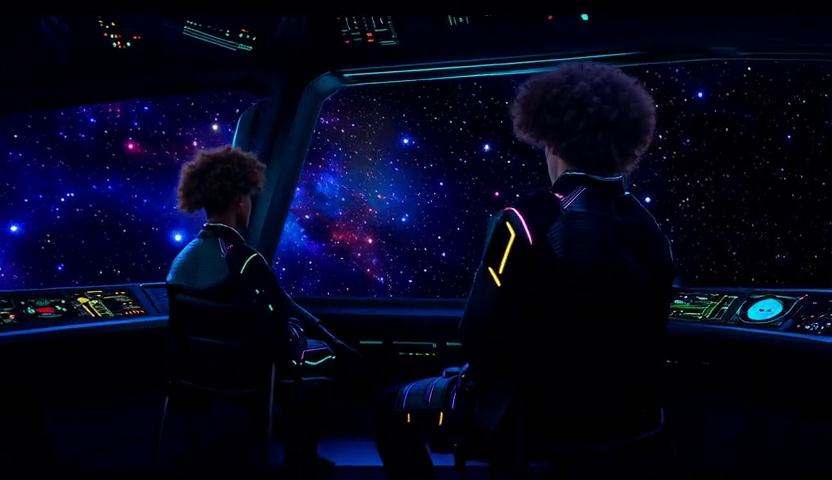} &
		\includegraphics[width=\moreresultwidthnew]{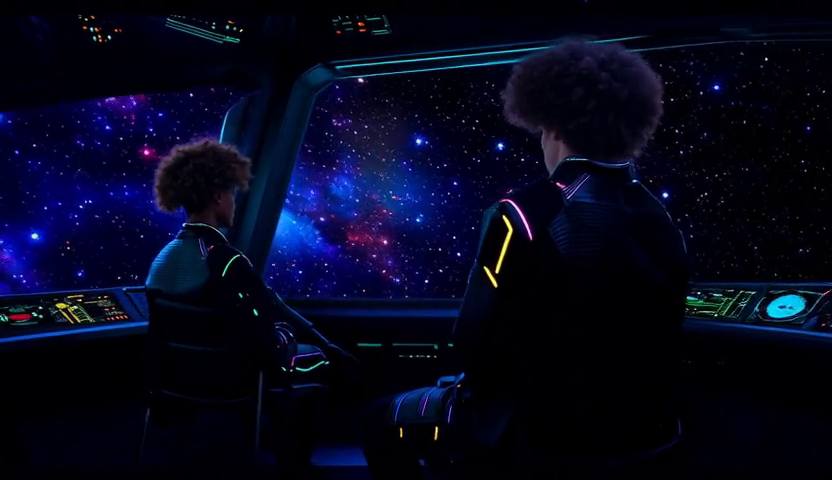} &
		\includegraphics[width=\moreresultwidthnew]{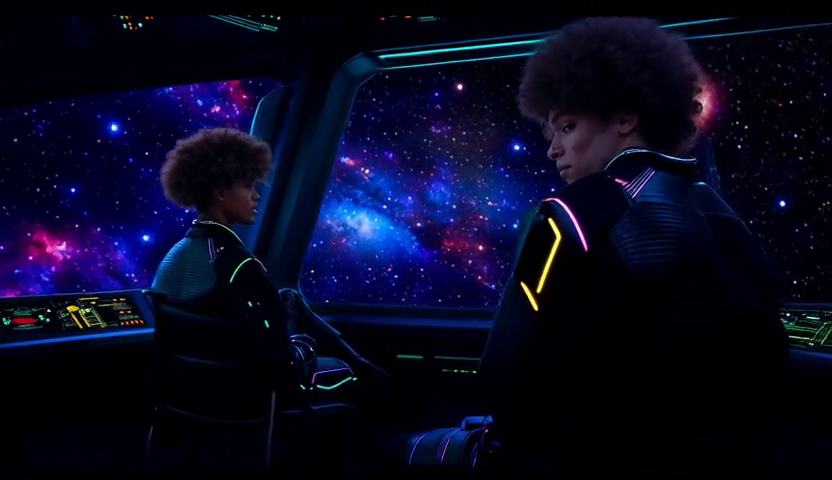} &
		\includegraphics[width=\moreresultwidthnew]{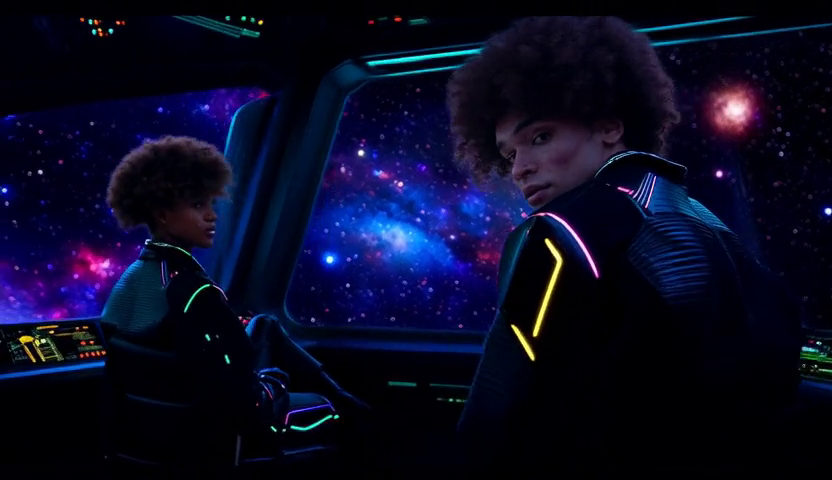} &
		\includegraphics[width=\moreresultwidthnew]{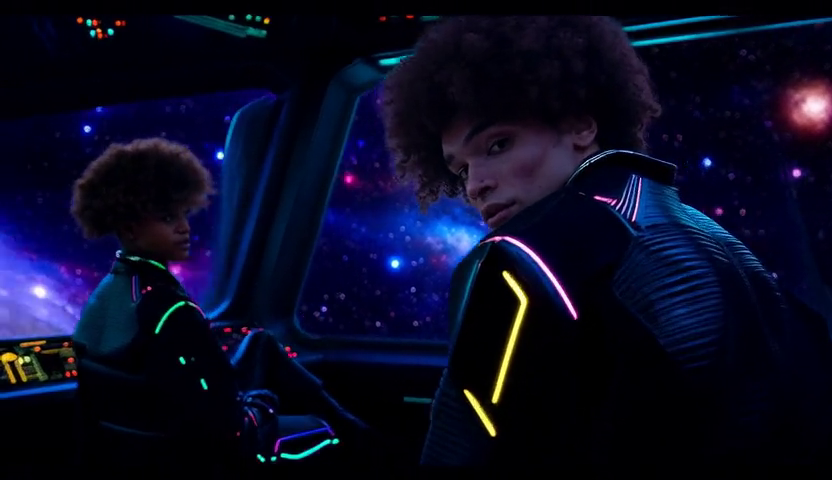} \\
		\multicolumn{7}{p{0.94\textwidth}}{\small Prompt: \textit{Two people with afro hairstyles sit on futuristic chairs inside a dark spaceship observation deck, facing a panoramic window filled with colorful nebulae and stars.}} \\
		\raisebox{0.05\height}{\rotatebox{90}{\scriptsize Input Video}} &
		\includegraphics[width=\moreresultwidthnew]{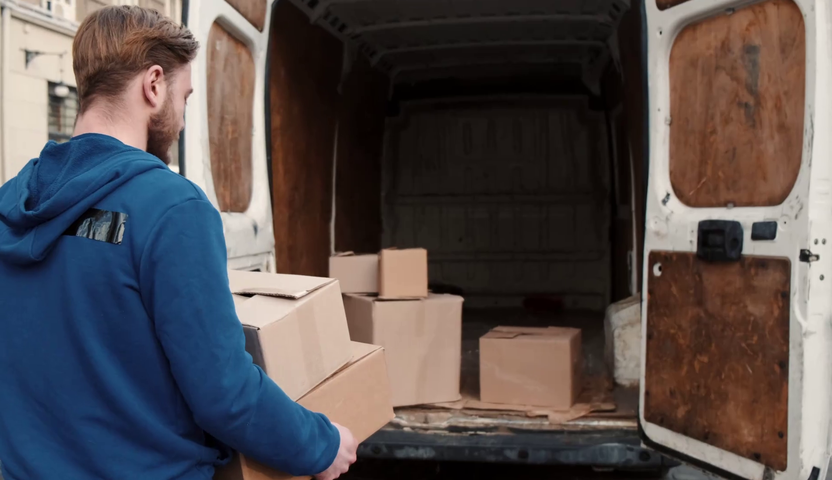} &
		\includegraphics[width=\moreresultwidthnew]{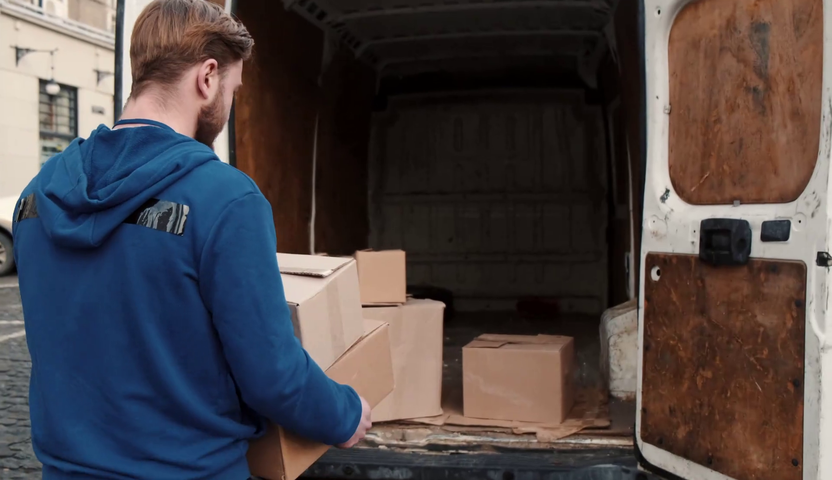} &
		\includegraphics[width=\moreresultwidthnew]{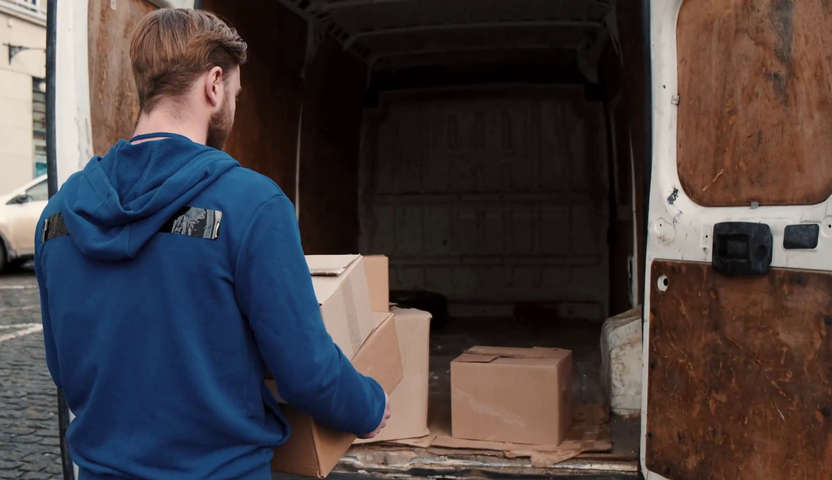} &
		\includegraphics[width=\moreresultwidthnew]{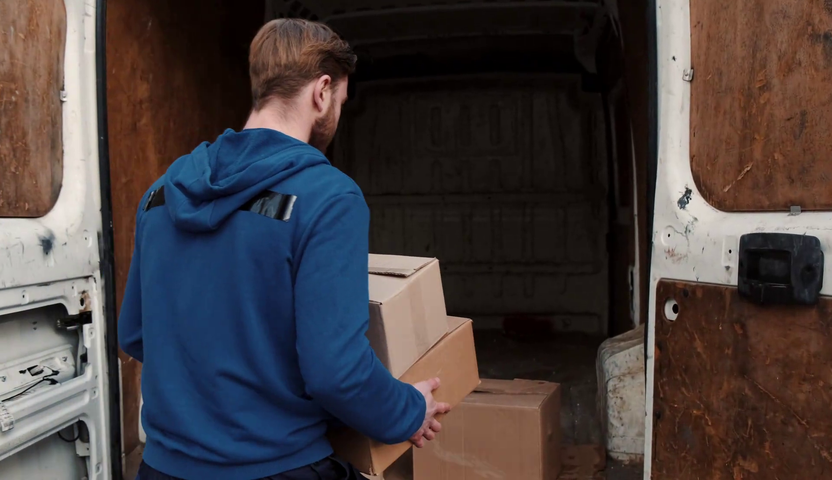} &
		\includegraphics[width=\moreresultwidthnew]{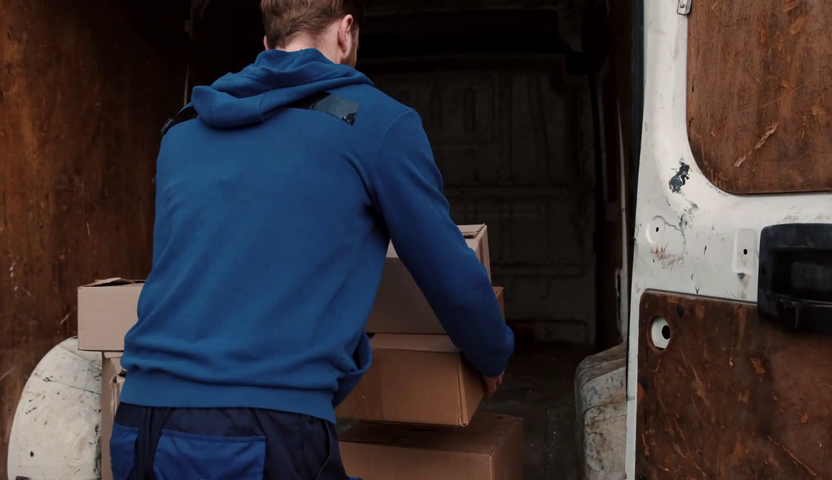} &
		\includegraphics[width=\moreresultwidthnew]{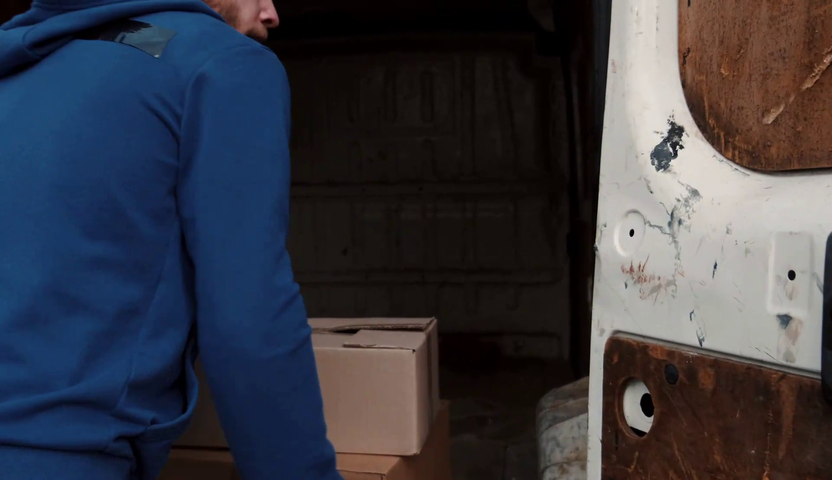} \\
		\raisebox{0.05\height}{\rotatebox{90}{\moreresultmasklabel}} &
		\includegraphics[width=\moreresultwidthnew]{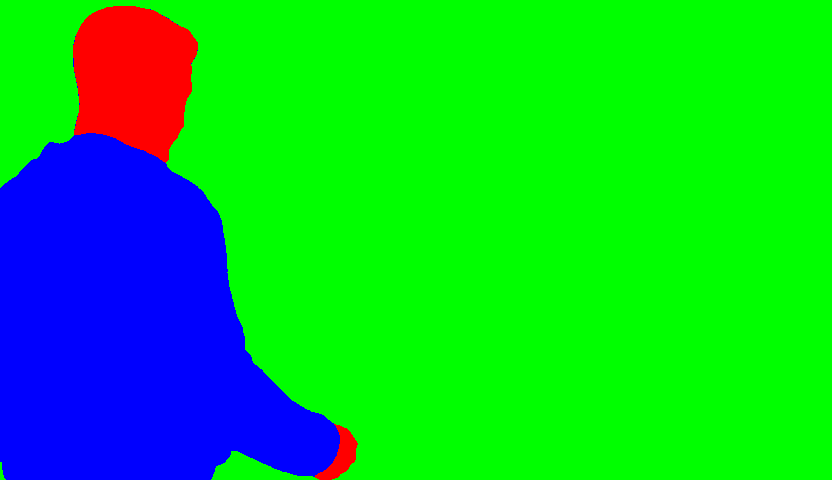} &
		\includegraphics[width=\moreresultwidthnew]{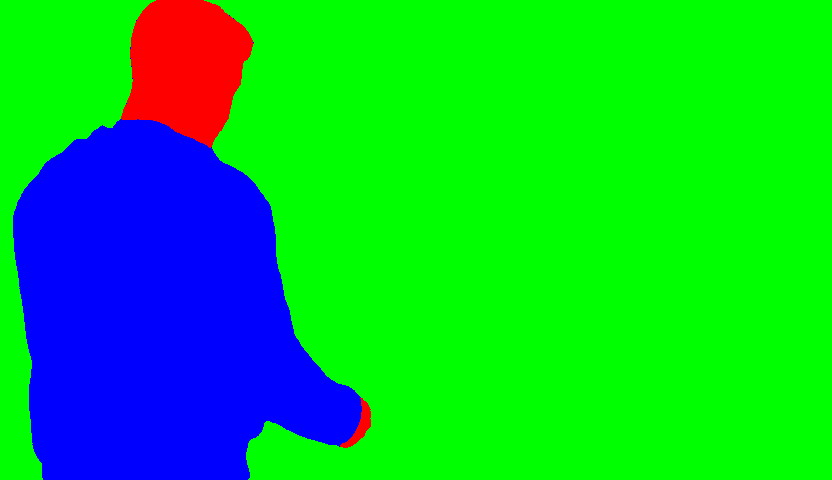} &
		\includegraphics[width=\moreresultwidthnew]{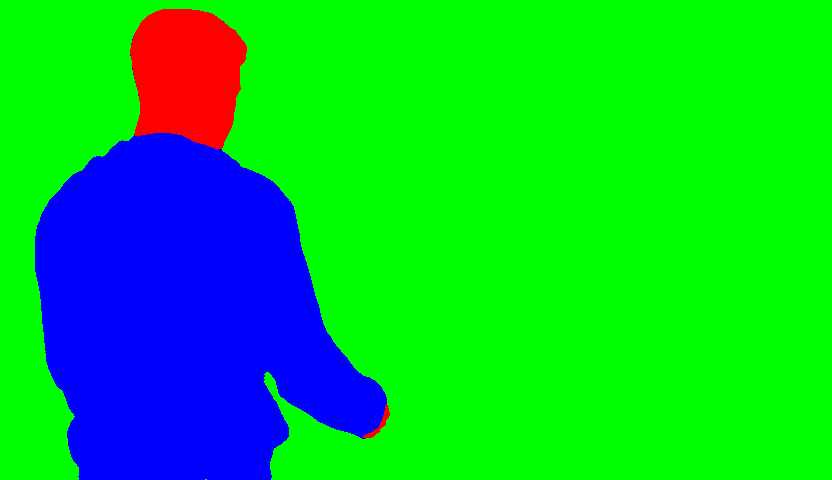} &
		\includegraphics[width=\moreresultwidthnew]{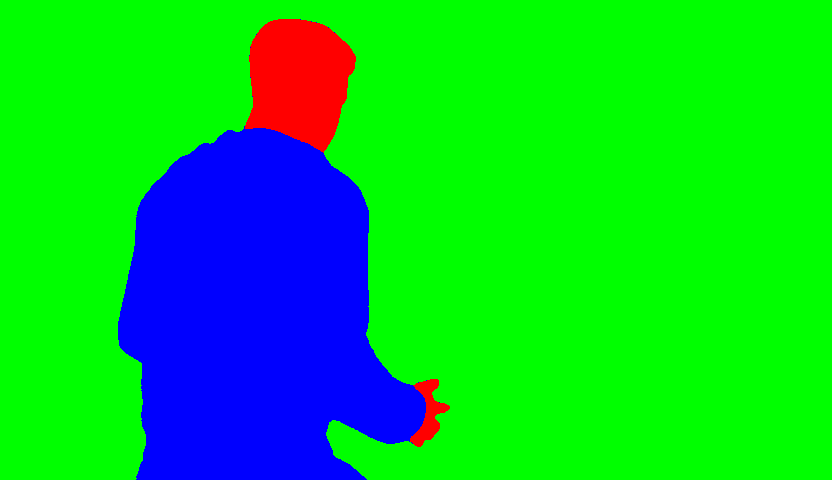} &
		\includegraphics[width=\moreresultwidthnew]{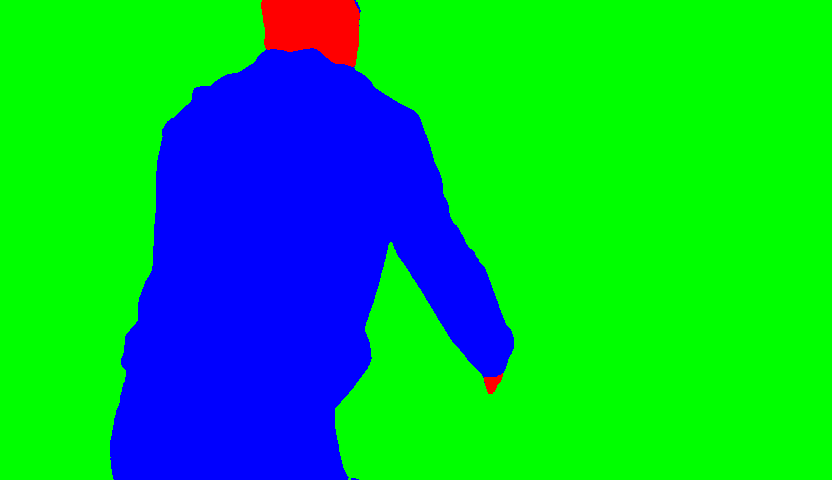} &
		\includegraphics[width=\moreresultwidthnew]{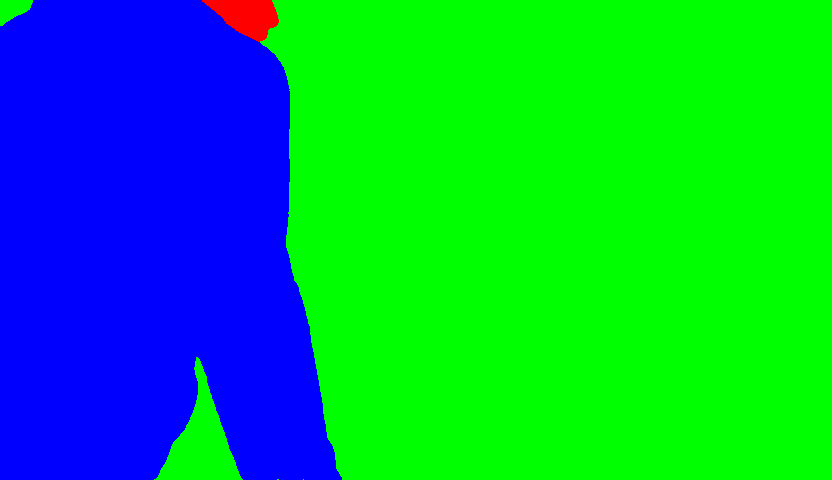} \\
		\raisebox{0.05\height}{\rotatebox{90}{\scriptsize Result}} &
		\includegraphics[width=\moreresultwidthnew]{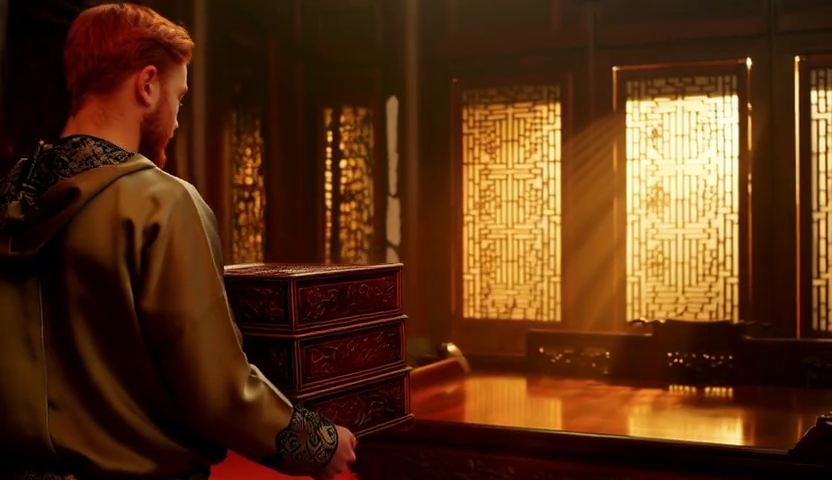} &
		\includegraphics[width=\moreresultwidthnew]{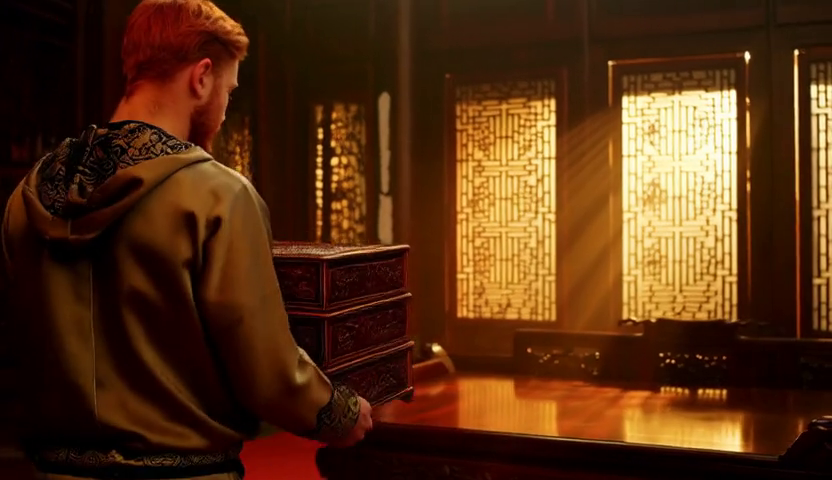} &
		\includegraphics[width=\moreresultwidthnew]{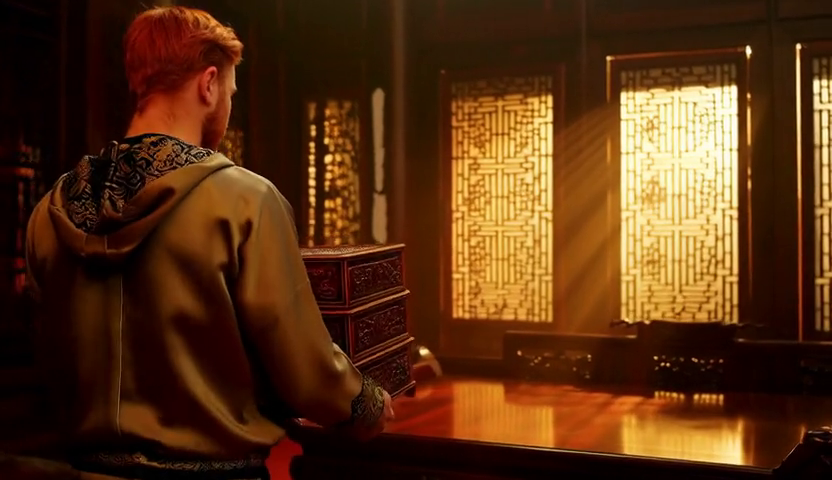} &
		\includegraphics[width=\moreresultwidthnew]{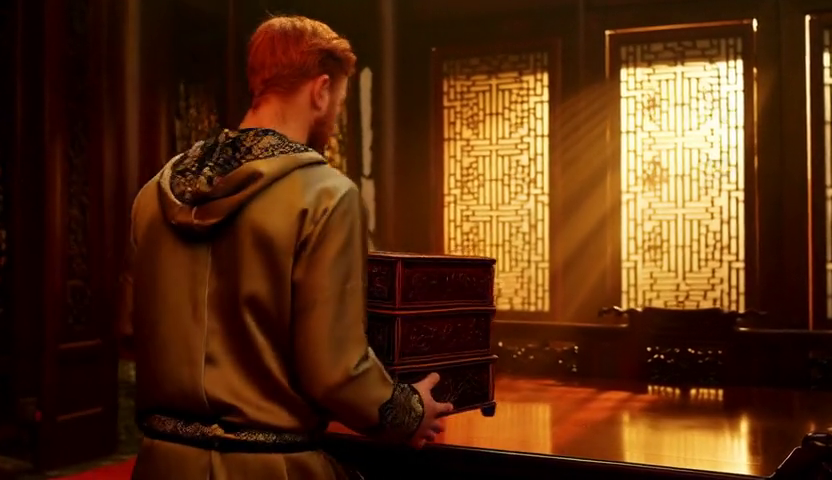} &
		\includegraphics[width=\moreresultwidthnew]{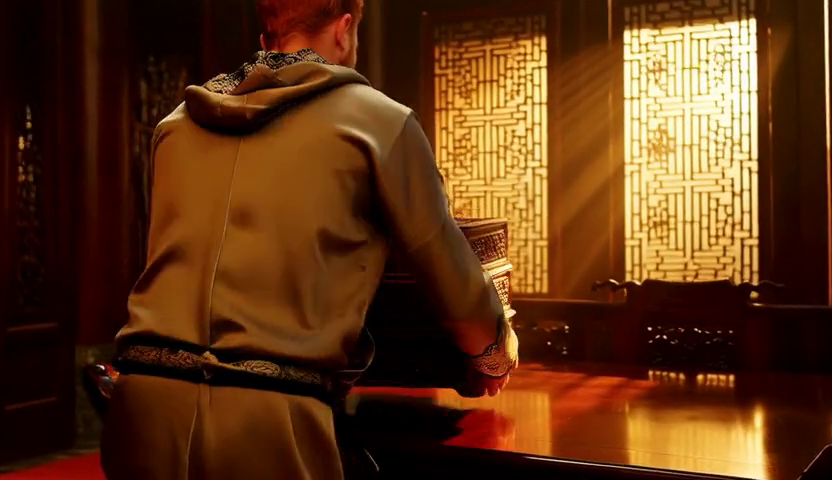} &
		\includegraphics[width=\moreresultwidthnew]{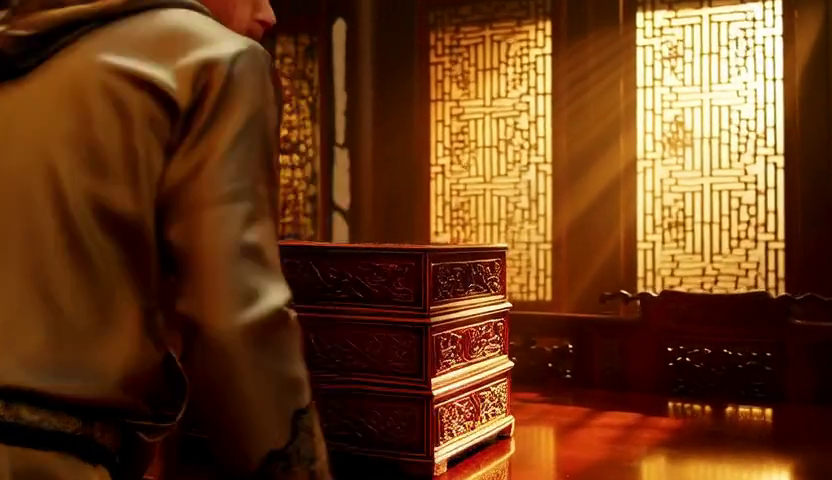} \\
		\multicolumn{7}{p{0.94\textwidth}}{\small Prompt: \textit{A man in elegant traditional Chinese robes carries carved wooden antique boxes through a grand ancient palace hall and places them on an ornate wooden table.}} \\
		\raisebox{0.05\height}{\rotatebox{90}{\scriptsize Input Video}} &
		\includegraphics[width=\moreresultwidthnew]{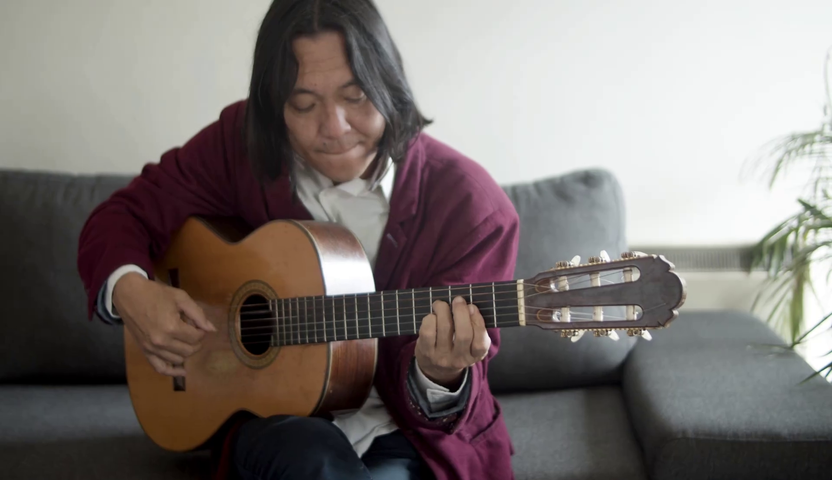} &
		\includegraphics[width=\moreresultwidthnew]{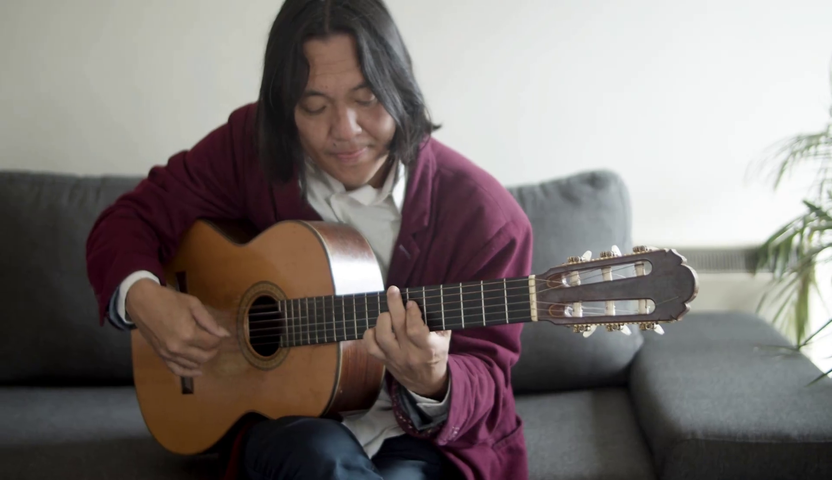} &
		\includegraphics[width=\moreresultwidthnew]{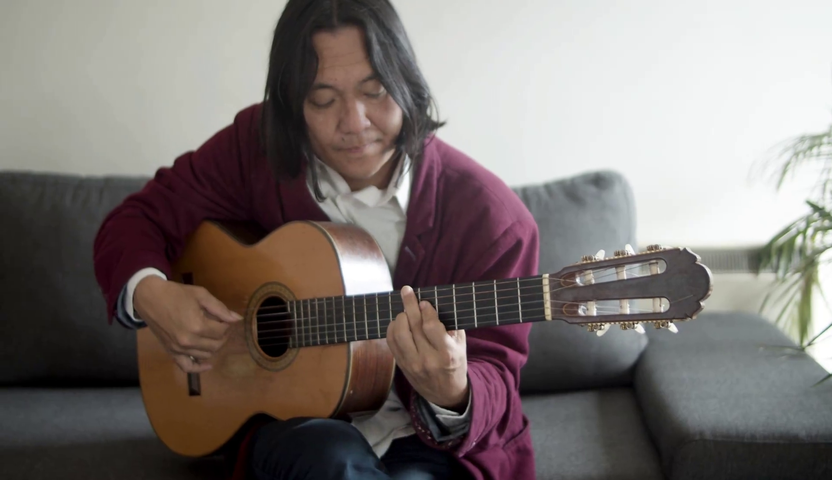} &
		\includegraphics[width=\moreresultwidthnew]{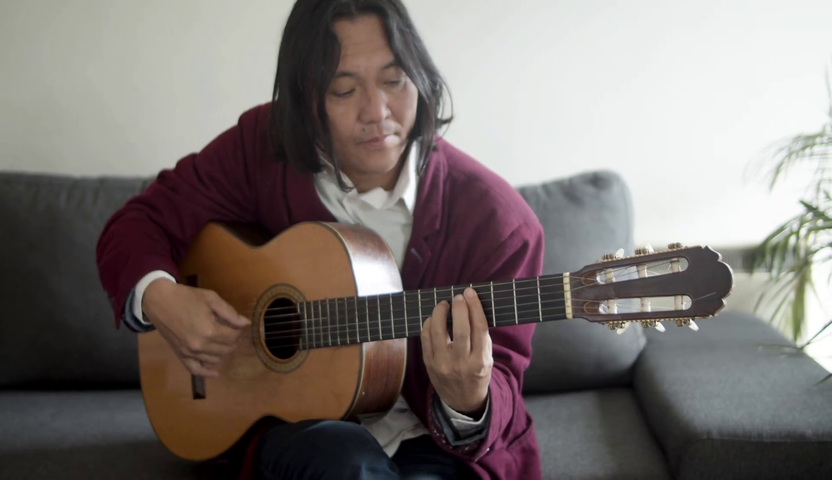} &
		\includegraphics[width=\moreresultwidthnew]{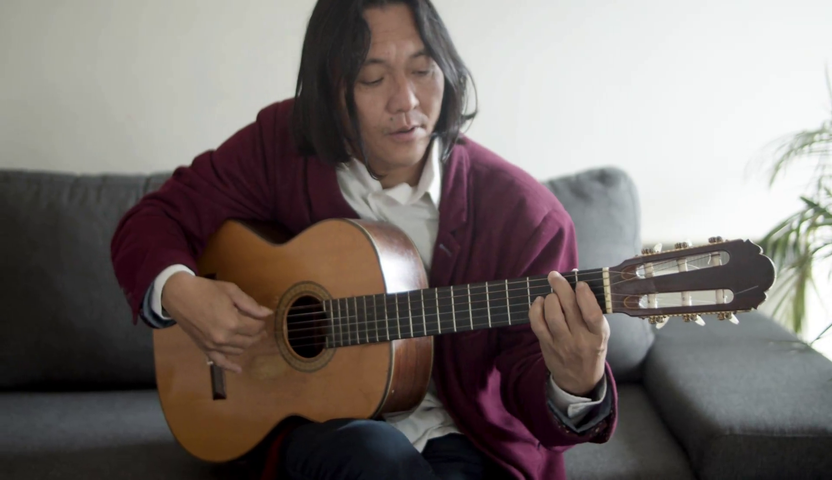} &
		\includegraphics[width=\moreresultwidthnew]{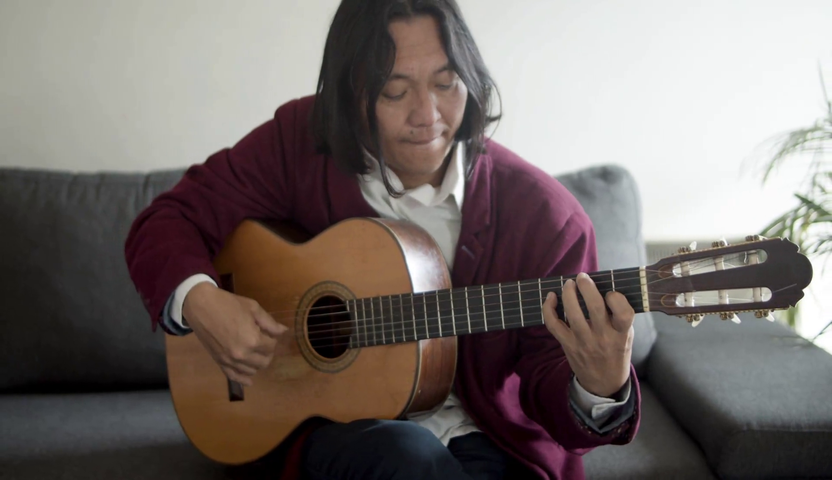} \\
		\raisebox{0.05\height}{\rotatebox{90}{\moreresultmasklabel}} &
		\includegraphics[width=\moreresultwidthnew]{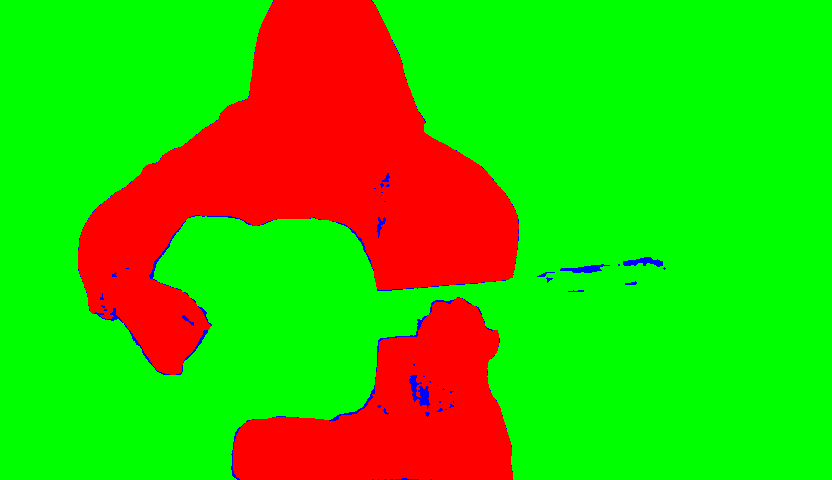} &
		\includegraphics[width=\moreresultwidthnew]{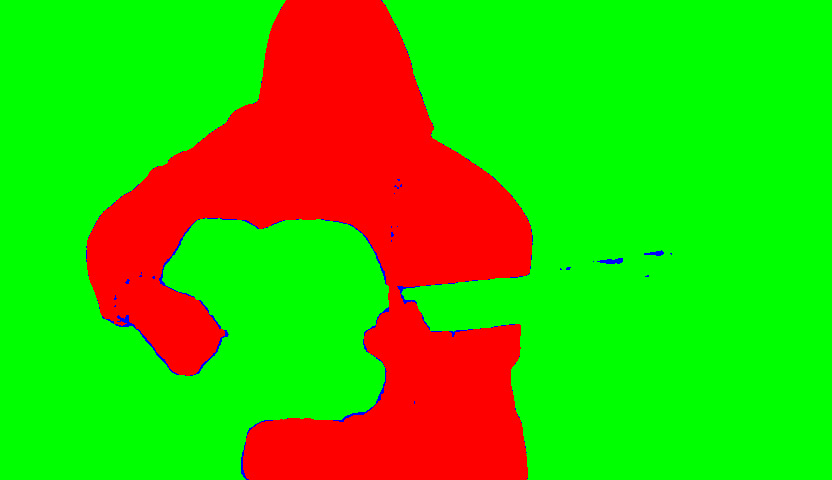} &
		\includegraphics[width=\moreresultwidthnew]{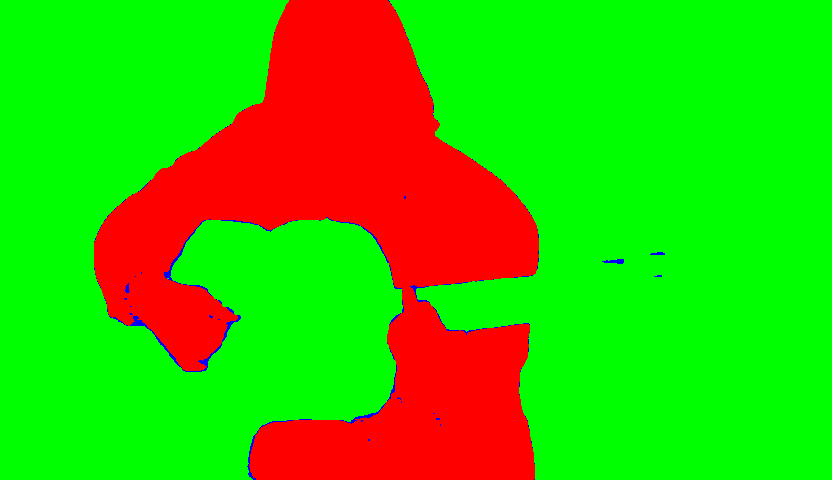} &
		\includegraphics[width=\moreresultwidthnew]{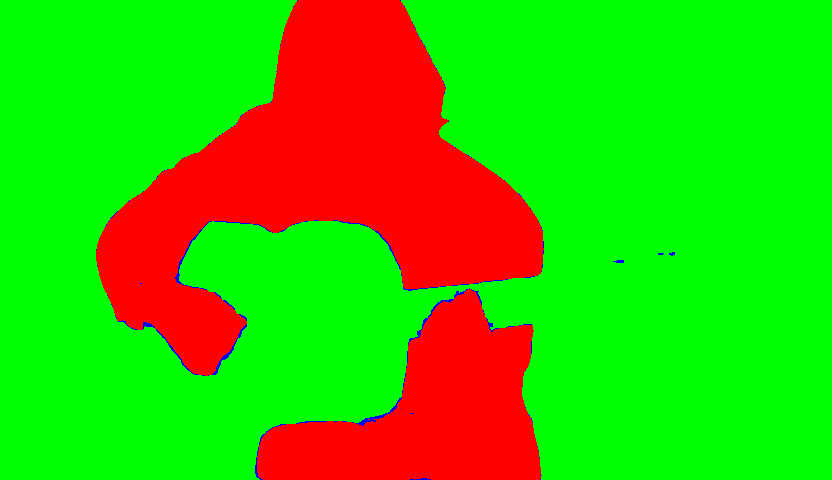} &
		\includegraphics[width=\moreresultwidthnew]{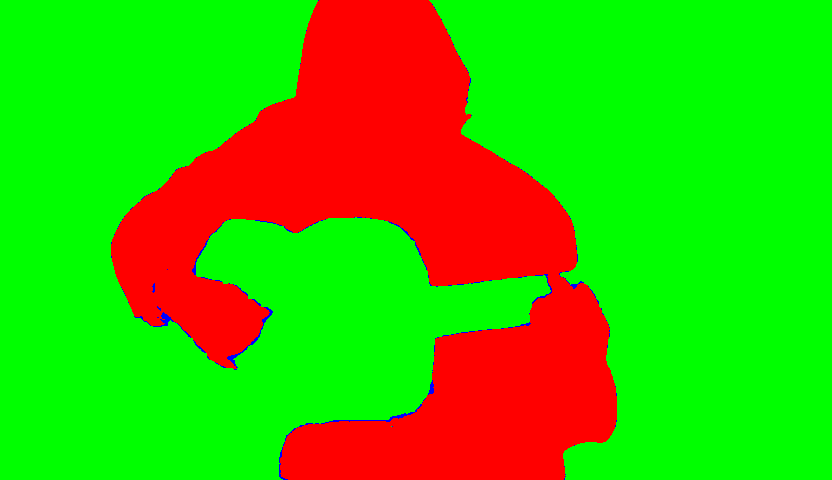} &
		\includegraphics[width=\moreresultwidthnew]{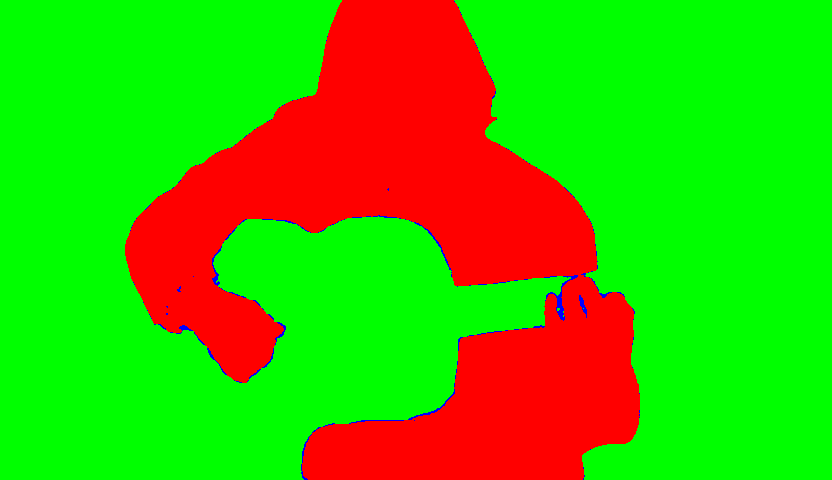} \\
		\raisebox{0.05\height}{\rotatebox{90}{\scriptsize Result}} &
		\includegraphics[width=\moreresultwidthnew]{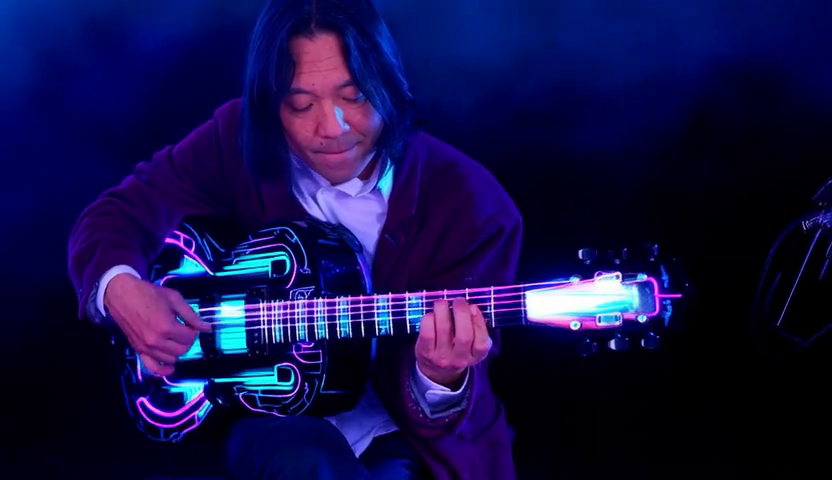} &
		\includegraphics[width=\moreresultwidthnew]{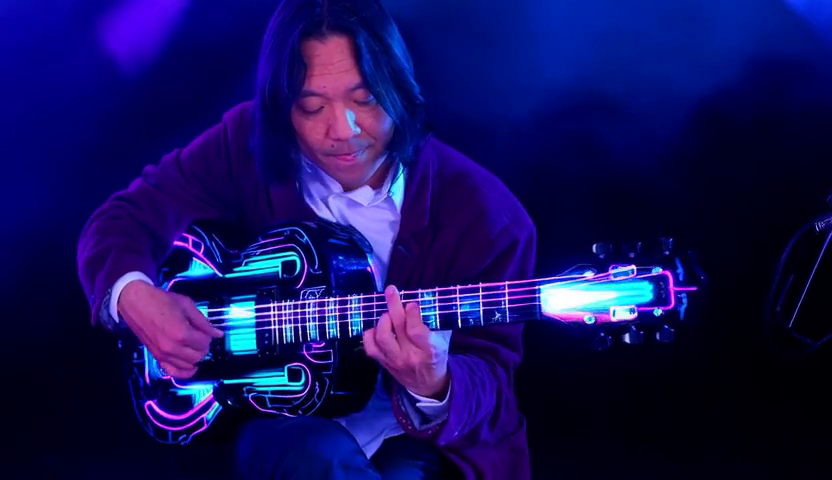} &
		\includegraphics[width=\moreresultwidthnew]{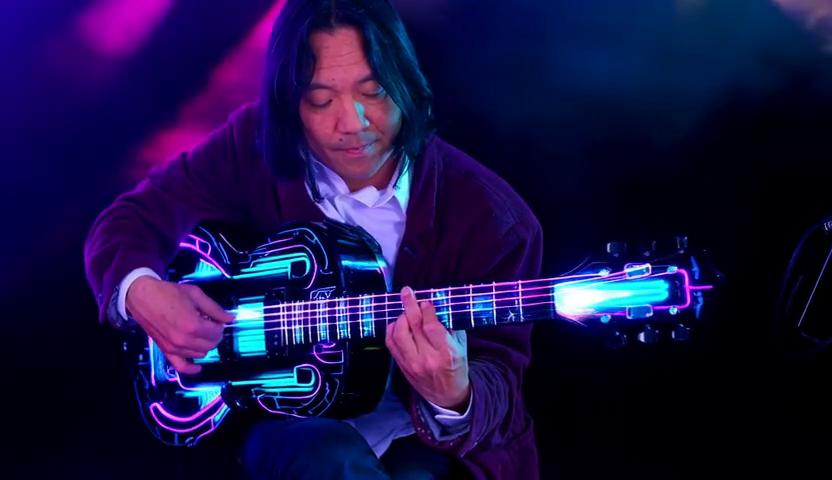} &
		\includegraphics[width=\moreresultwidthnew]{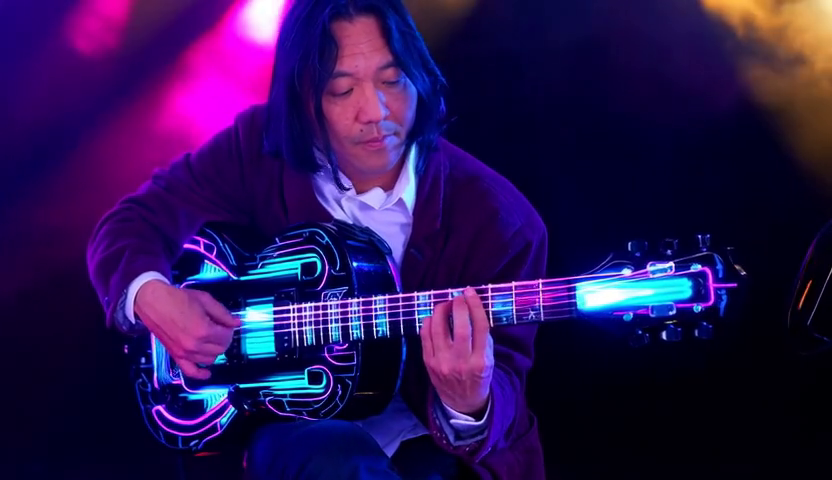} &
		\includegraphics[width=\moreresultwidthnew]{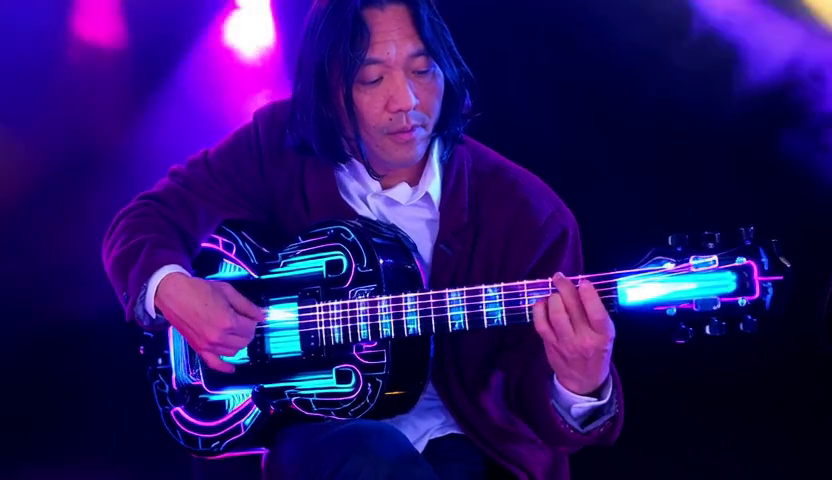} &
		\includegraphics[width=\moreresultwidthnew]{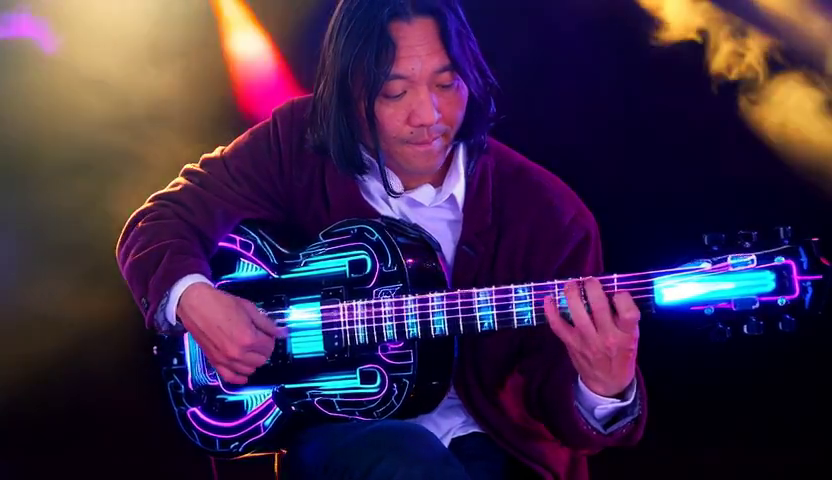} \\
		\multicolumn{7}{p{0.94\textwidth}}{\small Prompt: \textit{A man plays an advanced glowing cyberpunk electronic guitar on a dim concert stage, surrounded by sweeping magenta, cyan, and gold lights.}} \\
	\end{tabular}}
	\caption{\textbf{Applications on real-world videos.} Given different tri-masks, our model can flexibly replace or preserve arbitrary regions of a video while maintaining strong C2E and E2C interaction consistency.}
	
	\label{fig:more_result_0610_a}
\end{figure*}

\begin{figure*}[p]
	\centering
	\setlength{\tabcolsep}{0.5pt}
	\renewcommand{\arraystretch}{0.95}
	\scalebox{1}{\begin{tabular}{c@{\hspace{2pt}}cccccc}
		\raisebox{0.05\height}{\rotatebox{90}{\scriptsize Input Video}} &
		\includegraphics[width=\moreresultwidthnew]{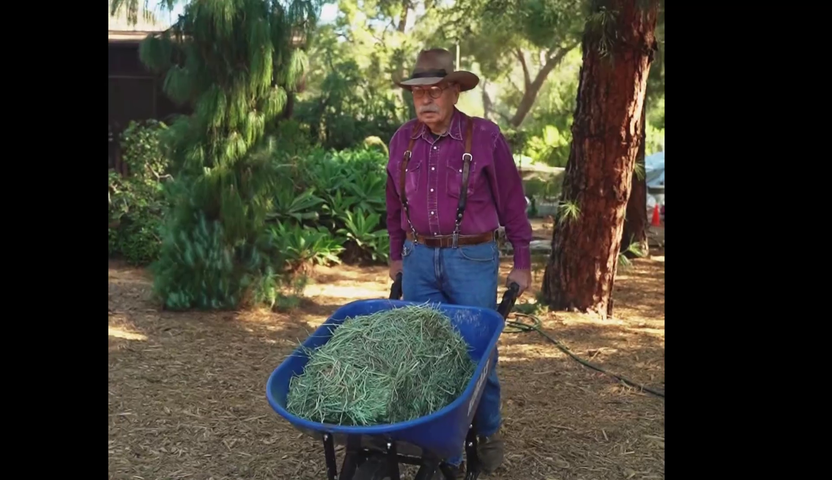} &
		\includegraphics[width=\moreresultwidthnew]{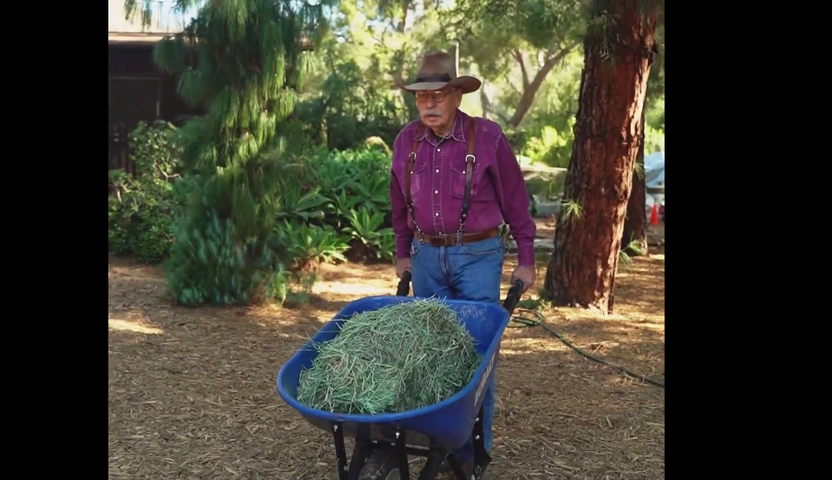} &
		\includegraphics[width=\moreresultwidthnew]{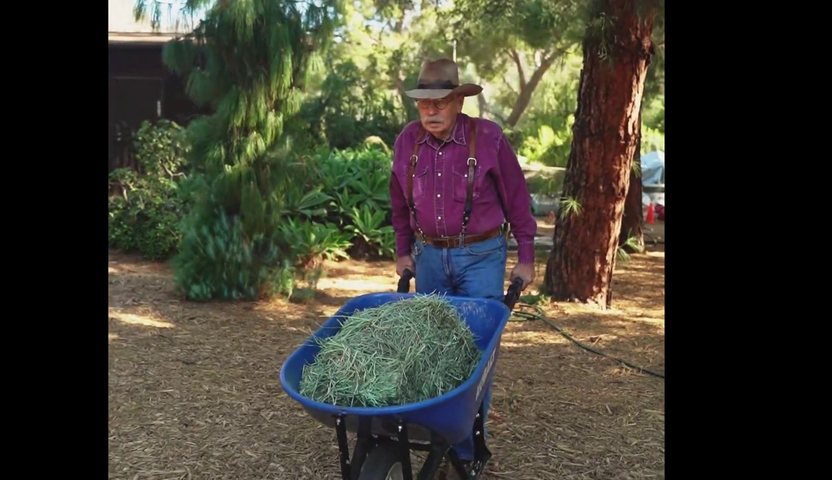} &
		\includegraphics[width=\moreresultwidthnew]{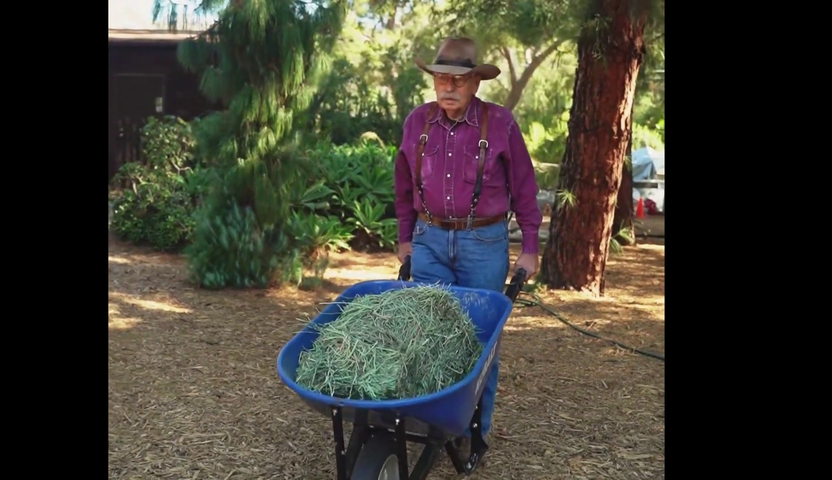} &
		\includegraphics[width=\moreresultwidthnew]{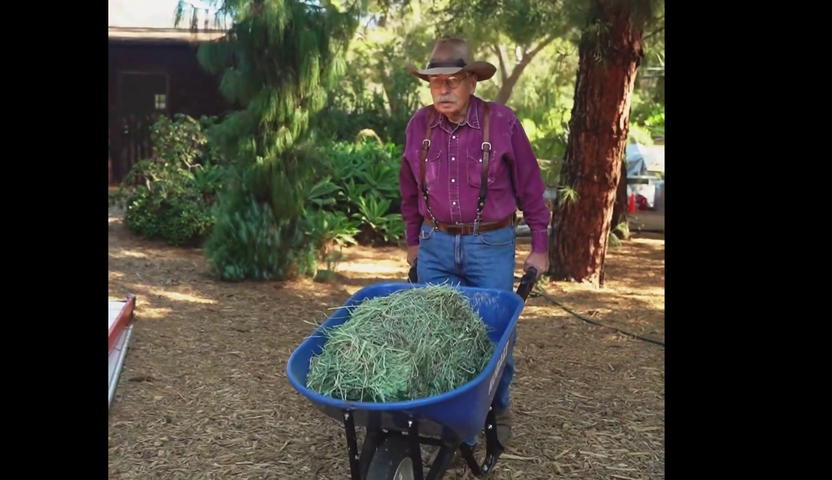} &
		\includegraphics[width=\moreresultwidthnew]{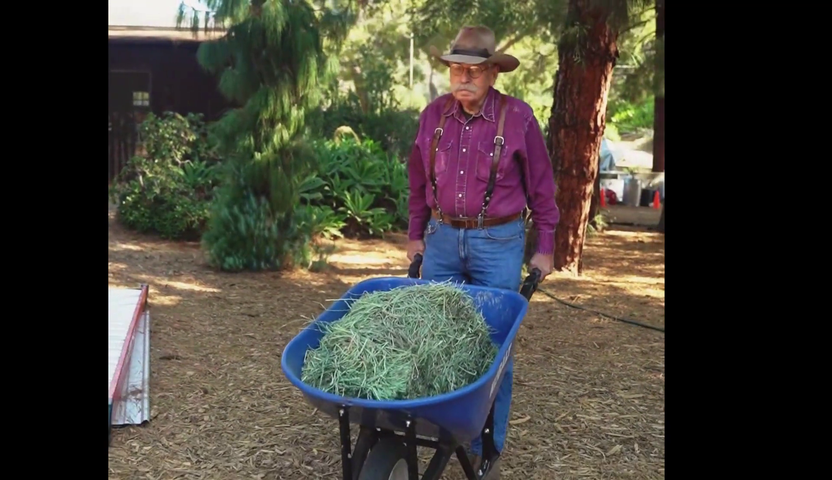} \\
		\raisebox{0.05\height}{\rotatebox{90}{\moreresultmasklabel}} &
		\includegraphics[width=\moreresultwidthnew]{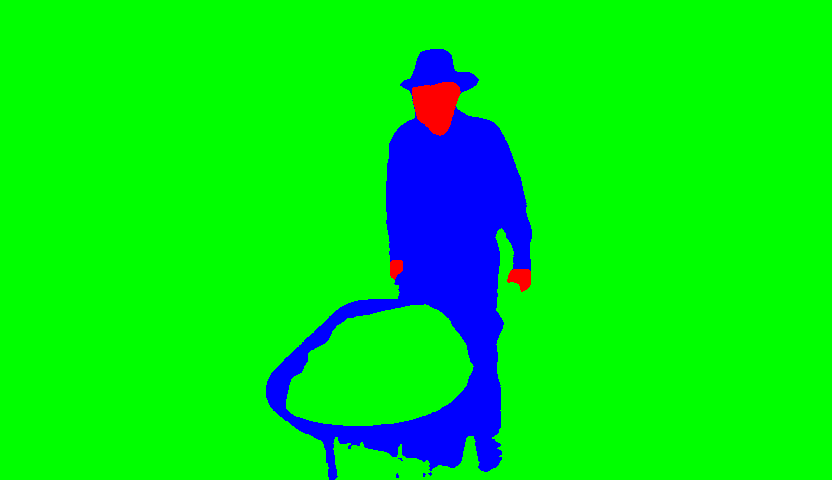} &
		\includegraphics[width=\moreresultwidthnew]{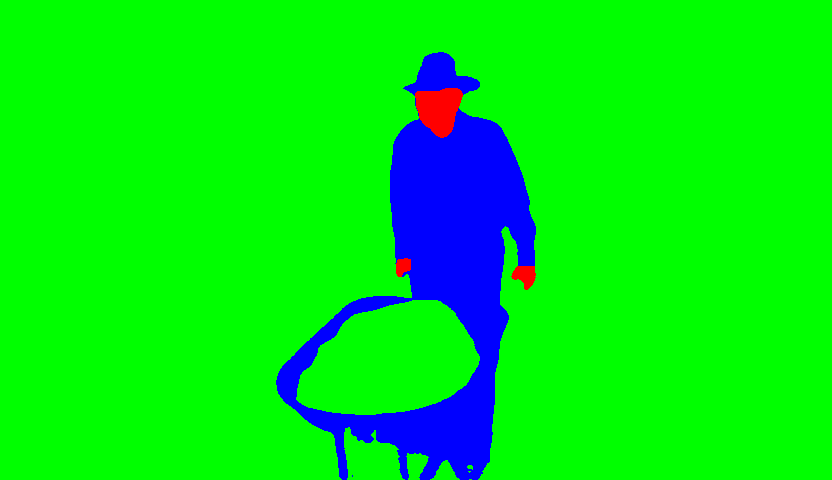} &
		\includegraphics[width=\moreresultwidthnew]{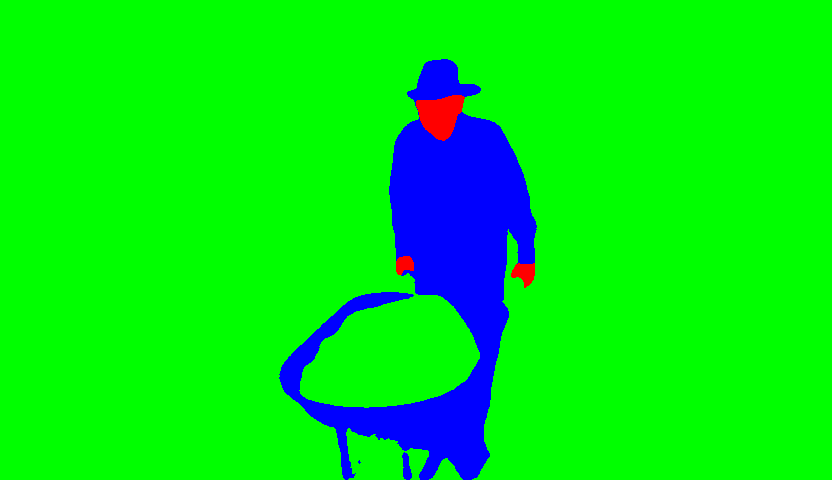} &
		\includegraphics[width=\moreresultwidthnew]{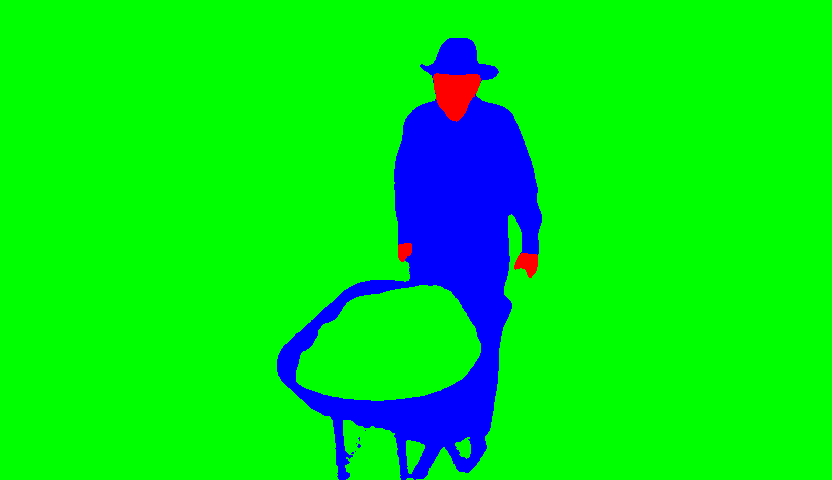} &
		\includegraphics[width=\moreresultwidthnew]{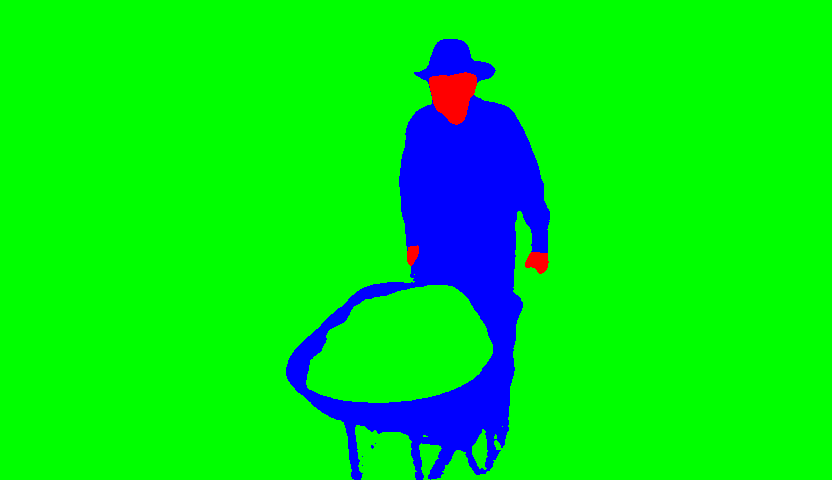} &
		\includegraphics[width=\moreresultwidthnew]{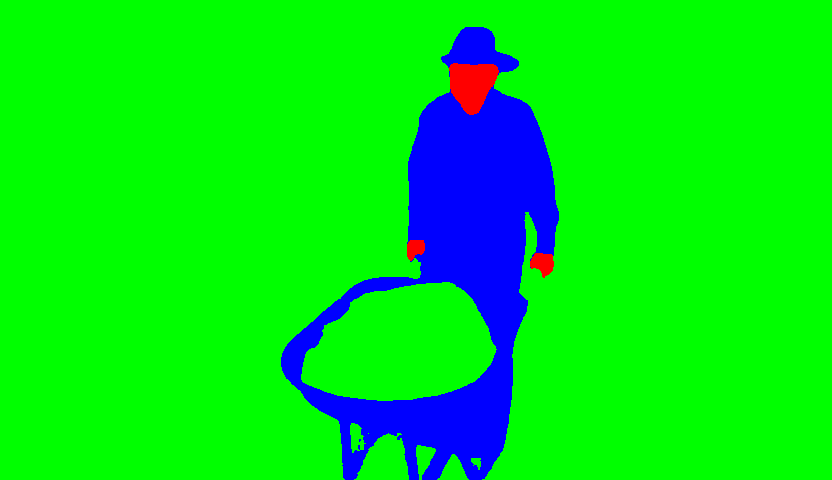} \\
		\raisebox{0.05\height}{\rotatebox{90}{\scriptsize Result}} &
		\includegraphics[width=\moreresultwidthnew]{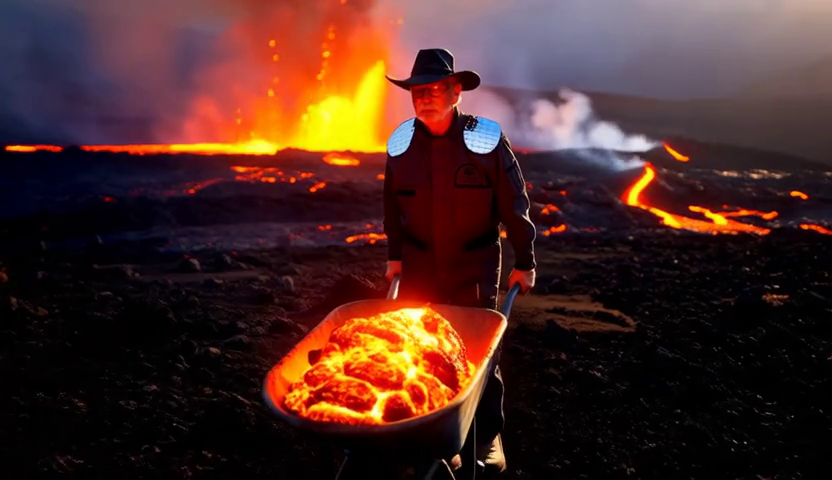} &
		\includegraphics[width=\moreresultwidthnew]{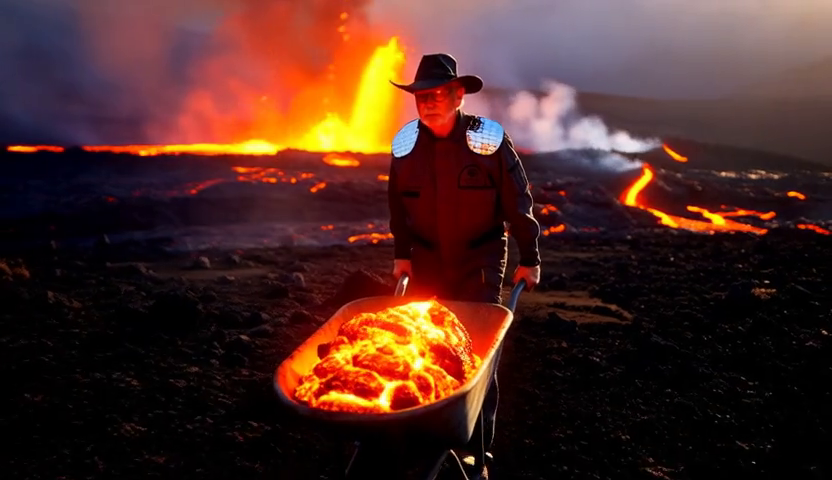} &
		\includegraphics[width=\moreresultwidthnew]{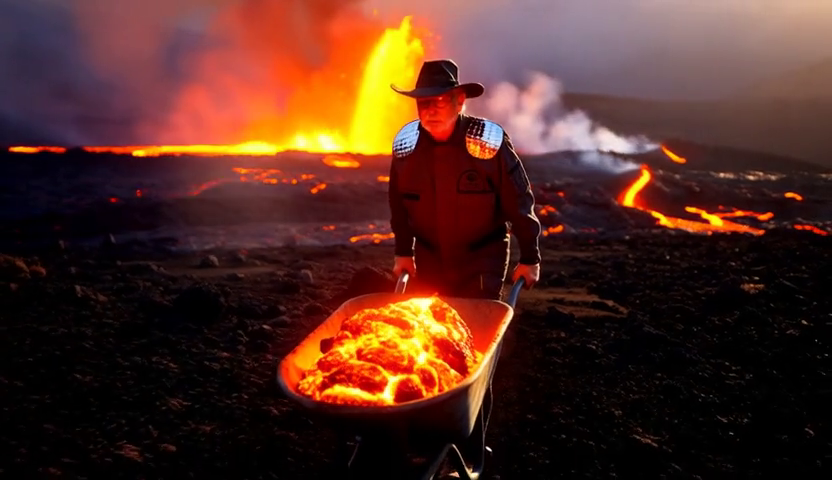} &
		\includegraphics[width=\moreresultwidthnew]{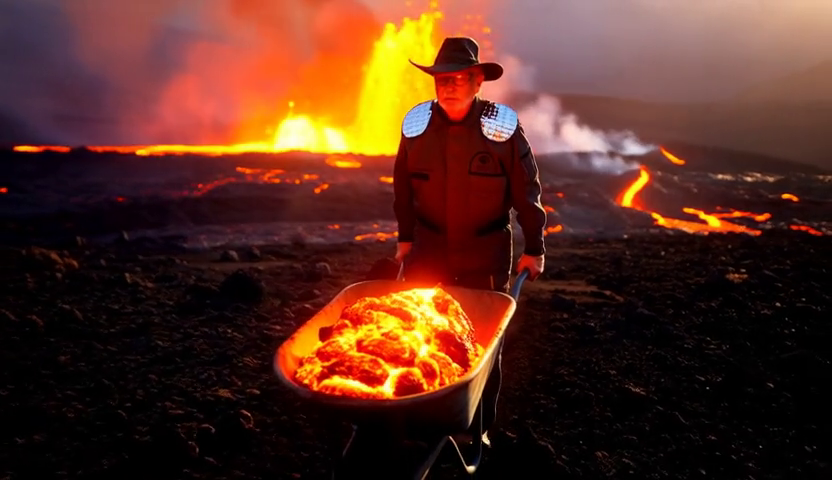} &
		\includegraphics[width=\moreresultwidthnew]{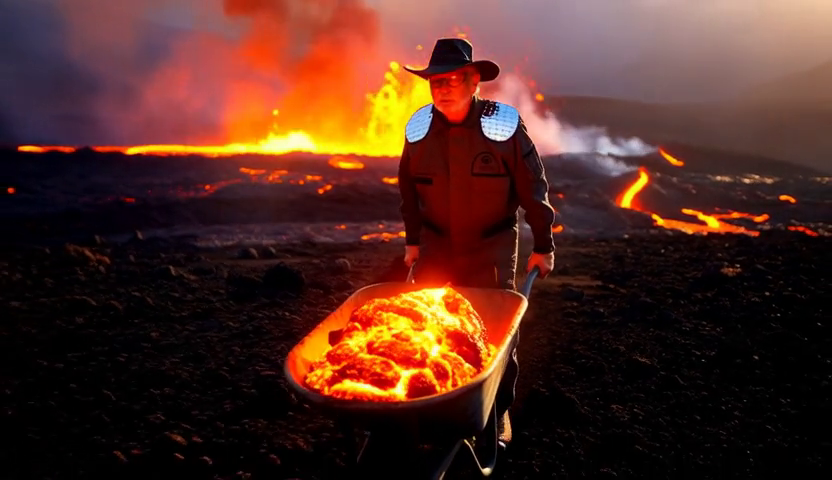} &
		\includegraphics[width=\moreresultwidthnew]{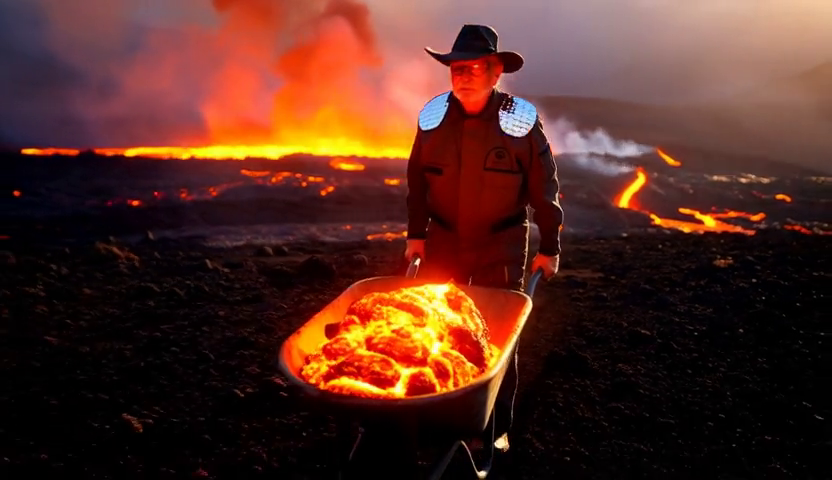} \\
		\multicolumn{7}{p{0.94\textwidth}}{\small Prompt: \textit{An elderly man in a sci-fi volcanologist suit pushes a heavy industrial wheelbarrow filled with glowing molten lava through a dramatic volcanic landscape.}} \\
		\raisebox{0.05\height}{\rotatebox{90}{\scriptsize Input Video}} &
		\includegraphics[width=\moreresultwidthnew]{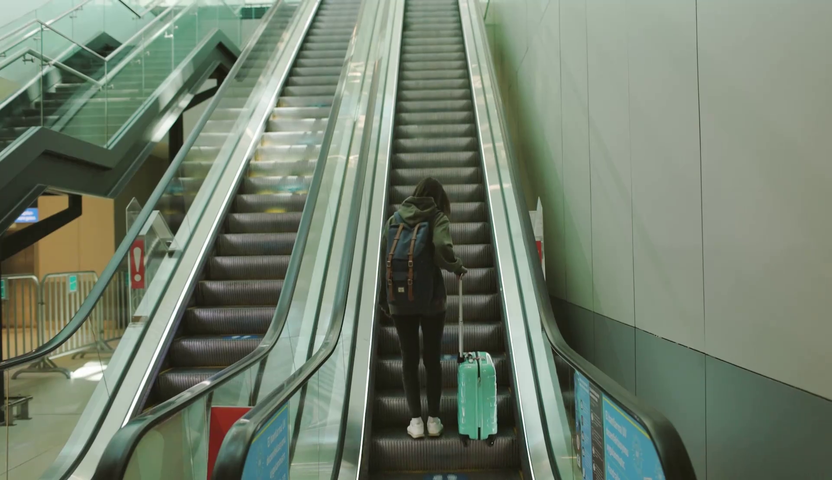} &
		\includegraphics[width=\moreresultwidthnew]{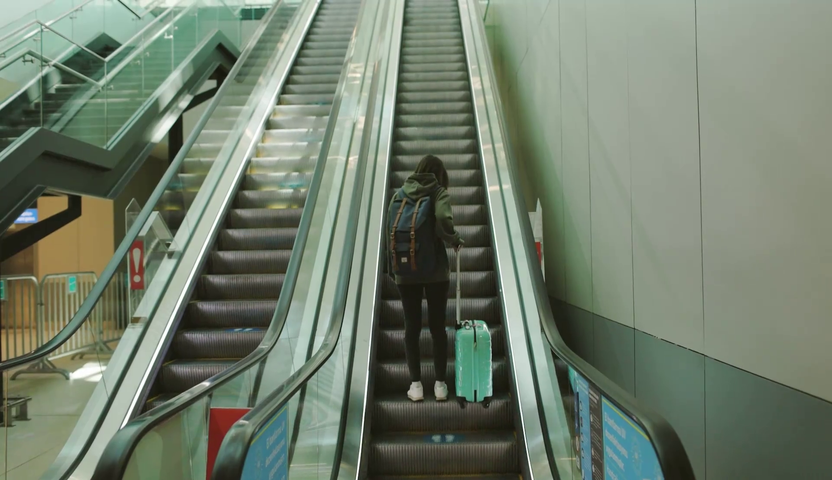} &
		\includegraphics[width=\moreresultwidthnew]{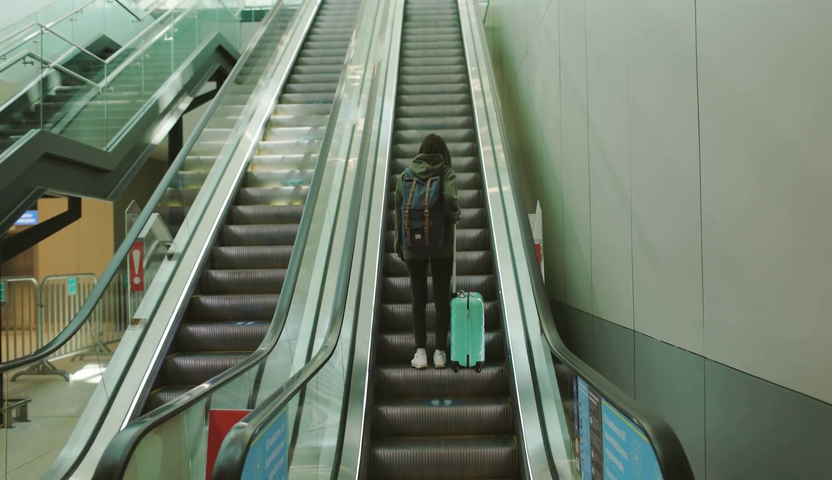} &
		\includegraphics[width=\moreresultwidthnew]{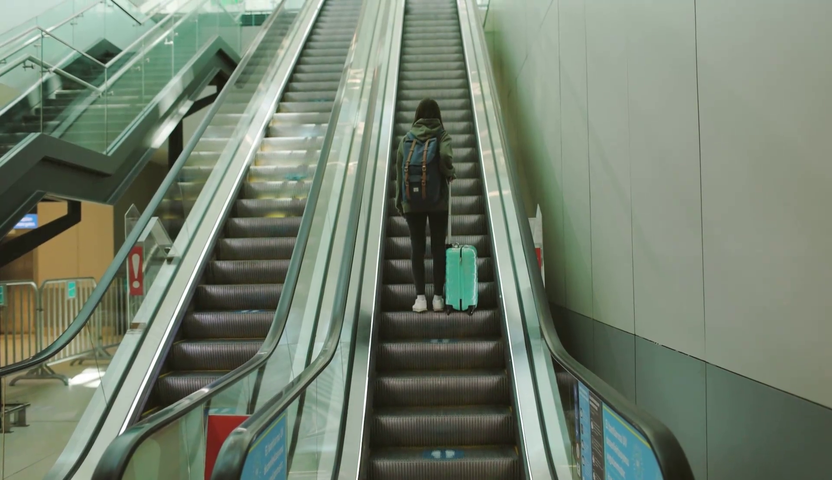} &
		\includegraphics[width=\moreresultwidthnew]{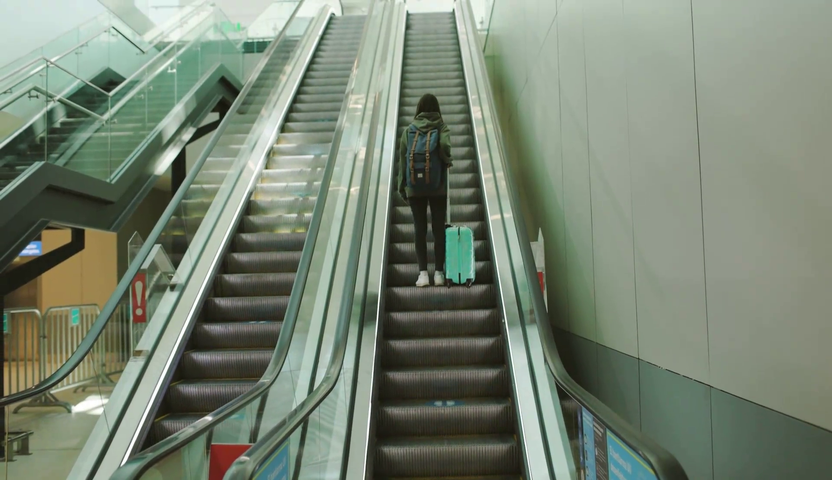} &
		\includegraphics[width=\moreresultwidthnew]{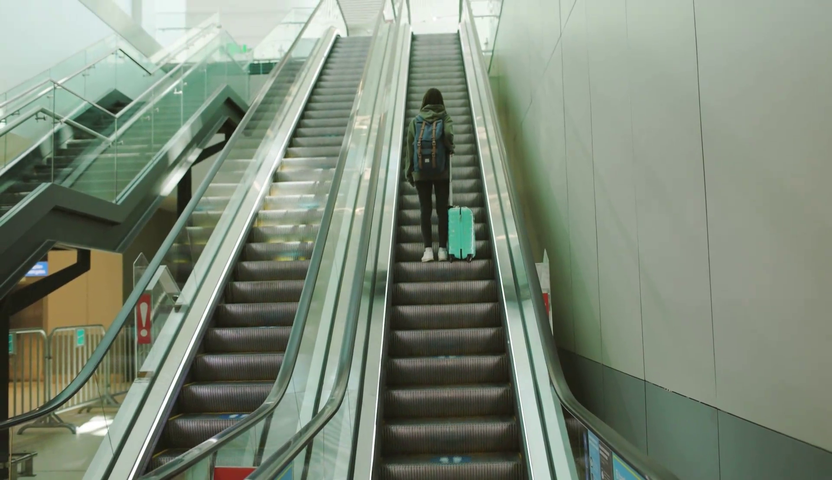} \\
		\raisebox{0.05\height}{\rotatebox{90}{\moreresultmasklabel}} &
		\includegraphics[width=\moreresultwidthnew]{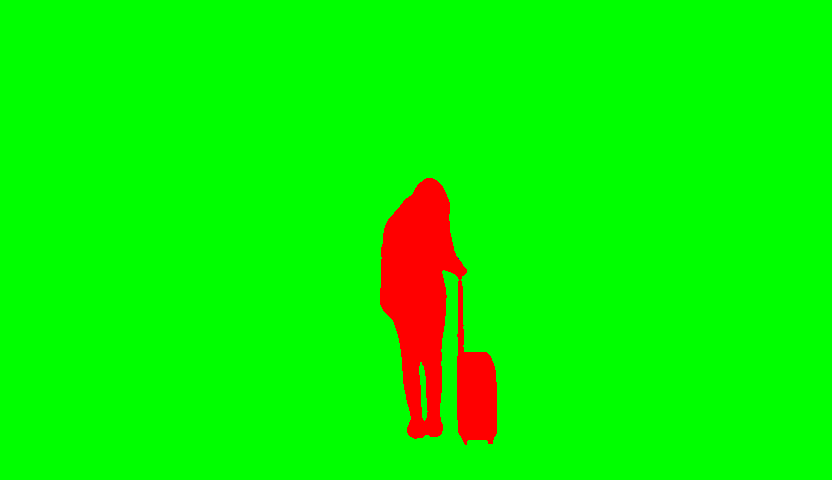} &
		\includegraphics[width=\moreresultwidthnew]{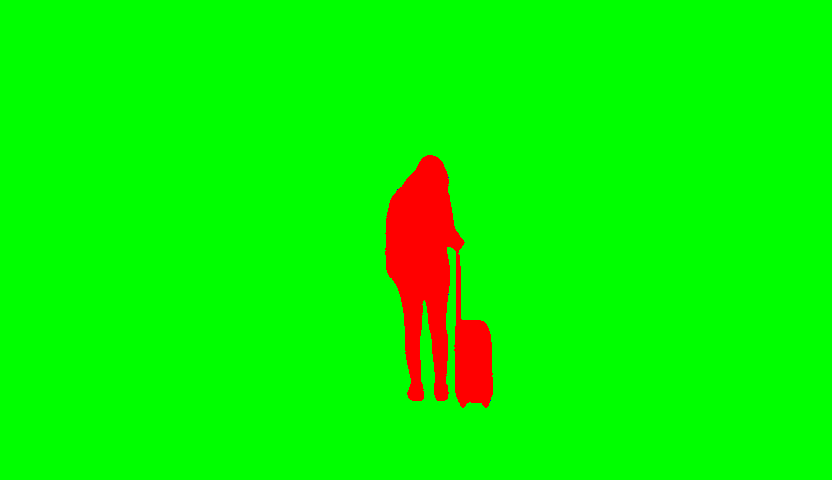} &
		\includegraphics[width=\moreresultwidthnew]{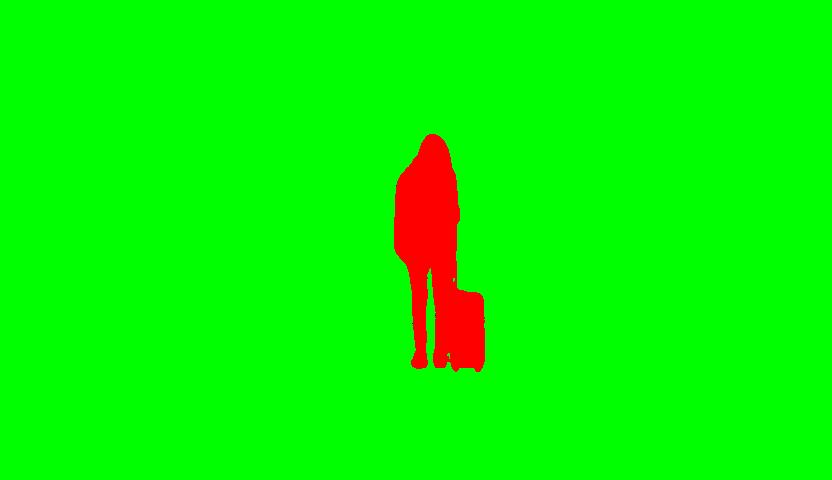} &
		\includegraphics[width=\moreresultwidthnew]{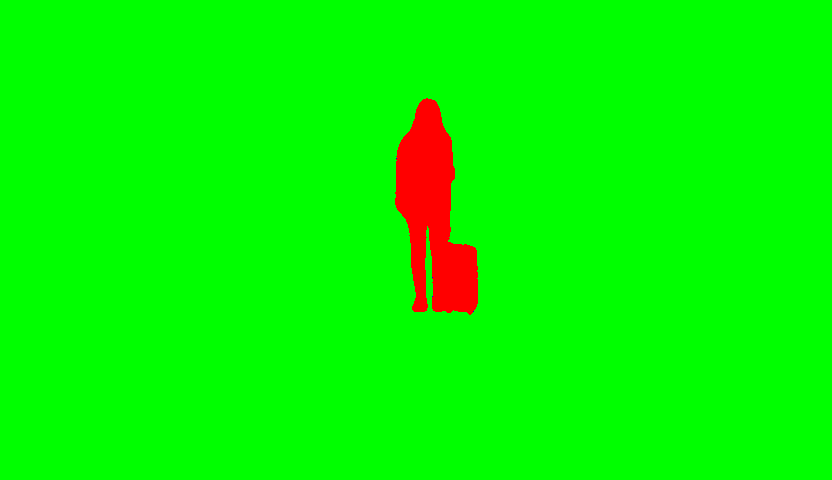} &
		\includegraphics[width=\moreresultwidthnew]{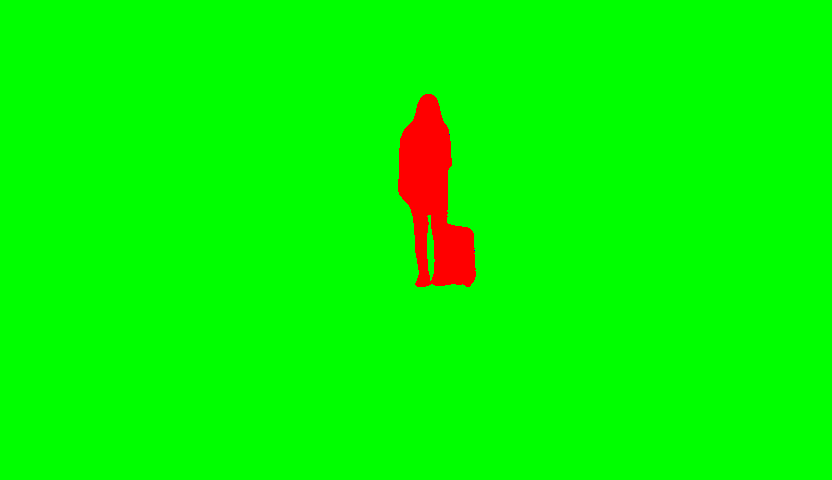} &
		\includegraphics[width=\moreresultwidthnew]{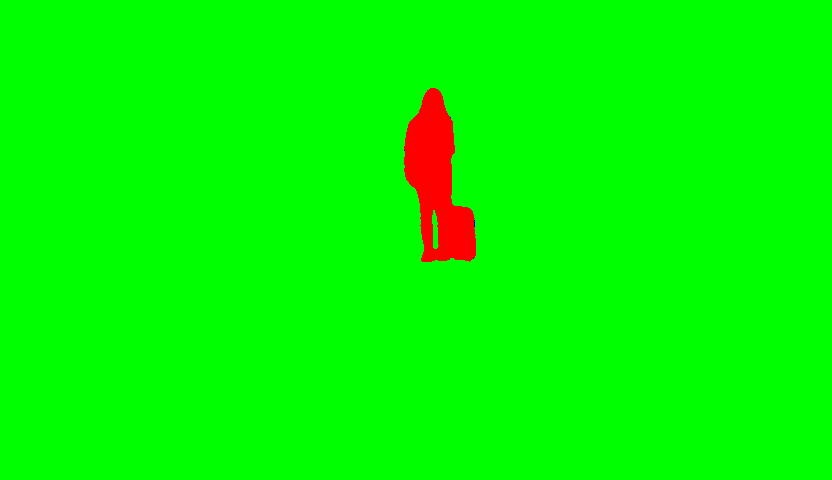} \\
		\raisebox{0.05\height}{\rotatebox{90}{\scriptsize Result}} &
		\includegraphics[width=\moreresultwidthnew]{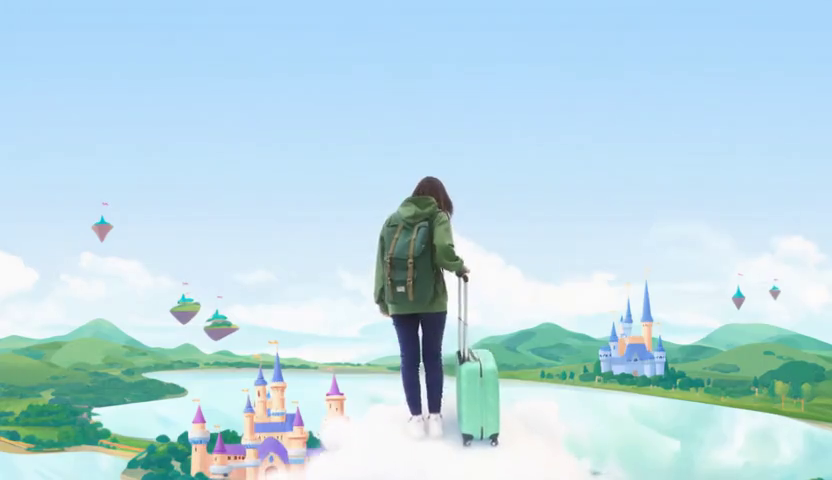} &
		\includegraphics[width=\moreresultwidthnew]{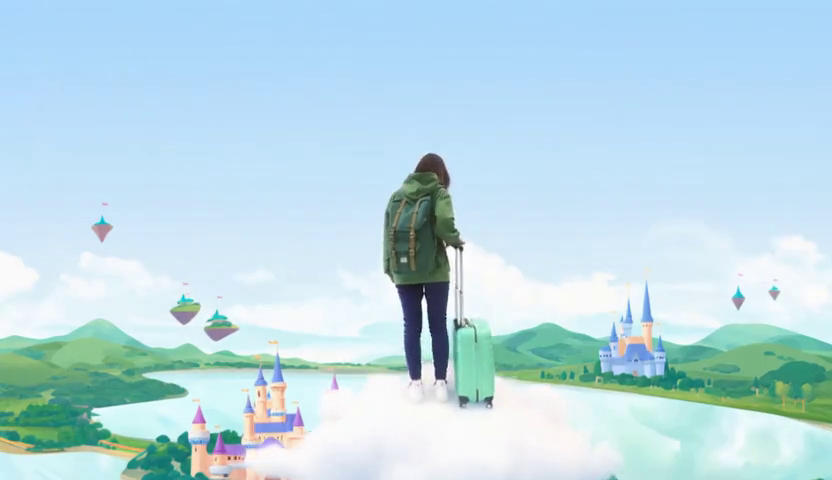} &
		\includegraphics[width=\moreresultwidthnew]{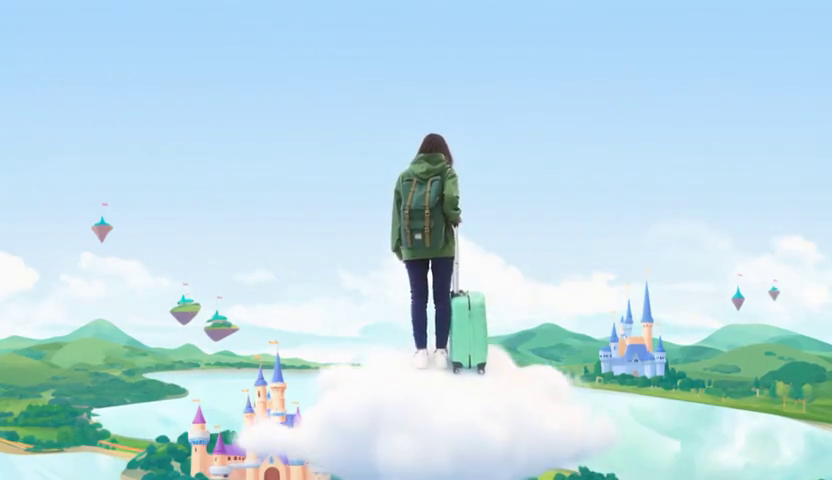} &
		\includegraphics[width=\moreresultwidthnew]{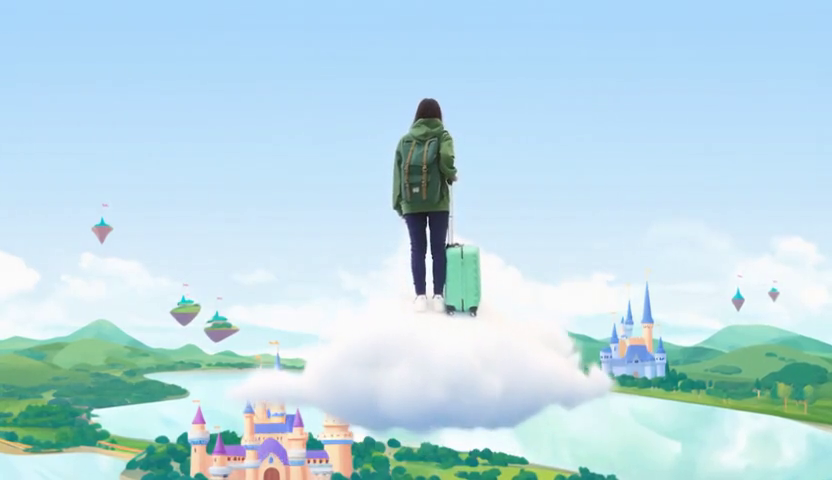} &
		\includegraphics[width=\moreresultwidthnew]{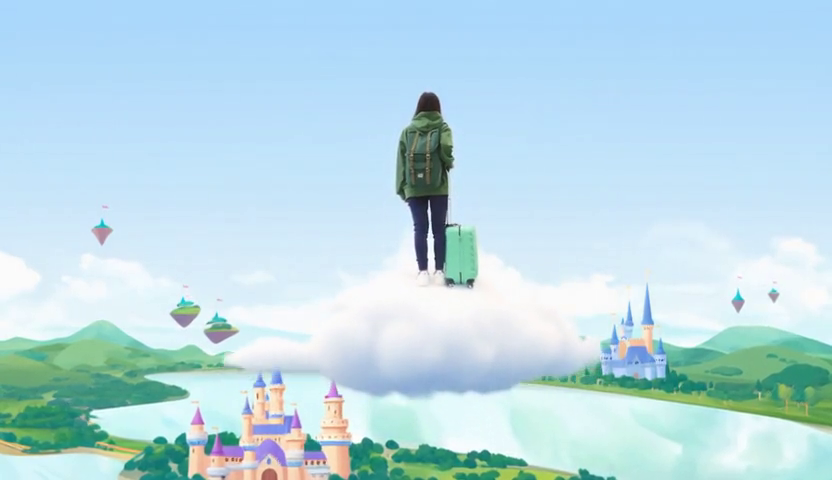} &
		\includegraphics[width=\moreresultwidthnew]{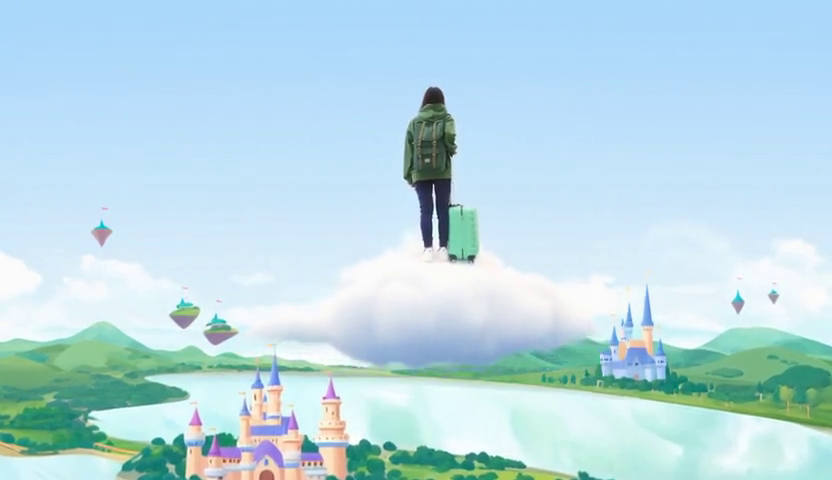} \\
		\multicolumn{7}{p{0.94\textwidth}}{\small Prompt: \textit{A woman with a backpack and rolling suitcase stands on a fluffy magical cloud rising through a whimsical fairytale landscape with castles and floating islands.}} \\
		\raisebox{0.05\height}{\rotatebox{90}{\scriptsize Input Video}} &
		\includegraphics[width=\moreresultwidthnew]{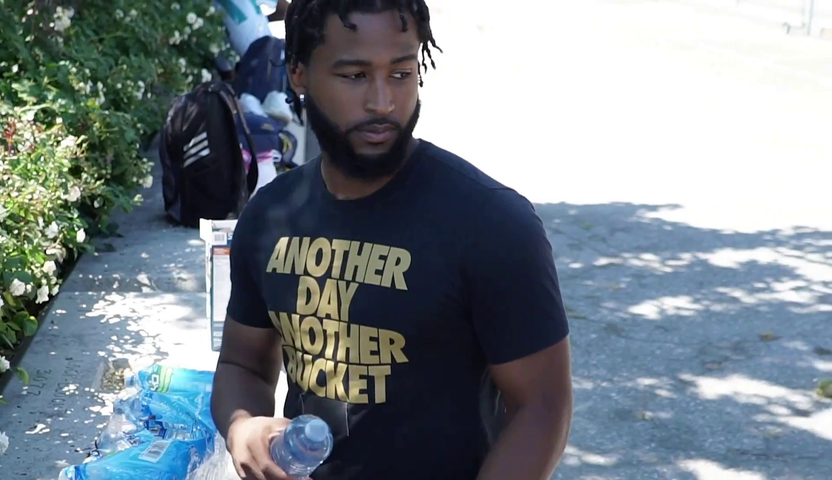} &
		\includegraphics[width=\moreresultwidthnew]{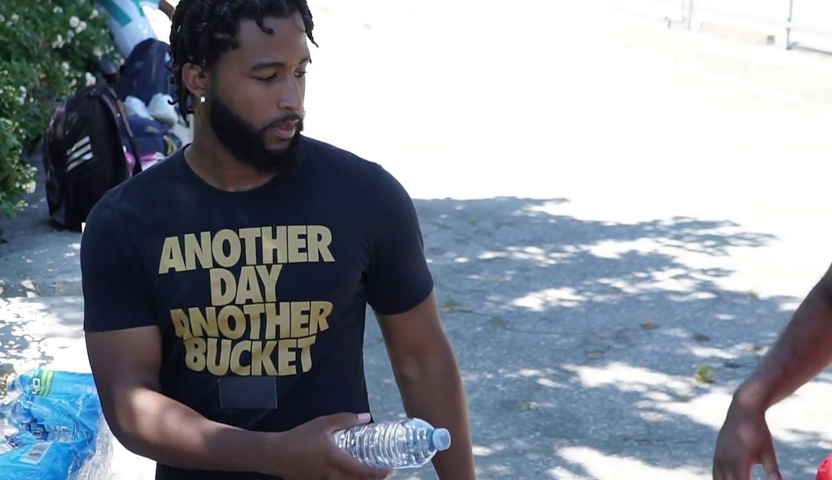} &
		\includegraphics[width=\moreresultwidthnew]{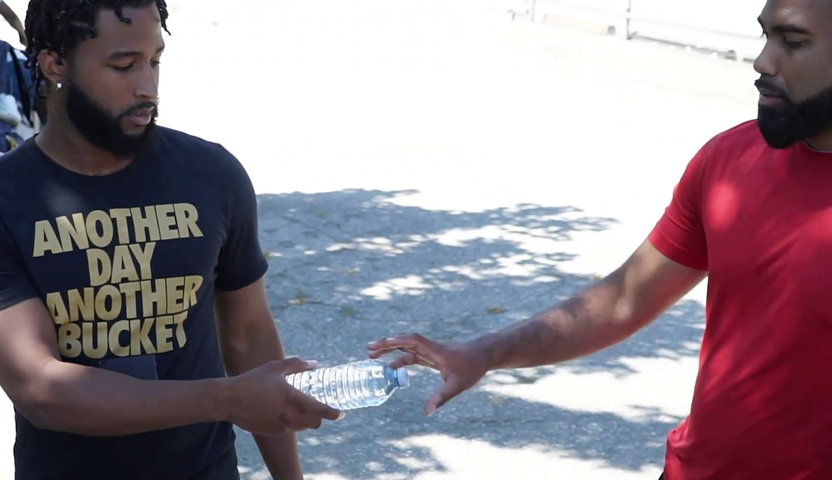} &
		\includegraphics[width=\moreresultwidthnew]{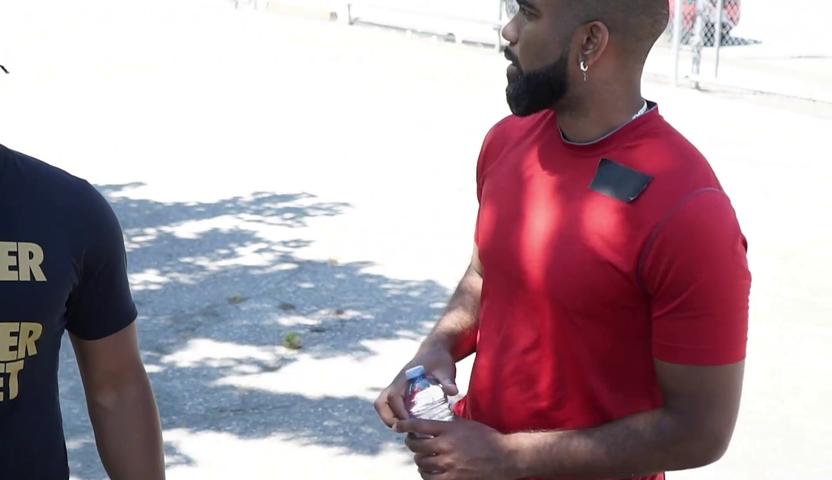} &
		\includegraphics[width=\moreresultwidthnew]{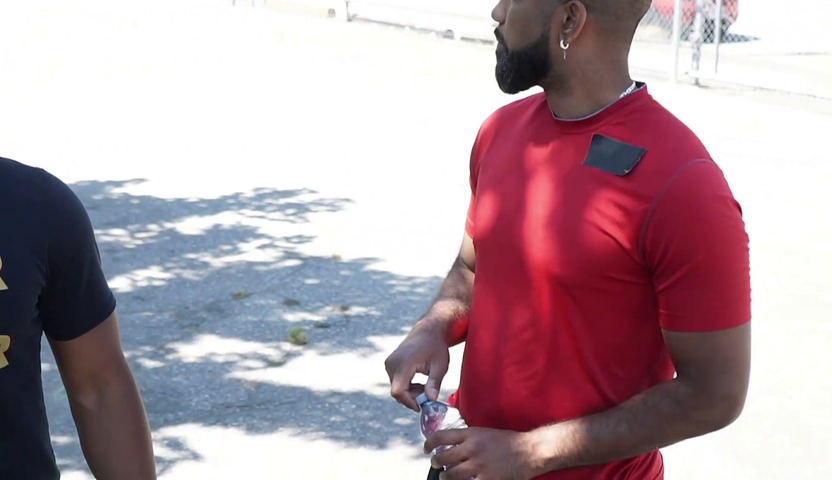} &
		\includegraphics[width=\moreresultwidthnew]{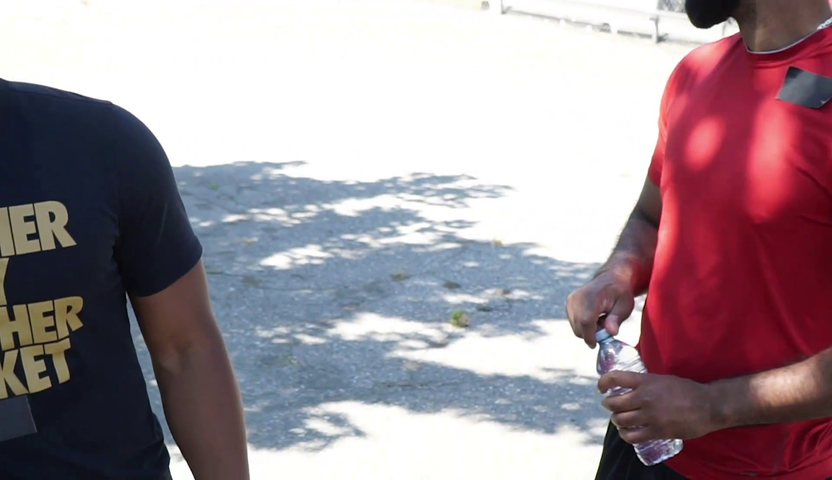} \\
		\raisebox{0.05\height}{\rotatebox{90}{\moreresultmasklabel}} &
		\includegraphics[width=\moreresultwidthnew]{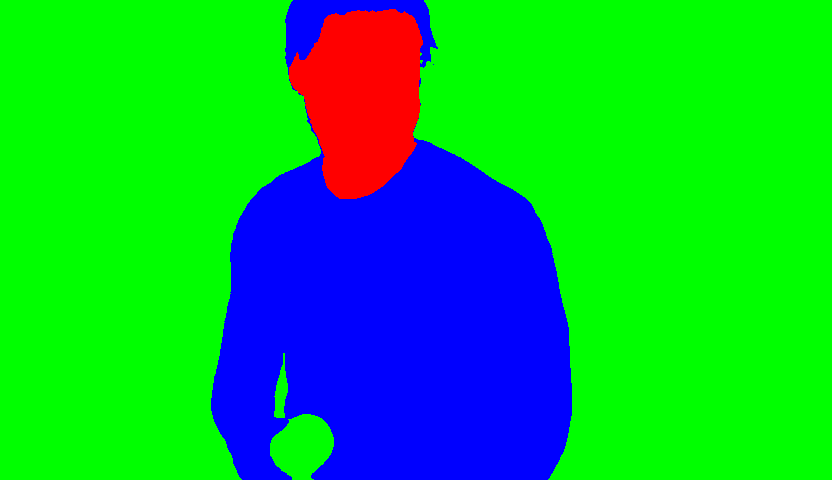} &
		\includegraphics[width=\moreresultwidthnew]{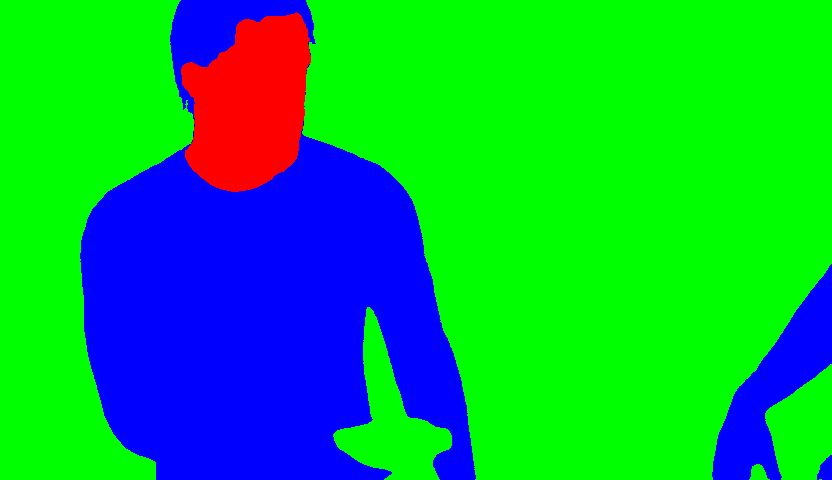} &
		\includegraphics[width=\moreresultwidthnew]{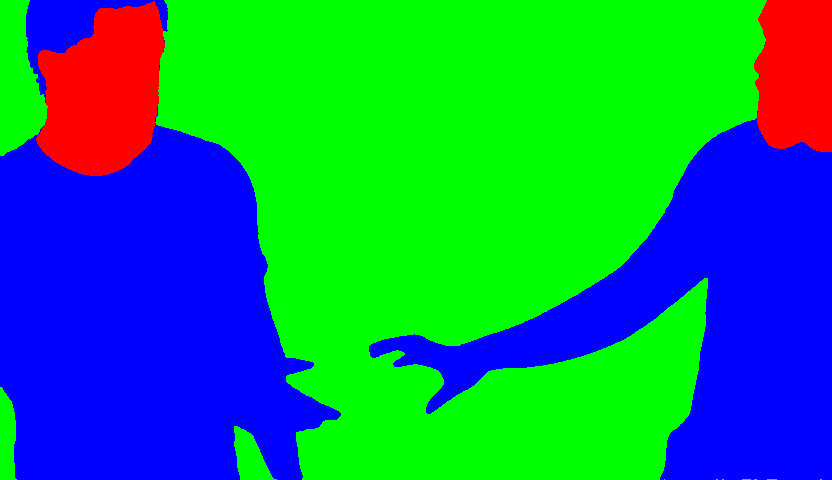} &
		\includegraphics[width=\moreresultwidthnew]{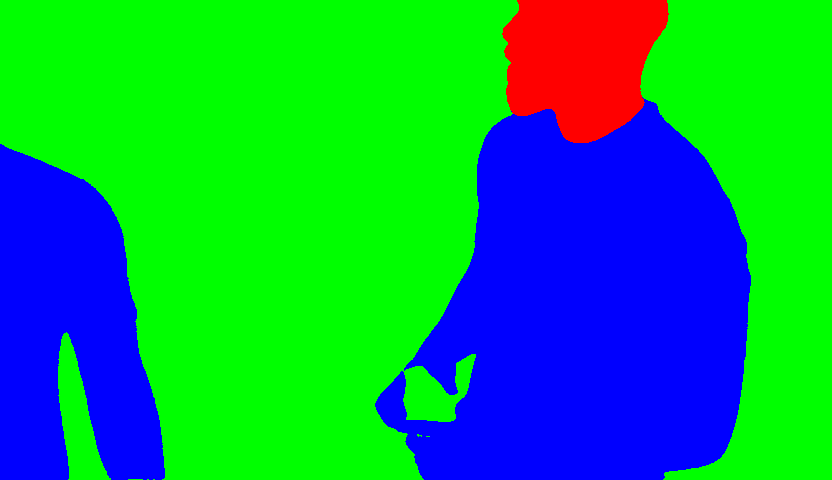} &
		\includegraphics[width=\moreresultwidthnew]{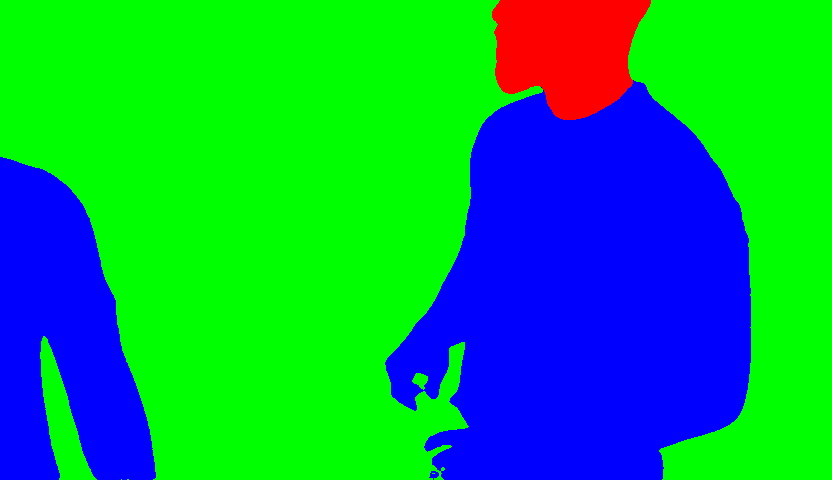} &
		\includegraphics[width=\moreresultwidthnew]{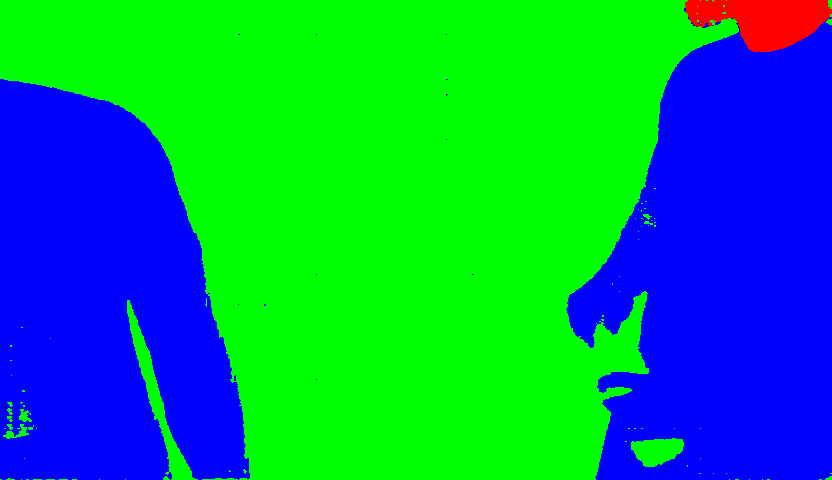} \\
		\raisebox{0.05\height}{\rotatebox{90}{\scriptsize Result}} &
		\includegraphics[width=\moreresultwidthnew]{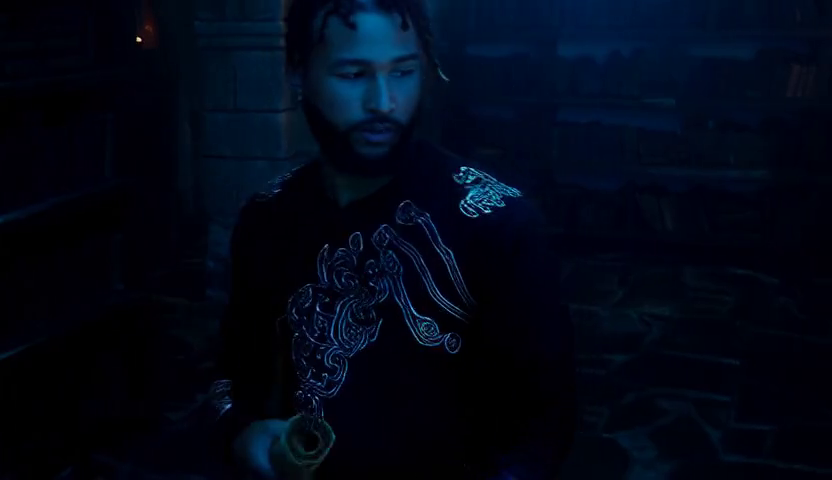} &
		\includegraphics[width=\moreresultwidthnew]{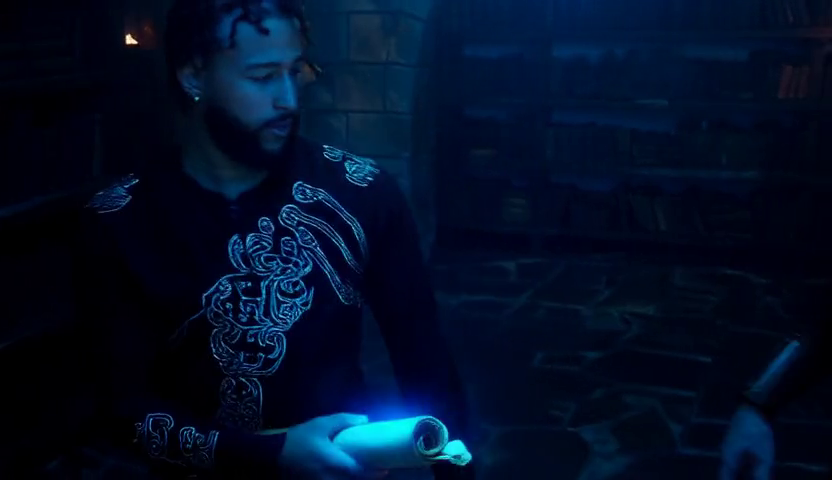} &
		\includegraphics[width=\moreresultwidthnew]{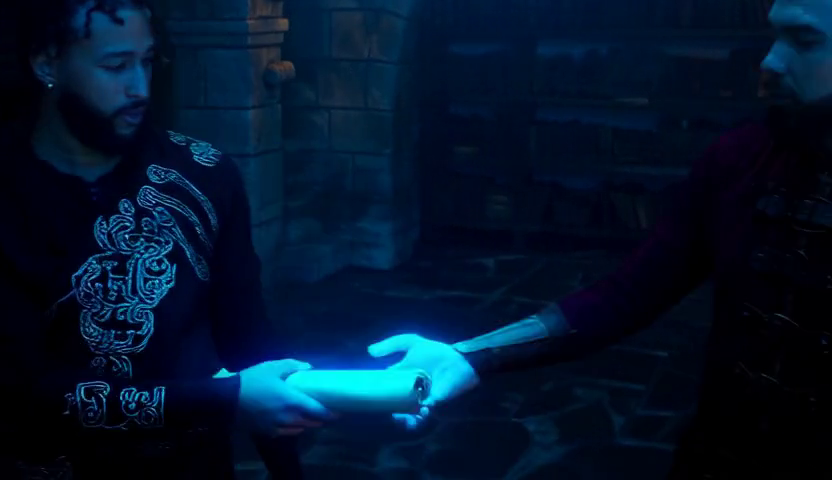} &
		\includegraphics[width=\moreresultwidthnew]{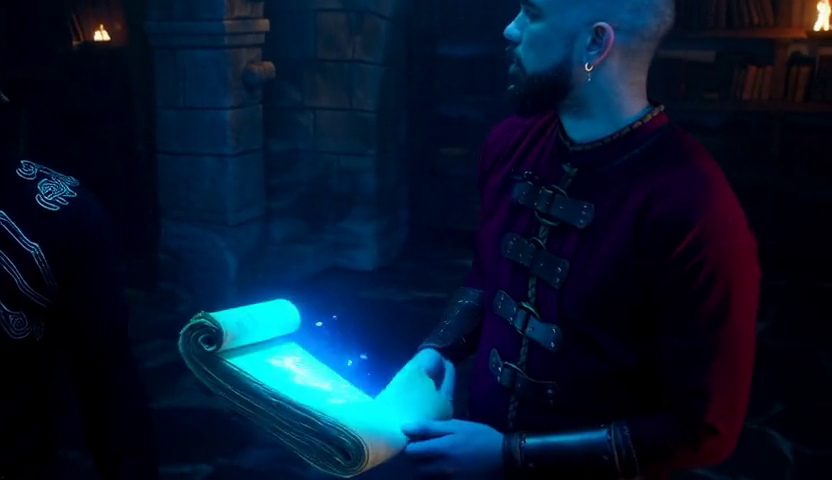} &
		\includegraphics[width=\moreresultwidthnew]{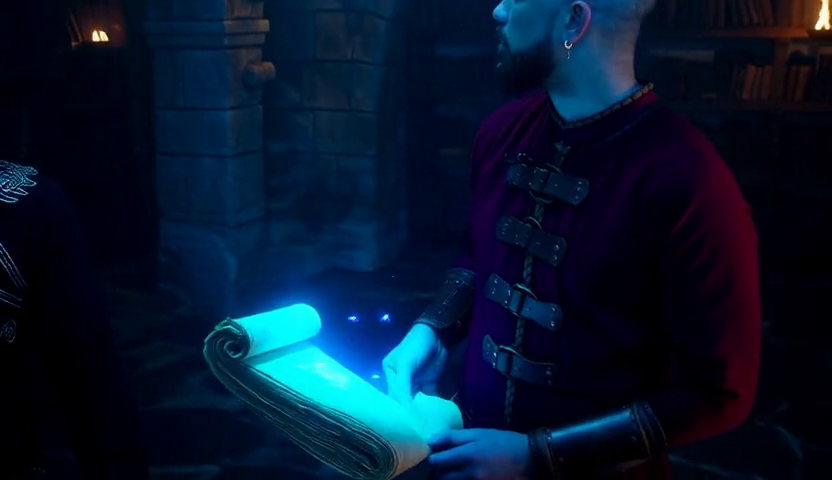} &
		\includegraphics[width=\moreresultwidthnew]{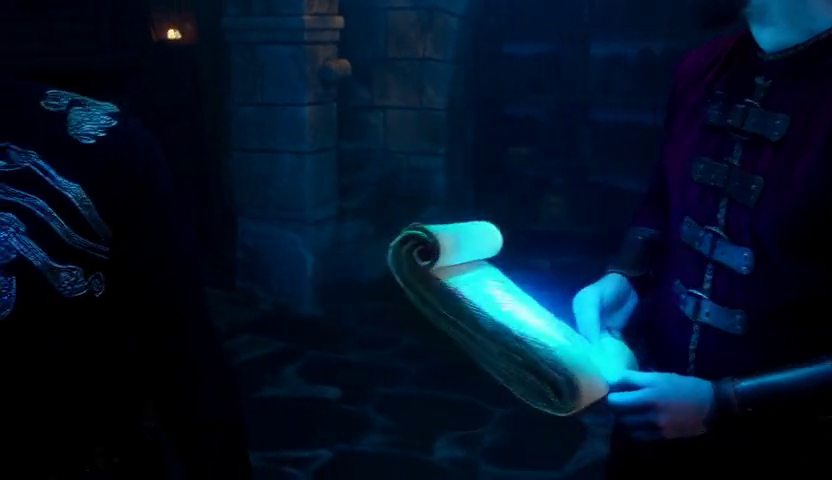} \\
		\multicolumn{7}{p{0.94\textwidth}}{\small Prompt: \textit{Two men exchange a glowing magical parchment scroll inside a mystical ancient library filled with stone arches, manuscripts, shadows, and torchlight.}} \\
		\raisebox{0.05\height}{\rotatebox{90}{\scriptsize Input Video}} &
		\includegraphics[width=\moreresultwidthnew]{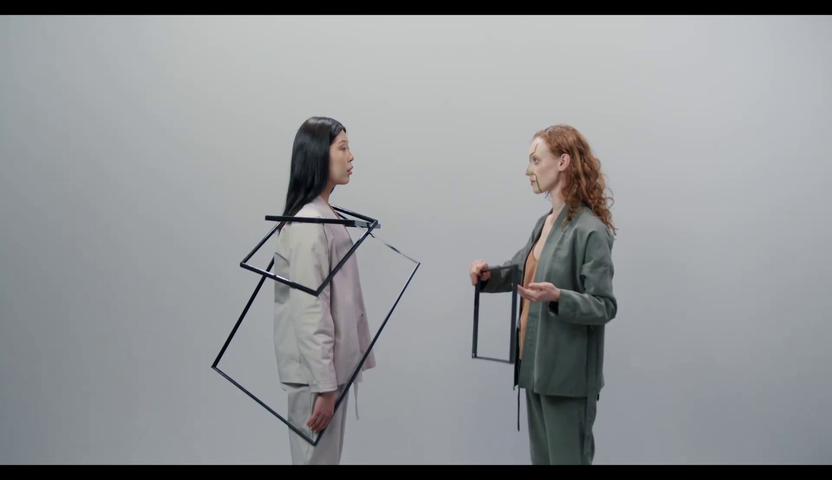} &
		\includegraphics[width=\moreresultwidthnew]{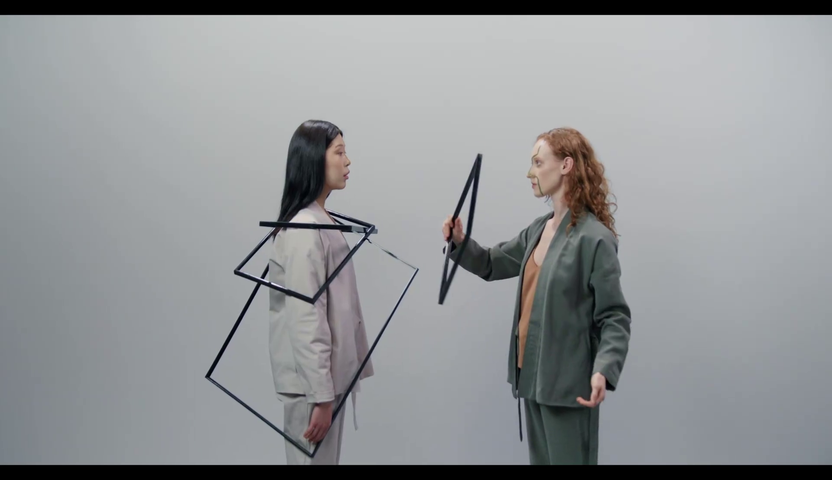} &
		\includegraphics[width=\moreresultwidthnew]{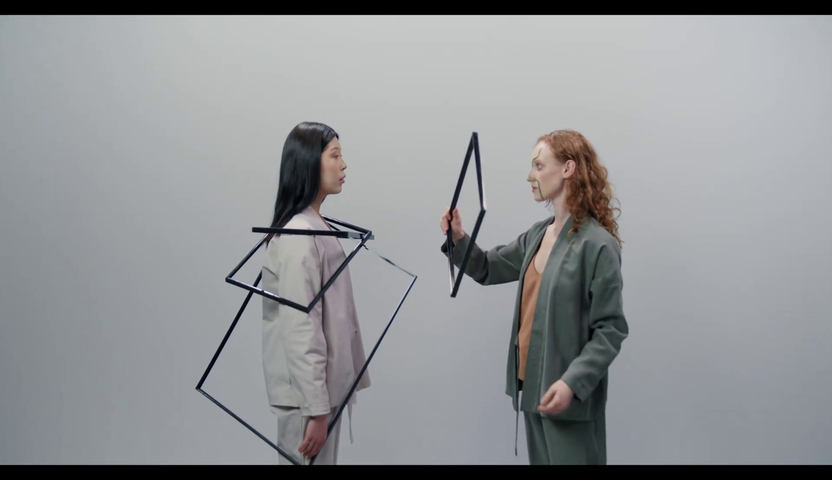} &
		\includegraphics[width=\moreresultwidthnew]{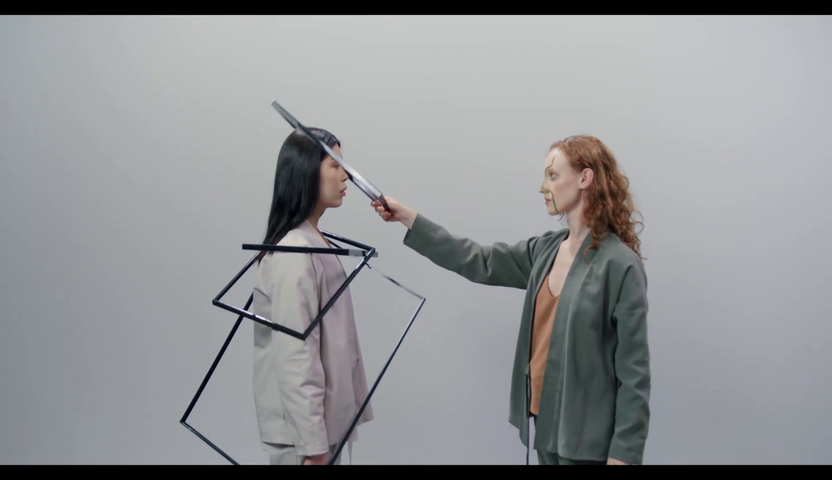} &
		\includegraphics[width=\moreresultwidthnew]{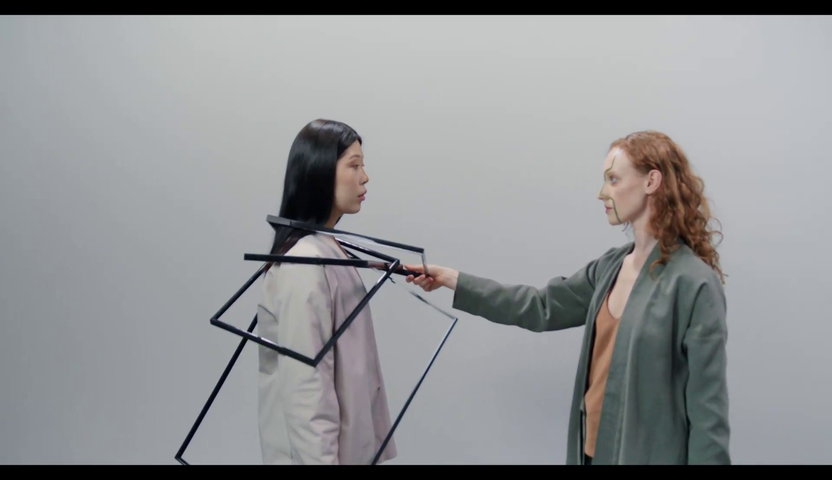} &
		\includegraphics[width=\moreresultwidthnew]{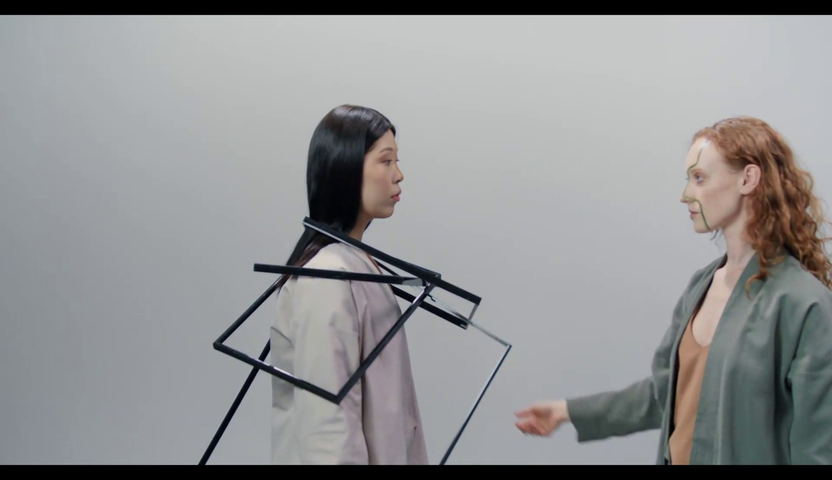} \\
		\raisebox{0.05\height}{\rotatebox{90}{\moreresultmasklabel}} &
		\includegraphics[width=\moreresultwidthnew]{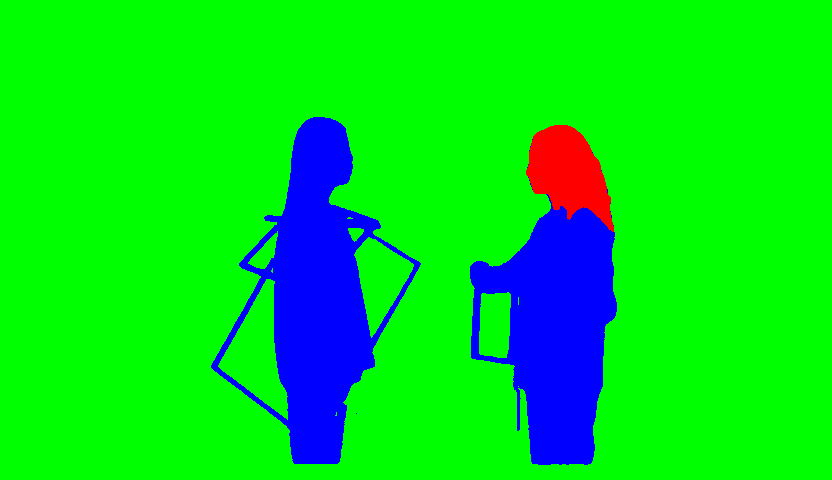} &
		\includegraphics[width=\moreresultwidthnew]{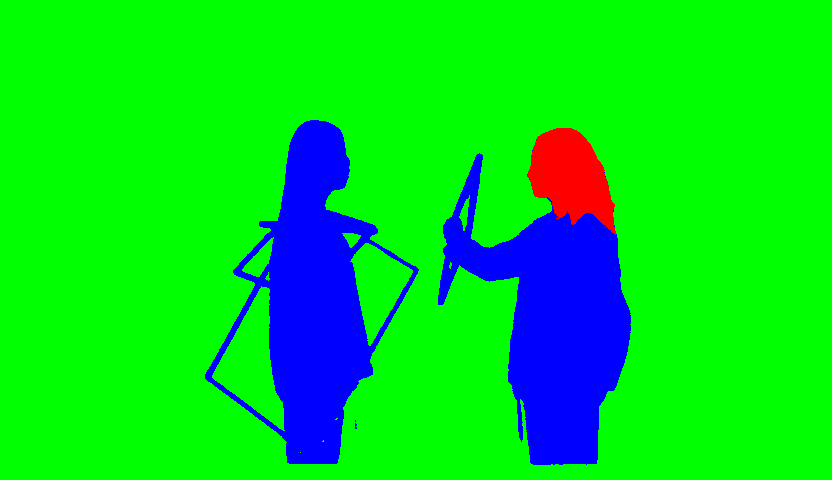} &
		\includegraphics[width=\moreresultwidthnew]{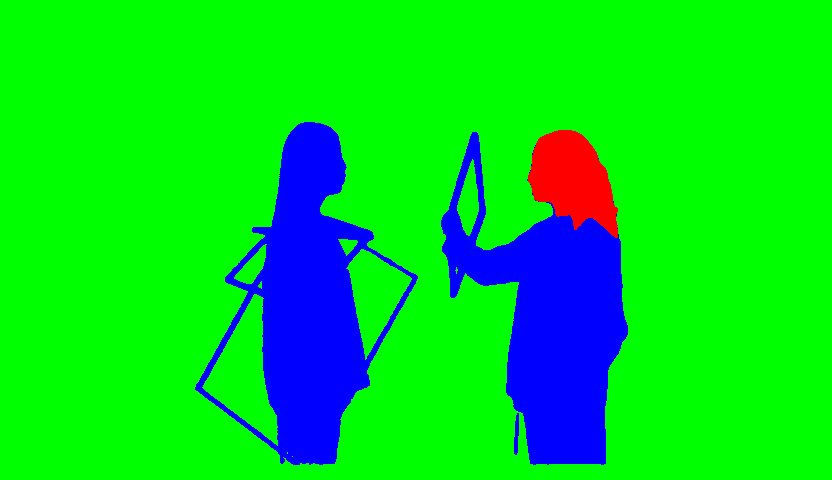} &
		\includegraphics[width=\moreresultwidthnew]{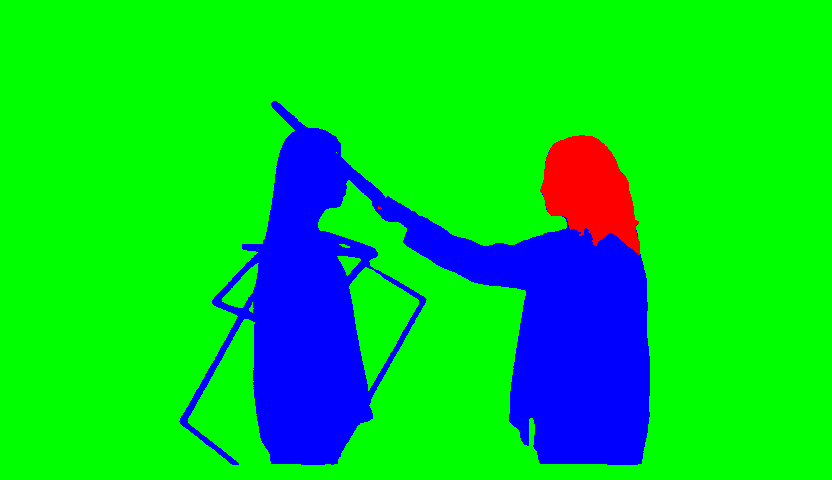} &
		\includegraphics[width=\moreresultwidthnew]{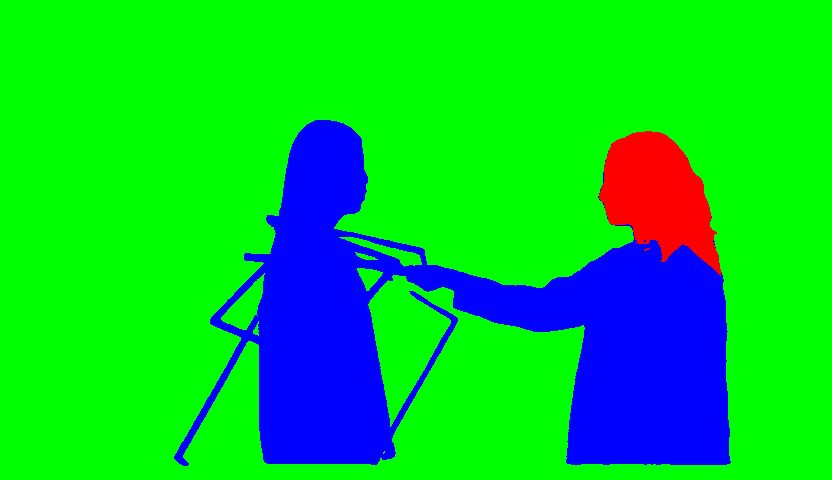} &
		\includegraphics[width=\moreresultwidthnew]{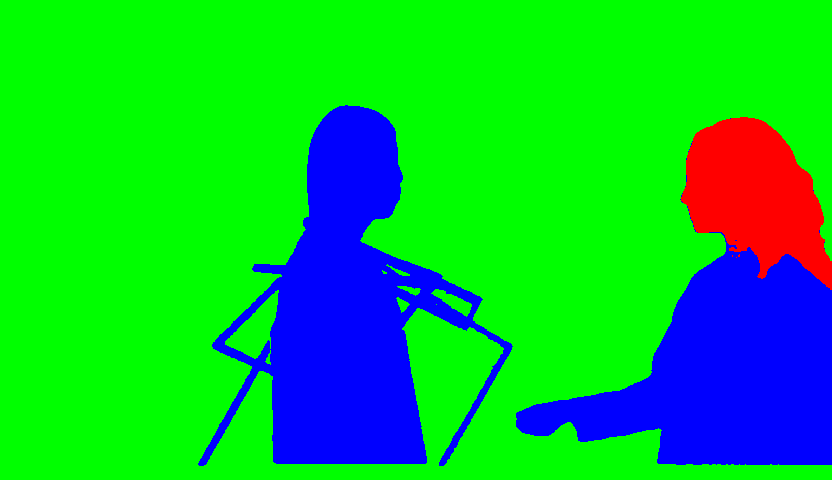} \\
		\raisebox{0.05\height}{\rotatebox{90}{\scriptsize Result}} &
		\includegraphics[width=\moreresultwidthnew]{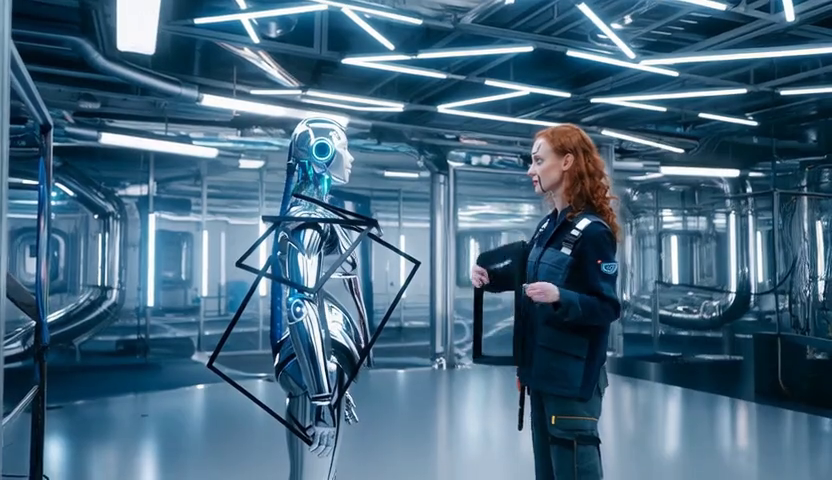} &
		\includegraphics[width=\moreresultwidthnew]{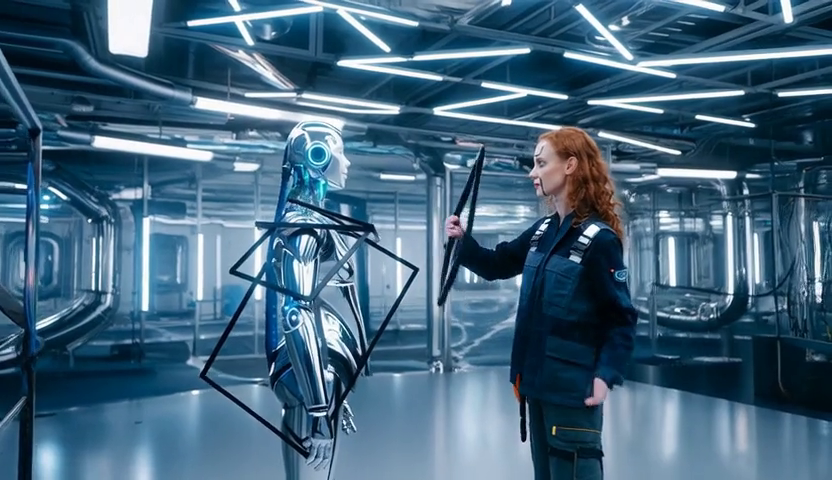} &
		\includegraphics[width=\moreresultwidthnew]{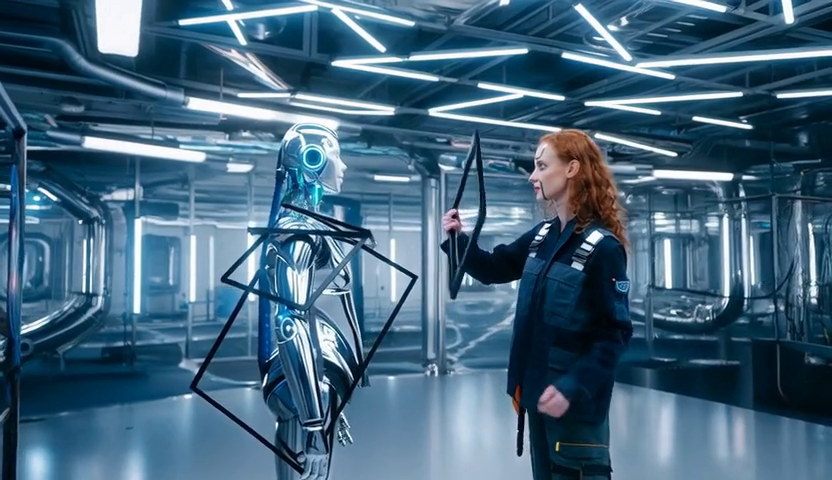} &
		\includegraphics[width=\moreresultwidthnew]{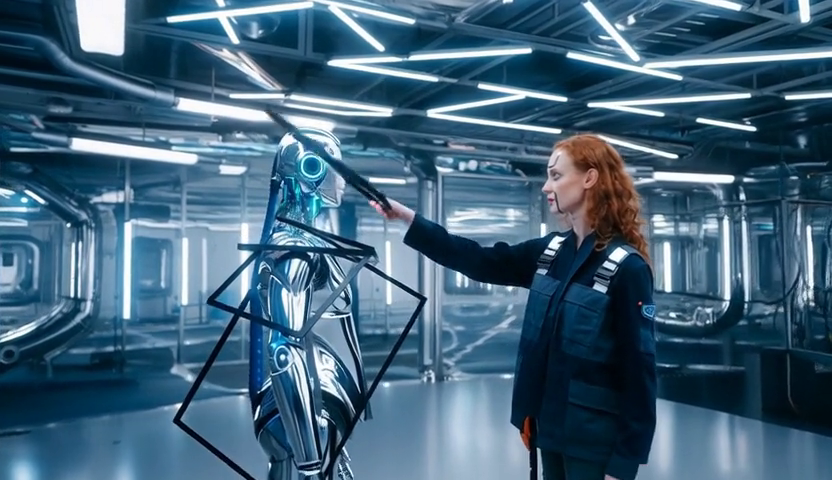} &
		\includegraphics[width=\moreresultwidthnew]{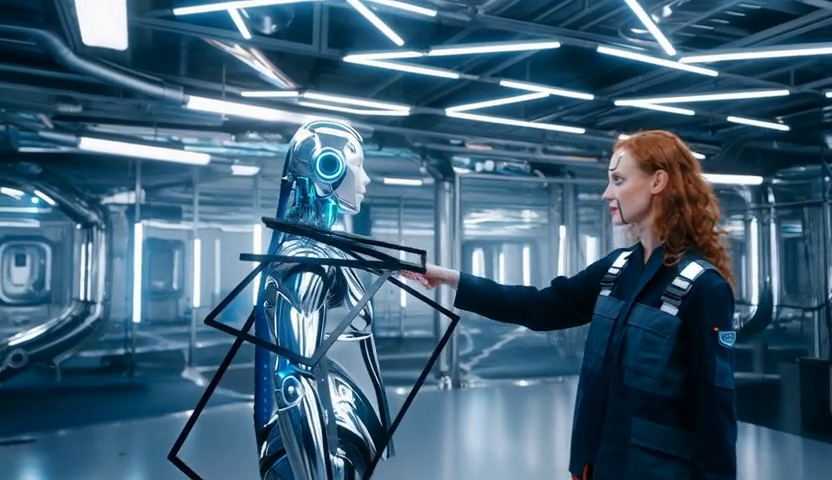} &
		\includegraphics[width=\moreresultwidthnew]{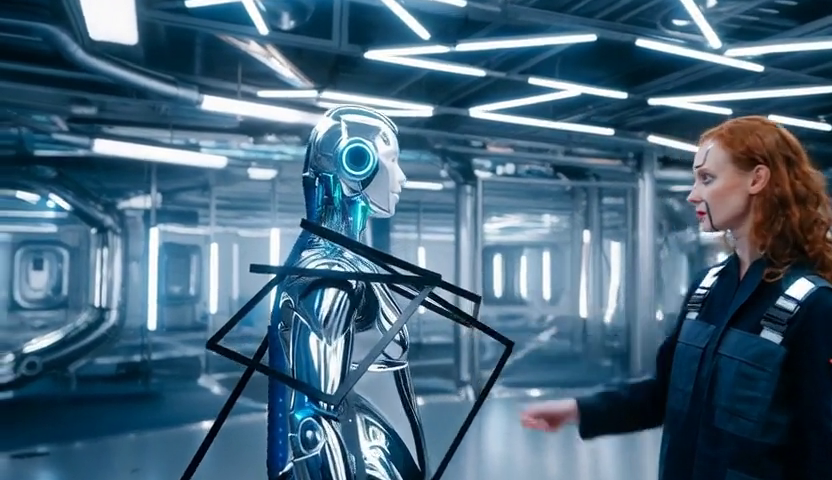} \\
		\multicolumn{7}{p{0.94\textwidth}}{\small Prompt: \textit{Inside a high-tech robotics laboratory, a cyberpunk mechanic lowers a hollow black geometric square frame over a reflective humanoid robot.}} \\
	\end{tabular}}
	\caption{\textbf{Applications on real-world videos.} Given different tri-masks, our model can flexibly replace or preserve arbitrary regions of a video while maintaining strong C2E and E2C interaction consistency.}
	
	\label{fig:more_result_0610_b}
\end{figure*}

\end{document}